\documentclass{article} 
\usepackage{iclr2025_conference,times}
\usepackage{subfiles}
\usepackage[table]{xcolor}
\usepackage{booktabs}   
\usepackage{multirow}   
\usepackage{caption}    
\usepackage{amsmath}    
\usepackage{enumitem}

\usepackage{amsmath,amsfonts,bm}

\def\eqref#1{equation~\ref{#1}}

\def\1{\bm{1}}

\DeclareMathAlphabet{\mathsfit}{\encodingdefault}{\sfdefault}{m}{sl}
\SetMathAlphabet{\mathsfit}{bold}{\encodingdefault}{\sfdefault}{bx}{n}

\usepackage{graphicx}
\usepackage{hyperref}
\usepackage{url}
\usepackage[ruled,vlined]{algorithm2e}
\usepackage{booktabs}
\usepackage{xcolor}
\definecolor{kqred}{RGB}{219,90,107}

\usepackage{xcolor}
\usepackage[most]{tcolorbox}
\usepackage{float}
\usepackage{booktabs}
\usepackage{multirow}
\usepackage{array}
\usepackage{subcaption}
\definecolor{DatasetBack}{RGB}{245,247,250}
\definecolor{DatasetFrame}{RGB}{200,210,225}
\newcommand{\fielditem}[1]{\par\noindent\textbullet\ \textbf{#1}: }

\newtcolorbox{datasetbox}{
  colback=DatasetBack,
  colframe=DatasetFrame,
  boxrule=0.4pt,
  arc=1pt,
  left=6pt,
  right=6pt,
  top=6pt,
  bottom=6pt,
  breakable,
  enhanced,
  before skip=10pt,
  after skip=10pt
}

\title{PepBenchmark: A Standardized Benchmark for Peptide Machine Learning}

\author{
Jiahui Zhang$^{1,2,*}$,
Rouyi Wang$^{1,3,*}$,
Kuangqi Zhou$^{1,*}$,
Tianshu Xiao$^{1,4}$,
\AND
Lingyan Zhu$^{3}$,
Yaosen Min$^{1,\dagger}$,
Yang Wang$^{2,\dagger}$ \\
\small
$^1$ Zhongguancun Academy \quad
$^2$ University of Science and Technology of China \\
$^3$ Nankai University \quad
$^4$ Tianjin University \\
$^*$ Equal contribution \quad
$\dagger$ Corresponding authors
}
\iclrfinalcopy 
\begin{document}

\maketitle

\begin{abstract}
Peptide therapeutics are widely regarded as the “third generation” of drugs, yet progress in peptide Machine Learning (ML) are hindered by the absence of standardized benchmarks. Here we present \textbf{PepBenchmark}, which unifies datasets, preprocessing, and evaluation protocols for peptide drug discovery. PepBenchmark comprises three components: (1) \textbf{PepBenchData}, a well-curated collection comprising 29 canonical-peptide and 6 non-canonical-peptide datasets across 7 groups, systematically covering key aspects of peptide drug development—representing, to the best of our knowledge, the most comprehensive AI-ready dataset resource to date; (2) \textbf{PepBenchPipeline}, a standardized preprocessing pipeline that ensures consistent dataset cleaning, construction, splitting, and feature transformation, mitigating quality issues common in ad hoc pipelines; and (3) \textbf{PepBenchLeaderboard}, a unified evaluation protocol and leaderboard with strong baselines across 4 major methodological families: Fingerprint-based, GNN-based, PLM-based, and SMILES-based models. Together, PepBenchmark provides the first standardized and comparable foundation for peptide drug discovery, facilitating methodological advances and translation into real-world applications. The data and code are publicly available at \href{https://github.com/ZGCI-AI4S-Pep/PepBenchmark/}{\texttt{https://github.com/ZGCI-AI4S-Pep/PepBenchmark/}}.

\end{abstract}

\section{Introduction}

Peptides—short chains of amino acids—are rapidly emerging as therapeutics, following small molecules and monoclonal antibodies as the “third generation" of drugs~\citep{Zheng2025TherapeuticPeptides}. 
They offer advantages such as synthetic accessibility, high biological specificity, and favorable safety profiles. 
With the growing availability of peptide data, Machine Learning (ML) has recently begun to transform peptide drug discovery, achieving early successes in anticancer~\citep{Arif2024PLMACPred} and antimicrobial~\citep{Wan2024MLforAMP,Du2023UniDL4BioPep}.

Despite recent successes, the development of peptide ML faces three major challenges. (1) Data sources are scattered across different databases and publications~\citep{ma2025dramp, li2023cycpeptmpdb, aguilera2023starpep, minkiewicz2008biopep, singh2016satpdb, cabas2024peptipedia}, forcing researchers to curate their own datasets, which is a labor-intensive process that often introduces inconsistency across studies. The problem is compounded by the heterogeneous representations of non-canonical peptides from different sources, which further limit progress in non-canonical peptide modeling~\citep{singh2025hemolytik2, li2023cycpeptmpdb}. (2) There are currently no standardized preprocessing pipelines for peptide data. As a result, steps such as redundancy removal, negative sampling, and dataset splitting vary considerably even when applied to the same raw data. Inappropriate preprocessing strategies may result in overly simplified datasets, which can artificially inflate model performance and thus fail to accurately reflect the true predictive capability of the models \citep{wang2025integrating, Wan2024MLforAMP}. (3) Evaluation metrics also differ substantially across studies. Collectively, these issues make it difficult to fairly compare different ML methods and to discern genuine methodological advances in this field.

\begin{figure}[h]
    \centering
    \includegraphics[width=1.0\textwidth]{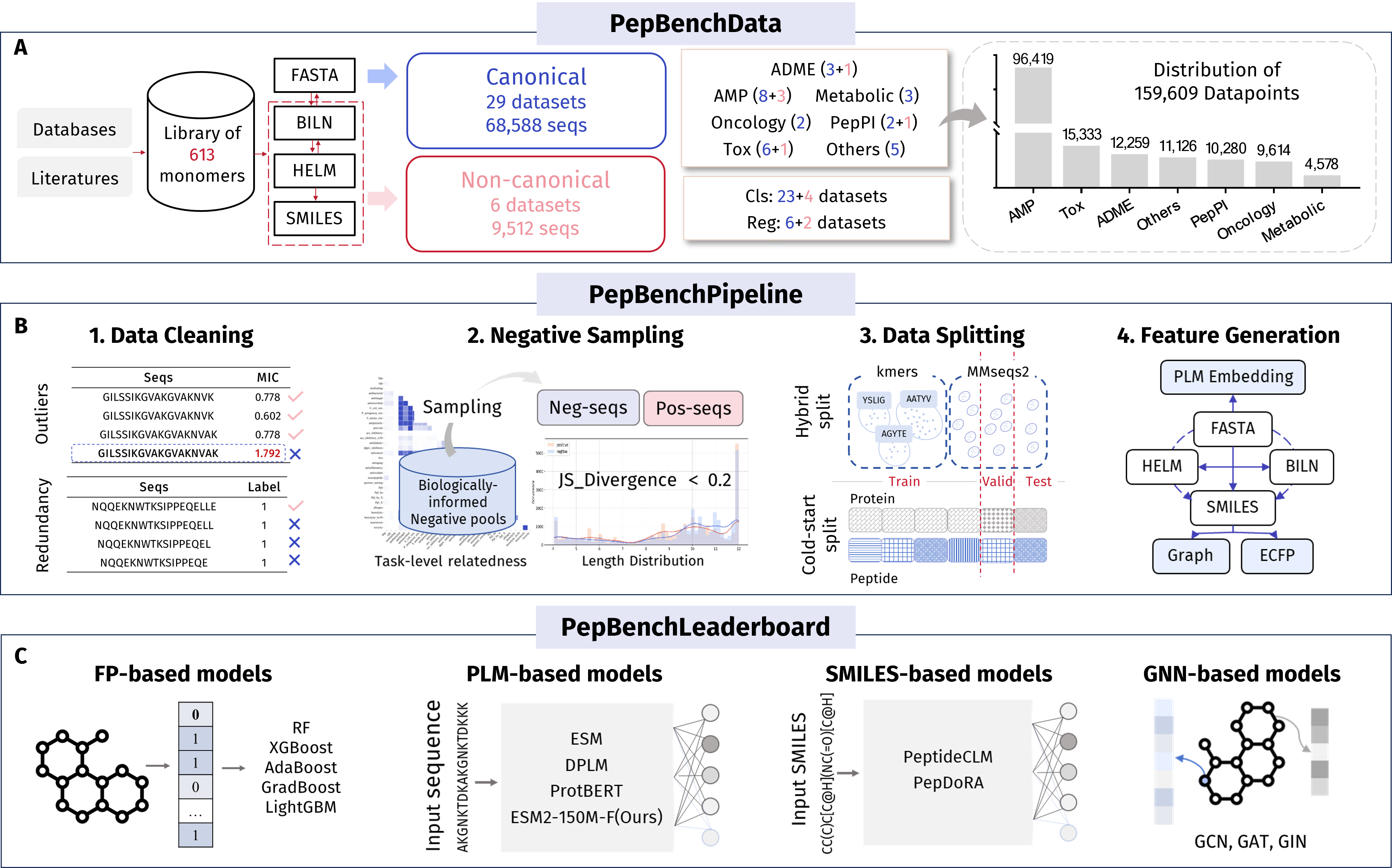}
    \caption{Overview of PepBenchmark, illustrating its three core components: PepBenchData, PepBenchPipeline, and PepBenchLeaderboard.}
    \label{fig:intro}
\end{figure}

To address these challenges, we introduce \textbf{PepBenchmark}, the most comprehensive peptide ML benchmark, offering a standardized and reproducible framework for unified datasets and model evaluation (Figure~\ref{fig:intro}). It comprises three core components:

\begin{itemize}[leftmargin=8pt, rightmargin=8pt]
\item \textbf{PepBenchData.} We comprehensively collect and process 35 datasets, providing the \emph{largest} AI-ready peptide database, which includes 29 canonical datasets (68,588 sequences) and 6 non-canonical datasets (9,512 sequences) across 7 groups related to the pharmacological properties of peptides. For non-canonical peptides, we collected 613 amino acid monomers and developed a tool that enables mutual conversion among commonly used representations, including BILN~\citep{fox2022biln}, HELM~\citep{zhang2012helm}, and SMILES, thereby providing a unified representation framework for non-canonical peptide modeling.
\item \textbf{PepBenchPipeline.} 
We propose a novel preprocessing pipeline, named \textit{PepBenchPipeline}, which standardizes the data curation process through four essential steps: data cleaning, negative sampling, dataset splitting, and feature generation. In \textbf{data cleaning}, we remove redundancy and filter outliers to ensure both data quality and representativeness. For \textbf{negative sampling}, we propose a \textbf{B}iologically-informed and \textbf{D}istribution-controlled strategy (\textbf{BDNegSamp}), which ensures that positive and negative samples are indistinguishable by shallow features while reducing the risk of pseudo-negatives. During \textbf{dataset splitting}, we address the overlooked issue of \textit{kmer leakage}, where frequent kmers disproportionately appear in positive samples and act as shortcut features, potentially inflating model performance. 
Finally, in \textbf{feature generation}, we provide a versatile toolkit for constructing diverse feature sets, enabling compatibility with different modeling paradigms and facilitating fair and comprehensive benchmarking.

\item \textbf{PepBenchLeaderboard.} 
Building on PepBenchData and PepBenchPipeline, we perform a large-scale comparison of four model families—fingerprint-based models, Graph Neural Networks (GNNs)~\citep{wu2020comprehensive}, Protein Language Models (PLMs)~\citep{ElnaggarProtBert}, and SMILES-based models—under a unified set of evaluation metrics. The comprehensive analysis not only benchmarks the current state of peptide ML models but also highlights directions for future development. Furthermore, by fine-tuning ESM2-150M on peptide-specific pretraining data, we develop \textbf{ESM-150M-F}, which achieves state-of-the-art performance.

\end{itemize}

Finally, to facilitate the use of our benchmark and future extensions, we release a Python package, also called \texttt{PepBenchmark}, which encapsulates all the three components. With just a few lines of code, users can load preprocessed datasets, perform standardized splitting, train models, and conduct evaluations (see Appendix~\ref{app:useage:sec}). 

Our paper is organized as follows. Section~\ref{sec:related work} reviews related work on peptide machine learning. Sections~\ref{sec:pbdata}, \ref{sec:pbpipeline}, and \ref{sec:pbleaderboard} present the three components of our benchmark—\textbf{PepBenchData}, \textbf{PepBenchPipeline}, and \textbf{PepBenchLeaderboard}, respectively. Finally, Section~\ref{sec:conclusion} concludes the paper.

\section{Related work}
\label{sec:related work}

\paragraph{ML in Peptide Prediction.}
ML models has been widely applied to predict therapeutically relevant properties of peptides, especially for canonical peptides. 
For example, amPEPpy~\citep{lawrence2021ampeppy} trains a Random Forest (RF) classifier on physicochemical descriptors to identify antimicrobial peptides, while AMPredictor~\citep{dong2025exploring} uses a graph convolutional network (GCN), incorporating both ESM embeddings and Morgan fingerprints as node features, for the prediction of minimum inhibitory concentration (MIC) of antimicrobial peptides. 
HemoPI2~\citep{rathore2025prediction} combines PLM embeddings and traditional ML to predict the hemolytic activity of peptides. Large language models (LLMs) are also widely applied to representation learning of canonical peptides. For example, AMPDesigner~\citep{Wang2025AMPDesigner} trains a GPT model on peptide sequences, whose embeddings can be used to achieve good MIC prediction performance. PepBERT~\citep{PepBert}, built on the BERT architecture, outperforms larger PLMs on a variety of downstream tasks. PeptideBERT~\citep{guntuboina2023peptidebert} fine-tunes ProtBert~\citep{ElnaggarProtBert} on peptide sequences and achieves state-of-the-art (SOTA) performance in predicting hemolysis and nonfouling properties. While these advances highlight the effectiveness of ML approaches for canonical peptides, modeling non-canonical peptides remains far less explored, largely due to the lack of unified and well-curated datasets. Current efforts mainly focus on transmembrane activity. For example, PeptideCLM~\citep{feller2025peptide}, a BERT-style transformer pretrained on large-scale synthetic non-canonical peptides, accurately predicts membrane diffusion ability of peptides. 
The Pepland~\citep{zhang2025pepland} pretraining framework uses a multi-view heterogeneous GNNs to generate representations of both canonical and non-canonical peptides, enabling prediction of transmembrane activity, Peptide-Protein Interaction (PepPI) and solubility. 
PepDoRA~\citep{wang2024pepdora} fine-tunes ChemBERTa to achieve accurate prediction of diverse therapeutic properties and PepPI. 

\paragraph{AI for Life Science Benchmarks}
Benchmarks play a crucial role in advancing machine learning by enabling standardized evaluation and comparison of models. 
In the field of AI for life science, numerous benchmarks have been established, with a focu   s on protein and small molecules~\citep{wu2018moleculenet, rao2019evaluating}.
For example, ProteinGym~\citep{notin2023proteingym} provides a large-scale, comprehensive benchmark for protein fitness prediction and design. 
For small molecules, TDC~\citep{huang2022artificial} offers a unified platform to access and evaluate machine learning across the entire drug development pipeline. In contrast, benchmarking efforts for peptide data remain scarce. To the best of our knowledge, only three attempts have been made to establish peptide benchmarks, none of which have seen wide adoption due to inherent limitations. 
The earliest effort, UniDL4BioPep~\citep{Du2023UniDL4BioPep}, covers 20 different peptide bioactivity data collected from a plethora of prior works, but is still limited in data scale and lacks standardized data processing pipelines. 
Peptipedia~\citep{cabas2024peptipedia}, collects data from over 70 databases covering 11 major biological activities, forming the largest collection of peptide activity data to our knowedge. However, the datasets in it are mostly not ready for ML models due to lack of preprocessing.
More relevant to our work is AutoPeptideML~\citep{fernandez2024autopeptideml}, which offers automated data-processing tools and a platform for building ML models for peptide bioactivity prediction. In comparison, we provide a larger-scale benchmark covering a broader set of ML tasks, and our data preprocessing pipelines enable more meaningful comparisons among models (as discussed in Section~\ref{app:exp:sec:addition_results}).

\section{PepbenchData}
\label{sec:pbdata}

PepbenchData represents the most comprehensive and systematic effort to date to organize machine learning (ML) tasks for peptide prediction. In total, 35 datasets have been collected and standardized (see Table~\ref{dataset_overview}). Each dataset has undergone unified cleaning and preprocessing, and is split into training, validation, and test sets according to a fixed ratio (8:1:1) to ensure comparability of model evaluation. The number of data points ranges from 277 to 52{,}941. 
Based on their roles in peptide development, the datasets are organized into seven groups across three stages of the pipeline.
\begin{itemize}[leftmargin=8pt, rightmargin=8pt]
        \item \textit{Activity Modeling:} This includes 5 groups related to peptide therapeutic activities, which can be further grouped into 2 types. The first comprises target-independent activities, including four groups: 11 antimicrobial peptide (AMP) datasets, two oncology-related datasets, four metabolic-disease-related datasets, and five datasets covering other bioactivities. The second category involves target-specific activities, represented by three peptide–protein interaction (PepPI) datasets that support the design of peptides targeting specific proteins.
        \item \textit{Pharmacokinetics Profiling:} This type includes a single group
        comprising 4 datasets that describes the pharmacokinetics-related properties of peptide drugs. In our paper, we follow the convention of the drug discovery field and call this type ``ADME'', which stands for Absorption, Distribution, Metabolism and Excretion.
        \item \textit{Safety Assessment:} This also consists of a single group, Tox, which includes 6 datasets capturing peptide toxicity-related properties for drug safety evaluation.
\end{itemize}

For varying requirements of various applications, we categorize the datasets from three orthogonal perspectives,
(1) \textbf{Input Type}: Single-input (32 datasets) for peptide-only dataset, and Multi-input (3 datasets) for peptide–protein interaction dataset; 
(2) \textbf{Peptide Type}: canonical (29 datasets), consisting of peptides composed solely of the 20 canonical amino acids, and non-canonical (6 datasets), consisting of peptides incorporate cyclic structures or non-canonical amino acid monomers; 
(3) \textbf{Task type}: classification tasks (27 datasets) or regression tasks (8 datasets).

In PepbenchData, canonical peptide datasets are mainly collected from benchmark datasets curated in prior literatures for specific tasks, supplemented with additional entries from the Peptipedia database~\citep{cabas2024peptipedia}. Non-canonical peptide datasets are primarily obtained from CycPeptMPDB~\citep{li2023cycpeptmpdb} and Hemolytic2~\citep{singh2025hemolytik2}, where raw non-canonical peptide sequences are originally encoded in HELM and MAP formats, respectively.
The two formats employ distinct monomer representations and cannot be directly converted into one another.
To unify these sources, we merge their monomer libraries and obtain a consolidated library of 613 unique monomers. 
All non-canonical peptide sequences are then converted into the SMILES format, which facilitates featurization.

Following Peptipedia, only sequences of length $\leq$150 are retained. 
We note that only 5 datasets contain $\geq$10\% of sequences longer than 50.
Therefore, given that a maximum length of 50 residues is commonly adopted for peptide drug candidates\citep{wang2025artificial, wang2025integrating}, we provide two versions: \textbf{PepbenchData-50} and \textbf{PepbenchData-150}.

\begin{table}[h!]
    \centering
    \scriptsize
    \caption{Overview of datasets in \textbf{PepBenchData}. 
    Here, ``bc'' denotes binary classification and ``reg'' denotes regression; 
    ``ca'' denotes canonical and ``n-ca'' denotes non-canonical. 
    Detailed data sources are provided in Table~\ref{app:des:tab:data_sources}.}

    \renewcommand{\arraystretch}{0}
    \setlength{\tabcolsep}{2pt}
    \resizebox{12cm}{!}{
    \begin{tabular}{cccccccc}
    \toprule
    Application & Dataset Name & Task Type & Peptide Type & Input Type & Unit & Size & Split \\
    \midrule
    \multirow{4}{*}{ADME} 
    & nonfouling        & bc  & ca   & single &           & 7200 & Hybrid\\
    & cpp               & bc  & ca   & single &           & 2296 & ECFP \\
    & bbp               & bc  & ca   & single &           & 665 & Hybrid \\
    & nc-cpp\_pampa     & reg & n-ca & single & log(cm/s) & 6970 & Hybrid\\
    \cmidrule{1-8}
    \multirow{11}{*}{AMP} 
    & antimicrobial     & bc  & ca   & single &           & 52941 & Hybrid\\
    & antibacterial     & bc  & ca   & single &           & 28591 & Hybrid\\
    & antifungal        & bc  & ca   & single &           & 12887 & Hybrid\\
    & antiparasitic     & bc  & ca   & single &           & 6755& Hybrid \\
    & antiviral         & bc  & ca   & single &           & 7785 & Hybrid \\
    & nc-antimicrobial  & bc  & n-ca & single &           & 2495 & ECFP  \\
    & nc-antibacterial  & bc  & n-ca & single &           & 1668 & ECFP  \\
    & nc-antifungal     & bc  & n-ca & single &           & 407 & ECFP  \\
    & E.coli\_mic       & reg & ca   & single & lg($\mu M$)    & 3204 & Hybrid \\
    & S.aureus\_mic     & reg & ca   & single & lg($\mu M$)    & 2822 & Hybrid\\
    & P.aeruginosa\_mic & reg & ca   & single & lg($\mu M$)    & 1490 & Hybrid\\
    \cmidrule{1-8}
    \multirow{4}{*}{Metabolic}
    & ace\_inhibitory       & bc  & ca   & single &        & 3537 & Hybrid \\
    & antidiabetic          & bc  & ca   & single &        & 3028 & Hybrid \\
    & dppiv\_inhibitors     & bc  & ca   & single &        & 1268 & Hybrid \\
    & ace\_inhibitory\_ic50 & reg & ca   & single & lg($\mu M$) & 337 & Random \\
    \cmidrule{1-8}
    \multirow{2}{*}{Oncology}
    & anticancer            & bc  & ca   & single &        & 12013 & Hybrid\\
    & ttca                  & bc  & ca   & single &        & 1182 & Hybrid \\
    \cmidrule{1-8}
    \multirow{5}{*}{Others}
    & neuropeptide      & bc  & ca   & single &        & 8687 & Hybrid \\
    & antiinflamatory   & bc  & ca   & single &        & 7665 & Hybrid \\
    & antioxidant       & bc  & ca   & single &        & 390 & Hybrid \\
    & antiaging         & bc  & ca   & single &        & 548 & Hybrid \\
    & quorum\_sensing   & bc  & ca   & single &        & 490 & Hybrid \\
    \cmidrule{1-8}
    \multirow{3}{*}{PepPI}
    & PpI               & bc  & ca   & multi  &        & 44148 & cold-start \\
    & PpI\_ba           & reg & ca   & multi  & -lg(M) & 1433 & cold-start  \\
    & nc-PpI\_ba        & reg & n-ca & multi  & -lg(M) & 277 & cold-start   \\
    \cmidrule{1-8}
    \multirow{6}{*}{Tox}
    & hemolytic         & bc  & ca   & single &        & 4306 & Hybrid\\
    & toxicity          & bc  & ca   & single &        & 4056 & Hybrid \\
    & neurotoxin        & bc  & ca   & single &        & 3159 & Hybrid \\
    & allergen          & bc  & ca   & single &        & 2538 & Hybrid \\
    & nc-hemolytic      & bc  & n-ca & single &        & 3971 & ECFP  \\
    & hemolytic\_hc50   & reg & ca   & single & lg($\mu M$) & 1926 & Hybrid\\
    \bottomrule
    \end{tabular}
    }
    \label{dataset_overview}
\end{table}


\section{PepBenchPipeline}
\label{sec:pbpipeline}

A standardized data preprocessing pipeline is essential for constructing reliable benchmark datasets. 
Here, we outline the 4 key components and the main design choices of each step, 
while detailed implementation can be found in Appendix~\ref{app:pipeline:sec:cls}, 
Appendix~\ref{app:pipeline:sec:reg}, and Appendix~\ref{app:pipeline:sec:ppi}.

\paragraph{Data Cleaning}

The data cleaning stage consists of 2 main procedures: \textit{outlier removal} for regression datasets and \textit{redundancy removal} for classification datasets.

(1) Outlier Removal: 
Many peptide datasets are derived from heterogeneous literature sources, where identical sequences often yield multiple experimental results due to variations in experiment conditions. 
Existing studies typically average all available measurements, while outlier detection is often overlooked~\citep{huang2023identification}, and thus the resulted label may be contaminated by outliers. 
To address this issue, we employ an InterQuartile Range (IQR)-based method to achieve robust outlier removal in regression datasets.

(2) Near-duplicate Redundancy Removal:
In peptide design, new candidates are often generated by introducing minor mutations into a known active peptide, which leads to substantial redundancy as many positive samples are nearly identical.
If not properly addressed, model may simply memorize the frequent patterns rather than learn a meaningful mapping. 
To mitigate this issue, \texttt{MMseqs2} is employed to remove sequences with more than 90\% similarity.

\paragraph{Negative Sampling (BDNegSamp)}  
In peptide-related studies, datasets typically contain only positive samples, since negative samples are extremely scarce. 
Consequently, negative samples must be manually constructed, and devising a principled sampling strategy remains a major challenge. 
Prior studies have mainly adopted two strategies: (1) randomly sampling sequences~\citep{Agrawal2021AntiCP, pinacho2021alignment, bin2020prediction}, or (2) using peptides from another activity dataset as negatives~\citep{Agrawal2021AntiCP, fernandez2025generalize}. 
The first strategy risks models exploiting superficial differences between random sequences and active peptides, while the second may lead models to simply learn activity-specific differences between the target property and the property used for negative sampling.
Moreover, both of them might introduce significant distributional shifts (e.g., in sequence length or charge), thereby simplifying the learning task and potentially leading to overly optimistic performance estimates. To address this issue, AutoPeptideML~\citep{fernandez2024autopeptideml} proposes sampling negatives from a negative sampling pool that consists of peptides with diverse bioactivities, while keeping the sequence length distribution close to that of the positive samples. However, this method does not consider task-to-task correlations. For instance, antimicrobial peptides often exhibit anticancer activity; thus, including them in the negative sampling pool for anticancer peptide prediction may introduce a substantial risk of false negatives. Based on above reasoning, we argue that a high-quality negative set must meet the two criteria below:
\begin{enumerate}
    \item There should not be dataset-specific artifacts (e.g., clear distributional differences, inherent contrasts between active and inactive peptides) between positive and negative data
    \item False negatives should be avoided as much as possible.  
\end{enumerate} 

This motivates us to propose the \textbf{B}iologically-informed and \textbf{D}istribution-controlled \textbf{N}egative Sampling (BDNegSamp), which includes 4 key steps:  
\begin{enumerate}[leftmargin=8pt, rightmargin=8pt]
    \item \textbf{Sampling pool construction.} Build a sampling pool by combining all the bioactivity data we collect. 
    \item \textbf{Task correlation filtering.} Use expert prior knowledge and statistical analysis to estimate correlation among tasks, and excluding data from tasks closely related to the target task.  
    \item \textbf{Similarity filtering and deduplication.} Remove sequences in the sampling pool that are highly similar to positive samples, and eliminate duplicate sequences.  
    \item \textbf{Distribution-controlled sampling.} Perform sampling from the filtered sampling pool while ensuring that multiple sequence-level features (such as length, net charge, and hydrophobicity) of the sampled negative data closely matches those of the positive samples.  
\end{enumerate}

The above 4 steps work well for canonical petides, but would fail for non-canonical peptide data, for which there are insufficient sequences to form a negative sampling pool. To address this limitation, we train a generative model that takes a canonical peptide as input and generates a chemically modified non-canonical peptide. This strategy effectively transforms the negative sampling problem for non-canonical peptides into one defined in the canonical peptide space (see Appendix~\ref{app:nc_gen:sec}).

\paragraph{Data Splitting}

Data splitting strategy critically affects the credibility of model evaluation. 
Conventional random splitting fail to capture the distribution shifts encountered in real-world drug discovery. Inspired by practices in protein modeling, recent studies~\citep{fernandez2024autopeptideml, pinacho2021alignment} attempt to lower similarity between training and test sets by similarity-based splitting using \texttt{MMseqs2}~\citep{steinegger2017mmseqs2} or CD-HIT. However, our analysis reveals that relying solely on sequence similarity is insufficient --- we identify an important but previously overlooked issue: \textbf{kmer leakage}.  

In particular, in most peptide datasets, some kmers show significant distributional difference betwen positive and negative samples (which we refer to as \textit{representative kmers}). Such kmers may correspond to activity-related motifs that experts intentionally introduce during peptide design. 
This would allow models to simply memorize these local shortcuts rather than learning biologically meaningful patterns, resulting in poor generalization to new active motifs. Importantly, \texttt{MMseqs2} cannot resolve this issue, since two sequences sharing the same kmer may exhibit low overall sequence similarity, and thus would not be clustered together.  To address this issue, we introduce two splitting strategies. 
The first is \textbf{kmer-split}, which contains three steps: (i) apply Fisher’s exact test to identify \textit{representative kmers}.; (ii) group peptides sharing the same \textit{representative kmer} into the same cluster; and (iii) assign clusters to the training, validation, and test sets. 

The second strategy is \textbf{hybrid-split}, which combines kmer-split with \texttt{MMseqs2}-based similarity clustering. This is achieved by first applying kmer-split and then using similarity-based cluster to further split sequences that does not contain \textit{representative kmers}. Hybrid-split simultaneously prevents \textbf{kmer leakage} and enforces similrity-aware grouping, yielding a more rigorous evaluation protocol. We recommend hybrid-split as the default splitting strategy.

So far, we have discussed dataset splitting for single-input canonical peptide datasets. We next consider the \texttt{PepPI} task involving multi-input data, as well as datasets of non-canonical peptides. For the \texttt{PepPI} task, we adopt a \textbf{cold-start split} following the paradigm used in previous studies, where proteins serve as the basis for data splitting: proteins are clustered based on similarity using \texttt{MMseqs2}, and the cold-start setting is enforced, meaning that proteins appearing in the training set are excluded from the test set. For non-canonical peptides, where FASTA sequences are not available, we adopt a splitting strategy based on ECFP fingerprints. Specifically, we compute pairwise similarities between peptides using molecular fingerprints, construct a similarity graph where edges connect pairs exceeding a similarity threshold, and then extract connected components as clusters. Each component is subsequently assigned to training, validation, or test sets.

\paragraph{Feature Generation}
Feature generation plays a crucial role in machine learning. To facilitate related research, we provide a variety of feature generation and transformation methods, including ECFP-based molecular fingerprint construction, pretrained language model (PLM) embeddings, as well as conversion tools among BILN, HELM, and SMILES representations.

\section{PepBenchLeaderboard}
\label{sec:pbleaderboard}

We introduce \textbf{PepBenchLeaderboard}, a standardized benchmark designed for the systematic evaluation of peptide property prediction models. The main body of our experiments is conducted on PepBenchData-50, while results on PepBenchData-150, which includes longer peptide sequences, are presented in Appendix~\ref{app:exp:sec:addition_results}. 
To facilitate the organization of experiments, we categorize all tasks based on two dimensions: input type (\textit{single peptide} or \textit{peptide--protein}) and peptide type (\textit{canonical} or \textit{non-canonical}). Accordingly, the experiments are grouped into three major parts:
\textbf{S}ingle-Input \textbf{C}anonical \textbf{P}eptide \textbf{P}rediction (\textbf{SCPP}), 
\textbf{S}ingle-Input \textbf{N}on-\textbf{C}anonical \textbf{P}eptide \textbf{P}rediction (\textbf{SNCPP}), 
and \textbf{Pep}tide--\textbf{P}rotein \textbf{I}nteraction (\textbf{PepPI}).
In the following sections, we first describe the experimental setup and then present the results and analyses for each of these three parts.

\subsection{Experimental Setup}

\paragraph{Datasets Description.}
SCPP includes 22 canonical peptide classification datasets and 4 canonical regression datasets. SNCPP includes 4 non-canonical peptide classification datasets and 1 non-canonical regression dataset. PepPI consists of three datasets: non-canonical regression (\texttt{nc-PPI\_ba}), canonical regression (\texttt{PPI\_ba}), and canonical classification (\texttt{PPI}). Detailed dataset descriptions are provided in Table~\ref{dataset_overview}.

\paragraph{Dataset Split.}
For each task, data are divided into training, validation, and test sets in an 8:1:1 ratio. For \textbf{SCPP}, to ensure a rigorous evaluation of generalization, we employ the \textit{hybrid-split} strategy. Sequences containing enriched $k$-mers are first assigned to the training set, and the remaining sequences are then clustered using MMseqs2 with a 30\% sequence identity threshold.  
For \textbf{SNCPP}, we use an ECFP-split strategy, where peptides are clustered based on fingerprint similarity (threshold set to 0.95). 
For \textbf{PepPI}, we adopt a strict protein-based cold-start split, ensuring that no protein appears across training, validation, and test sets. Additionally, to enforce sequence-level diversity and support robust generalization assessment, protein clusters (30\% identity) are assigned entirely to one partition using MMseqs2.

\paragraph{Benchmarking Model Families.}
For \textbf{SCPP}, we benchmark four families of models, each defined by a distinct peptide representation. \textbf{(1) Fingerprint-based (FP-based):} Traditional machine learning models—including Random Forest, AdaBoost, Gradient Boosting, XGBoost, and LightGBM--are trained on 1024-bit ECFP6 molecular fingerprints. \textbf{(2) GNN-based:}  
Graph Neural Networks (GNNs) represent peptides as atom-level molecular graphs. We evaluate classical architectures such as \texttt{GCN}~\citep{kipf2016semiGCN}, \texttt{GAT}~\citep{velickovic2018graphGAT}, and \texttt{GIN}~\citep{xu2019howGIN}, as well as \textit{Pepland}~\citep{zhang2025pepland}, a GNN framework pretrained on large-scale peptide data. \textbf{(3) SMILES-based:}  
This family encodes peptides as chemical language sequences in SMILES format. \texttt{ChemBERTa-77M-MLM} extends BERT with chemistry-specific adaptations to better capture the structural features of SMILES strings. \texttt{PepDoRA} fine-tunes \texttt{ChemBERTa-77M-MLM} with a masked language modeling objective, producing optimized embeddings for downstream property prediction involving both modified and unmodified peptides. \texttt{PeptideCLM} is a peptide-oriented chemical language model capable of encoding peptides with chemical modifications and cyclizations. \textbf{(4) PLM-based:}  
Protein Language Models (PLMs) leverage large-scale pretraining to generate sequence-based embeddings. We evaluate models from the \texttt{ESM2} family (8M--650M parameters), the \texttt{DPLM} family (150M, 650M)~\citep{wang2024dplm}, and \texttt{ProtBERT}~\citep{ElnaggarProtBert}. To further investigate the effect of pretraining, we include two variants of ESM2: \texttt{ESM2-8M-Scratch} (ESM-8M-S), which is trained from random initialization based on \texttt{ESM2-8M}, and \texttt{ESM2-150M-Finetune} (ESM-150M-F), which is obtained by further pretraining \texttt{ESM2-150M} on 1.9M short peptides (lengths $<50$) from UniRef50~\citep{Suzek2015UniRefClusters} (see Appendix~\ref{app:exp:esm_150_f}). For \textbf{SNCPP}, since non-canonical peptides lack FASTA representations, PLM-based approaches are inapplicable. Thus, evaluation is limited to FP-, GNN-, and SMILES-based models. For \textbf{PepPI}, peptide and protein embeddings are concatenated and fed into a multilayer perceptron (MLP). To isolate the contribution of the peptide encoder, the protein encoder is fixed as a frozen \texttt{ESM2-150M} model (see Appendix~\ref{app:exp:sec:addition_results} for results with a trainable protein encoder). The peptide encoders under evaluation cover the same four model families introduced above, with one distinction: for the FP-based family, both ECFP4 and ECFP6 fingerprints are employed as peptide representations.

\paragraph{Metrics.}  
We report the mean and standard deviation over five independent dataset splits. ROC-AUC is used for classification tasks, and Mean Absolute Error (MAE) for regression tasks.

\subsection{Results and Analysis}  
We next present experimental analyses for \textbf{SCPP}, \textbf{SNCPP}, and \textbf{PepPI}. Due to space constraints, only representative results are reported in the main text, while the complete results are provided in Appendix~\ref{app:exp:sec:addition_results}.

\subsubsection{Results and Analysis for SCPP}  
Tables~\ref{tab:natural_cls} and~\ref{tab:natural_reg} summarize the performance of SCPP on classification and regression tasks, respectively. The results reveal several noteworthy observations.



\begin{table}[h]
    \centering
    \caption{Performance of models on canonical peptide classification (ROC-AUC$\uparrow$, \%) with hybrid-split. Dataset sizes are shown separately; results are mean$_{\pm std}$. Best and second-best scores per row are in \textbf{bold} and gray shadow.}
    \label{tab:natural_cls}
    \resizebox{\textwidth}{!}{%
    \definecolor{lightgray}{gray}{0.9}
    \begin{tabular}{l|c|ccc|cc|cc|ccccccc}
    \toprule
    \multirow{2}{*}{Dataset} & \multirow{2}{*}{Size} & \multicolumn{3}{c|}{FP-based models} & \multicolumn{2}{c|}{GNN-based models} & \multicolumn{2}{c|}{SMILES-based models} & \multicolumn{7}{c}{PLM-based models} \\
    \cmidrule(lr){3-5} \cmidrule(lr){6-7} \cmidrule(lr){8-9} \cmidrule(lr){10-16}
    &  & RF & XGBoost & LightGBM & GIN & Pepland & ChemBERTa & PepDoRA & DPLM-150M & ESM2-8M & ESM2-35M & ESM2-150M & ESM2-650M & ESM2-8M-S & ESM2-150M-F \\
    \midrule
    
ace\_inhibitory & 3537  & $\mathbf{82.2_{\pm1.3}}$ & $80.4_{\pm1.8}$ & $\cellcolor{lightgray}{81.1_{\pm1.8}}$ & $78.8_{\pm1.4}$ & $72.4_{\pm3.6}$ & $77.3_{\pm2.5}$ & $70.3_{\pm2.0}$ & $79.7_{\pm1.0}$ & $79.2_{\pm1.4}$ & $80.7_{\pm1.9}$ & $80.2_{\pm0.5}$ & $79.6_{\pm1.9}$ & $79.1_{\pm1.6}$ & $80.2_{\pm2.3}$ \\
allergen & 2538  & $84.0_{\pm4.2}$ & $85.6_{\pm3.4}$ & $86.1_{\pm3.6}$ & $77.5_{\pm1.7}$ & $58.7_{\pm3.6}$ & $81.5_{\pm3.8}$ & $61.8_{\pm1.5}$ & $88.3_{\pm1.4}$ & $\cellcolor{lightgray}{88.3_{\pm2.6}}$ & $87.2_{\pm3.6}$ & $\mathbf{90.2_{\pm1.2}}$ & $87.4_{\pm3.8}$ & $83.2_{\pm2.5}$ & $86.8_{\pm1.8}$ \\
antiaging & 548  & $\cellcolor{lightgray}{66.5_{\pm1.9}}$ & $63.3_{\pm5.6}$ & $65.3_{\pm2.4}$ & $53.2_{\pm11.7}$ & $64.0_{\pm12.2}$ & $50.6_{\pm6.2}$ & $55.6_{\pm3.5}$ & $64.0_{\pm7.1}$ & $61.4_{\pm4.6}$ & $\mathbf{67.2_{\pm5.2}}$ & $62.5_{\pm3.3}$ & $63.6_{\pm5.8}$ & $58.1_{\pm2.2}$ & $61.9_{\pm3.9}$ \\
antibacterial & 28591  & $88.4_{\pm0.9}$ & $88.3_{\pm0.8}$ & $87.6_{\pm0.7}$ & $82.2_{\pm1.8}$ & $62.7_{\pm3.4}$ & $88.9_{\pm2.2}$ & $73.8_{\pm2.6}$ & $92.3_{\pm1.2}$ & $92.0_{\pm0.9}$ & $92.5_{\pm0.8}$ & $\cellcolor{lightgray}{93.0_{\pm1.2}}$ & $93.0_{\pm0.6}$ & $90.8_{\pm0.6}$ & $\mathbf{93.2_{\pm0.5}}$ \\
anticancer & 12013  & $87.7_{\pm1.0}$ & $87.6_{\pm0.9}$ & $87.1_{\pm0.6}$ & $78.9_{\pm2.6}$ & $69.5_{\pm4.6}$ & $87.5_{\pm1.2}$ & $71.4_{\pm2.8}$ & $92.2_{\pm0.9}$ & $\cellcolor{lightgray}{92.3_{\pm0.9}}$ & $92.3_{\pm0.9}$ & $92.3_{\pm0.8}$ & $\mathbf{92.4_{\pm0.6}}$ & $90.7_{\pm0.8}$ & $92.2_{\pm0.6}$ \\
antidiabetic & 3028  & $73.8_{\pm3.4}$ & $71.0_{\pm1.6}$ & $71.5_{\pm3.7}$ & $66.2_{\pm1.3}$ & $62.2_{\pm2.6}$ & $66.3_{\pm3.6}$ & $62.9_{\pm2.0}$ & $71.9_{\pm4.5}$ & $72.9_{\pm2.0}$ & $\cellcolor{lightgray}{74.1_{\pm4.1}}$ & $70.5_{\pm4.1}$ & $\mathbf{74.6_{\pm3.4}}$ & $69.8_{\pm2.5}$ & $72.7_{\pm2.5}$ \\
antifungal & 12887  & $87.3_{\pm0.8}$ & $87.3_{\pm0.8}$ & $87.1_{\pm0.8}$ & $78.5_{\pm3.6}$ & $53.5_{\pm3.5}$ & $87.0_{\pm1.1}$ & $70.3_{\pm1.2}$ & $90.7_{\pm1.4}$ & $90.6_{\pm0.6}$ & $90.7_{\pm1.1}$ & $\cellcolor{lightgray}{91.0_{\pm1.0}}$ & $90.6_{\pm0.8}$ & $87.8_{\pm2.2}$ & $\mathbf{91.1_{\pm0.7}}$ \\
antiinflamatory & 7665  & $74.3_{\pm1.4}$ & $72.4_{\pm1.0}$ & $73.7_{\pm1.2}$ & $71.1_{\pm3.0}$ & $57.7_{\pm2.7}$ & $73.5_{\pm2.4}$ & $62.4_{\pm1.8}$ & $77.4_{\pm2.3}$ & $77.3_{\pm1.9}$ & $\cellcolor{lightgray}{78.0_{\pm2.0}}$ & $76.8_{\pm2.3}$ & $77.3_{\pm1.9}$ & $74.6_{\pm1.7}$ & $\mathbf{79.9_{\pm2.0}}$ \\
antimicrobial & 52941  & $87.6_{\pm0.4}$ & $87.8_{\pm0.5}$ & $87.5_{\pm0.6}$ & $80.5_{\pm1.9}$ & $67.5_{\pm1.6}$ & $89.9_{\pm0.8}$ & $75.4_{\pm0.9}$ & $92.4_{\pm0.2}$ & $92.3_{\pm0.3}$ & $92.4_{\pm0.4}$ & $92.3_{\pm0.6}$ & $\cellcolor{lightgray}{92.6_{\pm1.1}}$ & $92.0_{\pm0.6}$ & $\mathbf{93.2_{\pm0.3}}$ \\
antioxidant & 390  & $\mathbf{68.1_{\pm2.7}}$ & $65.9_{\pm3.2}$ & $65.6_{\pm2.8}$ & $59.7_{\pm4.3}$ & $61.3_{\pm3.0}$ & $58.9_{\pm3.3}$ & $53.8_{\pm3.7}$ & $64.8_{\pm6.1}$ & $67.4_{\pm4.0}$ & $66.5_{\pm5.7}$ & $66.3_{\pm2.7}$ & $65.1_{\pm5.0}$ & $61.5_{\pm4.3}$ & $\cellcolor{lightgray}{68.0_{\pm3.5}}$ \\
antiparasitic & 6755  & $86.8_{\pm1.6}$ & $87.3_{\pm1.0}$ & $87.3_{\pm0.9}$ & $75.6_{\pm2.2}$ & $58.7_{\pm2.5}$ & $85.0_{\pm1.3}$ & $67.2_{\pm3.5}$ & $\mathbf{91.8_{\pm0.8}}$ & $90.7_{\pm1.4}$ & $90.8_{\pm1.1}$ & $91.1_{\pm0.9}$ & $91.4_{\pm1.0}$ & $88.0_{\pm1.4}$ & $\cellcolor{lightgray}{91.6_{\pm1.5}}$ \\
antiviral & 7785  & $84.2_{\pm0.9}$ & $83.6_{\pm1.0}$ & $84.5_{\pm1.3}$ & $74.6_{\pm4.9}$ & $71.0_{\pm3.1}$ & $81.2_{\pm2.9}$ & $59.7_{\pm2.2}$ & $86.3_{\pm2.2}$ & $86.0_{\pm1.1}$ & $85.2_{\pm1.6}$ & $86.2_{\pm1.5}$ & $\cellcolor{lightgray}{86.8_{\pm0.4}}$ & $84.1_{\pm2.8}$ & $\mathbf{87.1_{\pm1.1}}$ \\
bbp & 665  & $\cellcolor{lightgray}{69.3_{\pm6.4}}$ & $68.2_{\pm8.4}$ & $69.2_{\pm7.5}$ & $61.2_{\pm3.4}$ & $56.1_{\pm9.3}$ & $61.2_{\pm7.1}$ & $54.7_{\pm7.2}$ & $64.6_{\pm8.9}$ & $66.8_{\pm9.5}$ & $61.8_{\pm15.7}$ & $62.4_{\pm12.2}$ & $64.6_{\pm9.0}$ & $\mathbf{71.6_{\pm5.7}}$ & $64.6_{\pm12.5}$ \\
cpp & 2296  & $82.1_{\pm2.9}$ & $\cellcolor{lightgray}{82.3_{\pm4.9}}$ & $82.1_{\pm3.4}$ & $57.7_{\pm5.5}$ & $68.2_{\pm7.2}$ & $72.2_{\pm8.2}$ & $58.9_{\pm4.3}$ & $81.9_{\pm2.8}$ & $80.5_{\pm3.2}$ & $82.2_{\pm2.3}$ & $\mathbf{83.2_{\pm3.1}}$ & $80.1_{\pm6.0}$ & $79.6_{\pm8.0}$ & $81.9_{\pm3.1}$ \\
dppiv\_inhibitors & 1268  & $\cellcolor{lightgray}{82.4_{\pm3.0}}$ & $79.8_{\pm1.5}$ & $81.2_{\pm3.4}$ & $71.3_{\pm4.8}$ & $65.6_{\pm4.6}$ & $76.0_{\pm3.1}$ & $63.4_{\pm8.3}$ & $\mathbf{84.4_{\pm2.4}}$ & $81.5_{\pm2.2}$ & $80.5_{\pm3.8}$ & $79.9_{\pm2.3}$ & $80.5_{\pm0.7}$ & $73.2_{\pm4.0}$ & $81.8_{\pm2.9}$ \\
hemolytic & 4306  & $82.7_{\pm2.7}$ & $81.8_{\pm2.6}$ & $82.8_{\pm2.0}$ & $70.3_{\pm4.2}$ & $67.8_{\pm3.8}$ & $80.2_{\pm2.0}$ & $58.3_{\pm2.9}$ & $85.2_{\pm1.6}$ & $84.5_{\pm2.4}$ & $85.1_{\pm1.9}$ & $\cellcolor{lightgray}{85.3_{\pm1.4}}$ & $\mathbf{85.7_{\pm1.7}}$ & $82.9_{\pm2.9}$ & $84.3_{\pm2.7}$ \\
neuropeptide & 8687  & $84.1_{\pm1.7}$ & $84.0_{\pm1.8}$ & $84.3_{\pm1.7}$ & $72.5_{\pm1.2}$ & $67.8_{\pm3.2}$ & $78.8_{\pm2.6}$ & $66.8_{\pm1.8}$ & $\cellcolor{lightgray}{87.3_{\pm3.2}}$ & $85.2_{\pm3.6}$ & $86.7_{\pm2.3}$ & $86.3_{\pm2.3}$ & $\mathbf{88.0_{\pm1.1}}$ & $84.0_{\pm1.9}$ & $85.6_{\pm1.8}$ \\
neurotoxin & 3159  & $63.7_{\pm4.1}$ & $60.7_{\pm3.0}$ & $61.9_{\pm1.7}$ & $56.9_{\pm2.7}$ & $53.2_{\pm2.0}$ & $56.1_{\pm5.6}$ & $51.0_{\pm5.0}$ & $\cellcolor{lightgray}{71.0_{\pm4.1}}$ & $66.5_{\pm3.7}$ & $67.7_{\pm4.8}$ & $69.9_{\pm3.4}$ & $69.4_{\pm3.4}$ & $62.4_{\pm3.0}$ & $\mathbf{73.0_{\pm4.7}}$ \\
nonfouling & 7200  & $76.4_{\pm1.5}$ & $75.4_{\pm2.1}$ & $76.4_{\pm1.4}$ & $76.7_{\pm1.4}$ & $75.3_{\pm2.3}$ & $76.2_{\pm1.0}$ & $70.7_{\pm2.1}$ & $\cellcolor{lightgray}{78.0_{\pm0.8}}$ & $77.3_{\pm1.3}$ & $77.6_{\pm1.1}$ & $77.9_{\pm1.5}$ & $77.4_{\pm1.0}$ & $77.0_{\pm1.2}$ & $\mathbf{78.2_{\pm1.0}}$ \\
quorum\_sensing & 490  & $85.2_{\pm5.1}$ & $81.3_{\pm2.9}$ & $81.4_{\pm3.1}$ & $60.5_{\pm9.6}$ & $67.9_{\pm3.4}$ & $55.3_{\pm5.2}$ & $50.7_{\pm5.7}$ & $83.7_{\pm6.5}$ & $85.2_{\pm8.7}$ & $82.6_{\pm8.1}$ & $\cellcolor{lightgray}{85.4_{\pm5.8}}$ & $84.8_{\pm6.2}$ & $74.5_{\pm11.1}$ & $\mathbf{86.6_{\pm6.4}}$ \\
toxicity & 4056  & $64.3_{\pm1.4}$ & $63.2_{\pm2.1}$ & $64.1_{\pm1.2}$ & $55.4_{\pm4.0}$ & $53.4_{\pm3.9}$ & $59.5_{\pm5.8}$ & $53.1_{\pm2.6}$ & $\cellcolor{lightgray}{72.9_{\pm2.0}}$ & $69.6_{\pm3.0}$ & $71.6_{\pm2.7}$ & $72.3_{\pm2.6}$ & $72.8_{\pm3.3}$ & $65.8_{\pm1.8}$ & $\mathbf{78.0_{\pm3.0}}$ \\
ttca & 1182  & $\cellcolor{lightgray}{80.6_{\pm4.5}}$ & $78.1_{\pm8.6}$ & $78.8_{\pm7.0}$ & $70.5_{\pm8.1}$ & $71.4_{\pm4.1}$ & $67.0_{\pm4.9}$ & $54.8_{\pm4.7}$ & $79.2_{\pm7.1}$ & $79.7_{\pm6.2}$ & $79.0_{\pm5.7}$ & $79.2_{\pm4.4}$ & $75.7_{\pm7.3}$ & $66.8_{\pm6.9}$ & $\mathbf{81.7_{\pm5.8}}$ \\
\midrule
avg & -  & 79.2 & 78.0 & 78.5 & 69.5 & 63.9 & 73.2 & 62.1 & \cellcolor{lightgray}{80.9} & 80.3 & 80.6 & 80.6 & 80.6 & 77.2 & $\mathbf{81.5}$ \\
\bottomrule
\end{tabular}}
\end{table}


\begin{table}[h]
    \centering
    \caption{Performance of models on canonical peptide regression (MAE$\downarrow$, \%) with hybrid-split. Dataset sizes are shown separately; results are mean$_{\pm std}$. Best and second-best scores per row are in \textbf{bold} and gray shadow.}
    \label{tab:natural_reg}
    \resizebox{\textwidth}{!}{%
    \definecolor{lightgray}{gray}{0.9}
    \begin{tabular}{l|c|ccc|cc|cc|ccccccc}
    \toprule
    \multirow{2}{*}{Dataset} & \multirow{2}{*}{Size} & \multicolumn{3}{c|}{FP-based models} & \multicolumn{2}{c|}{GNN-based models} & \multicolumn{2}{c|}{SMILES-based models} & \multicolumn{7}{c}{PLM-based models} \\
    \cmidrule(lr){3-5} \cmidrule(lr){6-7} \cmidrule(lr){8-9} \cmidrule(lr){10-16}
    &  & RF & XGBoost & LightGBM & GIN & Pepland & ChemBERTa & PepDoRA & DPLM-150M & ESM2-8M & ESM2-35M & ESM2-150M & ESM2-650M & ESM2-8M-S & ESM2-150M-F \\
    \midrule
    
E.coli\_mic & 3204& $0.593_{\pm0.015}$& $0.604_{\pm0.019}$& $0.584_{\pm0.019}$& $0.631_{\pm0.015}$& $0.650_{\pm0.062}$& $0.603_{\pm0.007}$& $0.650_{\pm0.013}$& $0.525_{\pm0.006}$& $0.539_{\pm0.023}$& $0.527_{\pm0.004}$& $\cellcolor{lightgray}{0.509_{\pm0.010}}$& $\mathbf{0.488_{\pm0.008}}$& $0.548_{\pm0.024}$& $0.517_{\pm0.023}$ \\
P.aeruginosa\_mic & 1490& $0.540_{\pm0.030}$& $0.559_{\pm0.025}$& $0.550_{\pm0.026}$& $0.537_{\pm0.032}$& $0.566_{\pm0.037}$& $0.520_{\pm0.041}$& $0.546_{\pm0.029}$& $0.524_{\pm0.036}$& $0.506_{\pm0.046}$& $0.497_{\pm0.042}$& $0.496_{\pm0.046}$& $\mathbf{0.471_{\pm0.043}}$& $0.516_{\pm0.031}$& $\cellcolor{lightgray}{0.483_{\pm0.024}}$ \\
S.aureus\_mic & 2822& $0.569_{\pm0.035}$& $0.588_{\pm0.022}$& $0.572_{\pm0.025}$& $0.635_{\pm0.032}$& $0.648_{\pm0.016}$& $0.627_{\pm0.011}$& $0.657_{\pm0.029}$& $0.563_{\pm0.023}$& $0.545_{\pm0.030}$& $0.549_{\pm0.022}$& $\cellcolor{lightgray}{0.527_{\pm0.032}}$& $\mathbf{0.522_{\pm0.022}}$& $0.585_{\pm0.016}$& $0.533_{\pm0.028}$ \\
hemolytic\_hc50 & 1926& $0.517_{\pm0.042}$& $0.527_{\pm0.038}$& $0.523_{\pm0.035}$& $0.528_{\pm0.036}$& $0.516_{\pm0.033}$& $0.498_{\pm0.040}$& $0.535_{\pm0.027}$& $0.404_{\pm0.028}$& $0.422_{\pm0.042}$& $\cellcolor{lightgray}{0.400_{\pm0.021}}$& $0.413_{\pm0.047}$& $\mathbf{0.394_{\pm0.017}}$& $0.463_{\pm0.011}$& $0.412_{\pm0.020}$ \\
\midrule
avg & -& $0.555$& $0.569$& $0.557$& $0.583$& $0.595$& $0.562$& $0.597$& $0.504$& $0.503$& $0.493$& $0.486$& $\mathbf{0.469}$& $0.528$& $\cellcolor{lightgray}{0.486}$ \\
\bottomrule
\end{tabular}}
\end{table}

\begin{enumerate}[leftmargin=*, itemsep=0.5em]
    \item \textbf{Choice of Representation.} 
    We observe that PLM-based models consistently achieve the best performance, followed by FP-based models, whereas SMILES-based and GNN-based approaches perform the worst. The superior performance of PLM-based models underscores the critical role of large-scale pretraining on protein sequences, which can be effectively transferred to peptide property prediction. FP-based models, particularly RF, also exhibit competitive results. Although ECFP6 was originally designed for small molecules and is not necessarily optimal for peptides, its strong performance here suggests that developing peptide-specific fingerprint representations could be a promising research direction. In contrast, the relatively poor results of GNN- and SMILES-based models indicate that, for canonical peptides composed of the 20 standard amino acids, atom-level representations may introduce unnecessary redundancy.
    \item \textbf{Effect of Pretraining.} The pretrained \texttt{ESM2-8M} (80.3 ROC-AUC) outperforms its randomly initialized counterpart, \texttt{ESM2-8M-S} (77.2 ROC-AUC), confirming the benefits of large-scale protein pretraining. Moreover, the established performance hierarchy among protein PLMs ($\texttt{DPLM} > \texttt{ESM2} > \texttt{ProtBert}$) is consistently maintained on peptide prediction tasks, indicating effective cross-domain knowledge transfer. In contrast, \texttt{Pepland} and \texttt{PepDoRA}, which are pretrained on peptide data using graph and SMILES representations respectively, perform worse than their non-pretrained baselines (e.g., GIN, ChemBERTa). This suggests that naive, domain-specific pretraining with such representations can lead to negative transfer, the mitigation of which remains an open research question.

    \item \textbf{Scaling Laws of PLMs.}  
    For classification tasks, larger models (e.g., \texttt{ESM2-8M} $\to$ \texttt{ESM2-650M}) correspond to a general, albeit not uniform, increase in average performance. For regression tasks, the effect of scaling is more pronounced, suggesting that larger model capacity is particularly advantageous for quantitative prediction. 

    \item \textbf{Necessity of Peptide-Aware Finetuning for PLMs.}  
    PLMs are typically pretrained on large protein databases such as UniRef, in which short sequences ($\leq 50$ AAs) are severely underrepresented (~2.8\%). As a result, finetuning on a peptide-specific corpus becomes essential. Our finetuned model, \texttt{ESM2-150M-F}, achieves substantial gains over its parent model \texttt{ESM2-150M} on classification tasks, attaining state-of-the-art performance across all models. The improvements on regression tasks, however, are relatively modest. This suggests that peptide-aware finetuning is particularly critical for classification, while regression performance is more constrained by the intrinsic capacity of the parent PLM.

    \item \textbf{FP-based Methods in Low-Data Regimes.}
    On classification datasets, RF with ECFP6 fingerprints achieves an average ROC-AUC of 79.2, second only to PLMs. It frequently ranks first or second on smaller datasets such as \textit{ace\_inhibitory}, \textit{antiaging}, \textit{antioxidant}, \textit{bbp}, and \textit{dppiv\_inhibitors}. These results demonstrate that fingerprint-based methods remain a strong baseline in low-data regimes and further highlight the need to design peptide-specific descriptors, given that ECFP6 was originally developed for small molecules and is not ideally suited for peptides.

\end{enumerate}

\subsubsection{Results and Analysis For SNCPP}  
\begin{table}[h]
\centering
\caption{Performance of models on non-canonical peptide classification (ROC-AUC$\uparrow$, \%) and regression (MAE$\downarrow$, \%) with ECFP-split. Dataset sizes are shown separately; results are mean$_{\pm std}$. Best and second-best scores per row are in \textbf{bold} and gray shadow.}
\label{tab:nonnatural_cls_reg}
\resizebox{\textwidth}{!}{%
\definecolor{lightgray}{gray}{0.9}
\begin{tabular}{l|l|l|c|ccccc|cccc|ccc}
\toprule
\multirow{2}{*}{Task} & \multirow{2}{*}{Metric} & \multirow{2}{*}{Dataset} & \multirow{2}{*}{Size} & \multicolumn{5}{c|}{FP-based models} & \multicolumn{4}{c|}{GNN-based models} & \multicolumn{3}{c}{SMILES-based models} \\
\cmidrule(lr){5-9} \cmidrule(lr){10-13} \cmidrule(lr){14-16}
 & & & & RF & XGBoost & LightGBM & GradBoost & AdaBoost & GCN & GAT & GIN & Pepland & ChemBERTa & PeptideCLM & PepDoRA \\
\midrule

\multirow{6}{*}{cls} & \multirow{6}{*}{AUC ROC} & nc-antibacterial & 1668& $\mathbf{94.4_{\pm1.4}}$& $93.6_{\pm2.2}$& $\cellcolor{lightgray}{93.8_{\pm1.8}}$& $92.5_{\pm2.8}$& $90.9_{\pm3.4}$& $93.0_{\pm2.6}$& $81.5_{\pm3.5}$& $90.7_{\pm4.9}$& $84.9_{\pm3.9}$& $91.9_{\pm2.7}$& $71.4_{\pm1.8}$& $72.4_{\pm7.4}$ \\
& & nc-antifungal & 407& $\cellcolor{lightgray}{95.4_{\pm3.2}}$& $\mathbf{96.5_{\pm2.9}}$& $94.8_{\pm3.1}$& $95.2_{\pm3.8}$& $95.4_{\pm2.4}$& $70.1_{\pm12.0}$& $78.5_{\pm6.2}$& $86.1_{\pm2.8}$& $83.5_{\pm4.4}$& $78.0_{\pm18.3}$& $65.4_{\pm7.0}$& $65.9_{\pm12.0}$ \\
& & nc-antimicrobial & 2465& $97.6_{\pm0.9}$& $\cellcolor{lightgray}{97.7_{\pm0.9}}$& $\mathbf{97.8_{\pm0.6}}$& $97.3_{\pm0.7}$& $95.3_{\pm1.8}$& $94.9_{\pm1.6}$& $79.2_{\pm4.4}$& $91.5_{\pm2.3}$& $88.0_{\pm0.9}$& $95.2_{\pm1.9}$& $68.3_{\pm4.7}$& $80.0_{\pm1.9}$ \\
& & nc-hemolytic & 3971& $96.1_{\pm1.1}$& $\cellcolor{lightgray}{96.2_{\pm0.5}}$& $\mathbf{96.3_{\pm0.7}}$& $95.3_{\pm0.8}$& $93.7_{\pm1.6}$& $89.5_{\pm3.7}$& $85.7_{\pm2.6}$& $89.0_{\pm4.1}$& $82.8_{\pm2.1}$& $91.4_{\pm1.5}$& $76.1_{\pm4.5}$& $72.1_{\pm7.3}$ \\
\cmidrule(lr){3-16}
& & avg & -& $\cellcolor{lightgray}{95.9}$& $\mathbf{96.0}$& $95.7$& $95.1$& $93.8$& $86.9$& $81.2$& $89.3$& $84.8$& $89.1$& $70.3$& $72.6$ \\
\midrule
reg & MAE & nc-cpp & 6970& $\mathbf{0.649_{\pm0.006}}$& $0.705_{\pm0.026}$& $\cellcolor{lightgray}{0.651_{\pm0.016}}$& $0.683_{\pm0.007}$& $0.829_{\pm0.019}$& $0.754_{\pm0.027}$& $0.767_{\pm0.025}$& $0.701_{\pm0.027}$& $0.742_{\pm0.029}$ & $0.712_{\pm0.033}$& $0.822_{\pm0.035}$& $0.879_{\pm0.013}$ \\

\bottomrule
\end{tabular}}
\end{table}


As reported in Table~\ref{tab:nonnatural_cls_reg}, FP-based models achieve the best performance, while GNN- and SMILES-based models continue to lag behind. Intuitively, the chemical diversity provided by over 600 non-canonical monomers should favor atom-level representations; however, the empirical results do not support this hypothesis. We highlight two key directions for future research. First, finetuning existing PLMs may prove effective, as many non-canonical amino acids share structural backbones with the 20 canonical residues, suggesting that pretrained knowledge could transfer. Second, novel pretraining frameworks are needed for both canonical and non-canonical peptides. For sequence-level modeling, notations such as BILN or HELM present promising options, whereas for atom-level modeling the central challenge lies in capturing the modular structure of peptides. The main obstacle remains the scarcity of large-scale non-canonical peptide datasets, which we are beginning to address through rule-based generation of synthetic databases.

\subsubsection{Results and Analysis For PepPI}

\begin{table}[h]
    \centering
    \caption{Performance of models on PepPI classification (ROC-AUC$\uparrow$, \%) and regression (MAE$\downarrow$, \%) with cold-start-split. Dataset sizes are shown separately; results are mean$_{\pm std}$. Best and second-best scores per row are in \textbf{bold} and gray shadow.}
    \label{tab:ppi}
    \resizebox{\textwidth}{!}{%
    \definecolor{lightgray}{gray}{0.9}
    \begin{tabular}{ll|l|c|cc|cc|ccc|ccccc}
    \toprule
    \multirow{2}{*}{Task} & \multirow{2}{*}{Metric} & \multirow{2}{*}{Dataset} & \multirow{2}{*}{Size} & \multicolumn{2}{c|}{FP-based models} & \multicolumn{2}{c|}{GNN-based models} & \multicolumn{3}{c|}{SMILES-based models} & \multicolumn{5}{c}{PLM-based models} \\
    \cmidrule(lr){5-6} \cmidrule(lr){7-8} \cmidrule(lr){9-11} \cmidrule(lr){12-16}
    & & & & ECFP4 & ECFP6 & GIN & Pepland & ChemBERTa & PeptideCLM & PepDoRA & ESM2-8M & ESM2-35M & ESM2-150M & ESM2-650M & ESM2-150M-F \\
    \midrule
    
    cls & AUC ROC & ppi & 44148  & $54.4_{\pm2.3}$ & $53.7_{\pm2.3}$ & $\mathbf{61.3_{\pm7.2}}$ & $59.6_{\pm2.7}$ & $52.0_{\pm4.4}$ & $51.4_{\pm3.6}$ & $59.4_{\pm2.8}$ & \cellcolor{lightgray}$60.2_{\pm6.3}$  & $57.6_{\pm7.2}$ & $55.4_{\pm3.5}$ & $51.9_{\pm2.4}$ & $56.0_{\pm3.1}$ \\
    \midrule
    \multirow{2}{*}{reg} & \multirow{2}{*}{MAE} & ppi\_ba & 1433 & $1.043_{\pm0.050}$& $1.043_{\pm0.060}$& $1.189_{\pm0.140}$& $1.176_{\pm0.152}$& $\mathbf{1.034_{\pm0.013}}$& $1.128_{\pm0.042}$& $1.084_{\pm0.044}$& $1.061_{\pm0.043}$& $1.056_{\pm0.092}$& $1.079_{\pm0.038}$& $1.051_{\pm0.072}$& \cellcolor{lightgray}$1.038_{\pm0.030}$ \\
    & & nc\_ppi\_ba & 278 & $1.665_{\pm0.260}$& $1.647_{\pm0.286}$& $1.741_{\pm0.359}$& $1.705_{\pm0.366}$ & $1.613_{\pm0.189}$& \cellcolor{lightgray}$1.580_{\pm0.142}$& $\mathbf{1.465_{\pm0.234}}$& -& -& -& -& - \\

\bottomrule
\end{tabular}}
\end{table}


As shown in Table~\ref{tab:ppi}, both GNN- and SMILES-based models perform competitively on PepPI tasks. This outcome, which contrasts with their underperformance in single-peptide property prediction, suggests that atom-level representations are better suited for modeling peptide–protein interactions. PLM-based models achieve the second-best overall performance; however, on the larger \textit{ppi} dataset, they exhibit a counterintuitive trend—scaling up model size does not improve performance and in fact leads to degradation.

\subsection{Summary of Core Experimental Insights}
Across rigorously partitioned peptide benchmarks, PLMs emerge as the most effective approaches for single-peptide property prediction, while FP-based methods provide strong and data-efficient baselines. In particular, they dominate in non-canonical settings where PLMs are inapplicable. By contrast, SMILES- and GNN-based models generally lag behind. Considering that ECFP6, despite not being peptide-specific, already delivers solid performance, the development of peptide-tailored fingerprints appears highly promising. Pretraining and scale prove critical: pretrained ESM2 variants consistently outperform randomly initialized counterparts, and larger PLMs yield steadier improvements—especially for regression—while peptide-aware continued pretraining primarily benefits classification. In PepPI tasks, atom-level representations (GNN/SMILES) become competitive, suggesting that fine-grained molecular detail is more advantageous for modeling cross-molecular interactions than for single-peptide properties.


\section{Conclusion}
\label{sec:conclusion}
\textbf{PepBenchmark} establishes a unified, reproducible framework for peptide ML, resolving fragmentation in data curation, preprocessing, and evaluation. Empirically, PLM-based models achieve the strongest overall performance, FP-based methods remain competitive in low-data regimes, whereas current GNN/SMILES approaches lag behind; our \textit{hybrid-split} and \textit{ECFP-split} strategies mitigate leakage and yield more discriminative benchmarks. The accompanying Python package operationalizes standardized datasets, splits, and metrics, enabling fair comparison and rapid iteration across canonical and non-canonical peptide tasks. Together, these resources provide a practical foundation for advancing robust, generalizable peptide therapeutics.

\subsubsection*{Acknowledgments}
This work is supported by the Zhongguancun Academy, Beijing, China.

\bibliography{iclr2025_conference}

@article{Du2023UniDL4BioPep,
  author    = {Zhenjiao Du and Xingjian Ding and Yixiang Xu and Yonghui Li},
  title     = {UniDL4BioPep: a universal deep learning architecture for binary classification in peptide bioactivity},
  journal   = {Briefings in Bioinformatics},
  year      = {2023},
  volume    = {24},
  number    = {3},
  pages     = {bbad135},
  doi       = {10.1093/bib/bbad135},
  url       = {https://doi.org/10.1093/bib/bbad135}
}

@article{Wan2024MLforAMP,
  author    = {Fangqing Wan and Fei Wong and James J. Collins and Cesar de la Fuente-Nu{\~n}ez},
  title     = {Machine learning for antimicrobial peptide identification and design},
  journal   = {Nature Reviews Bioengineering},
  year      = {2024},
  volume    = {2},
  pages     = {392--407},
  doi       = {10.1038/s44222-024-00152-x},
  url       = {https://doi.org/10.1038/s44222-024-00152-x}
}

@article{zhang2012helm,
  title     = {HELM: a hierarchical notation language for complex biomolecule structure representation},
  author    = {Zhang, Tianhong and Li, Hongli and Xi, Hualin and Stanton, Robert V and Rotstein, Sergio H},
  journal   = {Journal of Chemical Information and Modeling},
  volume    = {52},
  pages     = {2796--2806},
  year      = {2012},
  publisher = {ACS Publications}
}

@article{fox2022biln,
  title={BILN: a human-readable line notation for complex peptides},
  author={Fox, Thomas and Bieler, Michael and Haebel, Peter and Ochoa, Rodrigo and Peters, Stefan and Weber, Alexander},
  journal={Journal of Chemical Information and Modeling},
  volume={62},
  number={17},
  pages={3942--3947},
  year={2022},
  publisher={ACS Publications}
}

@article{olsen2017tantigen,
  title={TANTIGEN: a comprehensive database of tumor T cell antigens},
  author={Olsen, Lars R{\o}nn and Tongchusak, Songsak and Lin, Honghuang and Reinherz, Ellis L and Brusic, Vladimir and Zhang, Guang Lan},
  journal={Cancer Immunology, Immunotherapy},
  volume={66},
  number={6},
  pages={731--735},
  year={2017},
  publisher={Springer}
}

@article{zhang2021tantigen,
  title={TANTIGEN 2.0: a knowledge base of tumor T cell antigens and epitopes},
  author={Zhang, Guanglan and Chitkushev, Lou and Olsen, Lars R{\o}nn and Keskin, Derin B and Brusic, Vladimir},
  journal={BMC bioinformatics},
  volume={22},
  number={Suppl 8},
  pages={40},
  year={2021},
  publisher={Springer}
}

@article{Suzek2015UniRefClusters,
  author       = {Suzek, B. E. and Wang, Y. and Huang, H. and McGarvey, P. B. and Wu, C. H. and UniProt Consortium},
  title        = {UniRef clusters: a comprehensive and scalable alternative for improving sequence similarity searches},
  journal      = {Bioinformatics},
  volume       = {31},
  number       = {6},
  pages        = {926--932},
  year         = {2015},
  doi          = {10.1093/bioinformatics/btu739}
}

@article{Arif2024PLMACPred,
  author    = {Muhammad Arif and Saleh Musleh and Huma Fida and Talam Alam},
  title     = {{PLMACPred prediction of anticancer peptides based on protein language model and wavelet denoising transformation}},
  journal   = {Scientific Reports},
  year      = {2024},
  volume    = {14},
  pages     = {16992},
  doi       = {10.1038/s41598-024-67433-8},
  url       = {https://doi.org/10.1038/s41598-024-67433-8}
}

@article{kardani2021cppsite,
  title={Cppsite 2.0: an available database of experimentally validated cell-penetrating peptides predicting their secondary and tertiary structures},
  author={Kardani, Kimia and Bolhassani, Azam},
  journal={Journal of molecular biology},
  volume={433},
  number={11},
  pages={166703},
  year={2021},
  publisher={Elsevier}
}

@article{Zheng2025TherapeuticPeptides,
  author    = {Bingyi Zheng and Xueting Wang and Meizhai Guo and Chi-Meng Tzeng},
  title     = {Therapeutic Peptides: Recent Advances in Discovery, Synthesis, and Clinical Translation},
  journal   = {International Journal of Molecular Sciences},
  year      = {2025},
  volume    = {26},
  number    = {11},
  pages     = {5131},
  doi       = {10.3390/ijms26115131},
  url       = {https://doi.org/10.3390/ijms26115131}
}

@article{roy2019biodadpep,
  title={BioDADPep: A Bioinformatics database for anti diabetic peptides},
  author={Roy, Susanta and Teron, Robindra},
  journal={Bioinformation},
  volume={15},
  number={11},
  pages={780},
  year={2019}
}

@article{tang2025peptune,
  title={Peptune: De novo generation of therapeutic peptides with multi-objective-guided discrete diffusion},
  author={Tang, Sophia and Zhang, Yinuo and Chatterjee, Pranam},
  journal={ArXiv},
  pages={arXiv--2412},
  year={2025}
}

@article{gautam2012cppsite,
  title={CPPsite: a curated database of cell penetrating peptides},
  author={Gautam, Ankur and Singh, Harinder and Tyagi, Atul and Chaudhary, Kumardeep and Kumar, Rahul and Kapoor, Pallavi and Raghava, GPS},
  journal={Database},
  volume={2012},
  pages={bas015},
  year={2012},
  publisher={Oxford University Press}
}

@article {PepBert,
	author = {Du, Zhenjiao and Li, Yonghui},
	title = {PepBERT: Lightweight language models for peptide representation},
	elocation-id = {2025.04.08.647838},
	year = {2025},
	doi = {10.1101/2025.04.08.647838},
	publisher = {Cold Spring Harbor Laboratory},
	URL = {https://www.biorxiv.org/content/early/2025/04/14/2025.04.08.647838},
	eprint = {https://www.biorxiv.org/content/early/2025/04/14/2025.04.08.647838.full.pdf},
	journal = {bioRxiv}
}

@article{Wang2025AMPDesigner,
  title = {Discovery of antimicrobial peptides with notable antibacterial potency by an LLM-based foundation model},
  author = {Wang, Jike and Feng, Jianwen and Kang, Yu and Pan, Peichen and Ge, Jingxuan and Wang, Yan and Wang, Mingyang and Wu, Zhenxing and Zhang, Xingcai and Yu, Jiameng and Zhang, Xujun and Wang, Tianyue and Wen, Lirong and Yan, Guangning and Deng, Yafeng and Shi, Hui and Hsieh, Chang-Yu and Jiang, Zhihui and Hou, Tingjun},
  journal = {Science Advances},
  year = {2025},
  volume = {11},
  number = {10},
  pages = {eads8932},
  doi = {10.1126/sciadv.ads8932},
  note = {Epub 2025 Mar 5}
}

@article{Agrawal2021AntiCP,
  title = {AntiCP 2.0: an updated model for predicting anticancer peptides},
  author = {Piyush Agrawal and Dhruv Bhagat and Manish Mahalwal and Neelam Sharma and Gajendra P. S. Raghava},
  journal = {Briefings in Bioinformatics},
  year = {2021},
  volume = {22},
  number = {3},
  pages = {bbaa153},
  doi = {10.1093/bib/bbaa153},
  url = {https://pubmed.ncbi.nlm.nih.gov/32770192/}
}

@article{bhat2025novo,
  title={De novo design of peptide binders to conformationally diverse targets with contrastive language modeling},
  author={Bhat, Suhaas and Palepu, Kalyan and Hong, Lauren and Mao, Joey and Ye, Tianzheng and Iyer, Rema and Zhao, Lin and Chen, Tianlai and Vincoff, Sophia and Watson, Rio and others},
  journal={Science Advances},
  volume={11},
  number={4},
  pages={eadr8638},
  year={2025},
  publisher={American Association for the Advancement of Science}
}

@article{bin2020prediction,
  title={Prediction of neuropeptides from sequence information using ensemble classifier and hybrid features},
  author={Bin, Yannan and Zhang, Wei and Tang, Wending and Dai, Ruyu and Li, Menglu and Zhu, Qizhi and Xia, Junfeng},
  journal={Journal of proteome research},
  volume={19},
  number={9},
  pages={3732--3740},
  year={2020},
  publisher={ACS Publications}
}

@article{abdin2022pepnn,
  title={PepNN: a deep attention model for the identification of peptide binding sites},
  author={Abdin, Osama and Nim, Satra and Wen, Han and Kim, Philip M},
  journal={Communications biology},
  volume={5},
  number={1},
  pages={503},
  year={2022},
  publisher={Nature Publishing Group UK London}
}

@article{cabas2024peptipedia,
  title={Peptipedia v2. 0: A peptide sequence database and user-friendly web platform. A major update},
  author={Cabas-Mora, Gabriel and Daza, Anamar{\'\i}a and Soto-Garc{\'\i}a, Nicole and Garrido, Valentina and Alvarez, Diego and Navarrete, Marcelo and Sarmiento-Var{\'o}n, Lindybeth and Sep{\'u}lveda Ya{\~n}ez, Julieta H and Davari, Mehdi D and Cadet, Frederic and others},
  journal={Database},
  volume={2024},
  pages={baae113},
  year={2024},
  publisher={Oxford University Press UK}
}

@article{steinegger2017mmseqs2,
  title={MMseqs2 enables sensitive protein sequence searching for the analysis of massive data sets},
  author={Steinegger, Martin and S{\"o}ding, Johannes},
  journal={Nature biotechnology},
  volume={35},
  number={11},
  pages={1026--1028},
  year={2017},
  publisher={Nature Publishing Group US New York}
}

@article{charoenkwan2020idppiv,
  title={iDPPIV-SCM: a sequence-based predictor for identifying and analyzing dipeptidyl peptidase IV (DPP-IV) inhibitory peptides using a scoring card method},
  author={Charoenkwan, Phasit and Kanthawong, Sakawrat and Nantasenamat, Chanin and Hasan, Md Mehedi and Shoombuatong, Watshara},
  journal={Journal of proteome research},
  volume={19},
  number={10},
  pages={4125--4136},
  year={2020},
  publisher={ACS Publications}
}

@article{wynendaele2013quorumpeps,
  title={Quorumpeps database: chemical space, microbial origin and functionality of quorum sensing peptides},
  author={Wynendaele, Evelien and Bronselaer, Antoon and Nielandt, Joachim and D’Hondt, Matthias and Stalmans, Sofie and Bracke, Nathalie and Verbeke, Frederick and Van De Wiele, Christophe and De Tr{\'e}, Guy and De Spiegeleer, Bart},
  journal={Nucleic acids research},
  volume={41},
  number={D1},
  pages={D655--D659},
  year={2013},
  publisher={Oxford University Press}
}

@article{wang2024neuropep,
  title={NeuroPep 2.0: an updated database dedicated to neuropeptide and its receptor annotations},
  author={Wang, Mingxia and Wang, Lei and Xu, Wei and Chu, Ziqiang and Wang, Hengzhi and Lu, Jingxiang and Xue, Zhidong and Wang, Yan},
  journal={Journal of Molecular Biology},
  volume={436},
  number={4},
  pages={168416},
  year={2024},
  publisher={Elsevier}
}

@article{fernandez2024autopeptideml,
  title={AutoPeptideML: a study on how to build more trustworthy peptide bioactivity predictors},
  author={Fern{\'a}ndez-D{\'\i}az, Ra{\'u}l and Cossio-P{\'e}rez, Rodrigo and Agoni, Clement and Lam, Hoang Thanh and Lopez, Vanessa and Shields, Denis C},
  journal={Bioinformatics},
  volume={40},
  number={9},
  pages={btae555},
  year={2024},
  publisher={Oxford University Press}
}

@misc{fernandez2025generalize,
  title = {How to Generalize Machine Learning Models to Both Canonical and Non-Canonical Peptides},
  author = {{Fern{\'a}ndez-D{\'i}az}, Ra{\'u}l and Ochoa, Rodrigo and Hoang, Thanh Lam and Lopez, Vanessa and Shields, Denis},
  year = {2025},
  month = mar,
  publisher = {ChemRxiv},
  doi = {10.26434/chemrxiv-2025-ggp8n},
  urldate = {2025-05-16},
  archiveprefix = {ChemRxiv},
  langid = {english}
}

@article{pirtskhalava2013transmembrane,
  title={Transmembrane and antimicrobial peptides. Hydrophobicity, amphiphilicity and propensity to aggregation},
  author={Pirtskhalava, M and Vishnepolsky, B and Grigolava, M},
  journal={arXiv preprint arXiv:1307.6160},
  year={2013}
}

@article{leo2016biopeptides,
  title={Biopeptides with antioxidant and anti-inflammatory potential in the prevention and treatment of diabesity disease},
  author={Leo, Edwin Enrique Mart{\'\i}nez and Fern{\'a}ndez, Juan Jos{\'e} Acevedo and Campos, Maira Rubi Segura},
  journal={Biomedicine \& Pharmacotherapy},
  volume={83},
  pages={816--826},
  year={2016},
  publisher={Elsevier}
}

@article{qiu2025amplyze,
  title={AmpLyze: A Deep Learning Model for Predicting the Hemolytic Concentration},
  author={Qiu, Peng and Feng, Hanqi and Zhang, Meng-Chun and Poczos, Barnabas},
  journal={arXiv preprint arXiv:2507.08162},
  year={2025}
}

@article{roudi2017antimicrobial,
  title={Antimicrobial peptides as biologic and immunotherapeutic agents against cancer: a comprehensive overview},
  author={Roudi, Raheleh and Syn, Nicholas L and Roudbary, Maryam},
  journal={Frontiers in immunology},
  volume={8},
  pages={1320},
  year={2017},
  publisher={Frontiers Media SA}
}

@article{guntuboina2023peptidebert,
  title={Peptidebert: A language model based on transformers for peptide property prediction},
  author={Guntuboina, Chakradhar and Das, Adrita and Mollaei, Parisa and Kim, Seongwon and Barati Farimani, Amir},
  journal={The Journal of Physical Chemistry Letters},
  volume={14},
  number={46},
  pages={10427--10434},
  year={2023},
  publisher={ACS Publications}
}

@article{yue2024discovery,
  title={Discovery of potential antidiabetic peptides using deep learning},
  author={Yue, Jianda and Xu, Jiawei and Li, Tingting and Li, Yaqi and Chen, Zihui and Liang, Songping and Liu, Zhonghua and Wang, Ying},
  journal={Computers in Biology and Medicine},
  volume={180},
  pages={109013},
  year={2024},
  publisher={Elsevier}
}

@article{charoenkwan2020ittca,
  title={iTTCA-Hybrid: Improved and robust identification of tumor T cell antigens by utilizing hybrid feature representation},
  author={Charoenkwan, Phasit and Nantasenamat, Chanin and Hasan, Md Mehedi and Shoombuatong, Watshara},
  journal={Analytical biochemistry},
  volume={599},
  pages={113747},
  year={2020},
  publisher={Elsevier}
}

@article{white2012decoding,
  title={Decoding nonspecific interactions from nature},
  author={White, Andrew D and Nowinski, Ann K and Huang, Wenjun and Keefe, Andrew J and Sun, Fang and Jiang, Shaoyi},
  journal={Chemical Science},
  volume={3},
  number={12},
  pages={3488--3494},
  year={2012},
  publisher={Royal Society of Chemistry}
}

@article{li2023cycpeptmpdb,
  title={CycPeptMPDB: a comprehensive database of membrane permeability of cyclic peptides},
  author={Li, Jianan and Yanagisawa, Keisuke and Sugita, Masatake and Fujie, Takuya and Ohue, Masahito and Akiyama, Yutaka},
  journal={Journal of Chemical Information and Modeling},
  volume={63},
  number={7},
  pages={2240--2250},
  year={2023},
  publisher={ACS Publications}
}

@article{yao2025dbamp,
  title={dbAMP 3.0: updated resource of antimicrobial activity and structural annotation of peptides in the post-pandemic era},
  author={Yao, Lantian and Guan, Jiahui and Xie, Peilin and Chung, Chia-Ru and Zhao, Zhihao and Dong, Danhong and Guo, Yilin and Zhang, Wenyang and Deng, Junyang and Pang, Yuxuan and others},
  journal={Nucleic acids research},
  volume={53},
  number={D1},
  pages={D364--D376},
  year={2025},
  publisher={Oxford University Press}
}

@article{iwaniak2024biopep,
  title={BIOPEP-UWM database—Present and future},
  author={Iwaniak, Anna and Minkiewicz, Piotr and Darewicz, Ma{\l}gorzata},
  journal={Current Opinion in Food Science},
  volume={55},
  pages={101108},
  year={2024},
  publisher={Elsevier}
}

@article{olsen2020anoxpepred,
  title={AnOxPePred: using deep learning for the prediction of antioxidative properties of peptides},
  author={Olsen, Tobias Hegelund and Yesiltas, Bet{\"u}l and Marin, Frederikke Isa and Pertseva, Margarita and Garc{\'\i}a-Moreno, Pedro J and Gregersen, Simon and Overgaard, Michael Toft and Jacobsen, Charlotte and Lund, Ole and Hansen, Egon Bech and others},
  journal={Scientific reports},
  volume={10},
  number={1},
  pages={21471},
  year={2020},
  publisher={Nature Publishing Group UK London}
}

@article{wang2025artificial,
  title={Artificial intelligence using a latent diffusion model enables the generation of diverse and potent antimicrobial peptides},
  author={Wang, Yeji and Song, Minghui and Liu, Fujing and Liang, Zhen and Hong, Rui and Dong, Yuemei and Luan, Huaizu and Fu, Xiaojie and Yuan, Wenchang and Fang, Wenjie and others},
  journal={Science Advances},
  volume={11},
  number={6},
  pages={eadp7171},
  year={2025},
  publisher={American Association for the Advancement of Science}
}

@article{pineda2018arachnoserver,
  title={ArachnoServer 3.0: an online resource for automated discovery, analysis and annotation of spider toxins},
  author={Pineda, Sandy S and Chaumeil, Pierre-Alain and Kunert, Anne and Kaas, Quentin and Thang, Mike WC and Le, Lien and Nuhn, Michael and Herzig, Volker and Saez, Natalie J and Cristofori-Armstrong, Ben and others},
  journal={Bioinformatics},
  volume={34},
  number={6},
  pages={1074--1076},
  year={2018},
  publisher={Oxford University Press}
}

@article{kaas2008conoserver,
  title={ConoServer, a database for conopeptide sequences and structures},
  author={Kaas, Quentin and Westermann, Jan-C and Halai, Reena and Wang, Conan KL and Craik, David J},
  journal={Bioinformatics},
  volume={24},
  number={3},
  pages={445--446},
  year={2008},
  publisher={Oxford University Press}
}

@article{wang2025integrating,
  title={Integrating Protein Language Models and Geometric Deep Learning for Peptide Toxicity Prediction},
  author={Wang, Yanling and Li, Na and Wang, Xiao and Cao, Feng and Xiong, Shuwen and Wei, Leyi},
  journal={Journal of Chemical Information and Modeling},
  volume={65},
  number={14},
  pages={7800--7810},
  year={2025},
  publisher={ACS Publications}
}

@article{ma2025dramp,
  title={DRAMP 4.0: an open-access data repository dedicated to the clinical translation of antimicrobial peptides},
  author={Ma, Tianyue and Liu, Yanchao and Yu, Bingxin and Sun, Xin and Yao, Huiyuan and Hao, Chen and Li, Jianhui and Nawaz, Maryam and Jiang, Xun and Lao, Xingzhen and others},
  journal={Nucleic Acids Research},
  volume={53},
  number={D1},
  pages={D403--D410},
  year={2025},
  publisher={Oxford University Press}
}

@article{manavalan2019mahtpred,
  title={mAHTPred: a sequence-based meta-predictor for improving the prediction of anti-hypertensive peptides using effective feature representation},
  author={Manavalan, Balachandran and Basith, Shaherin and Shin, Tae Hwan and Wei, Leyi and Lee, Gwang},
  journal={Bioinformatics},
  volume={35},
  number={16},
  pages={2757--2765},
  year={2019},
  publisher={Oxford University Press}
}

@article{kumar2015silico,
  title={An in silico platform for predicting, screening and designing of antihypertensive peptides},
  author={Kumar, Ravi and Chaudhary, Kumardeep and Singh Chauhan, Jagat and Nagpal, Gandharva and Kumar, Rahul and Sharma, Minakshi and Raghava, Gajendra PS},
  journal={Scientific reports},
  volume={5},
  number={1},
  pages={12512},
  year={2015},
  publisher={Nature Publishing Group UK London}
}

@article{rajput2015prediction,
  title={Prediction and analysis of quorum sensing peptides based on sequence features},
  author={Rajput, Akanksha and Gupta, Amit Kumar and Kumar, Manoj},
  journal={PLoS One},
  volume={10},
  number={3},
  pages={e0120066},
  year={2015},
  publisher={Public Library of Science San Francisco, CA USA}
}

@article{wei2021atse,
  title={ATSE: a peptide toxicity predictor by exploiting structural and evolutionary information based on graph neural network and attention mechanism},
  author={Wei, Lesong and Ye, Xiucai and Xue, Yuyang and Sakurai, Tetsuya and Wei, Leyi},
  journal={Briefings in bioinformatics},
  volume={22},
  number={5},
  pages={bbab041},
  year={2021},
  publisher={Oxford University Press}
}

@article{singh2025hemolytik2,
  title={Hemolytik2: An Updated Database of Hemolytic Peptides and Proteins},
  author={Singh, Ayushi and SA, Kavin Raj and Rathore, Anand Singh and Raghava, Gajendra PS},
  journal={bioRxiv},
  pages={2025--05},
  year={2025},
  publisher={Cold Spring Harbor Laboratory}
}

@article{witten2019deep,
  title={Deep learning regression model for antimicrobial peptide design},
  author={Witten, Jacob and Witten, Zack},
  journal={BioRxiv},
  pages={692681},
  year={2019},
  publisher={Cold Spring Harbor Laboratory}
}

@article{hikida2013systematic,
  title={Systematic analysis of a dipeptide library for inhibitor development using human dipeptidyl peptidase IV produced by a Saccharomyces cerevisiae expression system},
  author={Hikida, Aya and Ito, Keisuke and Motoyama, Takayasu and Kato, Ryuji and Kawarasaki, Yasuaki},
  journal={Biochemical and biophysical research communications},
  volume={430},
  number={4},
  pages={1217--1222},
  year={2013},
  publisher={Elsevier}
}

@article{pinacho2021alignment,
  title={Alignment-free antimicrobial peptide predictors: improving performance by a thorough analysis of the largest available data set},
  author={Pinacho-Castellanos, Sergio A and Garc{\'\i}a-Jacas, C{\'e}sar R and Gilson, Michael K and Brizuela, Carlos A},
  journal={Journal of Chemical Information and Modeling},
  volume={61},
  number={6},
  pages={3141--3157},
  year={2021},
  publisher={ACS Publications}
}

@article{aguilera2023starpep,
  title={StarPep Toolbox: an open-source software to assist chemical space analysis of bioactive peptides and their functions using complex networks},
  author={Aguilera-Mendoza, Longendri and Ayala-Ruano, Sebasti{\'a}n and Martinez-Rios, Felix and Chavez, Edgar and Garc{\'\i}a-Jacas, C{\'e}sar R and Brizuela, Carlos A and Marrero-Ponce, Yovani},
  journal={Bioinformatics},
  volume={39},
  number={8},
  pages={btad506},
  year={2023},
  publisher={Oxford University Press}
}

@article{zhang2025pepland,
  title={PepLand: a large-scale pre-trained peptide representation model for a comprehensive landscape of both canonical and non-canonical amino acids},
  author={Zhang, Ruochi and Wu, Haoran and Liu, Chang and Yang, Qian and Xiu, Yuting and Li, Kewei and Chen, Ningning and Wang, Yu and Wang, Yan and Gao, Xin and others},
  journal={Briefings in Bioinformatics},
  volume={26},
  number={4},
  pages={bbaf367},
  year={2025},
  publisher={Oxford University Press}
}

@article{lei2021deep,
  title={A deep-learning framework for multi-level peptide--protein interaction prediction},
  author={Lei, Yipin and Li, Shuya and Liu, Ziyi and Wan, Fangping and Tian, Tingzhong and Li, Shao and Zhao, Dan and Zeng, Jianyang},
  journal={Nature communications},
  volume={12},
  number={1},
  pages={5465},
  year={2021},
  publisher={Nature Publishing Group UK London}
}

@article{dong2025exploring,
  title={Exploring the repository of de novo-designed bifunctional antimicrobial peptides through deep learning},
  author={Dong, Ruihan and Liu, Rongrong and Liu, Ziyu and Liu, Yangang and Zhao, Gaomei and Li, Honglei and Hou, Shiyuan and Ma, Xiaohan and Kang, Huarui and Liu, Jing and others},
  journal={eLife},
  volume={13},
  pages={RP97330},
  year={2025},
  publisher={eLife Sciences Publications Limited}
}

@article{lawrence2021ampeppy,
  title={amPEPpy 1.0: a portable and accurate antimicrobial peptide prediction tool},
  author={Lawrence, Travis J and Carper, Dana L and Spangler, Margaret K and Carrell, Alyssa A and Rush, Tom{\'a}s A and Minter, Stephen J and Weston, David J and Labb{\'e}, Jessy L},
  journal={Bioinformatics},
  volume={37},
  number={14},
  pages={2058--2060},
  year={2021},
  publisher={Oxford University Press}
}

@article{huang2022artificial,
  title={Artificial intelligence foundation for therapeutic science},
  author={Huang, Kexin and Fu, Tianfan and Gao, Wenhao and Zhao, Yue and Roohani, Yusuf and Leskovec, Jure and Coley, Connor W and Xiao, Cao and Sun, Jimeng and Zitnik, Marinka},
  journal={Nature chemical biology},
  volume={18},
  number={10},
  pages={1033--1036},
  year={2022},
  publisher={Nature Publishing Group US New York}
}

@article{wang2024pepdora,
  title={Pepdora: A unified peptide language model via weight-decomposed low-rank adaptation},
  author={Wang, Leyao and Pulugurta, Rishab and Vure, Pranay and Zhang, Yinuo and Pal, Aastha and Chatterjee, Pranam},
  journal={arXiv preprint arXiv:2410.20667},
  year={2024}
}

@article{notin2023proteingym,
  title={Proteingym: Large-scale benchmarks for protein fitness prediction and design},
  author={Notin, Pascal and Kollasch, Aaron and Ritter, Daniel and Van Niekerk, Lood and Paul, Steffanie and Spinner, Han and Rollins, Nathan and Shaw, Ada and Orenbuch, Rose and Weitzman, Ruben and others},
  journal={Advances in Neural Information Processing Systems},
  volume={36},
  pages={64331--64379},
  year={2023}
}

@article{rathore2025prediction,
  title={Prediction of hemolytic peptides and their hemolytic concentration},
  author={Rathore, Anand Singh and Kumar, Nishant and Choudhury, Shubham and Mehta, Naman Kumar and Raghava, Gajendra PS},
  journal={Communications Biology},
  volume={8},
  number={1},
  pages={176},
  year={2025},
  publisher={Nature Publishing Group UK London}
}

@article{tanaka2019brain,
  title={Brain-transportable dipeptides across the blood-brain barrier in mice},
  author={Tanaka, Mitsuru and Dohgu, Shinya and Komabayashi, Genki and Kiyohara, Hayato and Takata, Fuyuko and Kataoka, Yasufumi and Nirasawa, Takashi and Maebuchi, Motohiro and Matsui, Toshiro},
  journal={Scientific reports},
  volume={9},
  number={1},
  pages={5769},
  year={2019},
  publisher={Nature Publishing Group UK London}
}

@article{shtatland2007pepbank,
  title={PepBank-a database of peptides based on sequence text mining and public peptide data sources},
  author={Shtatland, Timur and Guettler, Daniel and Kossodo, Misha and Pivovarov, Misha and Weissleder, Ralph},
  journal={BMC bioinformatics},
  volume={8},
  number={1},
  pages={280},
  year={2007},
  publisher={Springer}
}

@article{singh2016satpdb,
  title={SATPdb: a database of structurally annotated therapeutic peptides},
  author={Singh, Sandeep and Chaudhary, Kumardeep and Dhanda, Sandeep Kumar and Bhalla, Sherry and Usmani, Salman Sadullah and Gautam, Ankur and Tuknait, Abhishek and Agrawal, Piyush and Mathur, Deepika and Raghava, Gajendra PS},
  journal={Nucleic acids research},
  volume={44},
  number={D1},
  pages={D1119--D1126},
  year={2016},
  publisher={Oxford University Press}
}

@article{van2012brainpeps,
  title={Brainpeps: the blood--brain barrier peptide database},
  author={Van Dorpe, Sylvia and Bronselaer, Antoon and Nielandt, Joachim and Stalmans, Sofie and Wynendaele, Evelien and Audenaert, Kurt and Van De Wiele, Christophe and Burvenich, Christian and Peremans, Kathelijne and Hsuchou, Hung and others},
  journal={Brain Structure and Function},
  volume={217},
  number={3},
  pages={687--718},
  year={2012},
  publisher={Springer}
}

@article{dai2021bbppred,
  title={BBPpred: sequence-based prediction of blood-brain barrier peptides with feature representation learning and logistic regression},
  author={Dai, Ruyu and Zhang, Wei and Tang, Wending and Wynendaele, Evelien and Zhu, Qizhi and Bin, Yannan and De Spiegeleer, Bart and Xia, Junfeng},
  journal={Journal of Chemical Information and Modeling},
  volume={61},
  number={1},
  pages={525--534},
  year={2021},
  publisher={ACS Publications}
}

@inproceedings{kipf2016semiGCN,
  title        = {Semi-Supervised Classification with Graph Convolutional Networks},
  author       = {Kipf, Thomas N. and Welling, Max},
  booktitle    = {5th International Conference on Learning Representations (ICLR)},
  year         = {2017},
  note         = {arXiv preprint arXiv:1609.02907},
  url          = {https://arxiv.org/abs/1609.02907},
}

@inproceedings{velickovic2018graphGAT,
  title        = {Graph Attention Networks},
  author       = {Veličković, Petar and Cucurull, Guillem and Casanova, Arantxa and Romero, Adriana and Liò, Pietro and Bengio, Yoshua},
  booktitle    = {International Conference on Learning Representations (ICLR)},
  year         = {2018},
  note         = {arXiv preprint arXiv:1710.10903},
  url          = {https://arxiv.org/abs/1710.10903},
}

@article {ElnaggarProtBert,
    author = {Elnaggar, Ahmed and Heinzinger, Michael and Dallago, Christian and Rehawi, Ghalia and Wang, Yu and Jones, Llion and Gibbs, Tom and Feher, Tamas and Angerer, Christoph and Steinegger, Martin and BHOWMIK, DEBSINDHU and Rost, Burkhard},
    title = {ProtTrans: Towards Cracking the Language of Life{\textquoteright}s Code Through Self-Supervised Deep Learning and High Performance Computing},
    elocation-id = {2020.07.12.199554},
    year = {2020},
    doi = {10.1101/2020.07.12.199554},
    publisher = {Cold Spring Harbor Laboratory},
    journal = {bioRxiv}
}

@inproceedings{wang2024dplm,
  title     = {Diffusion Language Models Are Versatile Protein Learners},
  author    = {Wang, Xinyou and Zheng, Zaixiang and Ye, Fei and Xue, Dongyu and Huang, Shujian and Gu, Quanquan},
  booktitle = {Proceedings of the 41st International Conference on Machine Learning (ICML)},
  year      = {2024},
}

@inproceedings{xu2019howGIN,
  title        = {How Powerful are Graph Neural Networks?},
  author       = {Xu, Keyulu and Hu, Weihua and Leskovec, Jure and Jegelka, Stefanie},
  booktitle    = {International Conference on Learning Representations (ICLR)},
  year         = {2019},
  note         = {arXiv preprint arXiv:1810.00826},
  url          = {https://arxiv.org/abs/1810.00826},
}

@article{feller2025peptide,
  title={Peptide-aware chemical language model successfully predicts membrane diffusion of cyclic peptides},
  author={Feller, Aaron L and Wilke, Claus O},
  journal={Journal of Chemical Information and Modeling},
  volume={65},
  number={2},
  pages={571--579},
  year={2025},
  publisher={ACS Publications}
}

@article{zhang2022predapp,
  title={PredAPP: predicting anti-parasitic peptides with undersampling and ensemble approaches},
  author={Zhang, Wei and Xia, Enhua and Dai, Ruyu and Tang, Wending and Bin, Yannan and Xia, Junfeng},
  journal={Interdisciplinary Sciences: Computational Life Sciences},
  volume={14},
  number={1},
  pages={258--268},
  year={2022},
  publisher={Springer}
}

@article{wei2018acpred,
  title={ACPred-FL: a sequence-based predictor using effective feature representation to improve the prediction of anti-cancer peptides},
  author={Wei, Leyi and Zhou, Chen and Chen, Huangrong and Song, Jiangning and Su, Ran},
  journal={Bioinformatics},
  volume={34},
  number={23},
  pages={4007--4016},
  year={2018},
  publisher={Oxford University Press}
}

@article{aziz2022iacp,
  title={iACP-MultiCNN: Multi-channel CNN based anticancer peptides identification},
  author={Aziz, Abu Zahid Bin and Hasan, Md Al Mehedi and Ahmad, Shamim and Al Mamun, Md and Shin, Jungpil and Hossain, Md Rahat},
  journal={Analytical Biochemistry},
  volume={650},
  pages={114707},
  year={2022},
  publisher={Elsevier}
}

@article{vijayakumar2015acpp,
  title={ACPP: a web server for prediction and design of anti-cancer peptides},
  author={Vijayakumar, Saravanan and Ptv, Lakshmi},
  journal={International Journal of Peptide Research and Therapeutics},
  volume={21},
  number={1},
  pages={99--106},
  year={2015},
  publisher={Springer}
}

@article{yi2019acp,
  title={ACP-DL: a deep learning long short-term memory model to predict anticancer peptides using high-efficiency feature representation},
  author={Yi, Hai-Cheng and You, Zhu-Hong and Zhou, Xi and Cheng, Li and Li, Xiao and Jiang, Tong-Hai and Chen, Zhan-Heng},
  journal={Molecular Therapy Nucleic acids},
  volume={17},
  pages={1--9},
  year={2019},
  publisher={Elsevier}
}

@article{tyagi2015cancerppd,
  title={CancerPPD: a database of anticancer peptides and proteins},
  author={Tyagi, Atul and Tuknait, Abhishek and Anand, Priya and Gupta, Sudheer and Sharma, Minakshi and Mathur, Deepika and Joshi, Anshika and Singh, Sandeep and Gautam, Ankur and Raghava, Gajendra PS},
  journal={Nucleic acids research},
  volume={43},
  number={D1},
  pages={D837--D843},
  year={2015},
  publisher={Oxford University Press}
}

@article{kumar2015ahtpdb,
  title={AHTPDB: a comprehensive platform for analysis and presentation of antihypertensive peptides},
  author={Kumar, Ravi and Chaudhary, Kumardeep and Sharma, Minakshi and Nagpal, Gandharva and Chauhan, Jagat Singh and Singh, Sandeep and Gautam, Ankur and Raghava, Gajendra PS},
  journal={Nucleic acids research},
  volume={43},
  number={D1},
  pages={D956--D962},
  year={2015},
  publisher={Oxford University Press}
}

@article{wu2020comprehensive,
  title={A comprehensive survey on graph neural networks},
  author={Wu, Zonghan and Pan, Shirui and Chen, Fengwen and Long, Guodong and Zhang, Chengqi and Yu, Philip S},
  journal={IEEE transactions on neural networks and learning systems},
  volume={32},
  number={1},
  pages={4--24},
  year={2020},
  publisher={IEEE}
}

@article{minkiewicz2008biopep,
  title={BIOPEP database and other programs for processing bioactive peptide sequences},
  author={Minkiewicz, Piotr and Dziuba, Jerzy and Iwaniak, Anna and Dziuba, Marta and Darewicz, Magorzata},
  journal={Journal of AOAC International},
  volume={91},
  number={4},
  pages={965--980},
  year={2008},
  publisher={Oxford University Press}
}

@article{huang2023identification,
  title={Identification of potent antimicrobial peptides via a machine-learning pipeline that mines the entire space of peptide sequences},
  author={Huang, Junjie and Xu, Yanchao and Xue, Yunfan and Huang, Yue and Li, Xu and Chen, Xiaohui and Xu, Yao and Zhang, Dongxiang and Zhang, Peng and Zhao, Junbo and others},
  journal={Nature Biomedical Engineering},
  volume={7},
  number={6},
  pages={797--810},
  year={2023},
  publisher={Nature Publishing Group UK London}
}

@article{Mehta2014ParaPepAW,
  title={ParaPep: a web resource for experimentally validated antiparasitic peptide sequences and their structures},
  author={Divya Mehta and Priya Anand and Vineet Kumar and Anshika Joshi and Deepika Mathur and Sandeep Singh and Abhishek Tuknait and Kumardeep Chaudhary and Shailendra K. Gautam and Ankur Gautam and Grish C. Varshney and Gajendra P. Raghava},
  journal={Database: The Journal of Biological Databases and Curation},
  year={2014},
  volume={2014}
}

@article{Wang2015APD3TA,
  title={APD3: the antimicrobial peptide database as a tool for research and education},
  author={Guangshun Wang and Xia Li and Zhe Wang},
  journal={Nucleic Acids Research},
  year={2015},
  volume={44},
  pages={D1087 - D1093}
}

@article{Jhong2018dbAMPAI,
  title={dbAMP: an integrated resource for exploring antimicrobial peptides with functional activities and physicochemical properties on transcriptome and proteome data},
  author={Jhih-Hua Jhong and Yu-Hsiang Chi and Wen-Chi Li and Tsai-Hsuan Lin and Kai-Yao Huang and Tzong-Yi Lee},
  journal={Nucleic Acids Research},
  year={2018},
  volume={47},
  pages={D285 - D297}
}

@article{Thomas2009CAMPAU,
  title={CAMP: a useful resource for research on antimicrobial peptides},
  author={Shaini Thomas and Shreyas D. Karnik and Ram Shankar Barai and Vaidyanathan K. Jayaraman and Susan Idicula-Thomas},
  journal={Nucleic Acids Research},
  year={2009},
  volume={38},
  pages={D774 - D780}
}

@article{Fan2016DRAMPAC,
  title={DRAMP: a comprehensive data repository of antimicrobial peptides},
  author={Linlin Fan and Jian Sun and Meifeng Zhou and Jie Zhou and Xingzhen Lao and Heng Zheng and Hanmei Xu},
  journal={Scientific Reports},
  year={2016},
  volume={6}
}

@article{Nothaft2015RethinkingDS,
  title={Rethinking Data-Intensive Science Using Scalable Analytics Systems},
  author={Frank A. Nothaft and Matt Massie and Timothy Danford and Zhao Zhang and Uri Laserson and Carl Yeksigian and Jey Kottalam and Arun Ahuja and Jeff Hammerbacher and Michael D. Linderman and Michael J. Franklin and Anthony D. Joseph and David A. Patterson},
  journal={Proceedings of the 2015 ACM SIGMOD International Conference on Management of Data},
  year={2015}
}

@article{wu2018moleculenet,
  title={MoleculeNet: a benchmark for molecular machine learning},
  author={Wu, Zhenqin and Ramsundar, Bharath and Feinberg, Evan N and Gomes, Joseph and Geniesse, Caleb and Pappu, Aneesh S and Leswing, Karl and Pande, Vijay},
  journal={Chemical science},
  volume={9},
  number={2},
  pages={513--530},
  year={2018},
  publisher={Royal Society of Chemistry}
}

@article{rao2019evaluating,
  title={Evaluating protein transfer learning with TAPE},
  author={Rao, Roshan and Bhattacharya, Nicholas and Thomas, Neil and Duan, Yan and Chen, Peter and Canny, John and Abbeel, Pieter and Song, Yun},
  journal={Advances in neural information processing systems},
  volume={32},
  year={2019}
}
\bibliographystyle{iclr2025_conference}

\appendix

\section{Limitations}

A key limitation of the current work is the lack of structural datasets and the corresponding evaluation of structure-based prediction and generation tasks. At present, our benchmark focuses primarily on sequence-level data. Incorporating structural information, however, is essential for advancing realistic peptide modeling. The major challenge lies in data availability: existing peptide structure datasets are extremely scarce, with only about 2{,}000 peptide entries in the Protein Data Bank (PDB), among which fewer than 100 correspond to non-natural peptides. Consequently, structural expansion must rely heavily on computational simulation. Fortunately, compared with proteins, peptides are much smaller molecules, typically consisting of only a few hundred atoms. This size advantage makes high-accuracy simulations feasible, including advanced approaches such as quantum mechanics/molecular mechanics (QM/MM) hybrid methods, enhanced molecular dynamics (MD), or other physics-informed simulations. We are currently developing an efficient and accurate peptide structural simulation pipeline to generate reliable structural data, which will support structure-related tasks within \textit{PepBenchmark}. In future iterations, we plan to extend the benchmark to include the evaluation of peptide structure prediction and generative modeling methods.

\raggedbottom
\section{Details of PepBenchData}

\textit{PepBenchData} compiles datasets spanning 7 tasks that correspond to 3 stages of the drug discovery pipeline: activity modeling (i.e. AMP, PepPI, Oncology, Metabolic, Others), pharmacokinetic profiling (i.e. ADME), and safety assessment (i.e. Tox). The detailed descriptions of each dataset are presented in this section. The sources of our datasets are provided in Table~\ref{app:des:tab:data_sources}. Overviews of classification and regression datasets are provided in ~\ref{app:des:tab:pepbenchmark_cls} and ~\ref{app:des:tab:pepbenchmark_reg}, respectively.

\begin{table}[h]
    \centering
    \scriptsize
    \caption{Dataset sources.}
    \label{app:des:tab:data_sources}
    \renewcommand{\arraystretch}{1.0}
    \setlength{\tabcolsep}{2pt}
    \resizebox{\textwidth}{!}{%
    \begin{tabular}{lll}
    \toprule
    Application & Dataset Name & Data Source \\
    \midrule
    \multirow{4}{*}{ADME} 
    & nonfouling        & ~\cite{guntuboina2023peptidebert} \\
    & cpp               & ~\cite{zhang2025pepland}; Peptipedia~\citep{cabas2024peptipedia} \\
    & bbp               & ~\cite{dai2021bbppred}; Peptipedia~\citep{cabas2024peptipedia} \\
    & nc-cpp\_pampa     & CycPeptMPDB~\citep{li2023cycpeptmpdb} \\
    \cmidrule{1-3}
    \multirow{11}{*}{AMP} 
    & antimicrobial     & ~\cite{wang2025artificial}; Peptipedia~\citep{cabas2024peptipedia} \\
    & antibacterial     & ~\cite{pinacho2021alignment}; Peptipedia~\citep{cabas2024peptipedia} \\
    & antifungal        & ~\cite{pinacho2021alignment}; Peptipedia~\citep{cabas2024peptipedia} \\
    & antiparasitic     & ~\cite{zhang2022predapp}; Peptipedia~\citep{cabas2024peptipedia} \\
    & antiviral         & ~\cite{pinacho2021alignment}; Peptipedia~\citep{cabas2024peptipedia} \\
    & nc-antimicrobial  & Hemolytik 2.0~\citep{singh2025hemolytik2} \\
    & nc-antibacterial  & Hemolytik 2.0~\citep{singh2025hemolytik2} \\
    & nc-antifungal     & Hemolytik 2.0~\citep{singh2025hemolytik2} \\
    & E.coli\_mic       & Grampa~\citep{witten2019deep} \\
    & S.aureus\_mic     & Grampa~\citep{witten2019deep} \\
    & P.aeruginosa\_mic & Grampa~\citep{witten2019deep} \\
    \cmidrule{1-3}
    \multirow{4}{*}{Metabolic}
    & ace\_inhibitory       & ~\cite{manavalan2019mahtpred}; Peptipedia~\citep{cabas2024peptipedia} \\
    & antidiabetic          & ~\cite{yue2024discovery}; ~\cite{hikida2013systematic}; Peptipedia~\citep{cabas2024peptipedia} \\
    & dppiv\_inhibitors     & ~\cite{charoenkwan2020idppiv}; ~\cite{hikida2013systematic} \\
    & ace\_inhibitory\_ic50 & ~\cite{kumar2015silico} \\
    \cmidrule{1-3}
    \multirow{2}{*}{Oncology}
    & anticancer            & ~\cite{Agrawal2021AntiCP}; Peptipedia~\citep{cabas2024peptipedia} \\
    & ttca                  & ~\cite{charoenkwan2020ittca} \\
    \cmidrule{1-3}
    \multirow{5}{*}{Others}
    & neuropeptide      & Peptipedia~\citep{cabas2024peptipedia}; ~\cite{bin2020prediction} \\
    & antiinflamatory   & Peptipedia~\citep{cabas2024peptipedia} \\
    & antioxidant       & ~\cite{olsen2020anoxpepred}; Peptipedia~\citep{cabas2024peptipedia} \\
    & antiaging         & Peptipedia~\citep{cabas2024peptipedia} \\
    & quorum\_sensing   & Peptipedia~\citep{cabas2024peptipedia}; ~\cite{rajput2015prediction} \\
    \cmidrule{1-3}
    \multirow{3}{*}{PepPI}
    & PpI               & ~\cite{bhat2025novo}; ~\cite{abdin2022pepnn} \\
    & PpI\_ba           & ~\cite{zhang2025pepland} \\
    & nc-PpI\_ba        & ~\cite{zhang2025pepland} \\
    \cmidrule{1-3}
    \multirow{6}{*}{Tox}
    & hemolytic         & Hemolytik 2.0~\citep{singh2025hemolytik2}; Peptipedia~\citep{cabas2024peptipedia} \\
    & toxicity          & Peptipedia~\citep{cabas2024peptipedia}; ~\cite{wei2021atse} \\
    & neurotoxin        & Peptipedia~\citep{cabas2024peptipedia} \\
    & allergen          & Peptipedia~\citep{cabas2024peptipedia} \\
    \bottomrule
    \end{tabular}
    }
\end{table}



\begin{table}[h]
\centering
\caption{Overview of classification datasets.}
\label{app:des:tab:pepbenchmark_cls}
\resizebox{\textwidth}{!}{%
\begin{tabular}{llcc rrrrrrrrr}
\toprule
\textbf{Application} & \textbf{Dataset Name} & \textbf{Task Type} & \textbf{Peptide Type} & \textbf{Origin\_pos} & \textbf{Filt\_pos} & \textbf{Exp\_neg} & \textbf{Filt\_neg} & \textbf{Total\_Size} & \textbf{min\_len} & \textbf{max\_len} & \textbf{mean\_len} & \textbf{Len $>$ 50\%} \\
\midrule
\multirow{4}{*}{ADME} & bbp & bc & ca & 358 & 336 & 13 & 13 & 672 & 2 & 82 & 14.14 & 1.04 \\
 & cpp & bc & ca & 1318 & 1162 & 0 & - & 2324 & 3 & 61 & 17.24 & 1.20 \\
 & nonfouling & bc & ca & 3600 & 3600 & 0 & - & 7200 & 5 & 11 & 6.06 & 0.00 \\
 & solubility & bc & ca & 5096 & 5096 & 5384 & 5384 & 10480 & 19 & 150 & 112.34 & 97.86 \\
\midrule
\multirow{6}{*}{AMP} & antibacterial & bc & ca & 21220 & 15838 & 0 & - & 31676 & 2 & 150 & 25.33 & 9.74 \\
 & antifungal & bc & ca & 11087 & 8349 & 0 & - & 16698 & 2 & 148 & 34.17 & 22.82 \\
 & antimicrobial & bc & ca & 42800 & 30752 & 0 & - & 61504 & 2 & 150 & 28.55 & 13.92 \\
 & antiparasitic & bc & ca & 6041 & 4316 & 0 & - & 8632 & 2 & 140 & 35.97 & 21.74 \\
 & antiviral & bc & ca & 5210 & 4134 & 0 & - & 8268 & 2 & 138 & 22.75 & 5.84 \\
 & nc-antifungal & bc & n-ca & 207 & - & 0 & - & 410 & 4 & 76 & 18.18 & 1.93 \\
 & nc-antibacterial & bc & n-ca & 845 & - & 0 & - & 1666 & 4 & 76 & 16.82 & 0.95 \\
 & nc-antimicrobial & bc & n-ca & 1269 & - & 0 & - & 2496 & 4 & 130 & 17.79 & 0.24 \\
\midrule
\multirow{3}{*}{Metabolic} & ace\_inhibitory & bc & ca & 1833 & 1780 & 0 & - & 3560 & 4 & 81 & 8.08 & 0.65 \\
 & antidiabetic & bc & ca & 1599 & 1514 & 75 & 75 & 3028 & 2 & 46 & 10.41 & 0.00 \\
 & dppiv\_inhibitors & bc & ca & 650 & 634 & 85 & 85 & 1268 & 2 & 33 & 6.15 & 0.00 \\
\midrule
\multirow{2}{*}{Oncology} & anticancer & bc & ca & 9022 & 6926 & 0 & - & 13852 & 2 & 145 & 27.83 & 13.28 \\
 & ttca & bc & ca & 592 & 591 & 0 & - & 1182 & 8 & 20 & 9.36 & 0.00 \\
\midrule
\multirow{5}{*}{Others} & antiaging & bc & ca & 282 & 279 & 0 & - & 558 & 2 & 80 & 10.93 & 1.79 \\
 & antiinflamatory & bc & ca & 3902 & 3875 & 0 & - & 7750 & 2 & 107 & 16.95 & 1.10 \\
 & antioxidant & bc & ca & 1146 & 195 & 195 & 195 & 2242 & 2 & 11 & 4.10 & 0.00 \\
 & neuropeptide & bc & ca & 5336 & 4627 & 0 & - & 9254 & 2 & 150 & 19.57 & 6.13 \\
 & quorum\_sensing & bc & ca & 265 & 245 & 0 & - & 490 & 3 & 48 & 10.83 & 0.00 \\
\midrule
PepPI & PpI & bc & ca & 7358 & - & 0 & - & 44148 & 5 & 25 & 12.8 & 0 \\
\midrule
\multirow{5}{*}{Tox} & allergen & bc & ca & 2405 & 1677 & 0 & - & 3354 & 4 & 150 & 39.97 & 24.33 \\
 & hemolytic & bc & ca & 3096 & 2256 & 544 & 475 & 4512 & 2 & 144 & 23.69 & 4.57 \\
 & nc-hemolytic & bc & n-ca & 2002 & - & 342 & - & 3937 & 4 & 130 & 17.99 & 0.5 \\
 & neurotoxin & bc & ca & 2509 & 1753 & 0 & - & 3506 & 7 & 138 & 39.24 & 9.90 \\
 & toxicity & bc & ca & 2509 & 2204 & 0 & - & 4408 & 7 & 138 & 38.41 & 7.99 \\
\bottomrule
\end{tabular}%
}
\end{table}

\begin{table}[h]
\centering
\caption{Overview of regression datasets.}
\label{app:des:tab:pepbenchmark_reg}
\resizebox{\textwidth}{!}{%
\begin{tabular}{llcl rrrrr}
\toprule
\textbf{Application} & \textbf{Dataset Name} & \textbf{Peptide Type} & \textbf{Unit} & \textbf{Total\_Size} & \textbf{min\_len} & \textbf{max\_len} & \textbf{mean\_len} & \textbf{Len $>$ 50\%} \\
\midrule
ADME & nc-cpp\_pampa & n-ca & log(cm/s) & 6970 & 2 & 15 & 8.04 & 0 \\
\midrule
\multirow{3}{*}{AMP} & E.coli\_mic & ca & lg($\mu M$) & 3312 & 2 & 140 & 23.24 & 3.23 \\
 & P.aeruginosa\_mic & ca & lg($\mu M$) & 1531 & 2 & 140 & 22.01 & 2.61 \\
 & S.aureus\_mic & ca & lg($\mu M$) & 2900 & 2 & 140 & 22.70 & 2.66 \\
\midrule
Metabolic & ace\_inhibitory\_ic50 & ca & lg($\mu M$) & 337 & 2 & 3 & 2.61 & 0.00 \\
\midrule
\multirow{2}{*}{PepPI} & nc-PpI\_ba & n-ca & -lg(M) & 277 & 5 & 19 & 8.93 & 0 \\
 & PpI\_ba & ca & -lg(M) & 1433 & 5 & 50 & 16.14 & 0 \\
\midrule
Tox & hemolytic\_hc50 & ca & lg($\mu M$) & 1926 & 6 & 39 & 18.40 & 0.00 \\
\bottomrule
\end{tabular}%
}
\end{table}


\clearpage

\subsection{ADME}

\begin{datasetbox}
\paragraph{Definition.} 
Datasets in this group primarily focus on the pharmacokinetics-related properties of peptide drugs. In our paper, we follow the convention of the drug discovery field and call this type ``ADME'', which stands for Absorption, Distribution, Metabolism and Excretion.
\paragraph{Impact.} 
ADME properties are closely related to the stability, permeability, and bioavailability of peptide drugs, and therefore determine whether a therapeutically effective concentration of the peptide can reach its target site after administration, which is critical for both efficacy and safety.
\paragraph{Pipeline.} Pharmacokinetic Profiling
\end{datasetbox}

\subsubsection{nonfouling}

\fielditem{Property and Application} 
Nonfouling characterizes resistance to nonspecific interactions between peptides and other biomolecules. Nonfouling peptides are less likely to aggregate or trigger off-target effects as a result of nonspecific binding, which is essential for ensuring the specific bioactivity of peptides in complex biological environments.

\fielditem{Data Sources} 
Nonfouling peptides are sourced from ~\cite{white2012decoding}. The dataset is constructed by manually designing peptides resistance to nonspecific interaction by mimicking the amino acid distributions observed on protein surfaces and the inner surfaces of molecular chaperones. The original dataset contains 3,600 nonfouling peptides.

\fielditem{Dataset Statistics} 
The dataset contains 7,200 datapoints with sequences ranging from 5 to 11 amino acids (average length 6.06) in length.

\textbf{Task: Classification; Split: Hybrid; Evaluation: ROC-AUC}

\begin{center} 
\centering
\includegraphics[width=0.8\textwidth]{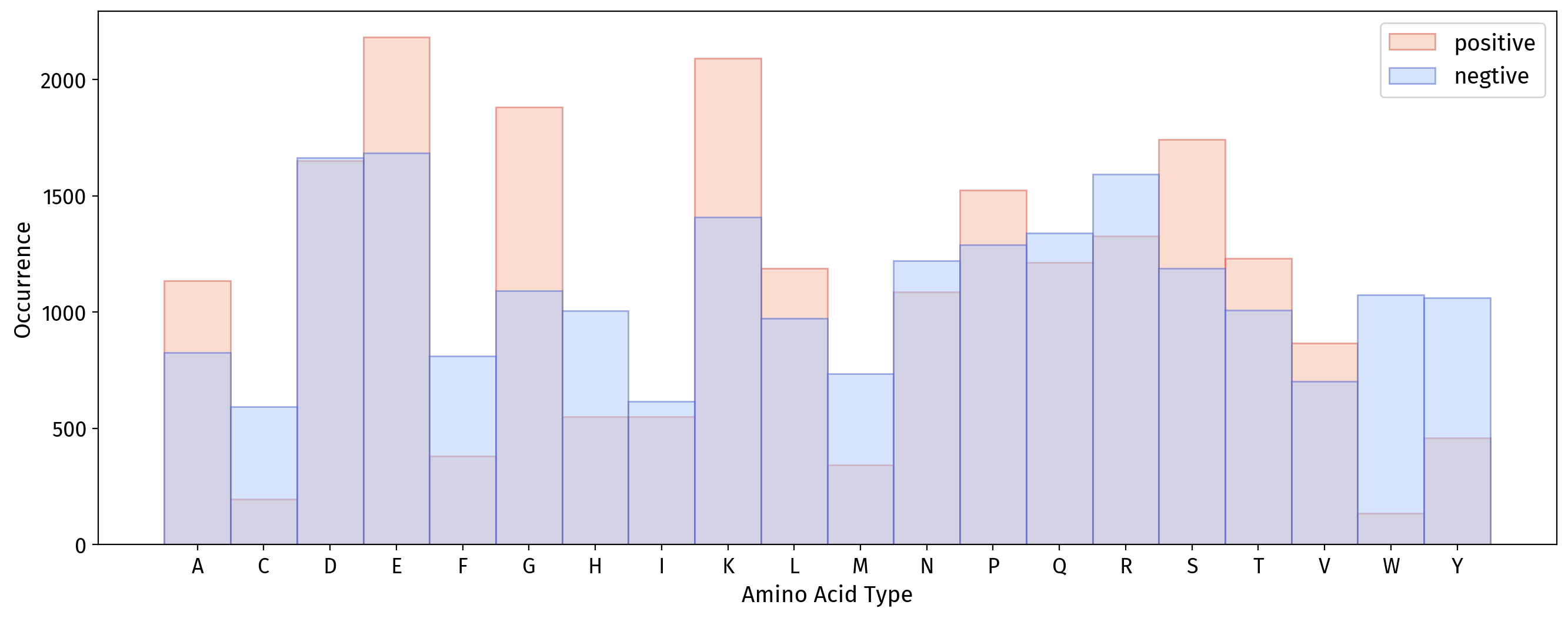}

   \captionof{figure}{Amino acid distribution comparison between positive and negative samples for nonfouling dataset.}
\label{fig:aa_dist_nonfouling}
\end{center}

\begin{center} 
\centering
\includegraphics[width=0.8\textwidth]{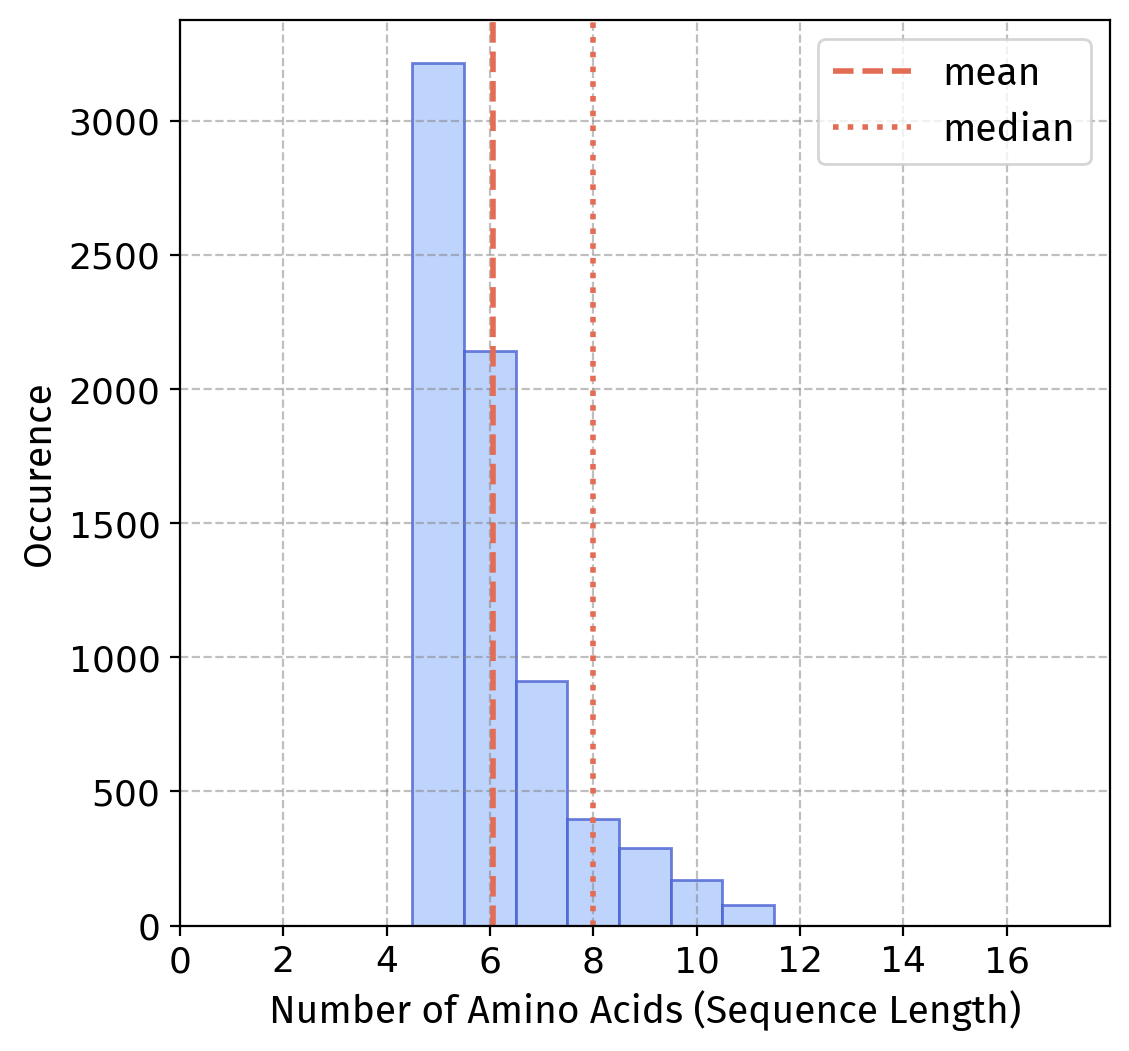}
   \captionof{figure}{Length distribution of nonfouling dataset.}
\label{fig:length_dist_nonfouling}
\end{center}

\begin{center} 
\centering
\includegraphics[width=0.8\textwidth]{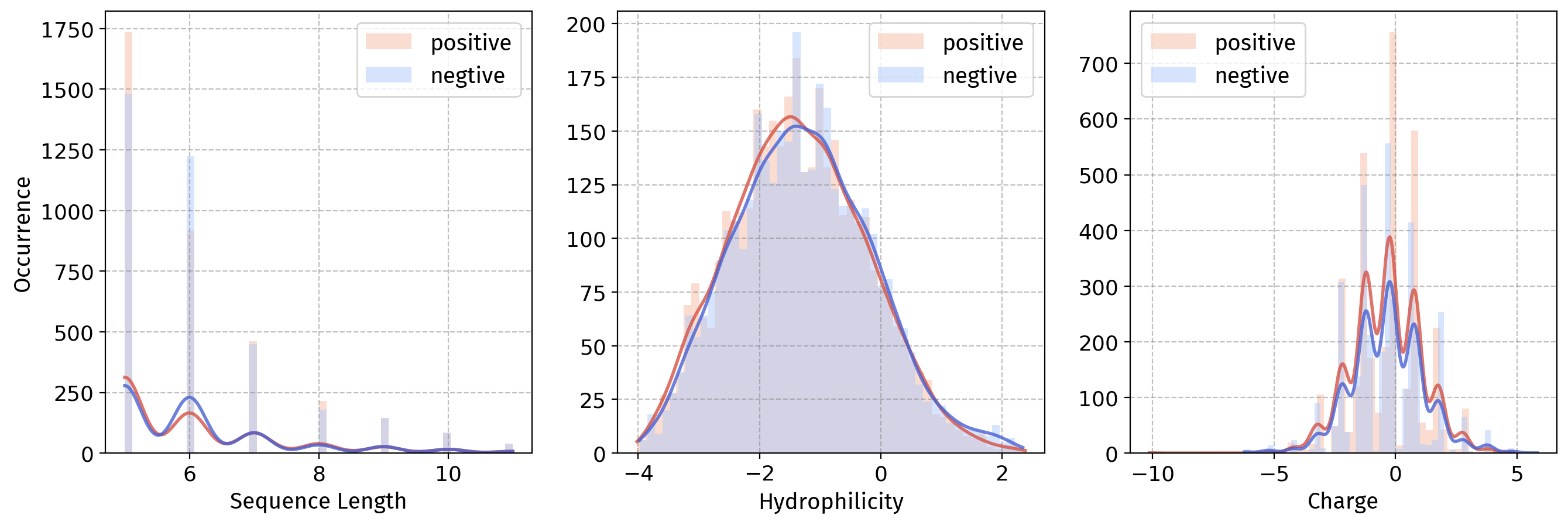}
   \captionof{figure}{Property comparison between positive and negative samples for nonfouling dataset.}
\label{fig:property_comp_nonfouling}
\end{center}

\subsubsection{Cell-Penetrating Peptides(cpp)}

\fielditem{Property and Application} 
The cpp dataset characterizes the ability of canonical peptides to penetrate cell membranes, a property crucial for designing efficient drug delivery systems targeting intracellular sites.

\fielditem{Data Source} 
The dataset is collected from Pepland~\citep{zhang2025pepland}, which extracts 1,162 transmembrane peptides from cell-penetrating peptide databases 
CPPsite and CPPsite2.0~\citep{kardani2021cppsite, gautam2012cppsite}, excluding any sequences with non-canonical amino 
acids. In addition, 156 canonical transmembrane peptides are supplemented from Peptipedia v2.0~\citep{cabas2024peptipedia}. 
Peptipedia v2.0 is constructed by integrating peptide records from literature, databases, public 
repositories and major protein databases such as UniProtKB and PDB using keyword searches. 
For each peptide, detailed information—including biological activities, sequence descriptions, 
experimental details, and associated publications or patents are extracted from all sources. 
Peptides are filtered by sequence length (3–150 residues) and annotated for biological activity 
through semantic analysis. 
The database comprises data from 76 sources, and it can be considered to comprehensively cover 
sequences with specific reported properties across all collected sources.

\fielditem{Dataset Statistics} 
The dataset contains 2,324 datapoints with sequences ranging from 3 to 61 amino acids (average length 17.24) in length.

\textbf{Task: Classification; Split: Hybrid; Evaluation: ROC-AUC}

\begin{center} 
\centering
\includegraphics[width=0.8\textwidth]{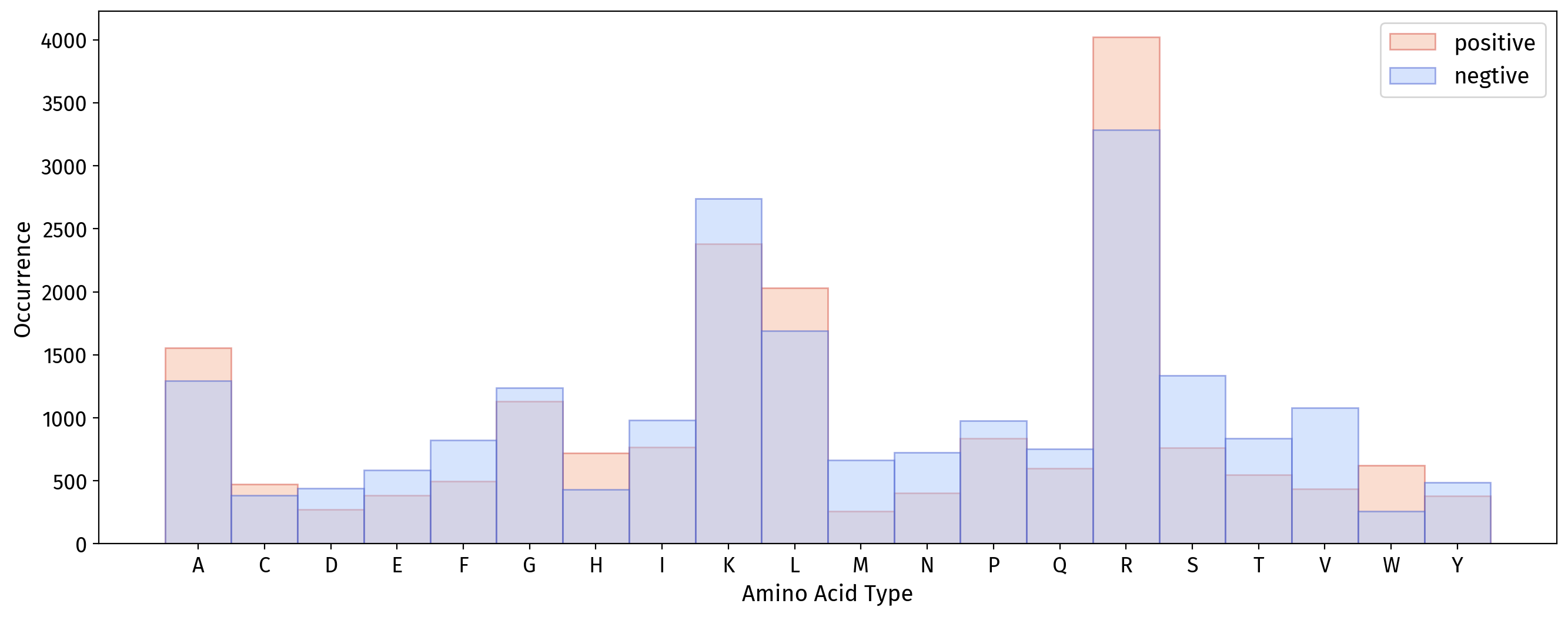}
   \captionof{figure}{Amino acid distribution comparison between positive and negative samples for cpp dataset.}
\label{fig:aa_dist_cpp}
\end{center}

\begin{center} 
\centering
\includegraphics[width=0.8\textwidth]{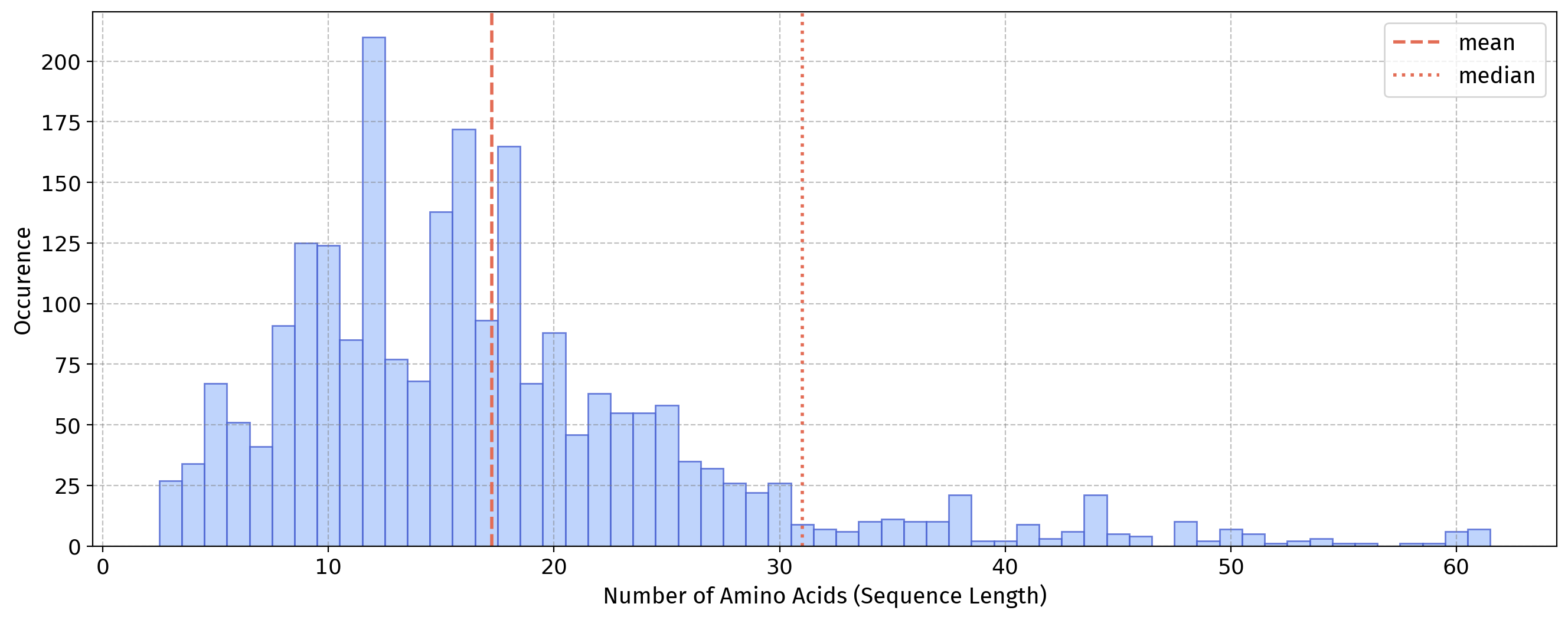}
   \captionof{figure}{Length distribution of cpp dataset.}
\label{fig:length_dist_cpp}
\end{center}

\begin{center} 
\centering
\includegraphics[width=0.8\textwidth]{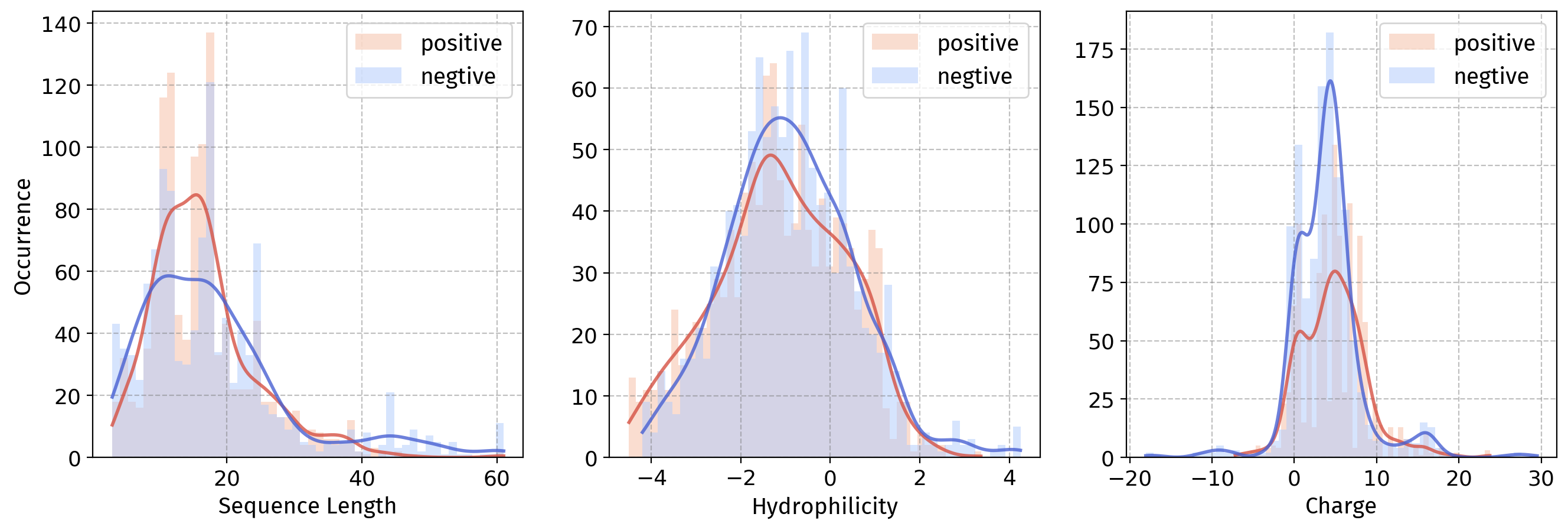}
   \captionof{figure}{Property comparison between positive and negative samples for cpp dataset.}
\label{fig:property_comp_cpp}
\end{center}

\subsubsection{Non-canonical Cell-Penetrating Peptides (nc-cpp\_pampa)}
\fielditem{Property and Application} 
The nc-cpp\_pampa dataset focuses on the membrane penetration ability of non-canonical cyclic peptides, providing insights into the design of synthetic peptides for drug delivery. This dataset enables the evaluation of how chemical modifications affect membrane permeability.

\fielditem{Data Source} 
The dataset is a subset of CycPeptMPDB~\citep{li2023cycpeptmpdb}, including only molecules that are measured using PAMPA (Parallel Artificial Membrane Permeability Assay). Data points categorized as "undetectable" (marked as -10) are removed to avoid potential errors due to aggregation or measurement artifacts. Labels are reported as log Pexp (log cm/s).

\fielditem{Dataset Statistics} 
The dataset contains 6,970 datapoints with sequences ranging from 2 to 15 amino acids (average length 8.04) in length.

\textbf{Task: Regression; Split: ECFP-based; Evaluation: MAE}

\begin{center} 
\centering
\includegraphics[width=0.8\textwidth]{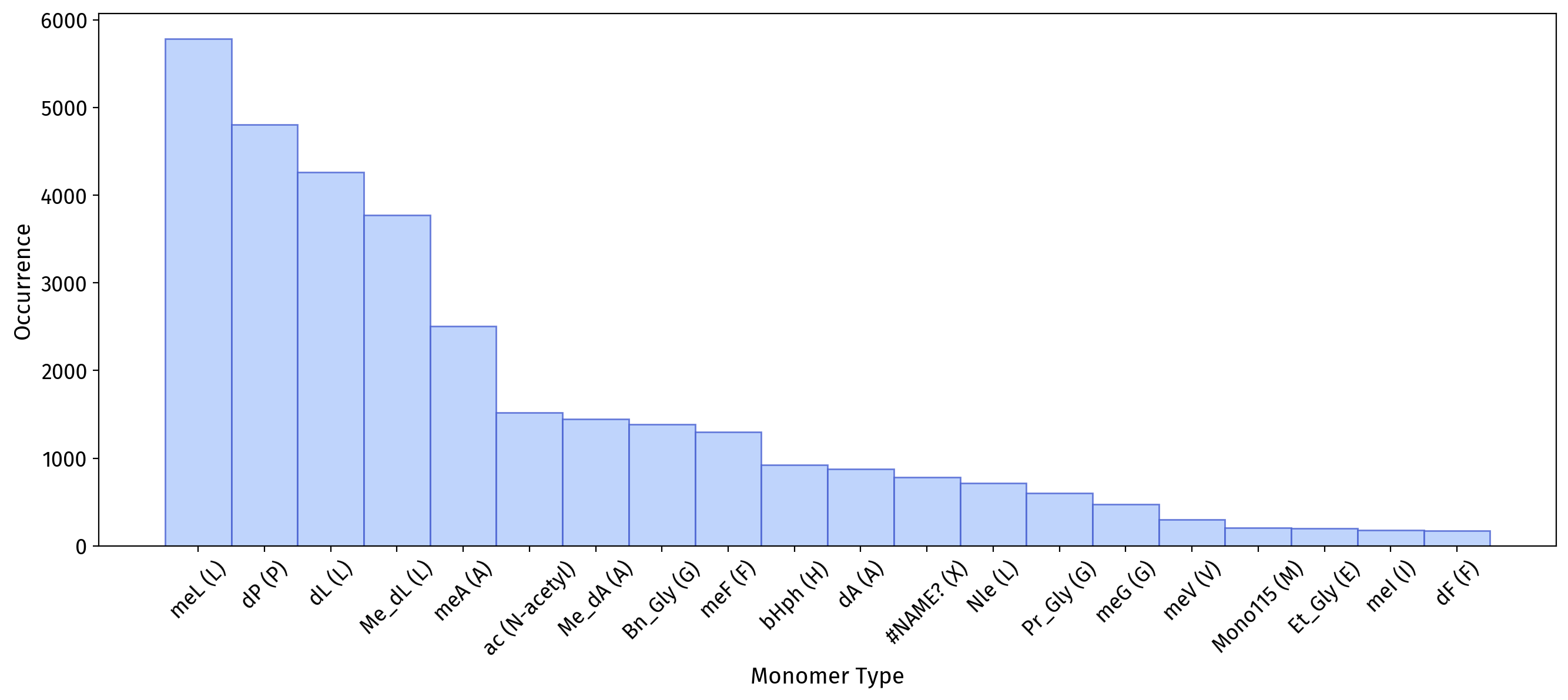}
   \captionof{figure}{Amino acid distribution for nc-cpp\_pampa dataset.}
\label{fig:aa_dist_nc-cpp}
\end{center}

\begin{center} 
\centering
\includegraphics[width=0.8\textwidth]{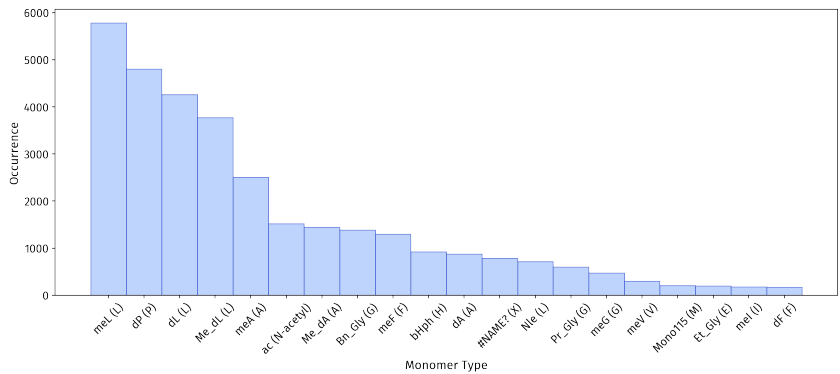}
   \captionof{figure}{non-canonical monomers distribution for nc-cpp\_pampa dataset.}
\label{fig:mon_nc-cpp}
\end{center}

\begin{center} 
\centering
\includegraphics[width=0.8\textwidth]{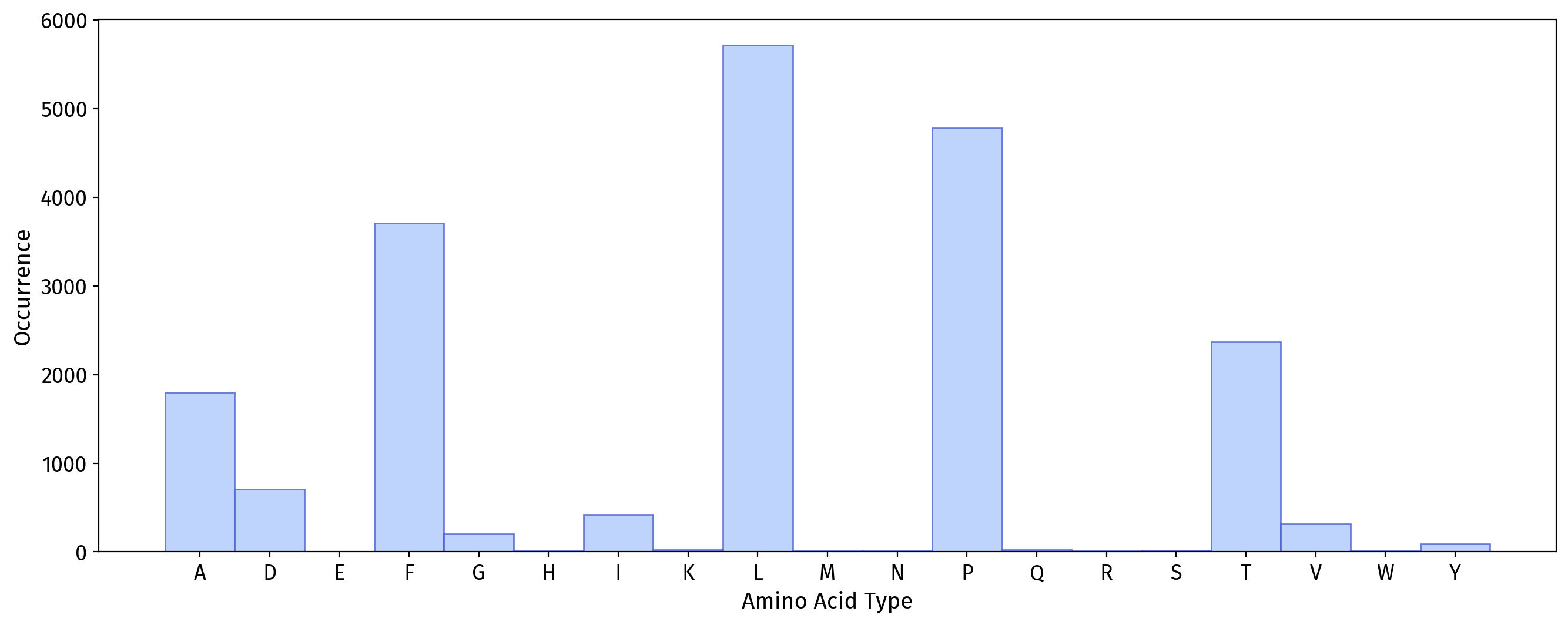}
   \captionof{figure}{Length distribution of nc-cpp\_pampa dataset.}
\label{fig:ncaa_dist_nc-cpp}
\end{center}


\subsubsection{Blood-Brain Barrier Peptides (bbp)}
\fielditem{Property and Application} 
The bbp dataset focuses on peptide's ability to penetrate the blood-brain barrier, a critical property for developing neuroactive peptides suitable for central nervous system diseases, such as Alzheimer's disease.

\fielditem{Data Source} 
The dataset is sourced from BBPpred~\citep{dai2021bbppred}, 
which collects BBP from Brainpeps~\citep{van2012brainpeps}, SATPdb~\citep{singh2016satpdb}, PepBank~\citep{shtatland2007pepbank}, 
and other literatures with experimental validation. Redundancy is removed using CD-HIT with a sequence identity threshold below 90\%, and obtains total 119 BBP. 
Additionally, 240 BBP from Peptipedia are merged into the dataset.

\paragraph{Experiment Negative Samples} 
Negative samples are generated through the official negative sampling procedure plus manually collected 14 dipeptide negative samples from ~\cite{tanaka2019brain}.

\fielditem{Dataset Statistics} 
The dataset contains 672 datapoints with sequences ranging from 2 to 82 amino acids (average length 14.14) in length.

\textbf{Task: Classification; Split: Hybrid; Evaluation: ROC-AUC}

\begin{center} 
\centering
\includegraphics[width=0.8\textwidth]{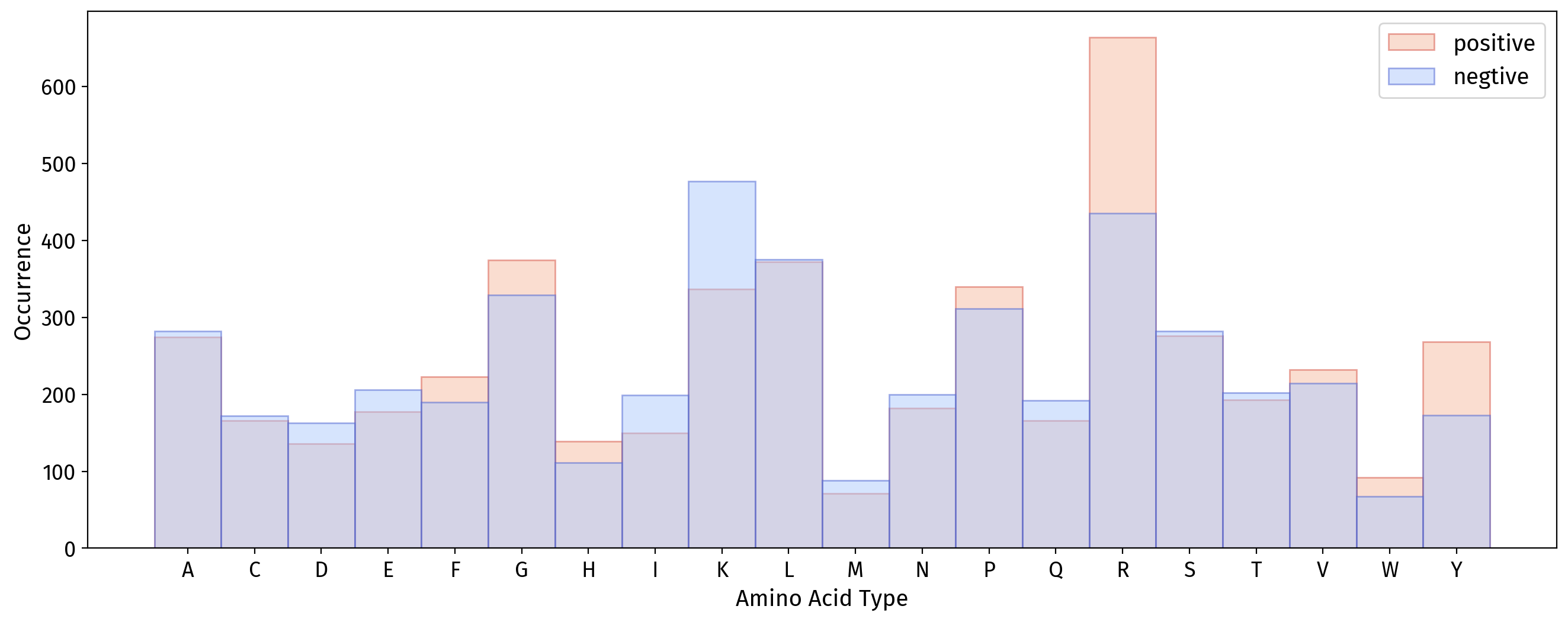}
   \captionof{figure}{Amino acid distribution comparison between positive and negative samples for bbp dataset.}
\label{fig:aa_dist_bbp}
\end{center}

\begin{center} 
\centering
\includegraphics[width=0.8\textwidth]{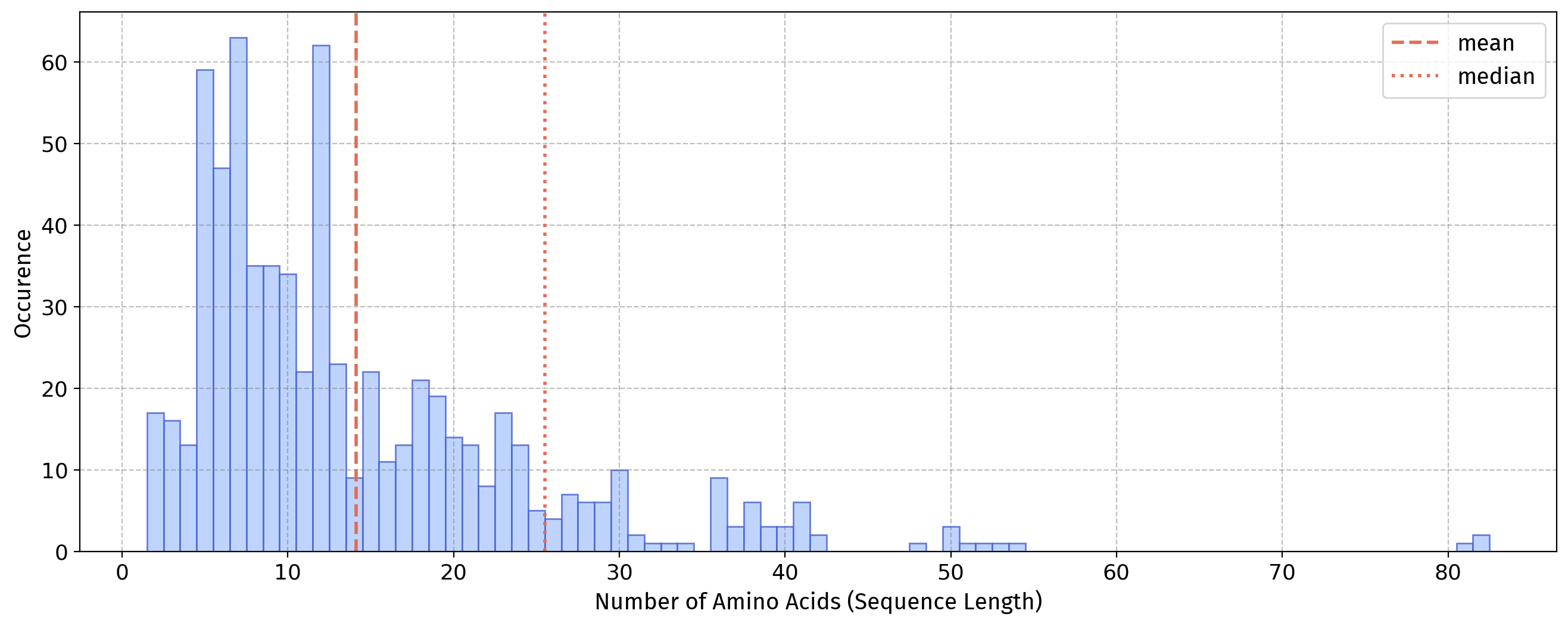}
   \captionof{figure}{Length distribution of bbp dataset.}
\label{fig:length_dist_bbp}
\end{center}

\begin{center} 
\centering
\includegraphics[width=0.8\textwidth]{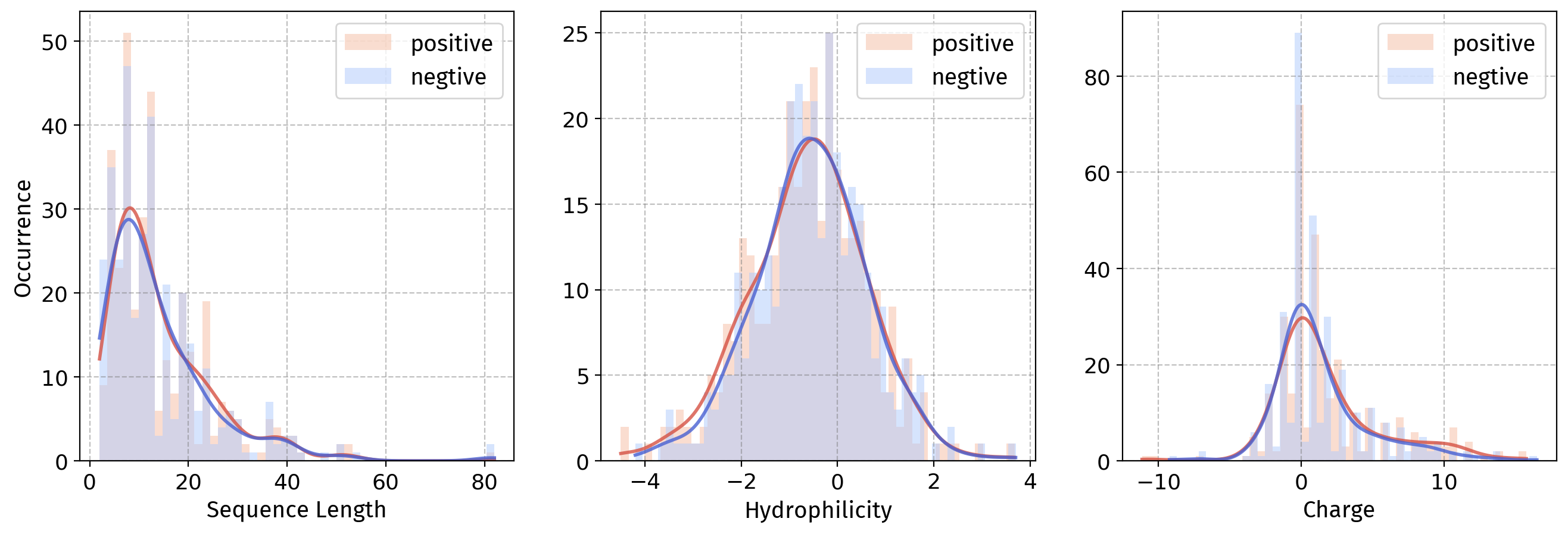}
   \captionof{figure}{Property comparison between positive and negative samples for bbp dataset.}
\label{fig:property_comp_bbp}
\end{center}

\subsection{AMP}

\begin{datasetbox}
\paragraph{Definition.} 
Datasets in this group focus on experimentally validated Antimicrobial peptides (AMP), covering peptides with either broad-spectrum antimicrobial activity or activity against specific types of microorganisms, such as bacteria, viruses, and fungi.
\paragraph{Impact.} 
Given the global health crisis posed by antimicrobial resistance, AMP represent promising alternatives or adjuncts to conventional antibiotics, offering broad-spectrum and evolutionarily conserved defense mechanisms.
\paragraph{Pipeline.} Activity Modeling
\end{datasetbox}

\subsubsection{antimicrobial}
\fielditem{Property and Application} 
The antimicrobial dataset encompasses peptides with broad-spectrum activity against bacteria, 
fungi, viruses, and parasites. These peptides represent a diverse class of 
natural defense molecules with potential therapeutic applications against resistant pathogens.

\fielditem{Data Source} 
This dataset originates from ~\cite{wang2025artificial},
which collects 22176 antimicrobial peptides from dbAMP~\citep{yao2025dbamp}, DRAMP~\citep{ma2025dramp}, 
GRAMPA~\citep{witten2019deep}, and starPep~\citep{aguilera2023starpep}, 
after removing duplicates. Additionally, another 16213 antimicrobial peptides from 
Peptipedia are merged, covering antibacterial, antifungal, antiviral, and antiparasitic peptides.

\fielditem{Dataset Statistics} 
The dataset contains 61,504 datapoints with sequences ranging from 2 to 150 amino acids (average length 28.55) in length.

\textbf{Task: Classification; Split: Hybrid; Evaluation: ROC-AUC}

\begin{center} 
\centering
\includegraphics[width=0.8\textwidth]{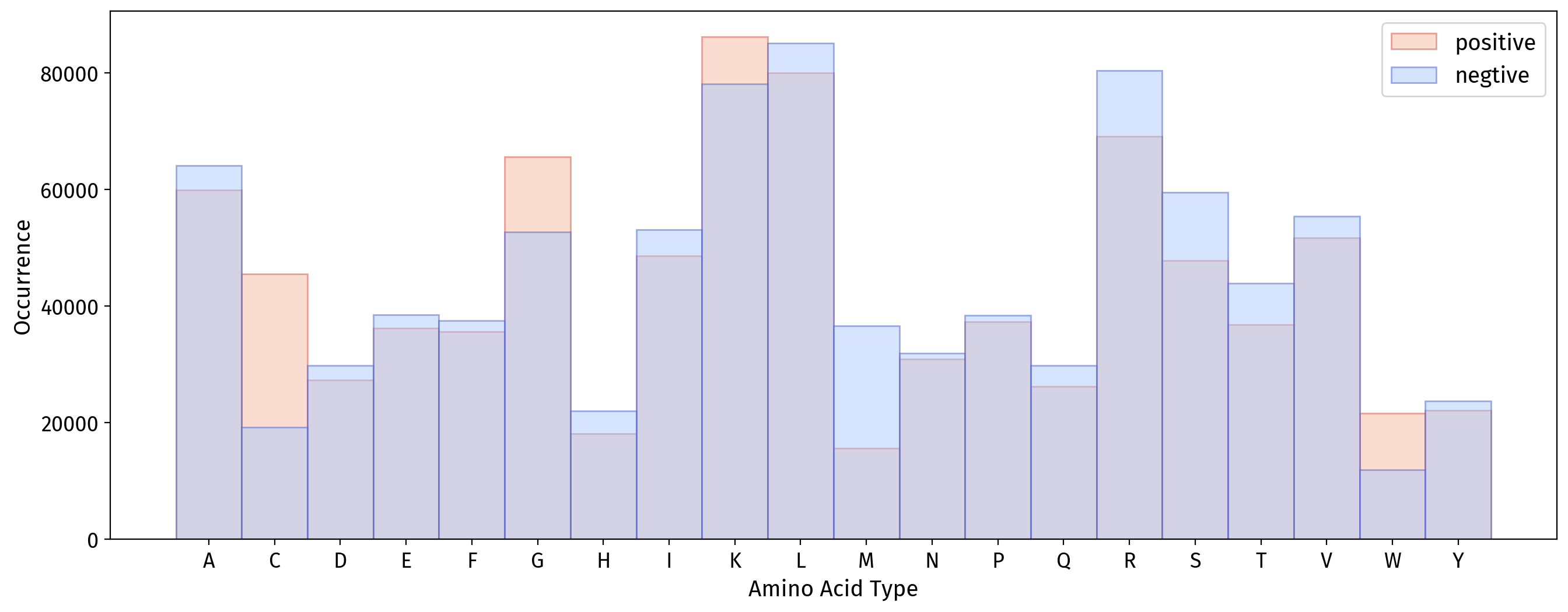}
   \captionof{figure}{Amino acid distribution comparison between positive and negative samples for antimicrobial dataset.}
\label{fig:aa_dist_antimicrobial}
\end{center}

\begin{center} 
\centering
\includegraphics[width=0.8\textwidth]{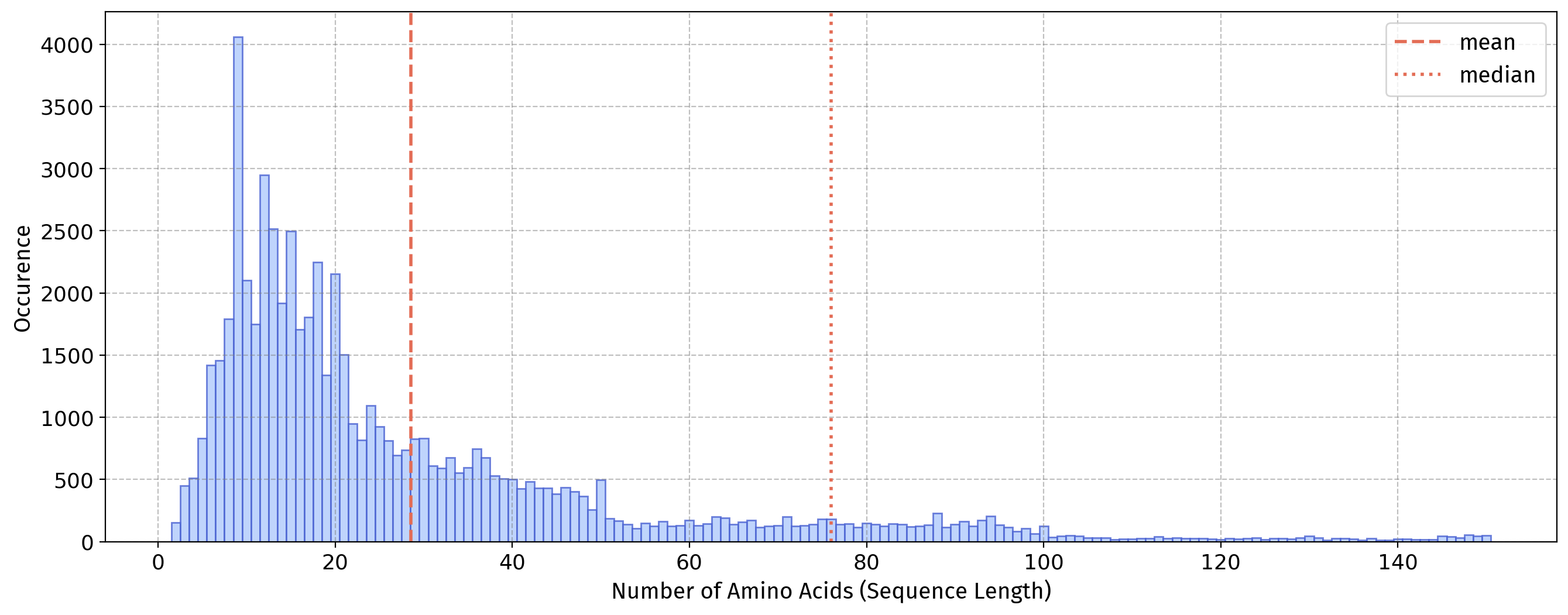}
   \captionof{figure}{Length distribution of antimicrobial dataset.}
\label{fig:length_dist_antimicrobial}
\end{center}

\begin{center} 
\centering
\includegraphics[width=0.8\textwidth]{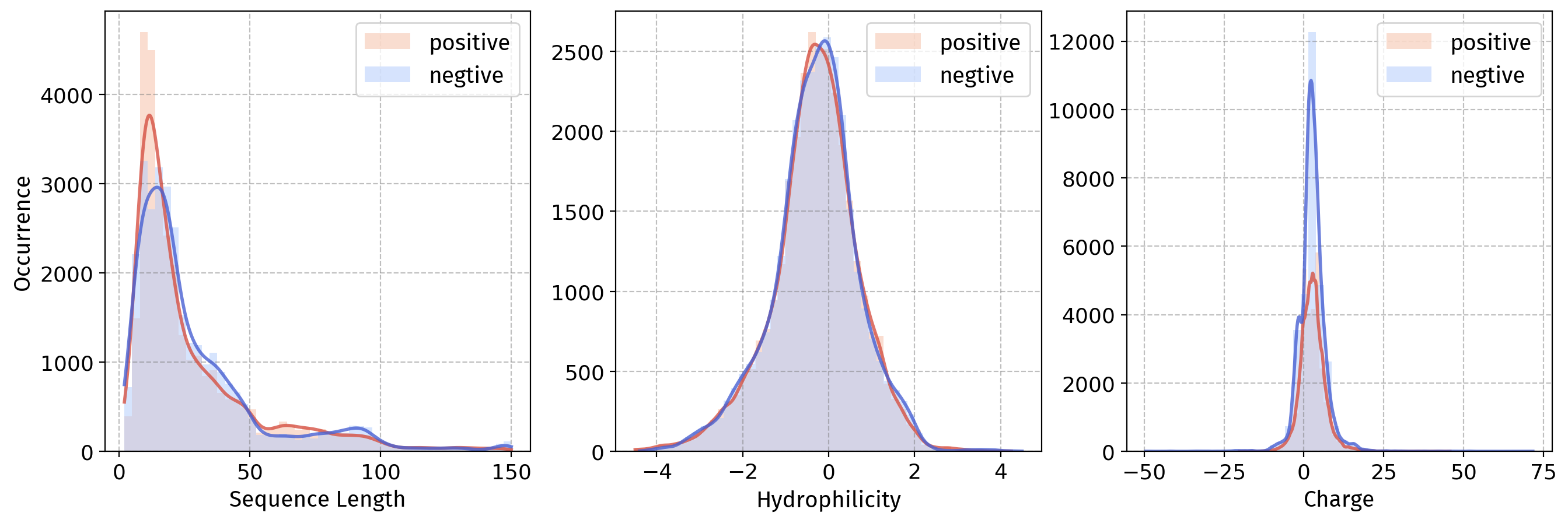}
   \captionof{figure}{Property comparison between positive and negative samples for antimicrobial dataset.}
\label{fig:property_comp_antimicrobial}
\end{center}

\subsubsection{antibacterial}
\fielditem{Property and Application} 
The antibacterial dataset focuses on peptides with activity against bacterial pathogens. These peptides are crucial for developing novel antibiotics to combat bacterial resistance.

\fielditem{Data Source} 
This dataset originates from ~\cite{pinacho2021alignment} 
which extracts a set of active AMP from starPepDB, comprising 8,278, 993, 130, and 2,944 AMPs 
with antibacterial, antifungal, antiparasitic, and antiviral activities, respectively. 
After retaining AMP with length between 5 and 100 natural amino acids and removing those listed as having more than one of the activities mentioned above, 
only 6010 peptides with single antibacterial activity are included in the training dataset. After adding 15210 peptides from Peptipedia, the dataset is expanded to 21220 peptides, which also includes antibacterial peptides with activity against other types of microorganisms.

\fielditem{Dataset Statistics} 
The dataset contains 31,676 datapoints with sequences ranging from 2 to 150 amino acids (average length 25.33) in length.

\textbf{Task: Classification; Split: Hybrid; Evaluation: ROC-AUC}

\begin{center} 
\centering
\includegraphics[width=0.8\textwidth]{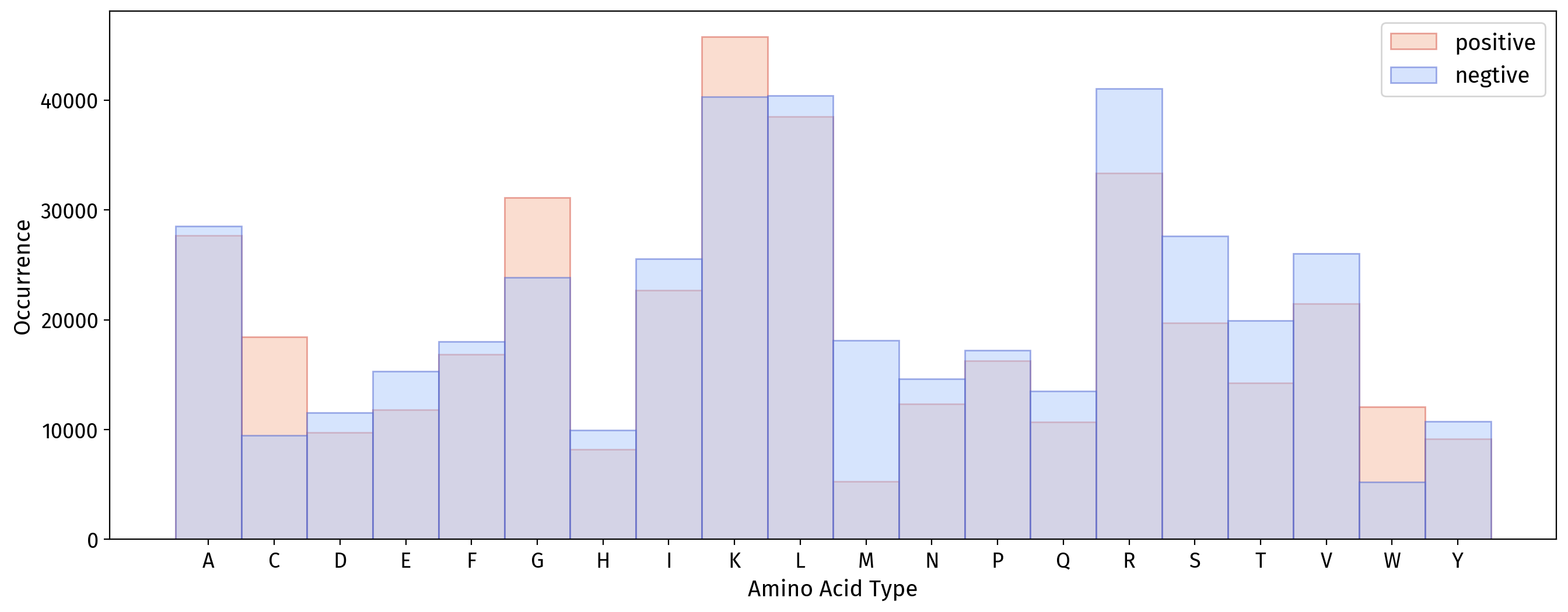}
   \captionof{figure}{Amino acid distribution comparison between positive and negative samples for antibacterial dataset.}
\label{fig:aa_dist_antibacterial}
\end{center}

\begin{center} 
\centering
\includegraphics[width=0.8\textwidth]{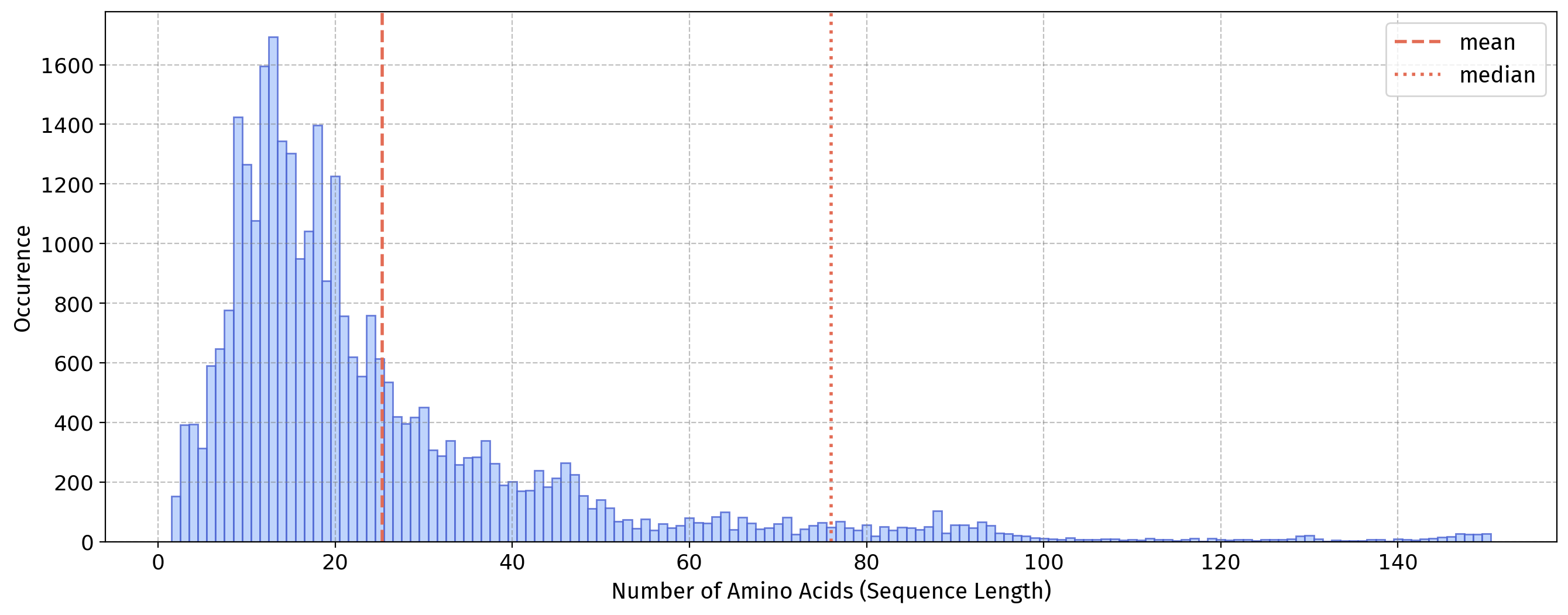}
   \captionof{figure}{Length distribution of antibacterial dataset.}
\label{fig:length_dist_antibacterial}
\end{center}

\begin{center} 
\centering
\includegraphics[width=0.8\textwidth]{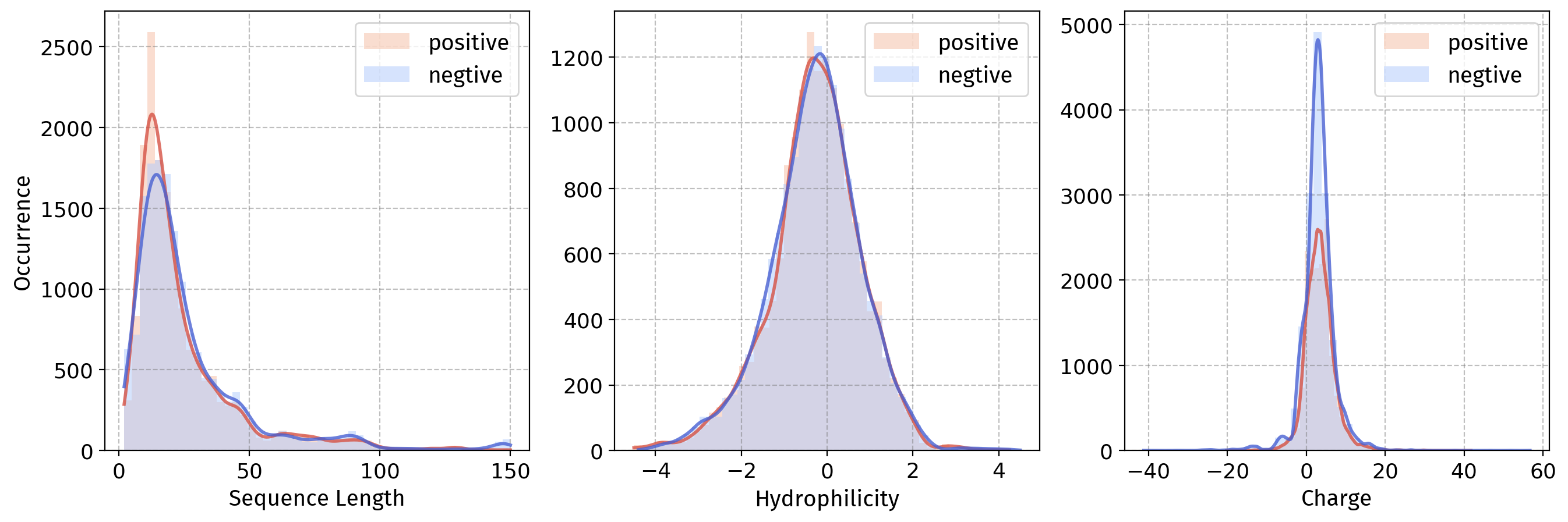}
   \captionof{figure}{Property comparison between positive and negative samples for antibacterial dataset.}
\label{fig:property_comp_antibacterial}
\end{center}

\subsubsection{antifungal}
\fielditem{Property and Application} 
The antifungal dataset collects peptides with activity against fungal pathogens. These peptides are important for developing treatments against fungal infections, which are increasingly problematic in immunocompromised patients.

\fielditem{Data Source} 
993 peptides are extracted from ~\cite{pinacho2021alignment} and 10094 antifungal peptides are merged from Peptipedia.

\fielditem{Dataset Statistics} 
The dataset contains 16,698 datapoints with sequences ranging from 2 to 148 amino acids (average length 34.17) in length.

\textbf{Task: Classification; Split: Hybrid; Evaluation: ROC-AUC}

\begin{center} 
\centering
\includegraphics[width=0.8\textwidth]{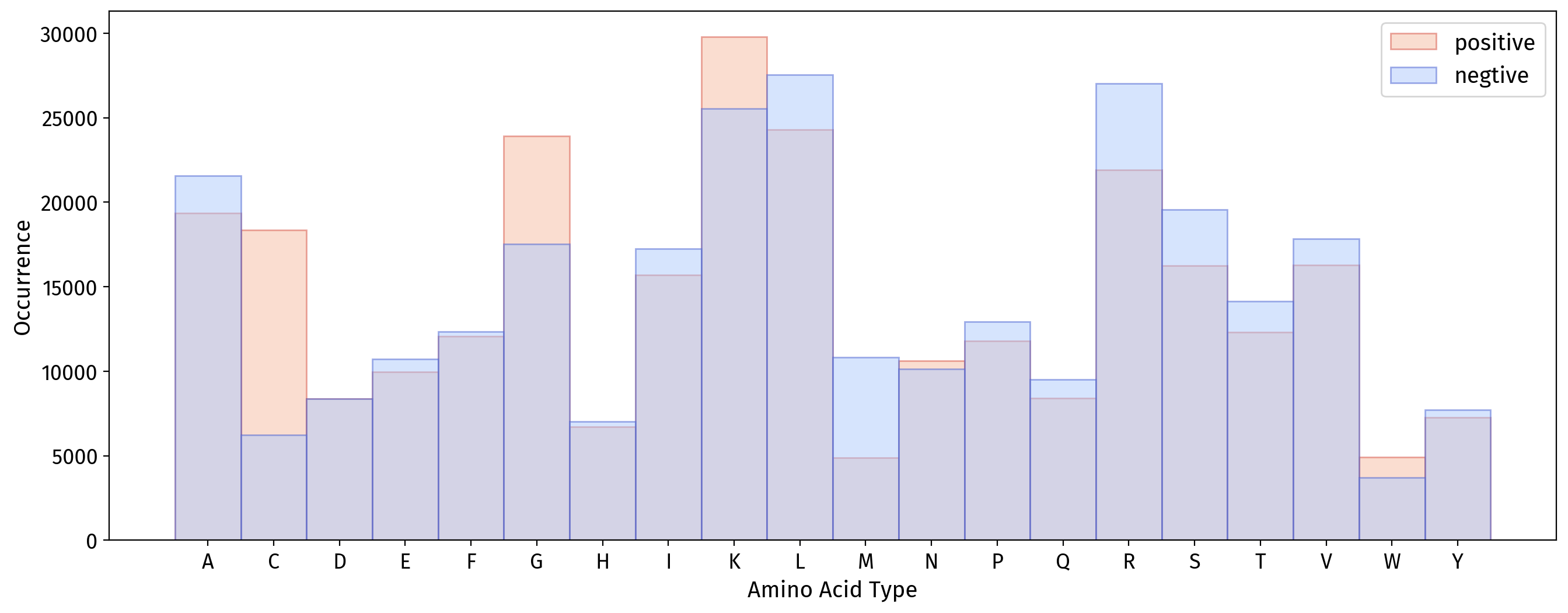}
   \captionof{figure}{Amino acid distribution comparison between positive and negative samples for antifungal dataset.}
\label{fig:aa_dist_antifungal}
\end{center}

\begin{center} 
\centering
\includegraphics[width=0.8\textwidth]{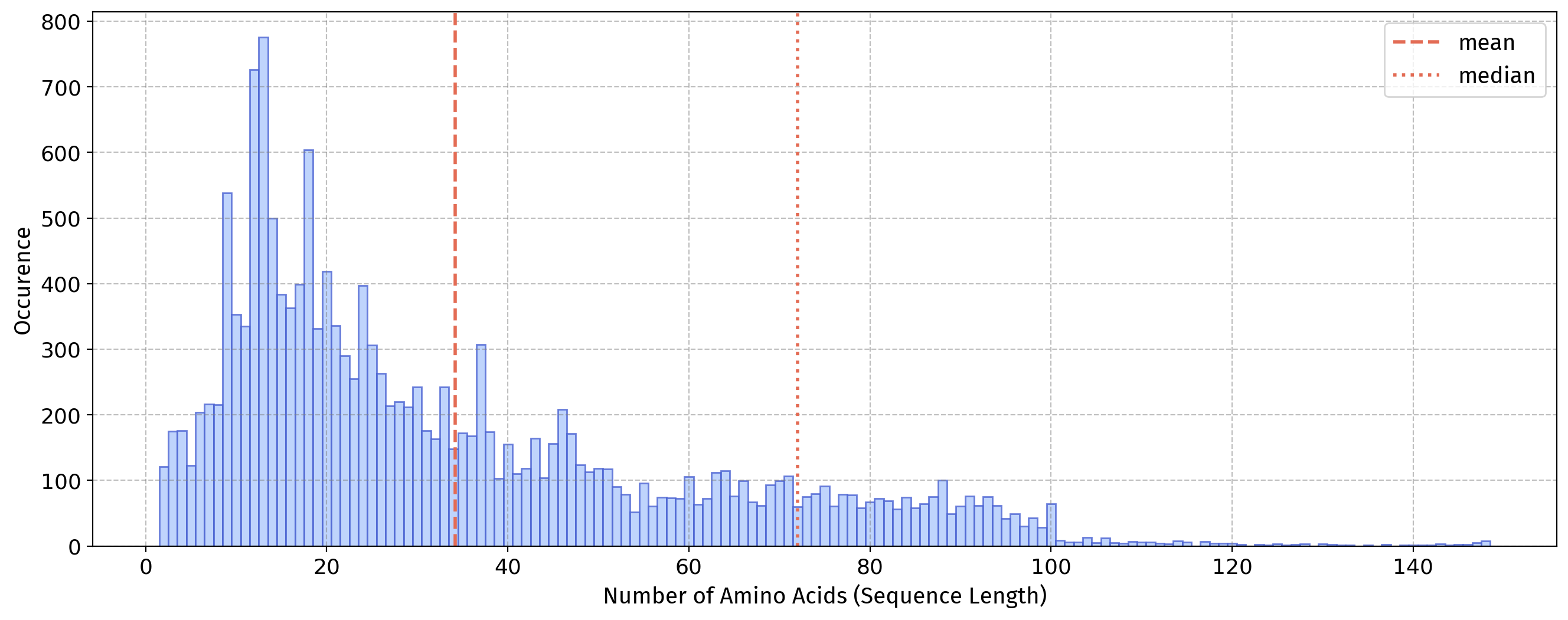}
   \captionof{figure}{Length distribution of antifungal dataset.}
\label{fig:length_dist_antifungal}
\end{center}

\begin{center} 
\centering
\includegraphics[width=0.8\textwidth]{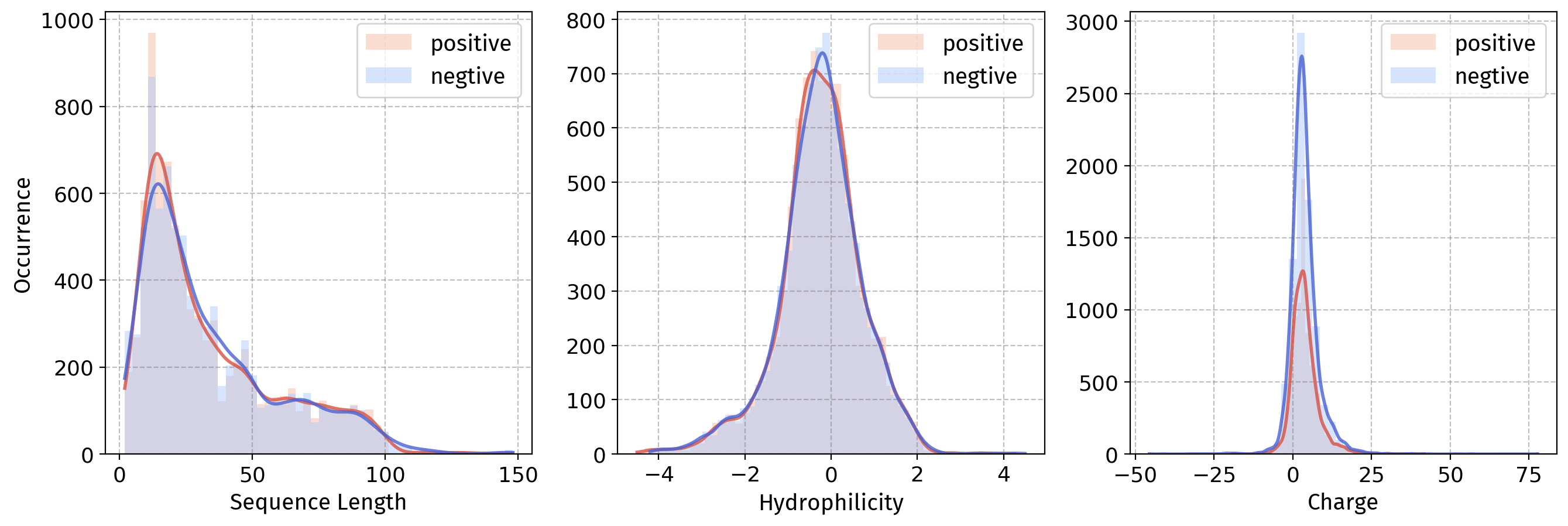}
   \captionof{figure}{Property comparison between positive and negative samples for antifungal dataset.}
\label{fig:property_comp_antifungal}
\end{center}

\subsubsection{antiparasitic}
\fielditem{Property and Application} 
The antiparasitic dataset contains peptides active against parasites. These peptides represent potential therapeutics for parasitic diseases that affect millions worldwide.

\fielditem{Data Source} 
The dataset originates from ~\cite{zhang2022predapp}, which collects peptides 
from two resources: (1) APP and AMP databases including ParaPep~\citep{Mehta2014ParaPepAW}, APD3~\citep{Wang2015APD3TA}, 
dbAMP~\citep{Jhong2018dbAMPAI}, CAMP~\citep{Thomas2009CAMPAU}, DRAMP~\citep{Fan2016DRAMPAC}, and ADAM~\citep{Nothaft2015RethinkingDS}; 
(2) APP-related articles during 2015-01-01 to 2019-10-31 in PubMed. 
Homologous sequences at a threshold of 90\% are removed for positive samples, obtaining 301 peptides in total.
After adding 5741 antiparasitic peptides from Peptipedia, the dataset is expanded to 6042 peptides.

\fielditem{Dataset Statistics} 
The dataset contains 8,632 datapoints with sequences ranging from 2 to 140 amino acids (average length 35.97) in length.

\textbf{Task: Classification; Split: Hybrid; Evaluation: ROC-AUC}

\begin{center} 
\centering
\includegraphics[width=0.8\textwidth]{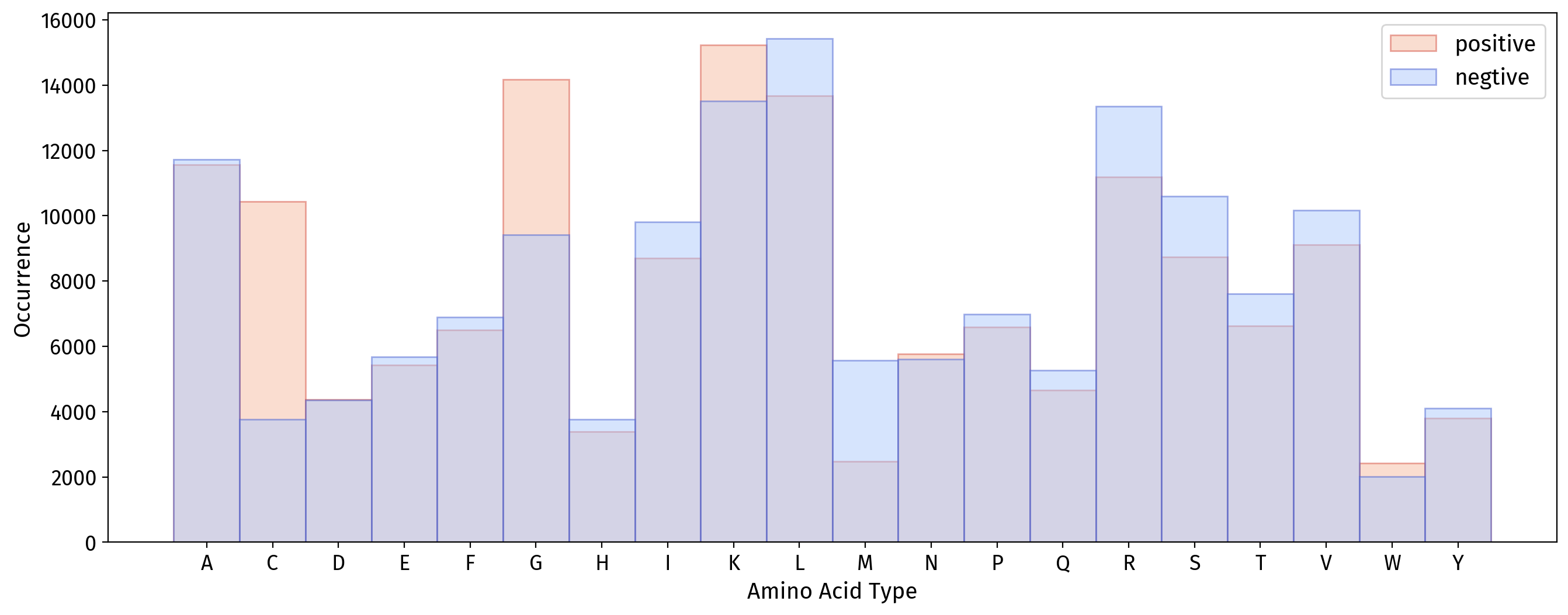}
   \captionof{figure}{Amino acid distribution comparison between positive and negative samples for antiparasitic dataset.}
\label{fig:aa_dist_antiparasitic}
\end{center}

\begin{center} 
\centering
\includegraphics[width=0.8\textwidth]{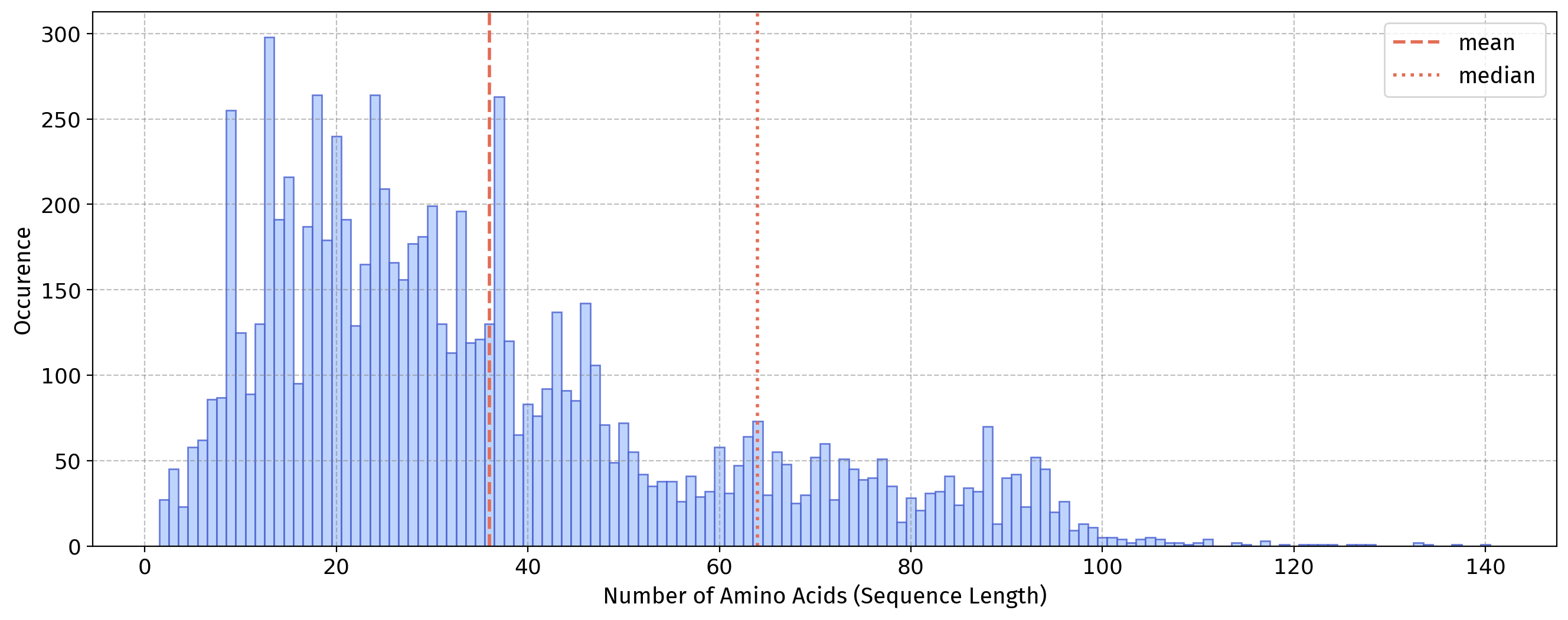}
   \captionof{figure}{Length distribution of antiparasitic dataset.}
\label{fig:length_dist_antiparasitic}
\end{center}

\begin{center} 
\centering
\includegraphics[width=0.8\textwidth]{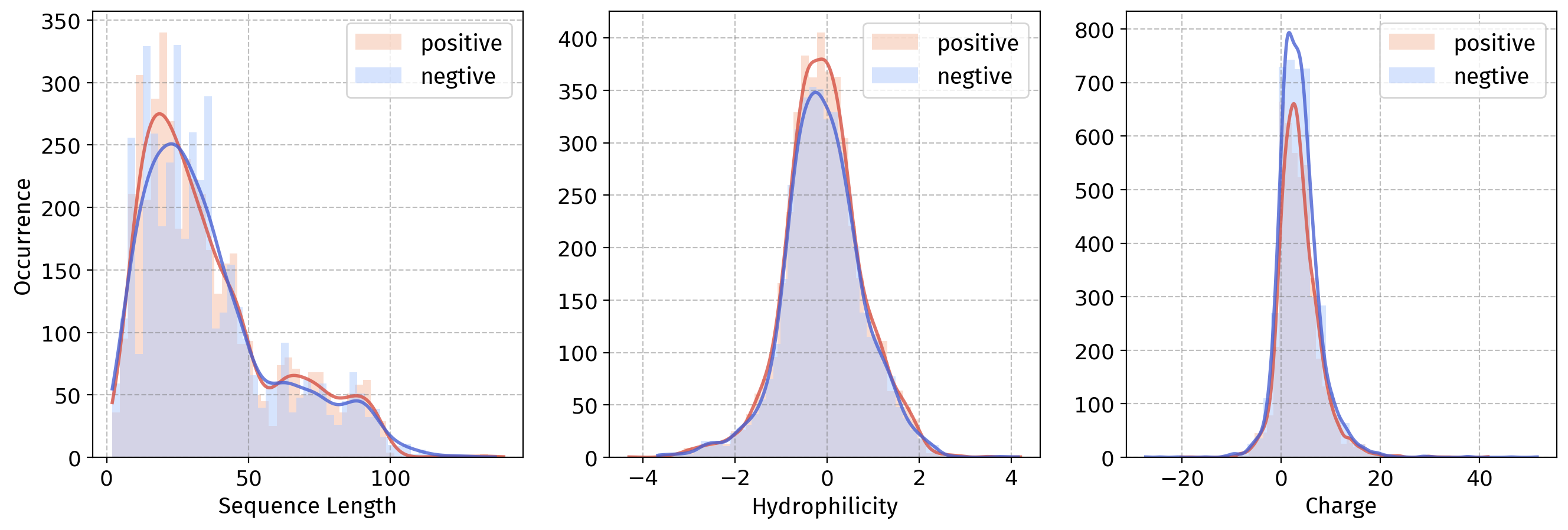}
   \captionof{figure}{Property comparison between positive and negative samples for antiparasitic dataset.}
\label{fig:property_comp_antiparasitic}
\end{center}

\subsubsection{antiviral}
\fielditem{Property and Application} 
The antiviral dataset focuses on peptides with activity against viral pathogens. These peptides are crucial for developing novel antiviral therapeutics, especially important given the emergence of new viral threats.

\fielditem{Data Source} 
2994 antifungal peptides are extracted from ~\cite{pinacho2021alignment} and 2266 antifungal peptides are merged from Peptipedia. 

\fielditem{Dataset Statistics} 
The dataset contains 8,268 datapoints with sequences ranging from 2 to 138 amino acids (average length 22.75) in length.

\textbf{Task: Classification; Split: Hybrid; Evaluation: ROC-AUC}

\begin{center} 
\centering
\includegraphics[width=0.8\textwidth]{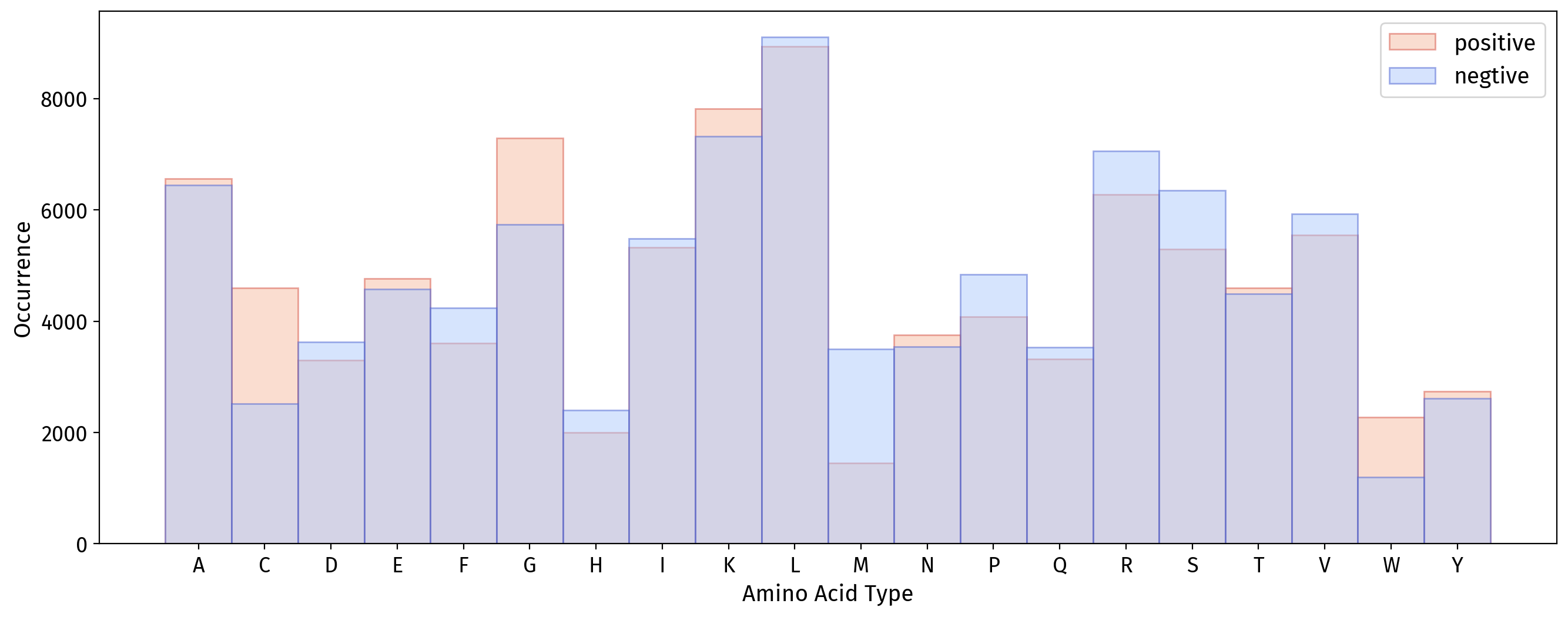}
   \captionof{figure}{Amino acid distribution comparison between positive and negative samples for antiviral dataset.}
\label{fig:aa_dist_antiviral}
\end{center}

\begin{center} 
\centering
\includegraphics[width=0.8\textwidth]{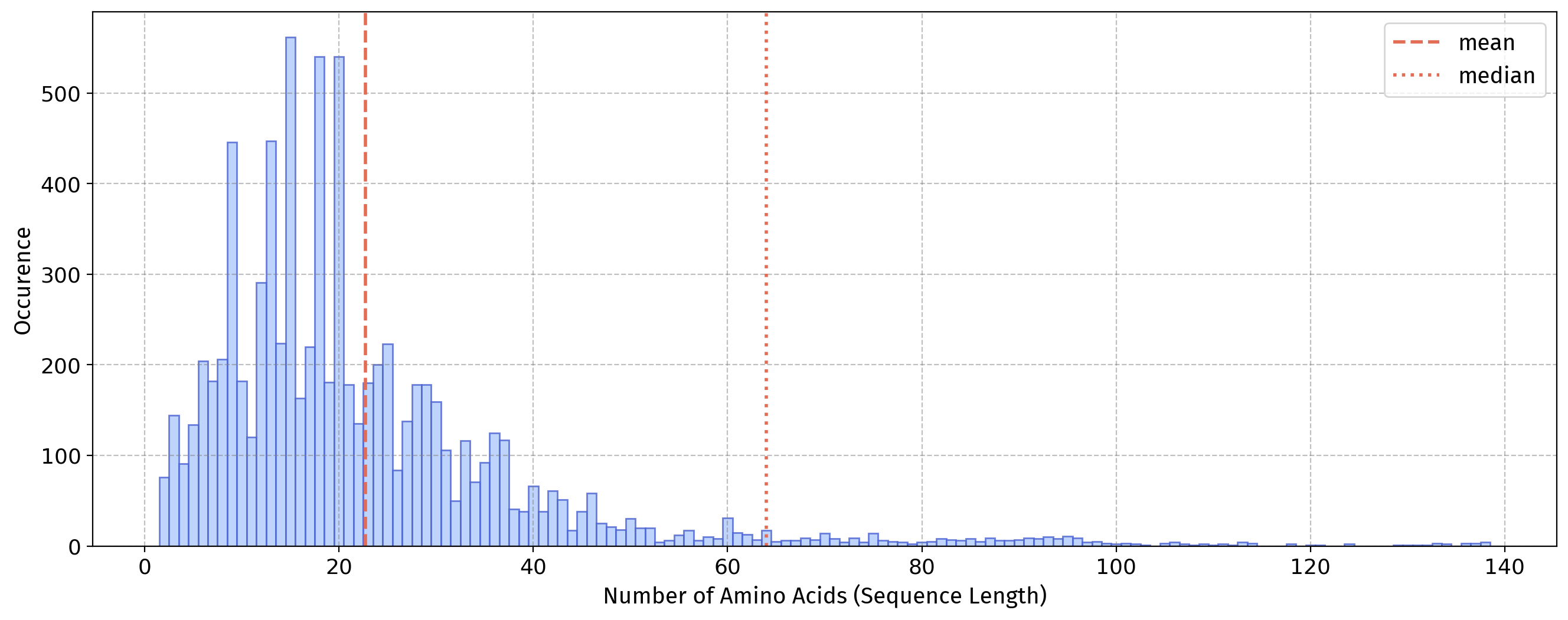}
   \captionof{figure}{Length distribution of antiviral dataset.}
\label{fig:length_dist_antiviral}
\end{center}

\begin{center} 
\centering
\includegraphics[width=0.8\textwidth]{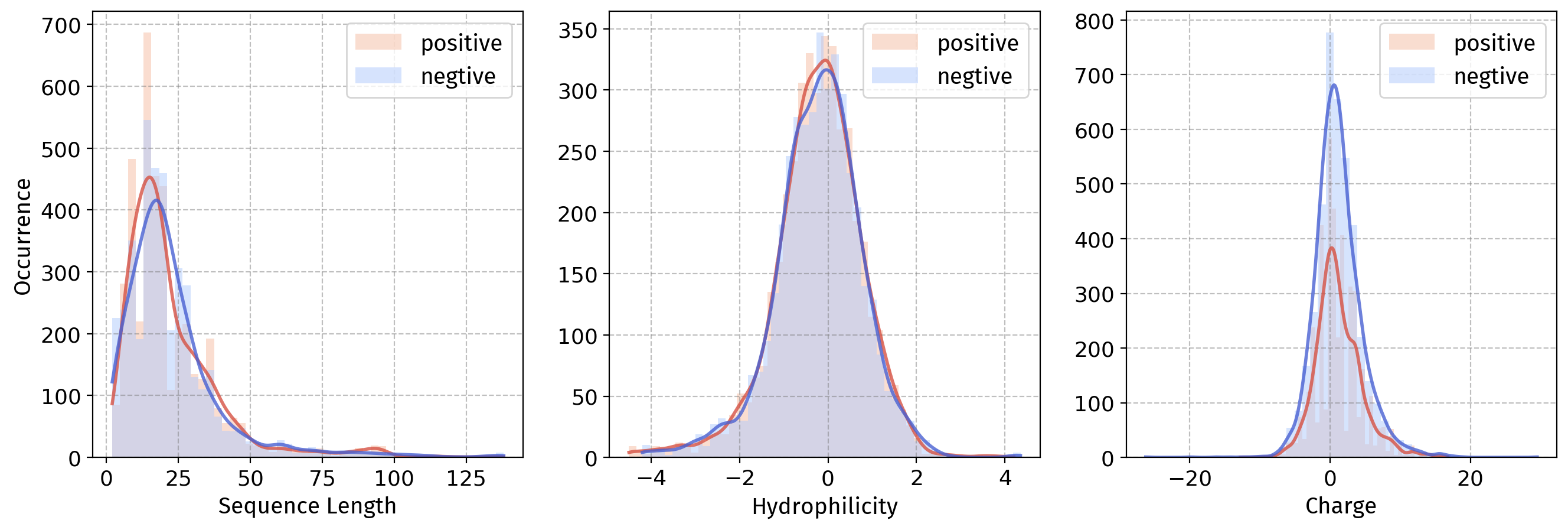}
   \captionof{figure}{Property comparison between positive and negative samples for antiviral dataset.}
\label{fig:property_comp_antiviral}
\end{center}

\subsubsection{E. coli MIC}
\fielditem{Property and Application} 
The E. coli MIC dataset reports the minimum inhibitory concentrations (MICs) of peptides against \textit{Escherichia coli}. MIC, defined as the lowest drug concentration that inhibits pathogen growth \textit{in vitro} after 18-24 hours of bacterial culture, provides a quantitative measure of antibacterial potency.

\fielditem{Data Source} 
The Dataset is sourced from GRAMPA~\citep{witten2019deep}, 
which records peptide MIC values against different bacterial strains. 
To minimize experimental variation caused by different experimental environments and batches, 
data points outside the range [Q1-1.5×IQR, Q3+1.5×IQR] are identified as outliers and removed, 
with the average of remaining values used as the final label. The MIC values are reported as $\log(\mu\text{M})$.

\fielditem{Dataset Statistics} 
The dataset contains 3,312 datapoints with sequences ranging from 2 to 140 amino acids (average length 23.24) in length.

\textbf{Task: Regression; Split: Hybrid; Evaluation: MAE}

\begin{center} 
\centering
\includegraphics[width=0.8\textwidth]{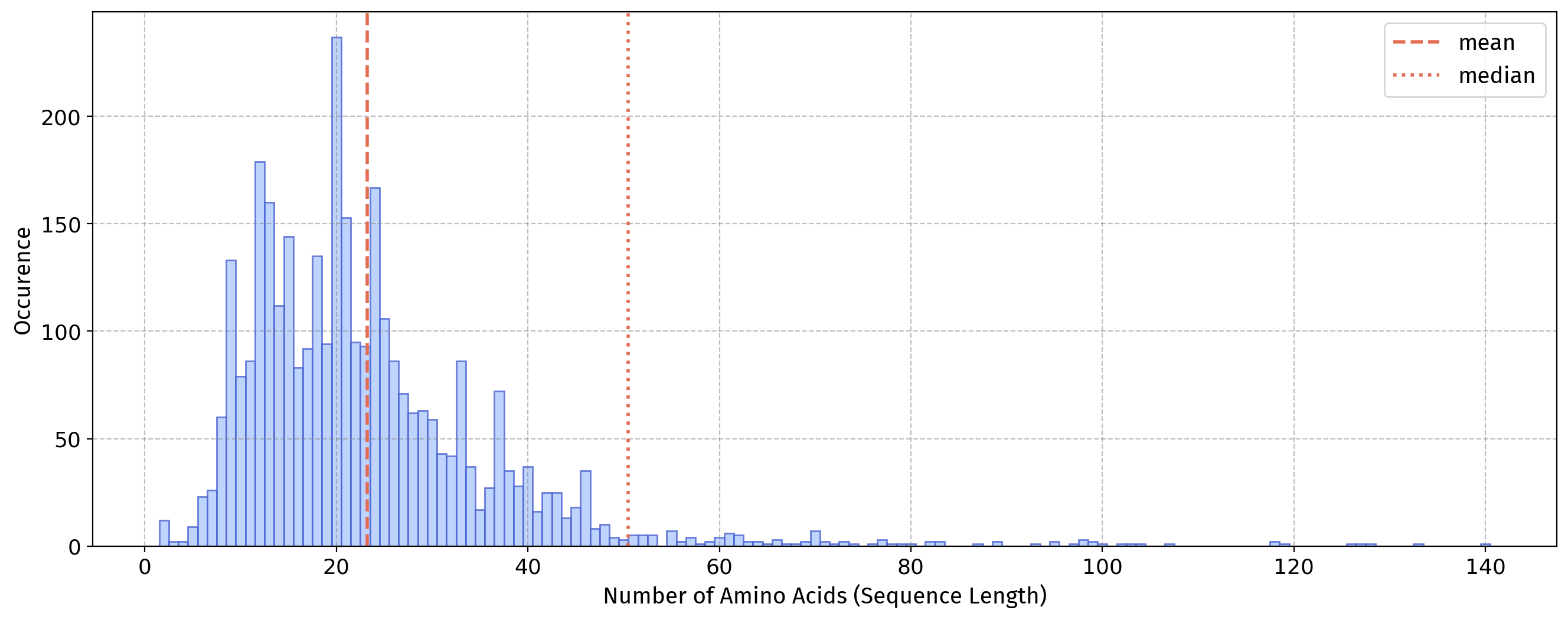}
   \captionof{figure}{Length distribution of E. coli MIC dataset.}
\label{fig:length_dist_E_coli_mic}
\end{center}

\begin{center} 
\centering
\includegraphics[width=0.8\textwidth]{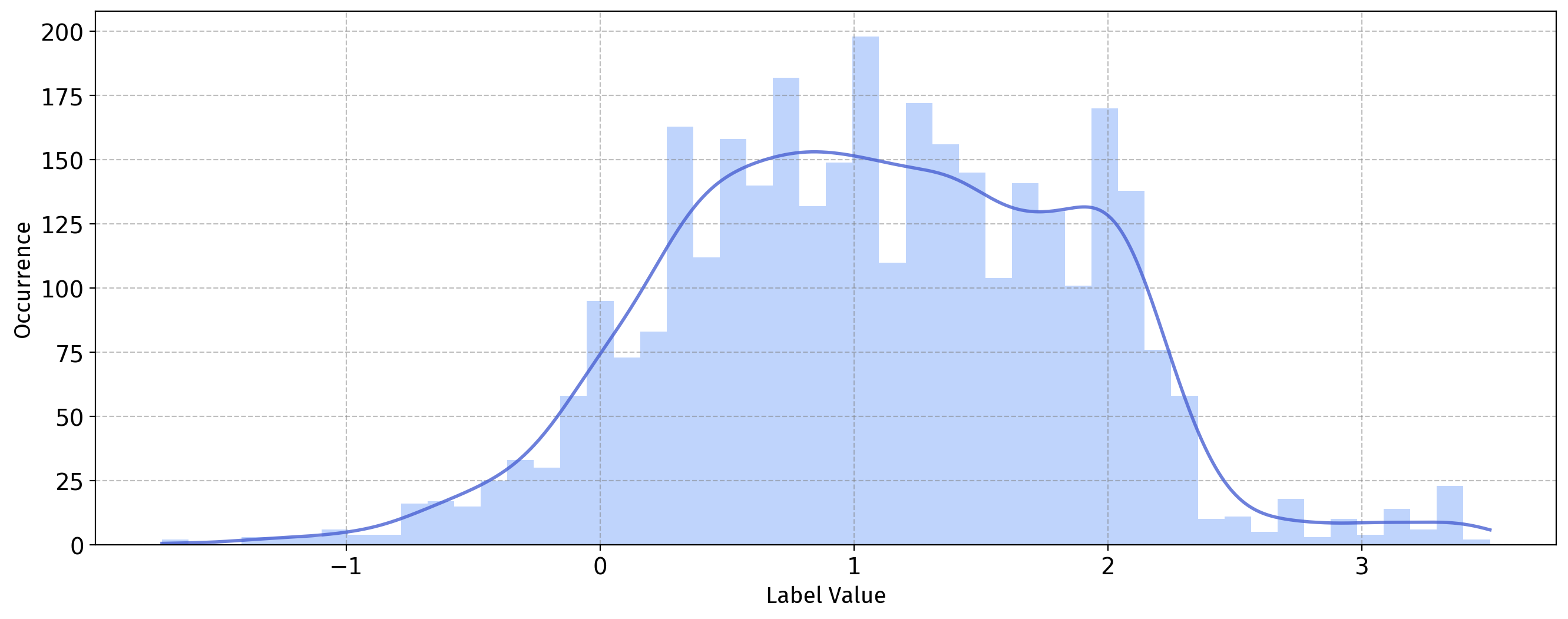}
   \captionof{figure}{Label distribution of E. coli MIC dataset.}
\label{fig:label_dist_E_coli_mic}
\end{center}

\begin{center} 
  \centering
  \includegraphics[width=0.8\textwidth]{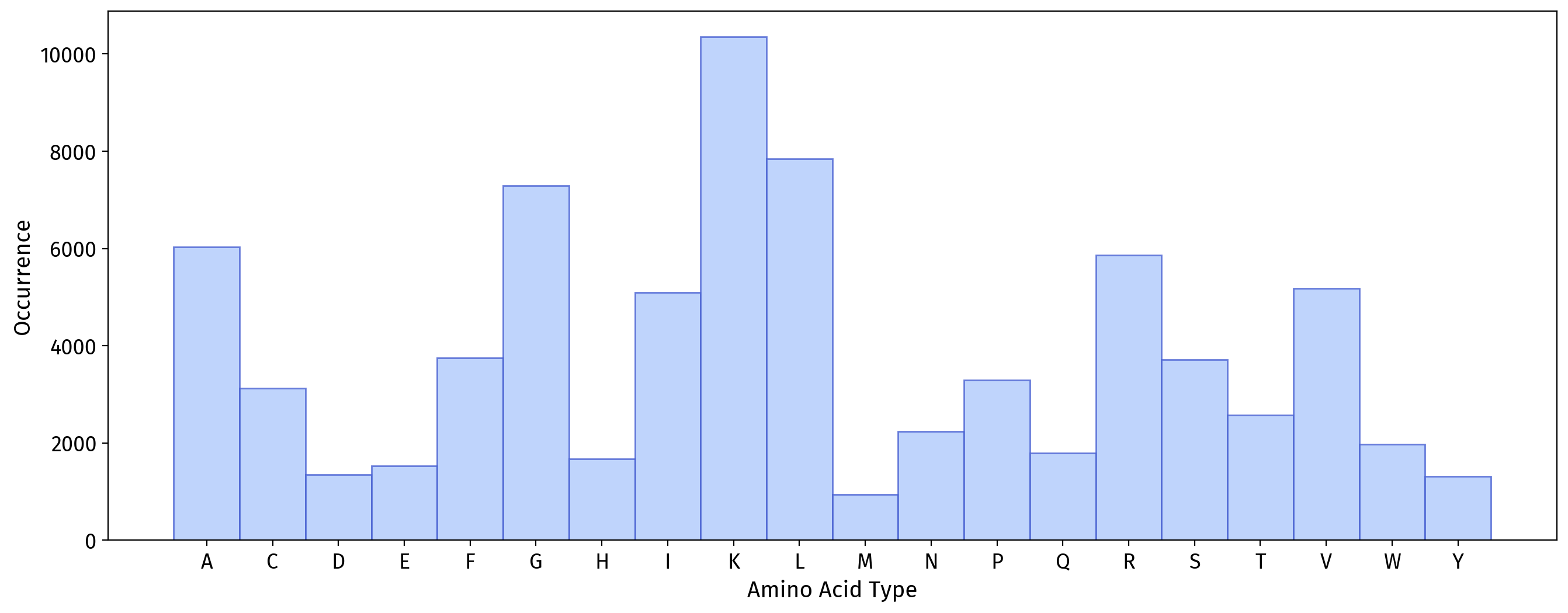}
     \captionof{figure}{Amino acid distribution of E. coli MIC dataset.}
  \label{fig:aa_dist_E_coli_mic}
  \end{center}

\subsubsection{P. aeruginosa MIC}
\fielditem{Property and Application} 
This dataset provides quantitative measurements of peptide activity against \textit{Pseudomonas aeruginosa}, a clinically important Gram-negative pathogen known for its resistance to many antibiotics.

\fielditem{Data Source} 
The Dataset is sourced from GRAMPA~\citep{witten2019deep}, following the same data processing methodology as the E. coli MIC dataset to ensure consistency and reliability.

\fielditem{Dataset Statistics} 
The dataset contains 1,531 datapoints with sequences ranging from 2 to 140 amino acids (average length 22.01) in length.

\textbf{Task: Regression; Split: Hybrid; Evaluation: MAE}

\begin{center} 
\centering
\includegraphics[width=0.8\textwidth]{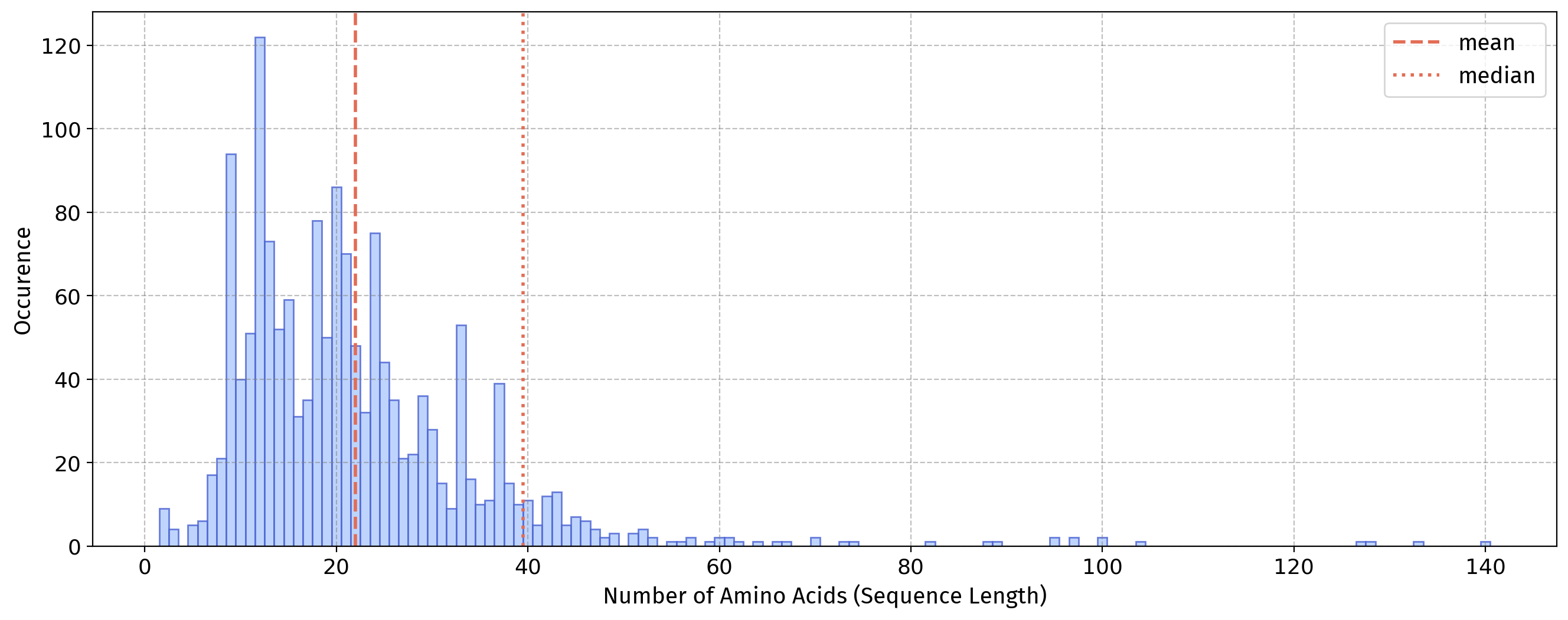}
   \captionof{figure}{Length distribution of P. aeruginosa MIC dataset.}
\label{fig:length_dist_P_aeruginosa_mic}
\end{center}

\begin{center} 
\centering
\includegraphics[width=0.8\textwidth]{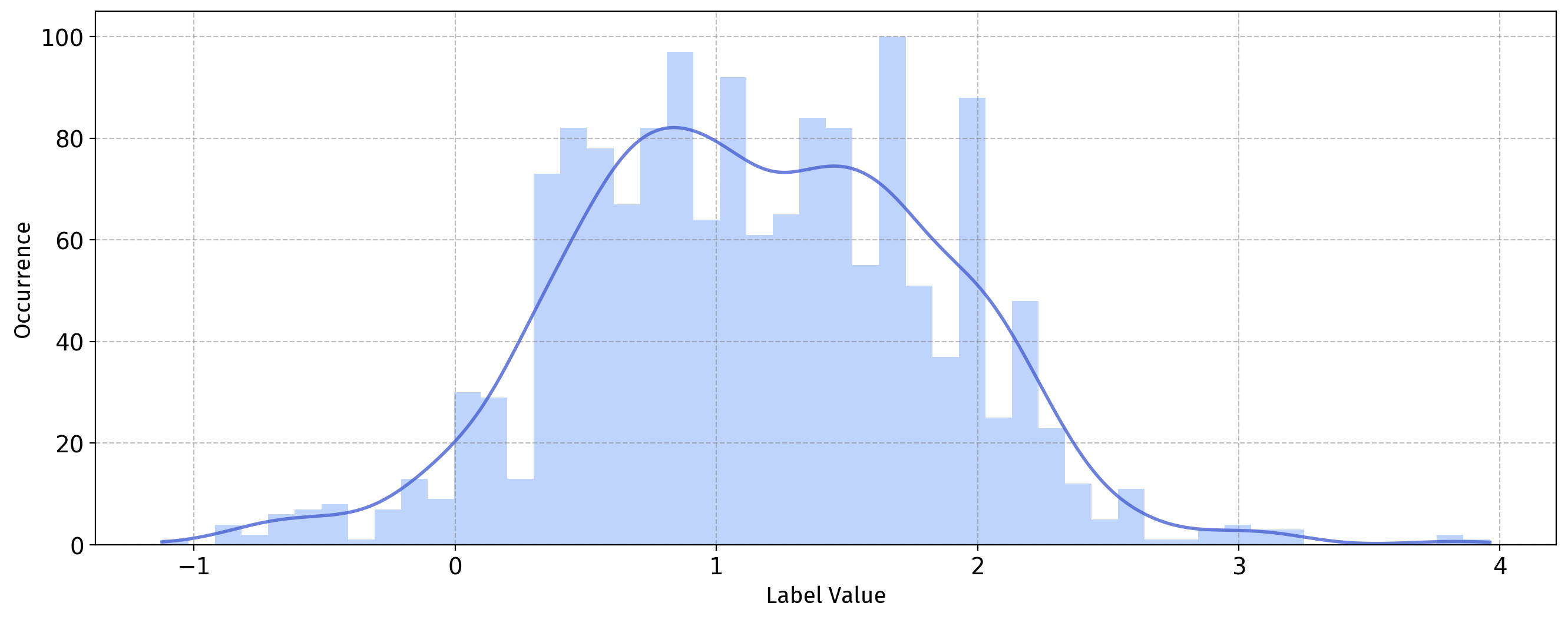}
   \captionof{figure}{Label distribution of P. aeruginosa MIC dataset.}
\label{fig:label_dist_P_aeruginosa_mic}
\end{center}

\begin{center} 
  \centering
  \includegraphics[width=0.8\textwidth]{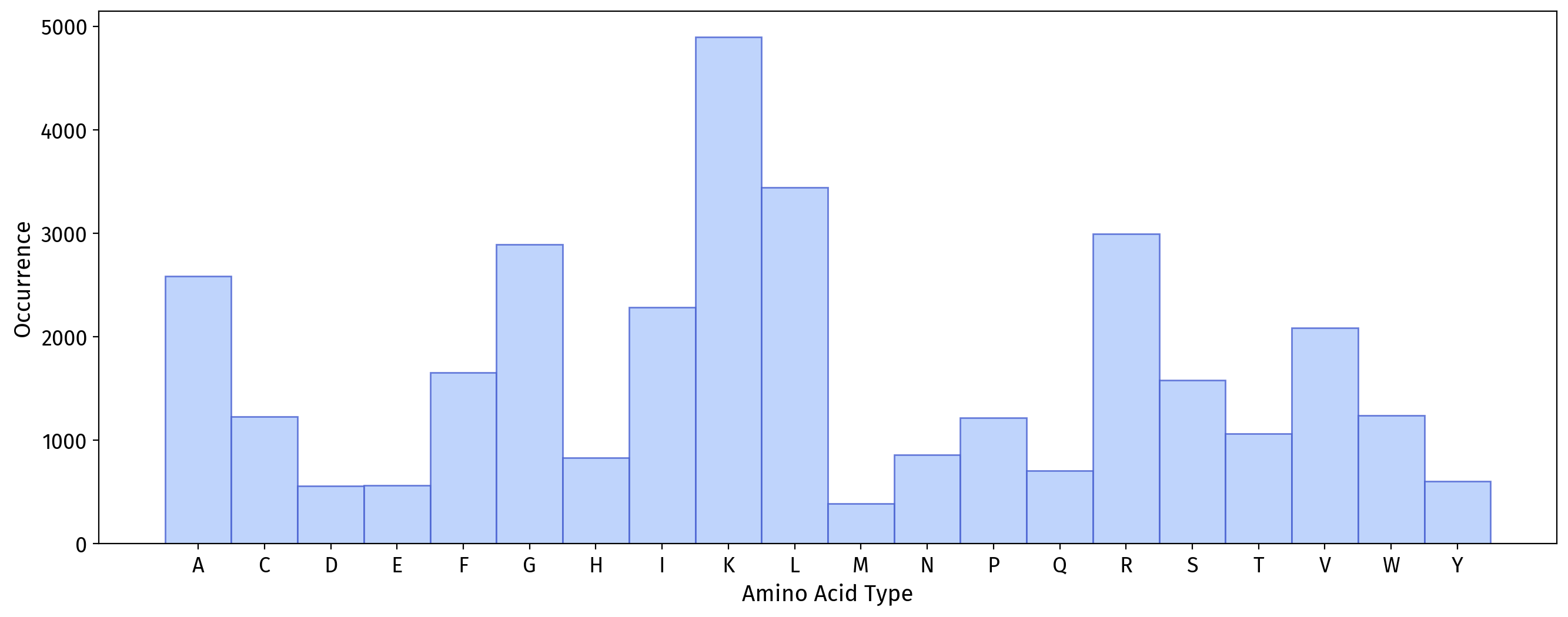}
     \captionof{figure}{Amino acid distribution of P. aeruginosa MIC dataset.}
  \label{fig:aa_dist_P_aeruginosa_mic}
  \end{center}

\subsubsection{S. aureus MIC}
\fielditem{Property and Application} 
This dataset provides quantitative measurements of peptide activity against \textit{Staphylococcus aureus}, a representative Gram-positive pathogen that causes various infections.

\fielditem{Data Source} 
The Dataset is sourced from GRAMPA~\citep{witten2019deep}, following the same data processing methodology as other MIC datasets to ensure consistency and reliability.

\fielditem{Dataset Statistics} 
The dataset contains 2,900 datapoints with sequences ranging from 2 to 140 amino acids (average length 22.70) in length.

\textbf{Task: Regression; Split: Hybrid; Evaluation: MAE}

\begin{center} 
\centering
\includegraphics[width=0.8\textwidth]{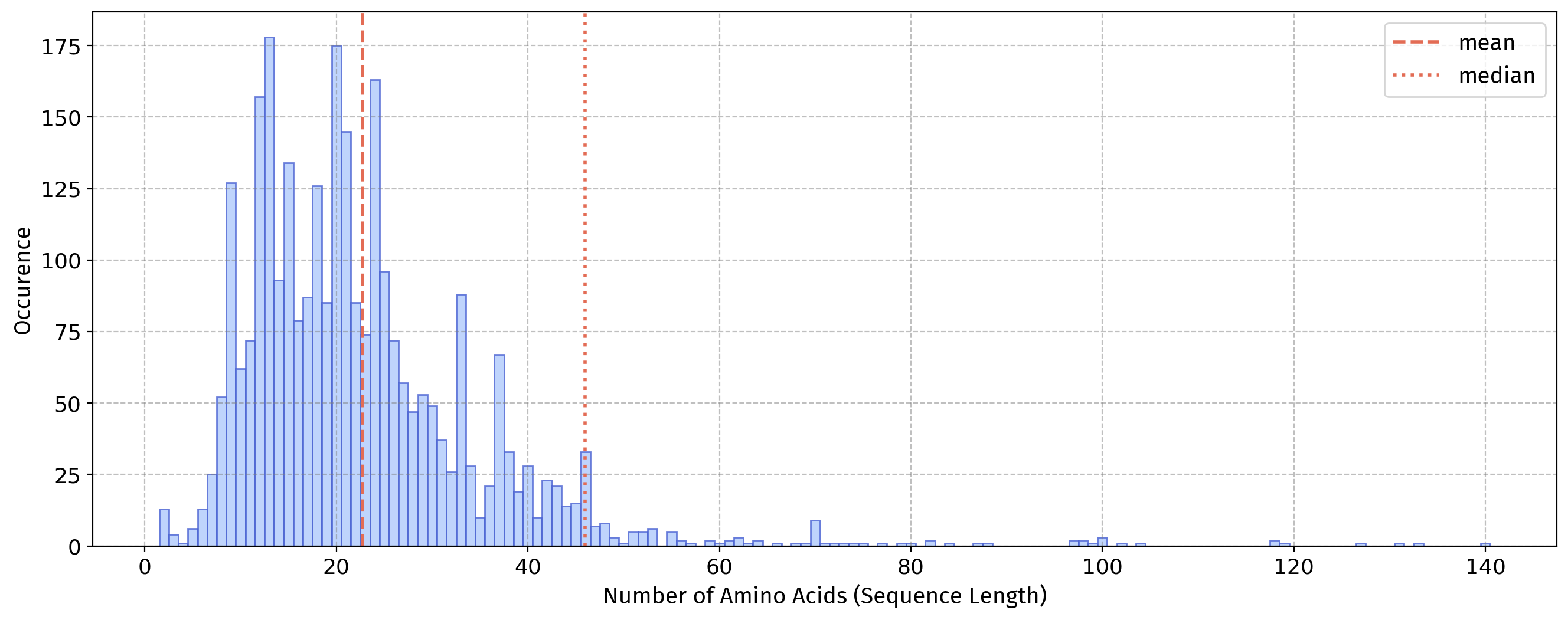}
   \captionof{figure}{Length distribution of S. aureus MIC dataset.}
\label{fig:length_dist_S_aureus_mic}
\end{center}

\begin{center} 
\centering
\includegraphics[width=0.8\textwidth]{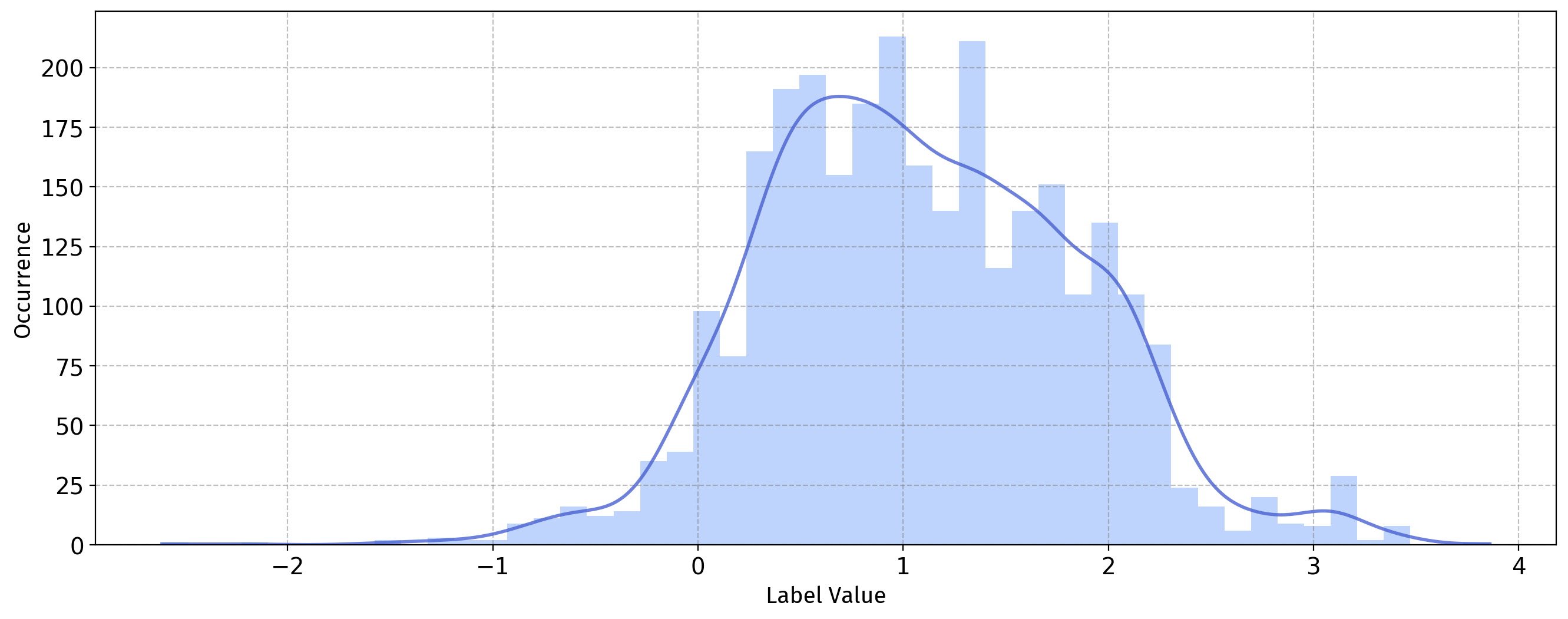}
   \captionof{figure}{Label distribution of S. aureus MIC dataset.}
\label{fig:label_dist_S_aureus_mic}
\end{center}

\begin{center} 
  \centering
  \includegraphics[width=0.8\textwidth]{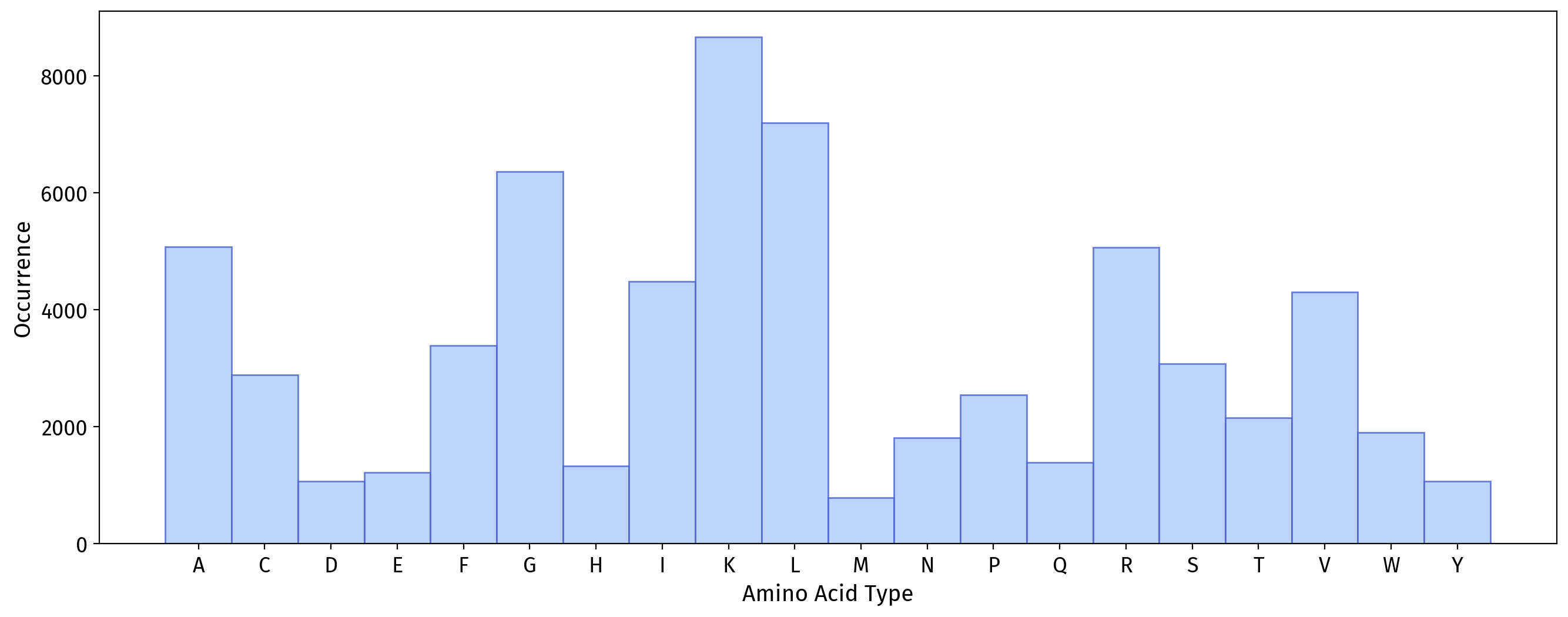}
     \captionof{figure}{Amino acid distribution of S. aureus MIC dataset.}
  \label{fig:aa_dist_S_aureus_mic}
  \end{center}

\subsubsection{Non-canonical Antimicrobial (nc-antimicrobial)}
\fielditem{Property and Application} 
The nc-antimicrobial dataset includes non-canonical peptides with broad-spectrum antimicrobial activity. 
These peptides represent the potential of synthetic modifications to enhance antimicrobial properties.

\fielditem{Data Source} 
The Dataset is sourced from Hemolytik 2.0~\citep{singh2025hemolytik2}, which contains manually collected data 
from published literature and various databases. 
Hemolytik 2.0 is subset to only include sequences that have antimicrobial records. Sequences in Modified Amino acid Peptide (MAP) format stored in the original database are converted into BILN format~\citep{fox2022biln} supported by this project.

\fielditem{Dataset Statistics} 
The dataset contains 2,496 datapoints with sequences ranging from 4 to 130 amino acids (average length 17.79) in length.

\textbf{Task: Classification; Split: ECFP-based; Evaluation: ROC-AUC}

\begin{center} 
\centering
\includegraphics[width=0.8\textwidth]{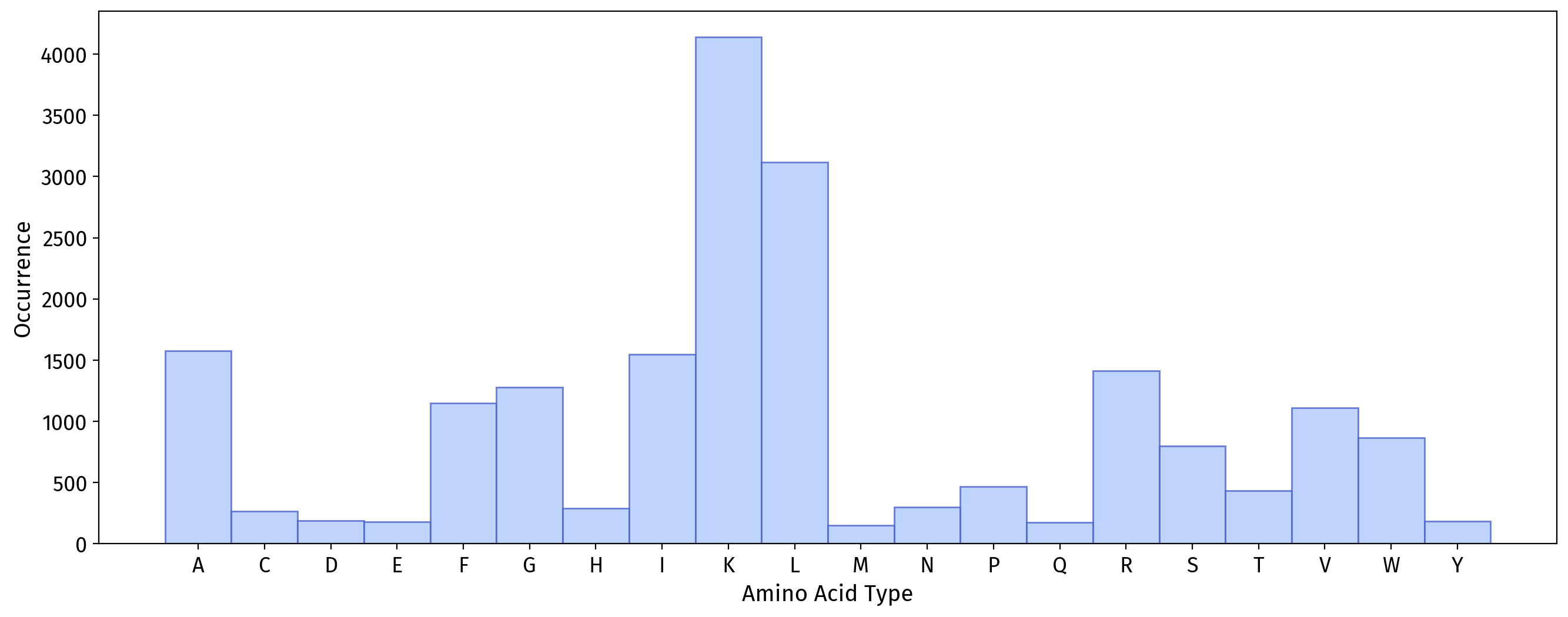}
   \captionof{figure}{Canonical Amino acid distribution comparison between positive and negative samples for nc-antimicrobial dataset.}
\label{fig:aa_dist_nc-antimicrobial}
\end{center}

\begin{center} 
\centering
\includegraphics[width=0.8\textwidth]{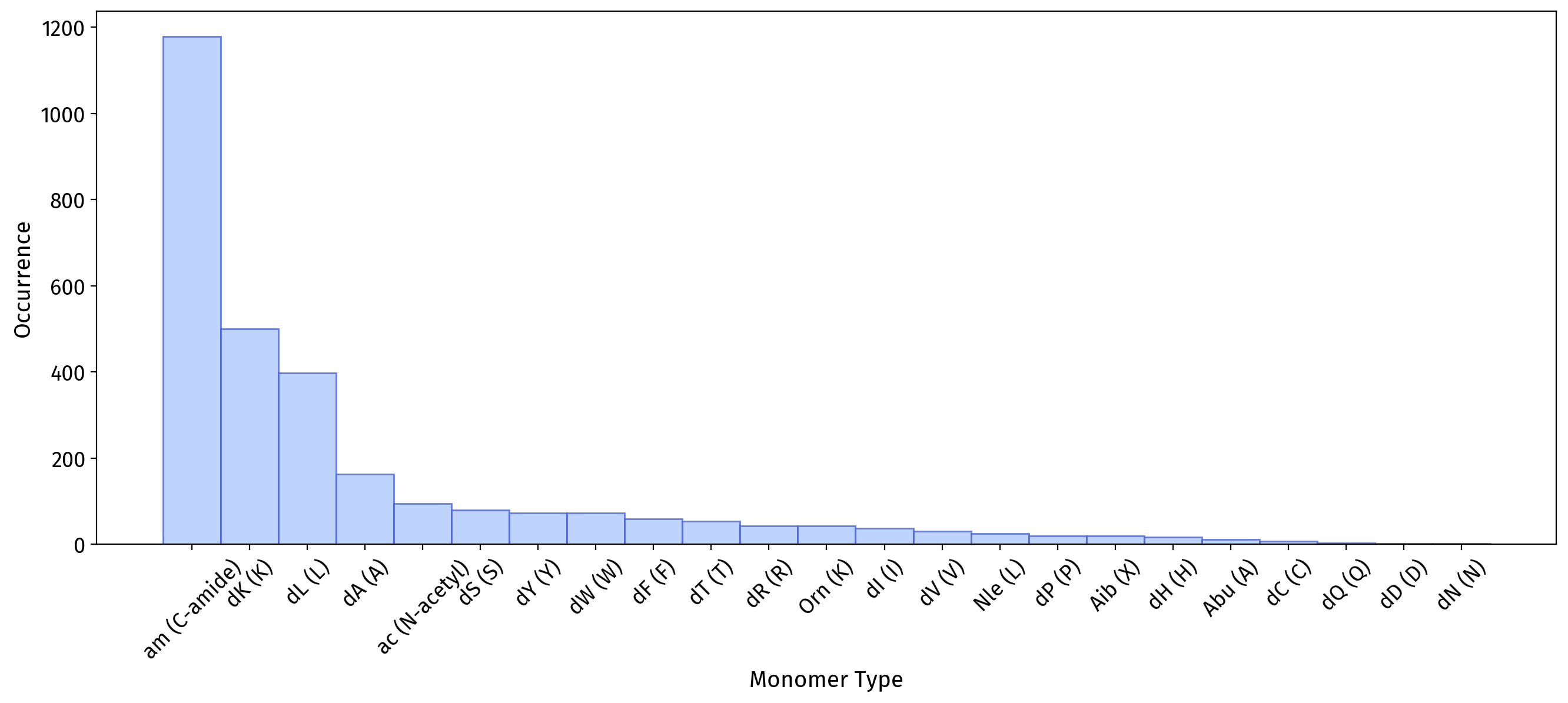}
   \captionof{figure}{non-canonical Amino acid distribution comparison between positive and negative samples for nc-antimicrobial dataset.}
\label{fig:ncaa_dist_nc-antimicrobial}
\end{center}

\begin{center} 
\centering
\includegraphics[width=0.8\textwidth]{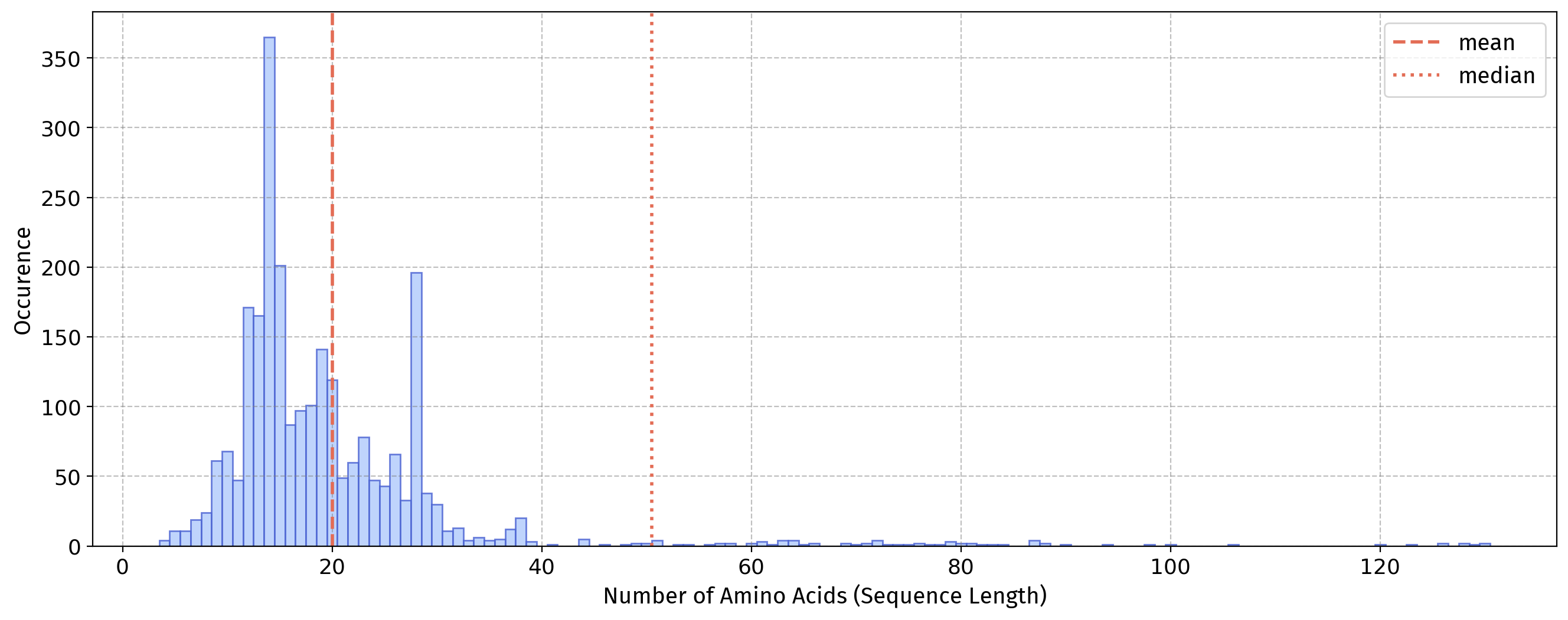}
   \captionof{figure}{Length distribution of nc-antimicrobial dataset.}
\label{fig:length_dist_nc-antimicrobial}
\end{center}

\subsubsection{Non-canonical Antibacterial (nc-antibacterial)}
\fielditem{Property and Application} 
The nc-antibacterial dataset includes non-canonical peptides with activity against bacterial pathogens. 
This dataset enables evaluation of how synthetic modifications affect antibacterial activity.

\fielditem{Data Source} 
The Dataset is sourced from Hemolytik 2.0~\citep{singh2025hemolytik2}. 
Hemolytik 2.0 is subset to only include sequences that have antibacterial records. 
Sequences in MAP format stored in the original database are converted into BILN format supported by this project.

\fielditem{Dataset Statistics} 
The dataset contains 1,666 datapoints with sequences ranging from 4 to 76 amino acids (average length 16.82) in length.

\textbf{Task: Classification; Split: ECFP-based; Evaluation: ROC-AUC}

\begin{center} 
\centering
\includegraphics[width=0.8\textwidth]{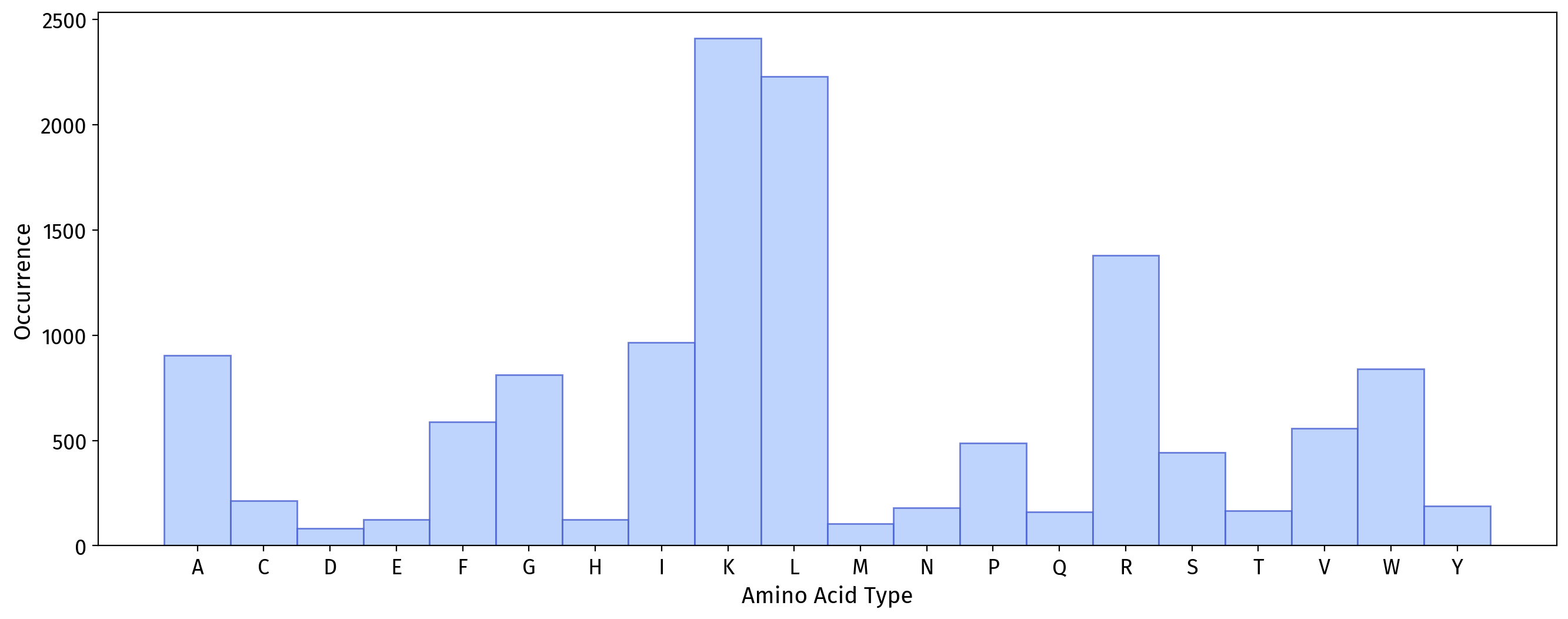}
   \captionof{figure}{Canonical Amino acid distribution comparison between positive and negative samples for nc-antibacterial dataset.}
\label{fig:aa_dist_nc-antibacterial}
\end{center}

\begin{center} 
\centering
\includegraphics[width=0.8\textwidth]{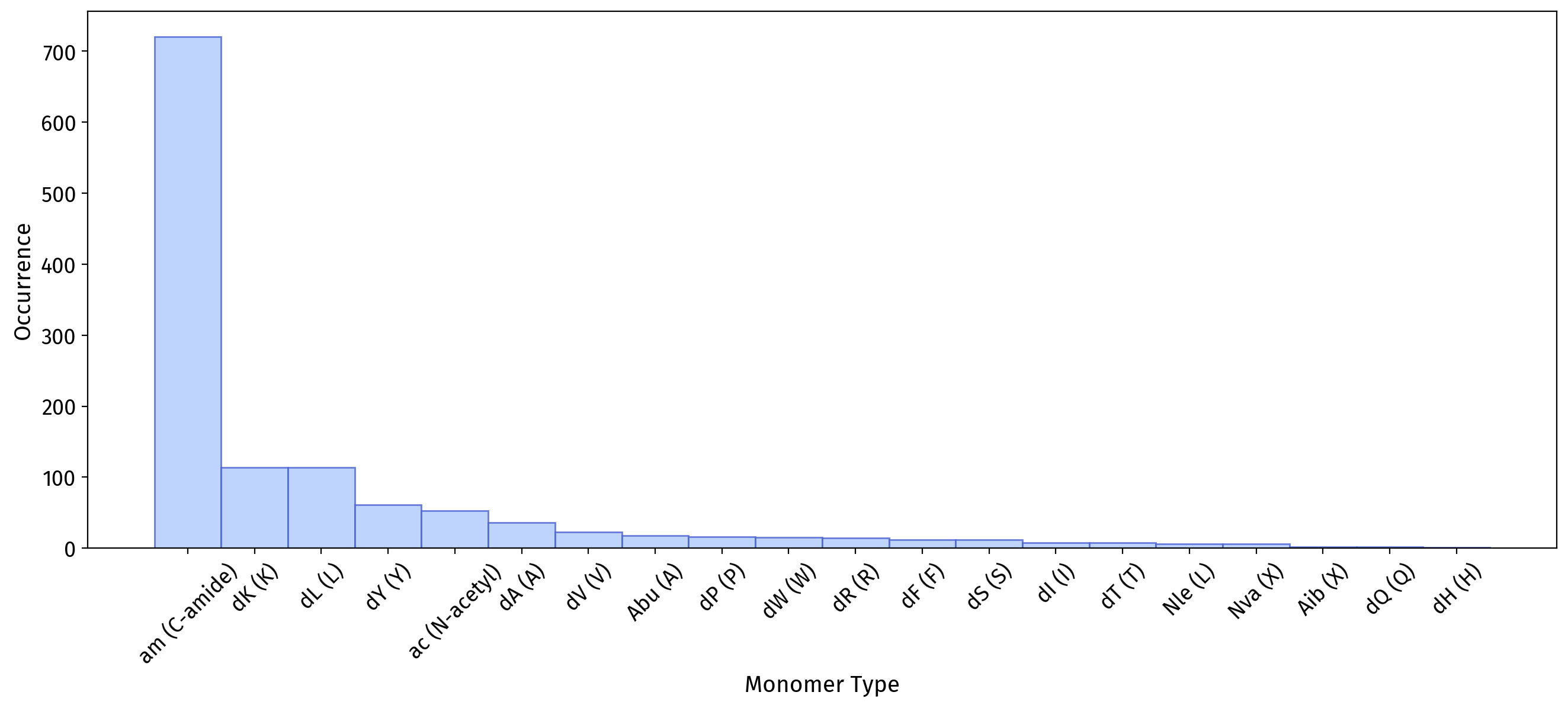}
   \captionof{figure}{non-canonical Amino acid distribution comparison between positive and negative samples for nc-antibacterial dataset.}
\label{fig:ncaa_dist_nc-antibacterial}
\end{center}

\begin{center} 
\centering
\includegraphics[width=0.8\textwidth]{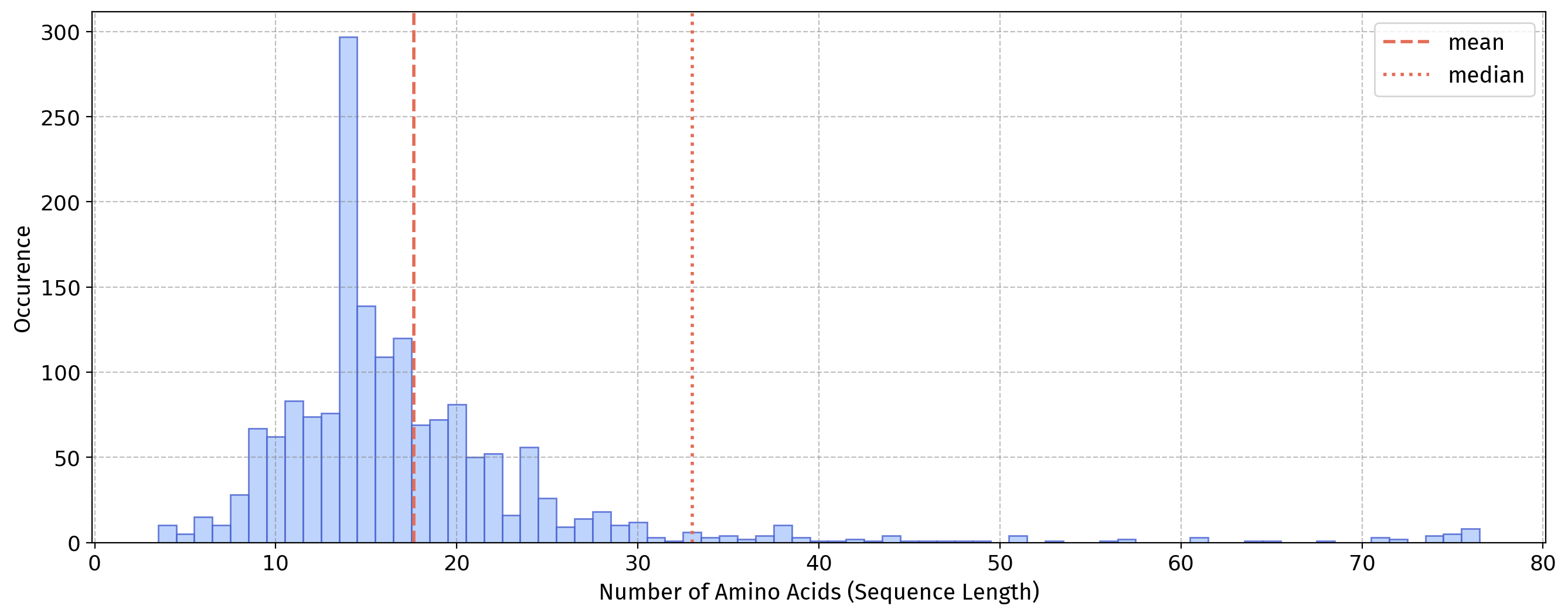}
   \captionof{figure}{Length distribution of nc-antibacterial dataset.}
\label{fig:length_dist_nc-antibacterial}
\end{center}

\subsubsection{Non-canonical Antifungal (nc-antifungal)}
\fielditem{Property and Application} 
The nc-antifungal dataset collects non-canonical peptides targeting fungal pathogens. This dataset enables evaluation of synthetic modifications for antifungal activity.

\fielditem{Data Source} 
The Dataset is sourced from Hemolytik 2.0~\citep{singh2025hemolytik2}. Hemolytik 2.0 is subset to only include sequences that have antifungal records. Sequences in MAP format stored in the original database are converted into BILN format supported by this project.

\fielditem{Dataset Statistics} 
The dataset contains 410 datapoints with sequences ranging from 4 to 76 amino acids (average length 18.18) in length.

\textbf{Task: Classification; Split: ECFP-based; Evaluation: ROC-AUC}

\begin{center} 
\centering
\includegraphics[width=0.8\textwidth]{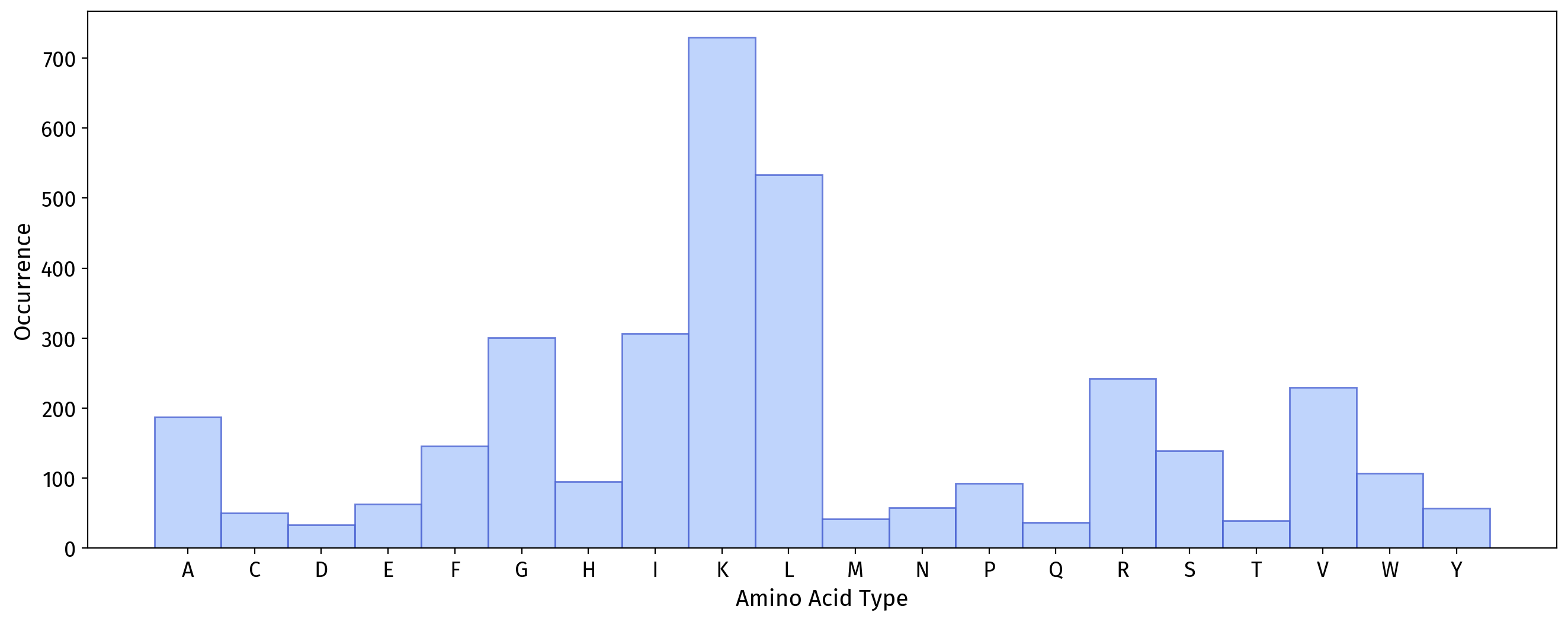}
   \captionof{figure}{Canonical Amino acid distribution comparison between positive and negative samples for nc-antifungal dataset.}
\label{fig:aa_dist_nc-antifungal}
\end{center}

\begin{center} 
\centering
\includegraphics[width=0.8\textwidth]{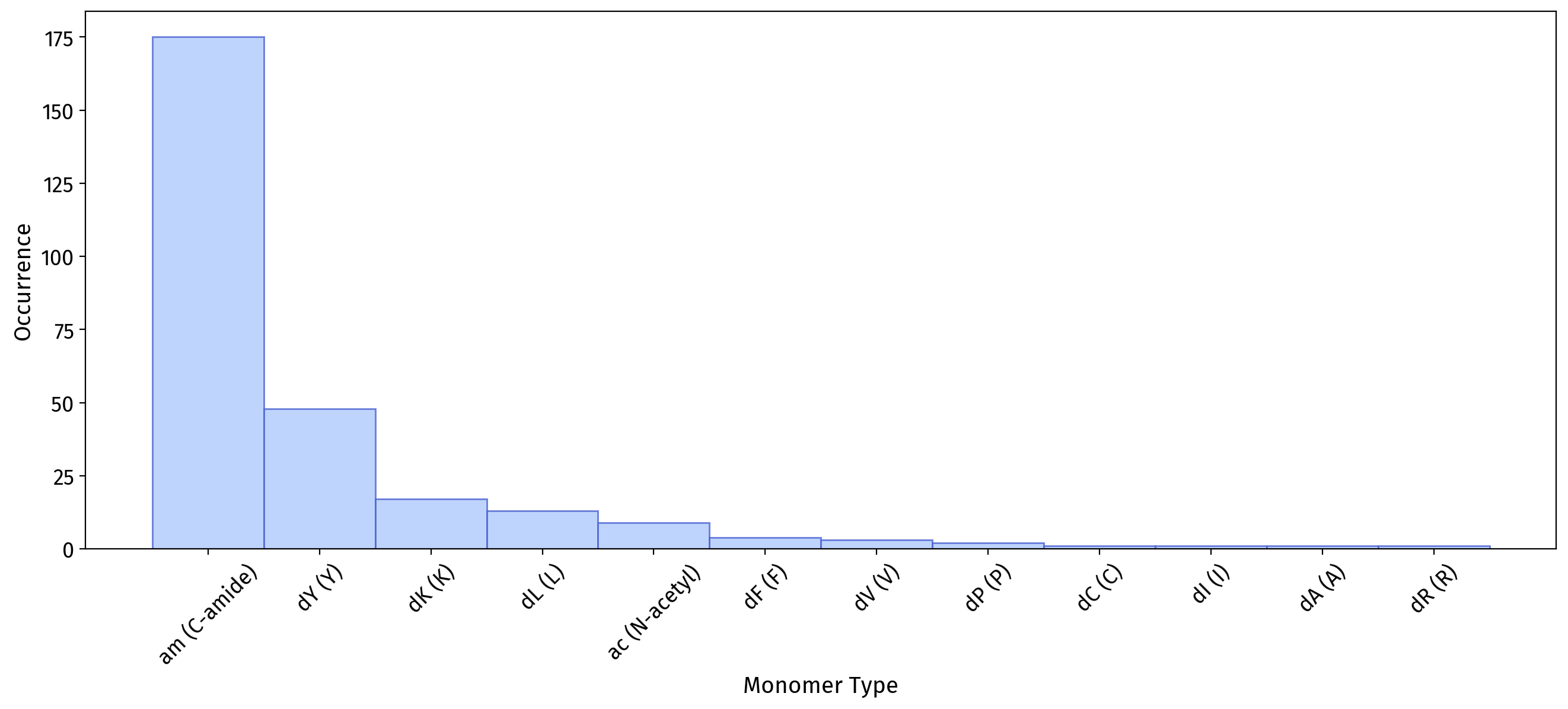}
   \captionof{figure}{non-canonical Amino acid distribution comparison between positive and negative samples for nc-antifungal dataset.}
\label{fig:ncaa_dist_nc-antifungal}
\end{center}

\begin{center} 
\centering
\includegraphics[width=0.8\textwidth]{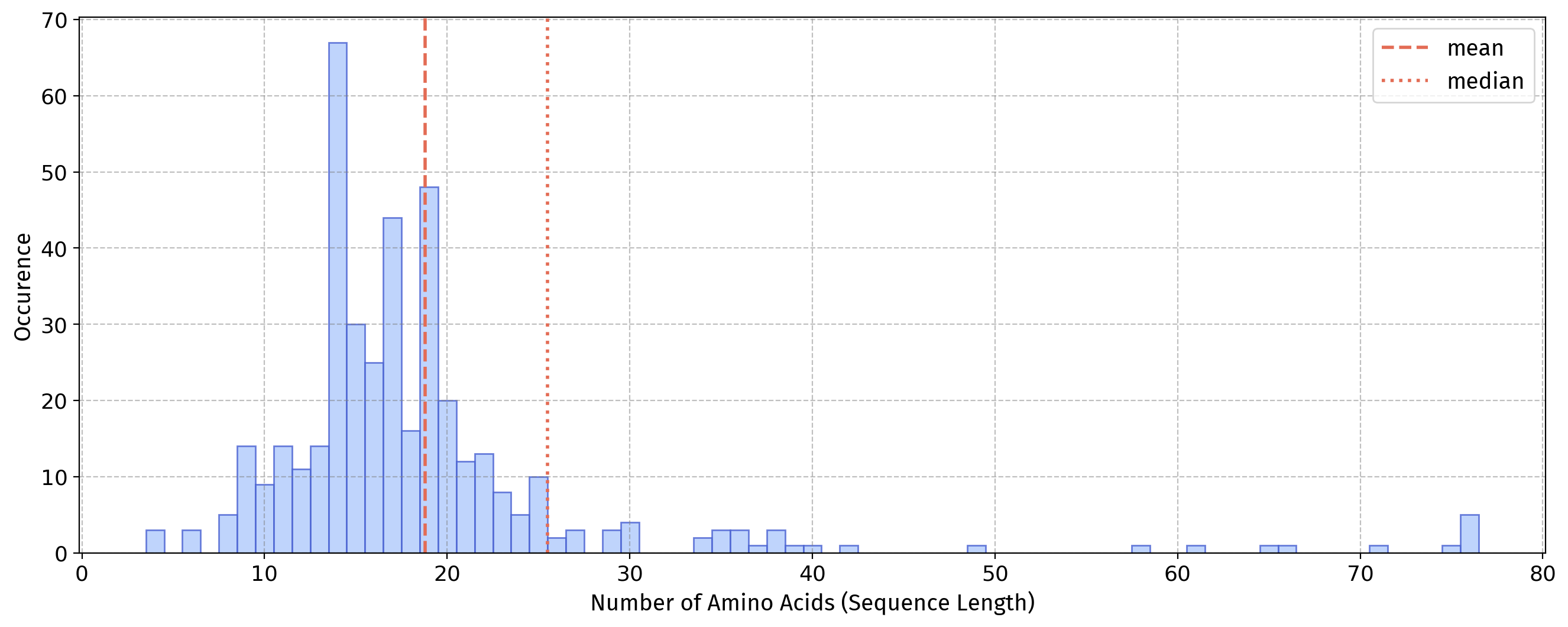}
   \captionof{figure}{Length distribution of nc-antifungal dataset.}
\label{fig:length_dist_nc-antifungal}
\end{center}

\subsection{Oncology}

\begin{datasetbox}
\paragraph{Definition.} 
Datasets in this group include peptides with experimentally validated anticancer activity. The mechanisms of the activity include cytotoxic effects and immune regulatory functions.
\paragraph{Impact.} 
Peptides offer unique advantages in cancer therapy due to their high target specificity, low off-target toxicity, and potential to modulate the immune system. These features make them promising candidates for both direct anticancer treatments and peptide-based vaccines, addressing limitations of conventional chemotherapies.
\paragraph{Pipelines.} Activity Modeling
\end{datasetbox}

\subsubsection{anticancer}
\fielditem{Property and Application} 
The anticancer dataset includes peptides that exert anticancer activity by targeting tumor cells through mechanisms such as membrane disruption, interference with specific intracellular processes, or induction of apoptosis and programmed cell death. Anticancer Peptides (ACPs) represent promising therapeutic agents for cancer treatment.

\fielditem{Data Source} 
The dataset originates from AntiCP 2.0~\citep{Agrawal2021AntiCP}.The original literature obtained experimentally validated anticancer peptides from datasets of previous studies including ACP-DL~\citep{yi2019acp}, ACPP~\citep{vijayakumar2015acpp}, ACPred-FL~\citep{wei2018acpred}, and iACP~\citep{aziz2022iacp}. In addition, data are also extracted from the ACP database CancerPPD~\citep{tyagi2015cancerppd}. After removing small, long, identical and non-natural peptides, 970 unique ACPs having 4 or more residues and 50 or fewer residues were obtained.
Additionally, 8161 ACPs from Peptipedia are merged.

\fielditem{Dataset Statistics} 
The dataset contains 13,852 datapoints with sequences ranging from 2 to 145 amino acids (average length 27.83) in length.

\textbf{Task: Classification; Split: Hybrid; Evaluation: ROC-AUC}

\begin{center} 
\centering
\includegraphics[width=0.8\textwidth]{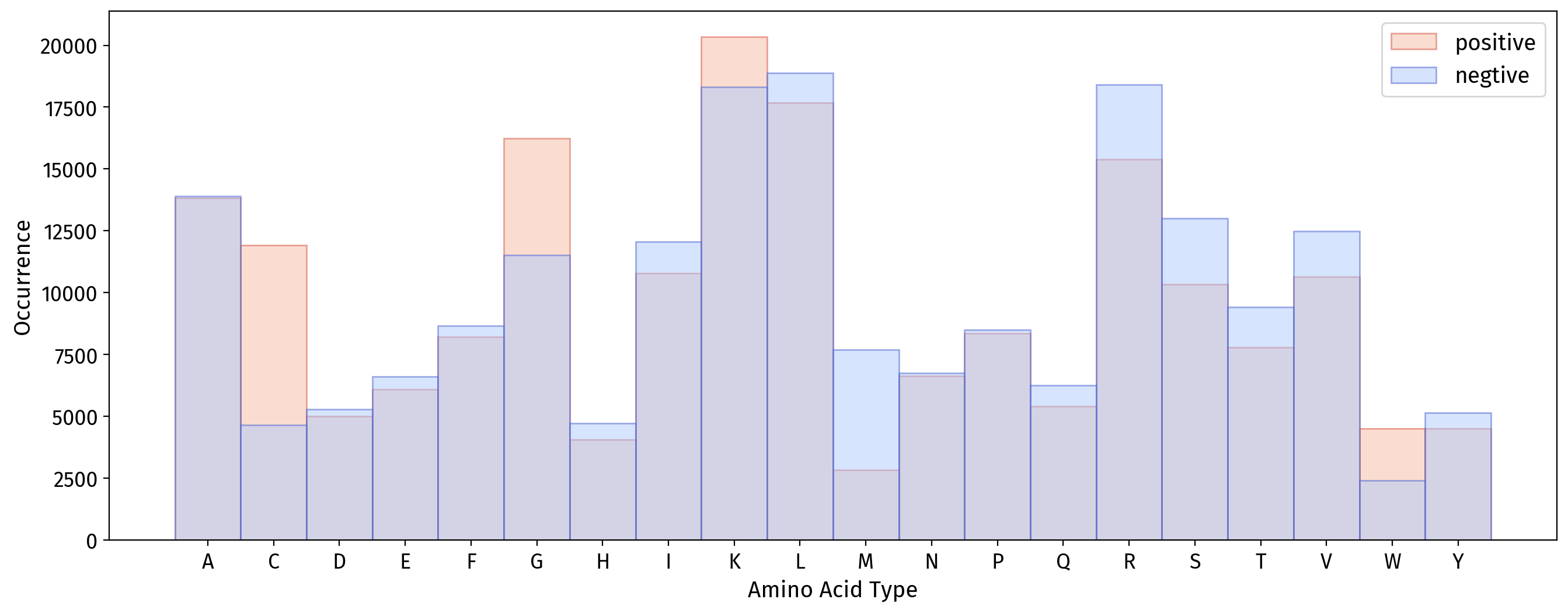}
   \captionof{figure}{Amino acid distribution comparison between positive and negative samples for anticancer dataset.}
\label{fig:aa_dist_anticancer}
\end{center}

\begin{center} 
\centering
\includegraphics[width=0.8\textwidth]{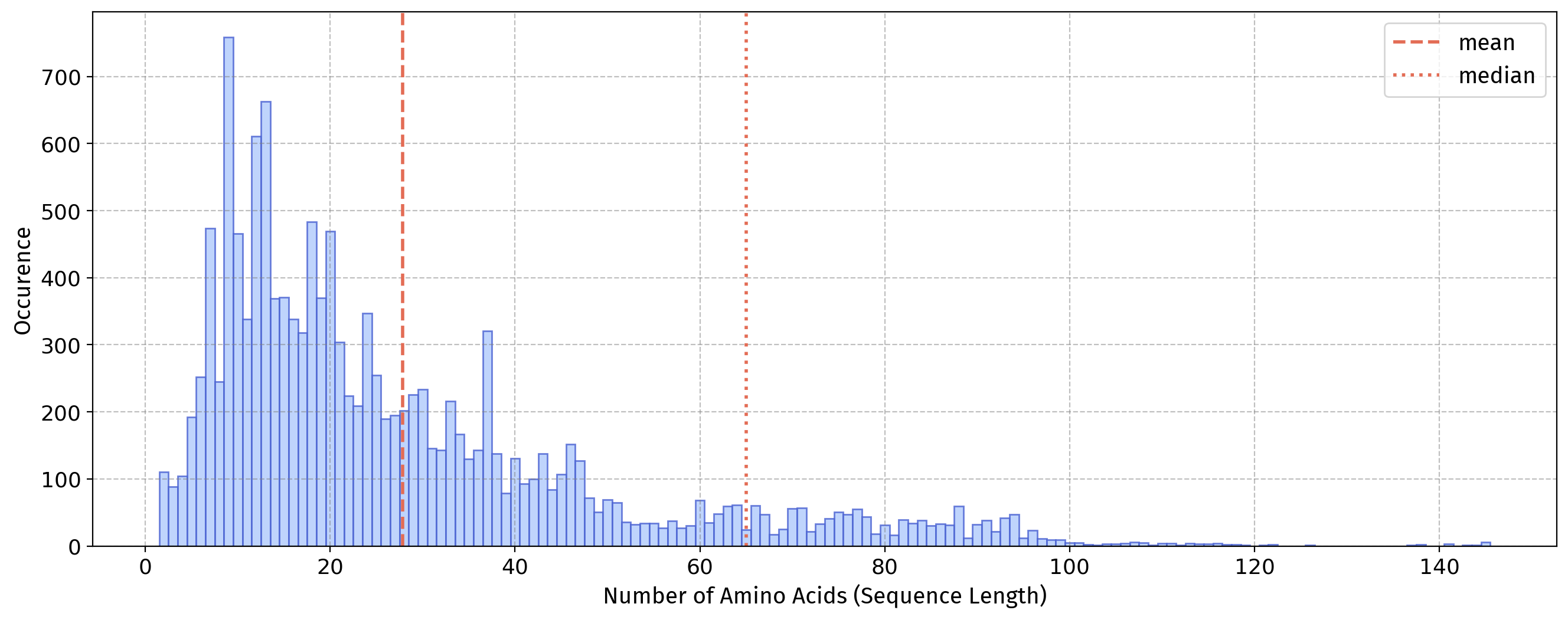}
   \captionof{figure}{Length distribution of anticancer dataset.}
\label{fig:length_dist_anticancer}
\end{center}

\begin{center} 
\centering
\includegraphics[width=0.8\textwidth]{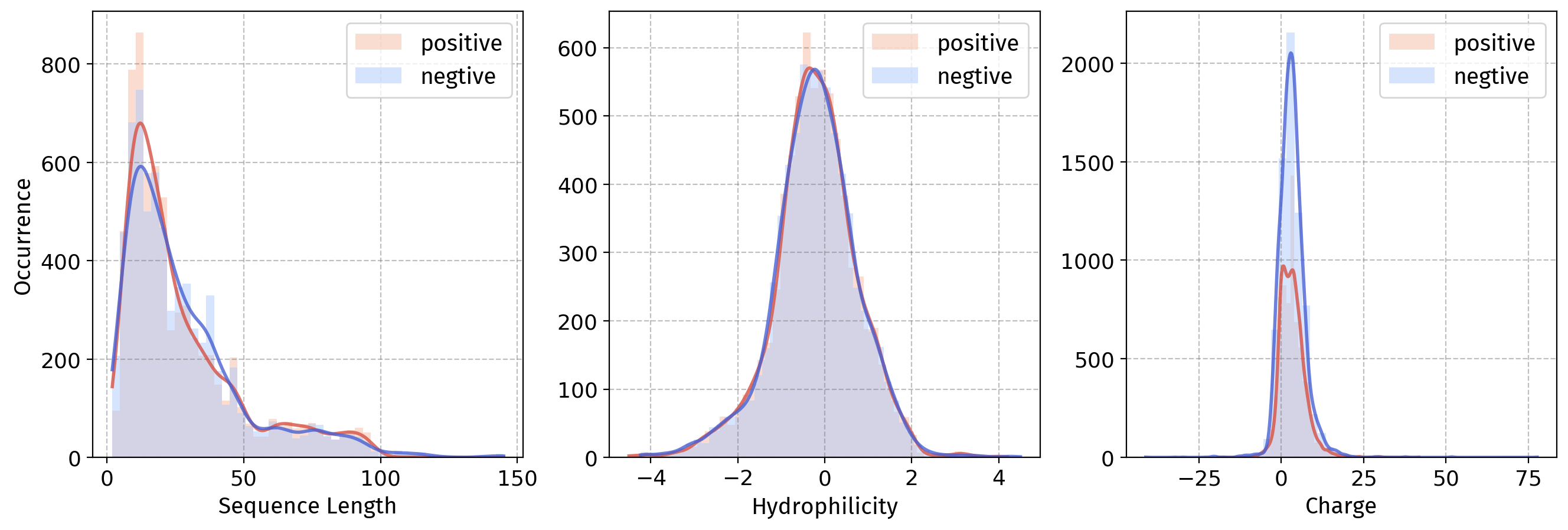}
   \captionof{figure}{Property comparison between positive and negative samples for anticancer dataset.}
\label{fig:property_comp_anticancer}
\end{center}

\subsubsection{Tumor T-cell Antigens (ttca)}
\fielditem{Property and Application} 
The ttca dataset comprises tumor T-cell antigen (ttca) peptides, which are capable of stimulating antitumor immune responses. These peptides function by presenting tumor-associated epitopes to T cells, thereby activating cellular immunity against cancer cells.

\fielditem{Data Source} 
The dataset originates from ~\cite{charoenkwan2020ittca}. The literature constructs a benchmark dataset, where ttca peptides are obtained from TANTIGEN~\citep{olsen2017tantigen} and TANTIGEN 2.0~\citep{zhang2021tantigen}, whereby a total of 529 unique MHC class I peptides are collected and considered as positive samples.

\fielditem{Dataset Statistics} 
The dataset contains 1,182 datapoints with sequences ranging from 8 to 20 amino acids (average length 9.36) in length.

\textbf{Task: Classification; Split: Hybrid; Evaluation: ROC-AUC}

\begin{center} 
\centering
\includegraphics[width=0.8\textwidth]{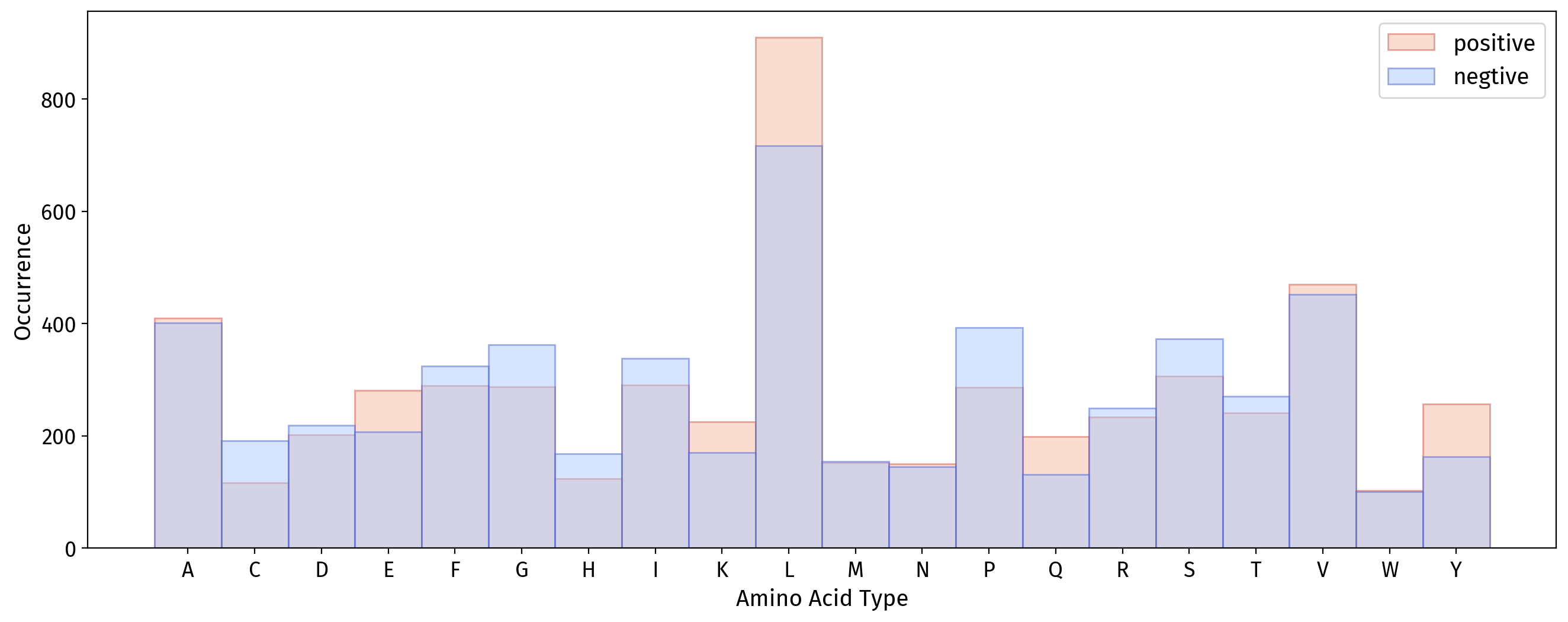}
   \captionof{figure}{Amino acid distribution comparison between positive and negative samples for tumor T-cell antigen dataset.}
\label{fig:aa_dist_ttca}
\end{center}

\begin{center} 
\centering
\includegraphics[width=0.8\textwidth]{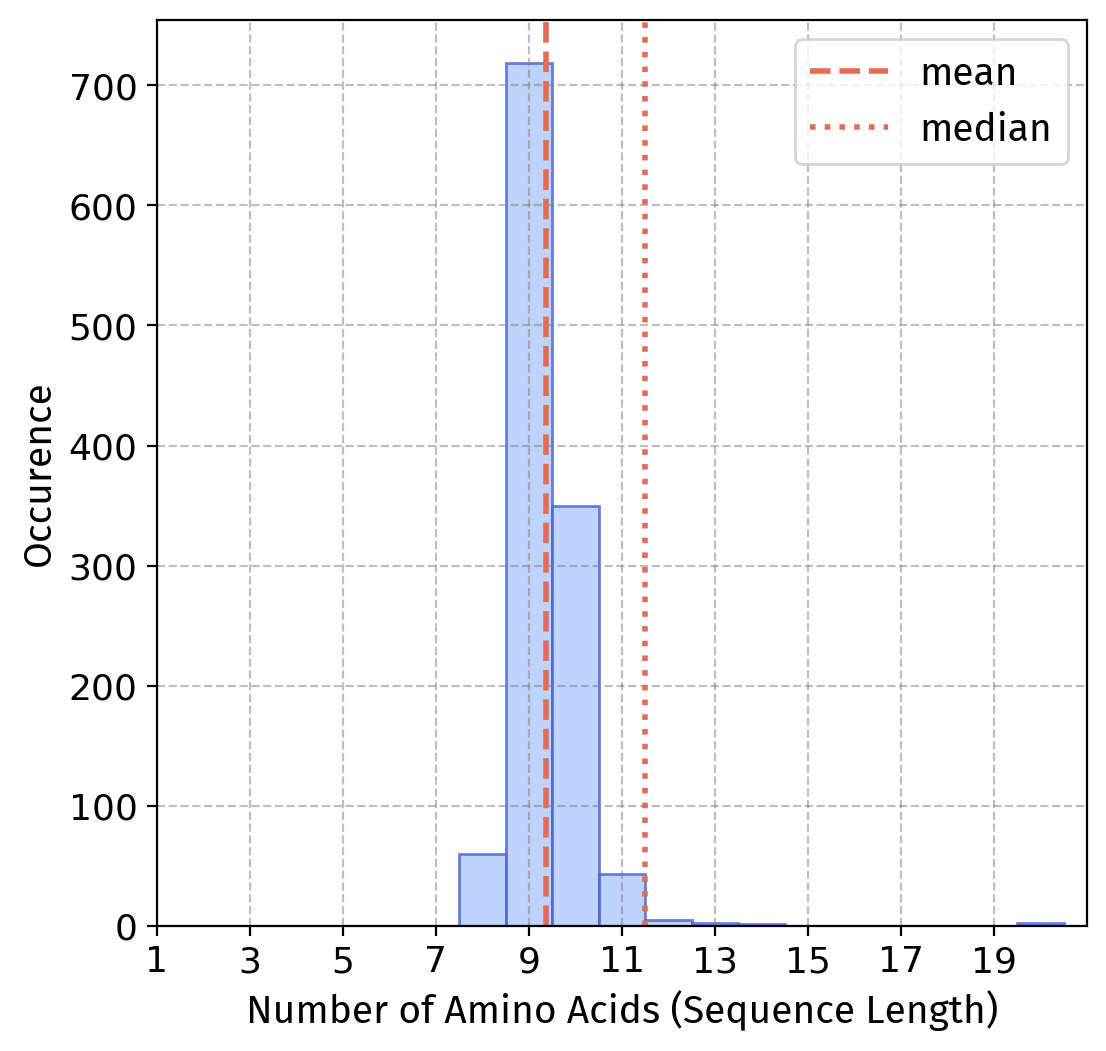}
   \captionof{figure}{Length distribution of tumor T-cell antigen dataset.}
\label{fig:length_dist_ttca}
\end{center}

\begin{center} 
\centering
\includegraphics[width=0.8\textwidth]{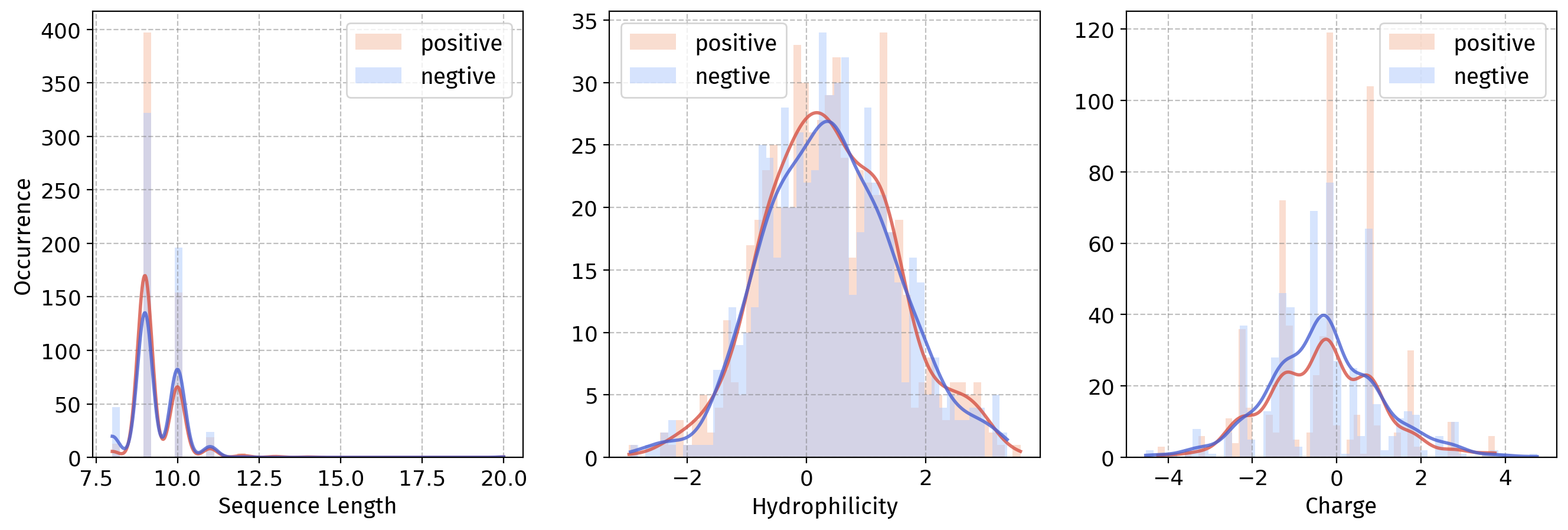}
   \captionof{figure}{Property comparison between positive and negative samples for tumor T-cell antigen dataset.}
\label{fig:property_comp_ttca}
\end{center}

\subsection{Metabolic Group}

\begin{datasetbox}
\paragraph{Definition.} 
Datasets in this group comprise peptides with therapeutic potential for metabolic disorders, including diabetes and hypertension. These peptides act via mechanisms such as enzyme inhibition (e.g., DPP-IV or ACE inhibitors) or modulation of hormone-like signaling pathways.
\paragraph{Impact.}
Peptides are particularly well-suited for the long-term management of metabolic diseases, as they possess high target specificity, favorable biocompatibility, and reduced off-target effects, which can help minimize cumulative toxicity and improve patient adherence over prolonged treatment periods.
\paragraph{Pipeline.} Activity Modeling
\end{datasetbox}

\subsubsection{ace inhibitory}
\fielditem{Property and Application} 
The ace inhibitory dataset comprises peptides that inhibit Angiotensin-Converting Enzyme (ACE), which normally converts angiotensin I into angiotensin II, a potent vasoconstrictor. By preventing this conversion, ACE inhibitory peptides help lower blood pressure and reduce cardiovascular risk.

\fielditem{Data Source} 
The dataset originates from ~\cite{manavalan2019mahtpred}, which extracts ACE inhibitory peptides from literature and publicly available databases like AHTPDB~\citep{kumar2015ahtpdb}, BIOPEP~\citep{minkiewicz2008biopep}, and PDB~\citep{kumar2015ahtpdb}. The original literature provides inhibitory activity (IC50) for dipeptides and tripeptides, while peptides with more than 3 residues are only provided classification labels. Since classification labels of longer peptide are collected from different literature sources, making it difficult to unify the positive/negative sample division threshold for dipeptides and tripeptides. 
Therefore, peptides with lengths < 5 amino acid residues are excluded from the classification task, leaving 1,053 peptides. Additionally, 780 ACE inhibitory peptides from Peptipedia are merged.

\fielditem{Dataset Statistics} 
The dataset contains 3,560 datapoints with sequences ranging from 4 to 81 amino acids (average length 8.08) in length.

\textbf{Task: Classification; Split: Hybrid; Evaluation: ROC-AUC}

\begin{center} 
\centering
\includegraphics[width=0.8\textwidth]{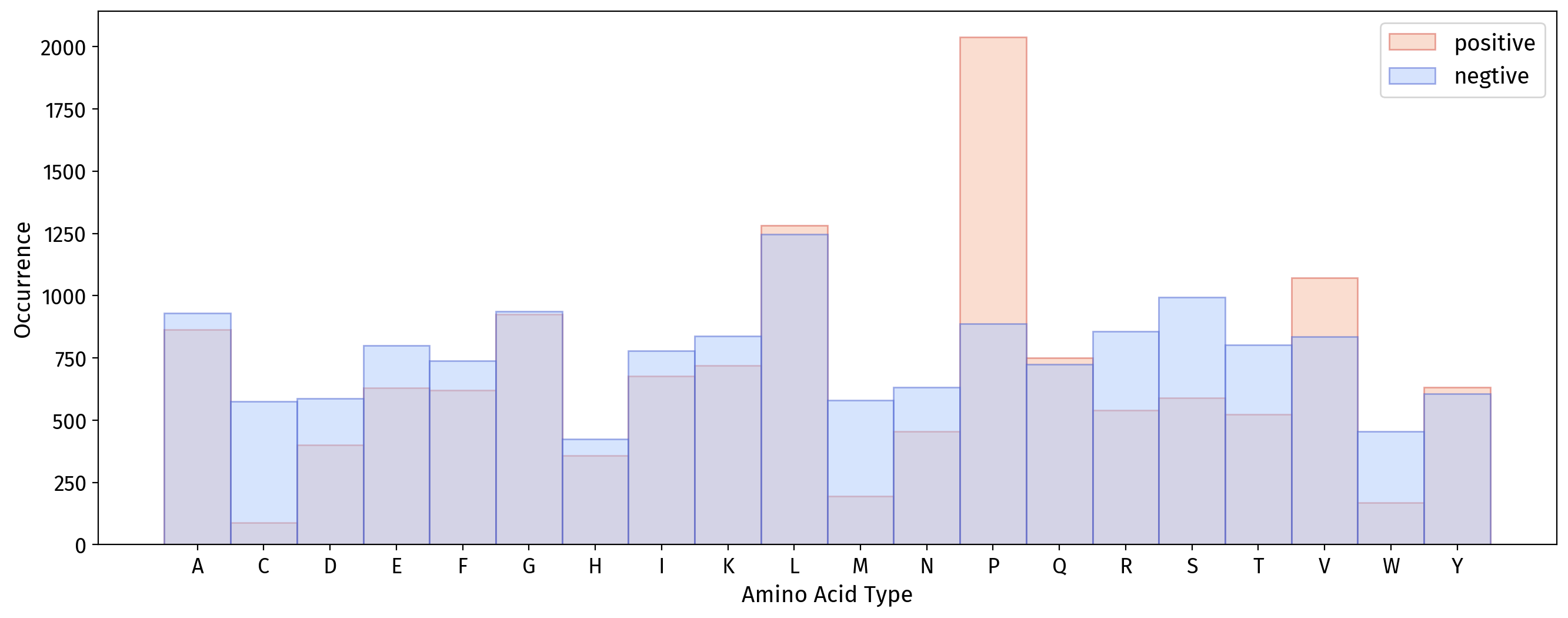}
   \captionof{figure}{Amino acid distribution comparison between positive and negative samples for ace inhibitory dataset.}
\label{fig:aa_dist_ace_inhibitory}
\end{center}

\begin{center} 
\centering
\includegraphics[width=0.8\textwidth]{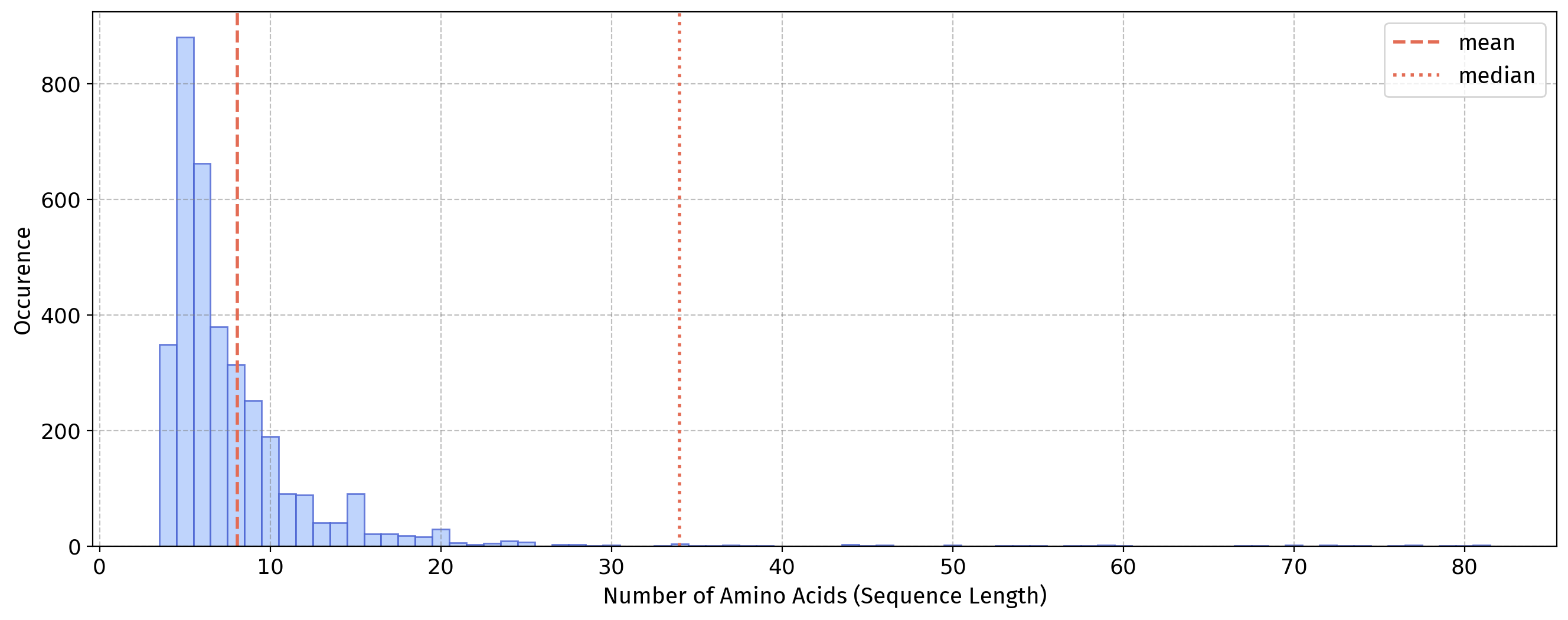}
   \captionof{figure}{Length distribution of ace inhibitory dataset.}
\label{fig:length_dist_ace_inhibitory}
\end{center}

\begin{center} 
\centering
\includegraphics[width=0.8\textwidth]{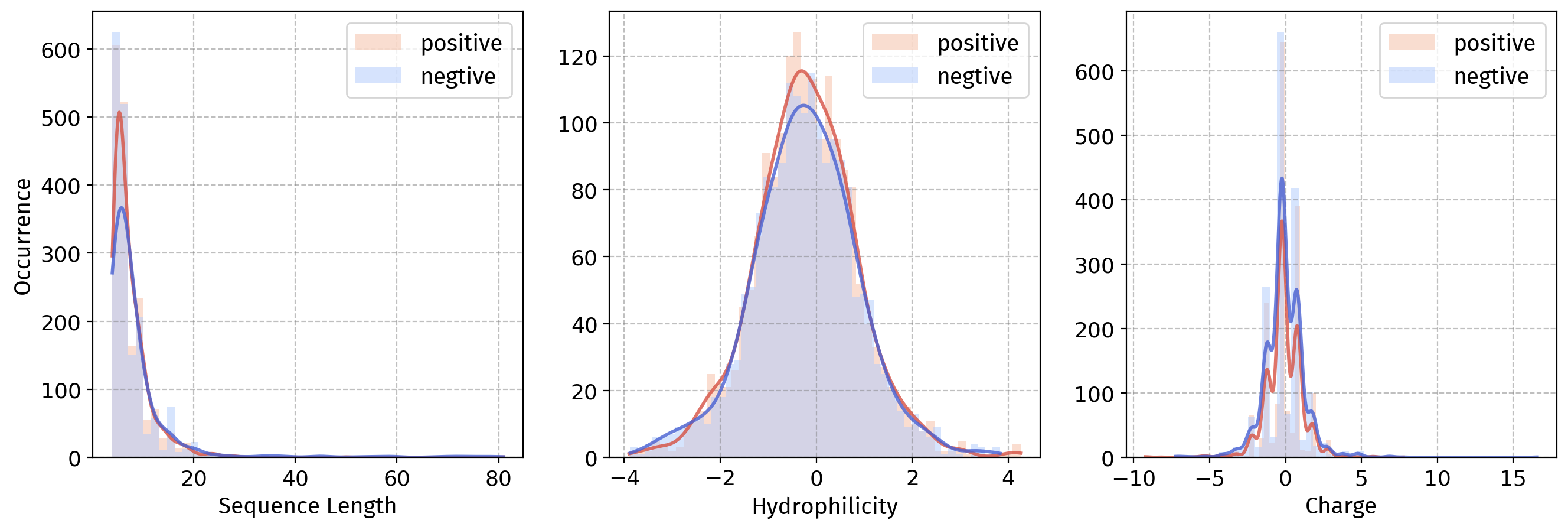}
   \captionof{figure}{Property comparison between positive and negative samples for ace inhibitory dataset.}
\label{fig:property_comp_ace_inhibitory}
\end{center}

\subsubsection{ace inhibitory ic50}
\fielditem{Property and Application} 
This dataset comprises dipeptides and tripeptides with ACE inhibitory ability. Developing predictive models specifically for these short peptides can help identify structural motifs that are most relevant for ACE inhibition, providing valuable leads for peptide-based drug development.

\fielditem{Data Source} 
The dataset originates from ~\cite{manavalan2019mahtpred}. The dataset contains 131 dipeptides having inhibitory activity (IC50) between 0.92 to 17,000 $\mu$M; 205 tripeptides having IC50 between 0.04 to 2,700 $\mu$M. IC50 values were converted into normalized pIC50 values {=-log ($\mu$M)} to narrow down the scale.

\fielditem{Dataset Statistics} 
The dataset contains 337 datapoints with sequences ranging from 2 to 3 amino acids (average length 2.61) in length.

\textbf{Task: Regression; Split: Hybrid; Evaluation: MAE}

\begin{center} 
\centering
\includegraphics[width=0.8\textwidth]{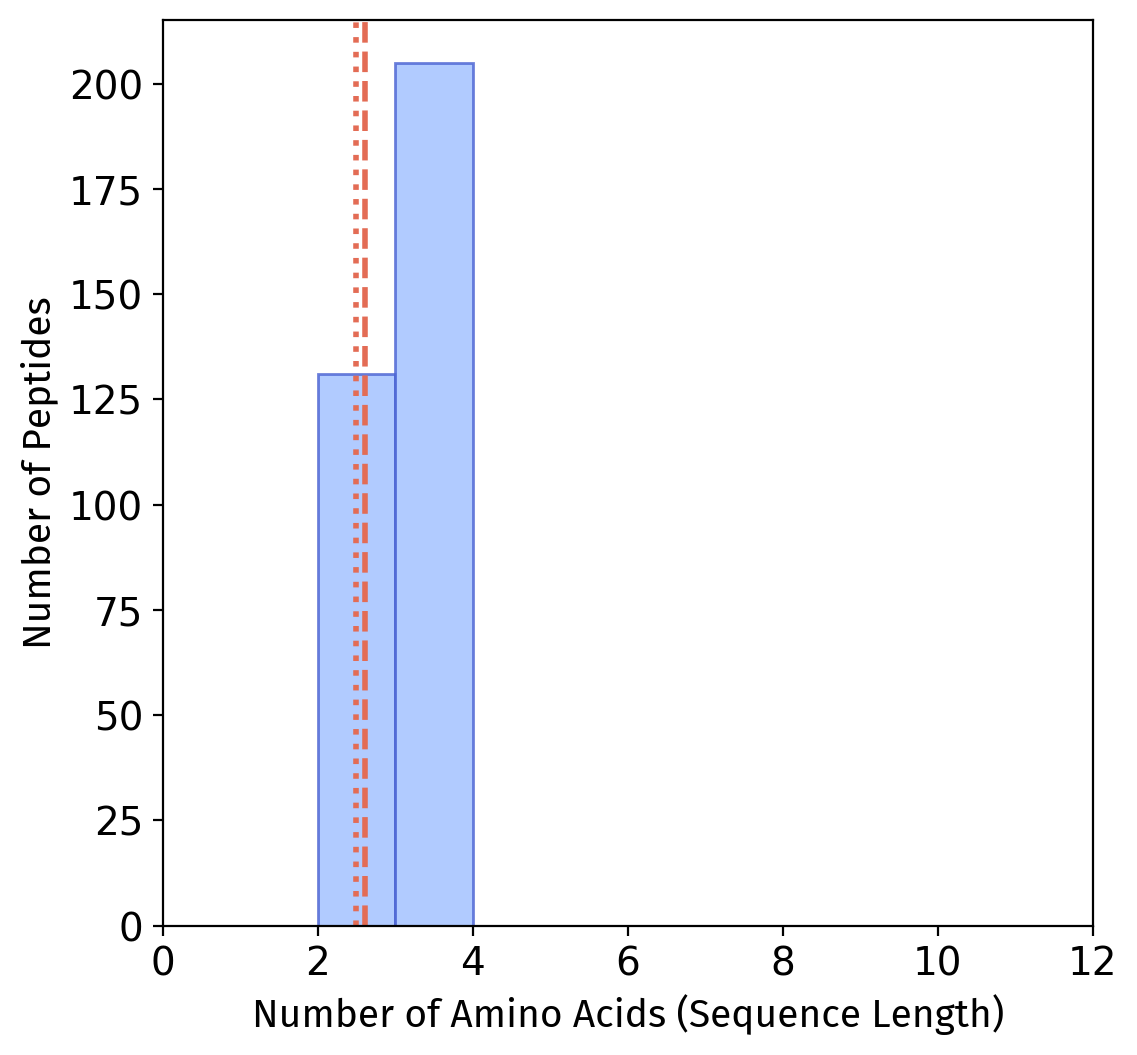}
   \captionof{figure}{Length distribution of ace inhibitory ic50 dataset.}
\label{fig:length_dist_ace_inhibitory_ic50}
\end{center}

\begin{center} 
\centering
\includegraphics[width=0.8\textwidth]{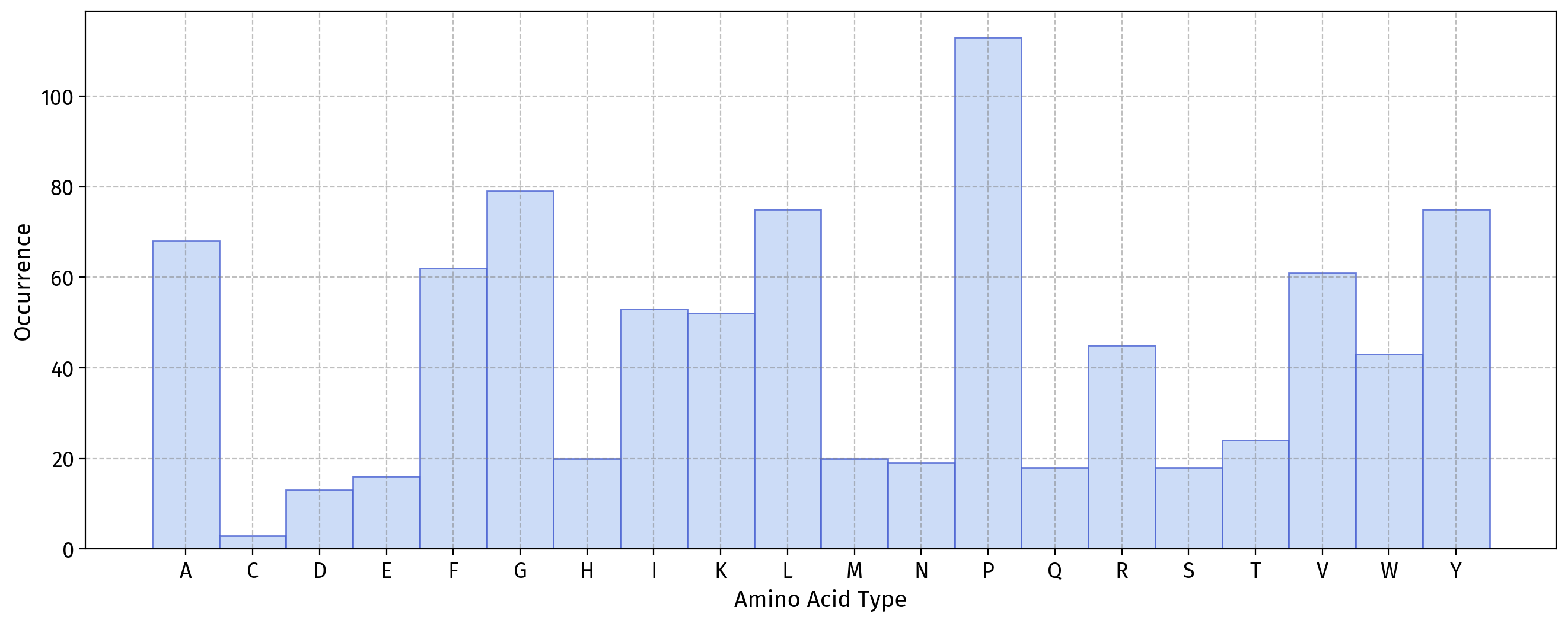}
   \captionof{figure}{Label distribution of ACE inhibitory IC50 dataset.}
\label{fig:label_dist_ace_inhibitory_ic50}
\end{center}

\begin{center} 
  \centering
  \includegraphics[width=0.8\textwidth]{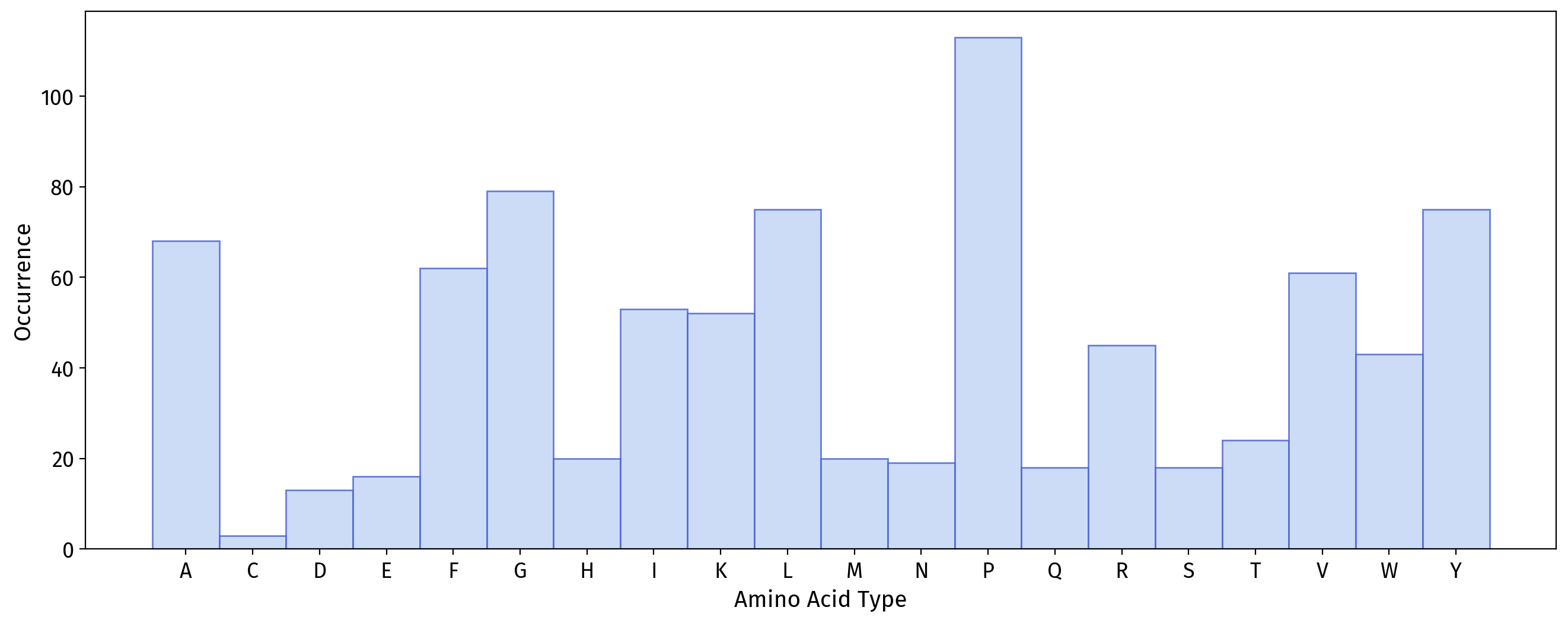}
     \captionof{figure}{Amino acid distribution of ace inhibitory ic50 dataset.}
  \label{fig:aa_dist_ace_inhibitory_ic50}
  \end{center}

\subsubsection{dpp-iv inhibitors}
\fielditem{Property and Application} 
The dpp-iv inhibitor dataset includes peptides that inhibit dipeptidyl peptidase-4 (DPP-4), thereby preventing the degradation of incretins such as Glucagon-Like Peptide 1 (GLP-1) and Gastric Inhibitory Polypeptide (GIP). This inhibition enhances insulin secretion and improves glycemic control in diabetic patients. Clinically, this mechanism helps manage blood glucose levels while minimizing the risk of hypoglycemia and weight gain.

\fielditem{Data Source} 
The positive samples are collected from ~\cite{charoenkwan2020idppiv}, Which extractes 665 unique DPP-IV inhibitors from literatures and publically available databases (i.e., BIOPEP-UWM).

\paragraph{Experiment Negative Samples} 
Dipeptides with DPPIV inhibition rate <5\% at 0.5 mM from ~\cite{hikida2013systematic} are added as negative samples to the dataset. After removing 13 duplicate sequences between positive and negative samples in the merged dataset, the final dataset contains 86 dipeptide negative samples.

\fielditem{Dataset Statistics} 
The dataset contains 1,268 datapoints with sequences ranging from 2 to 33 amino acids (average length 6.15) in length.

\textbf{Task: Classification; Split: Hybrid; Evaluation: ROC-AUC}

\begin{center} 
\centering
\includegraphics[width=0.8\textwidth]{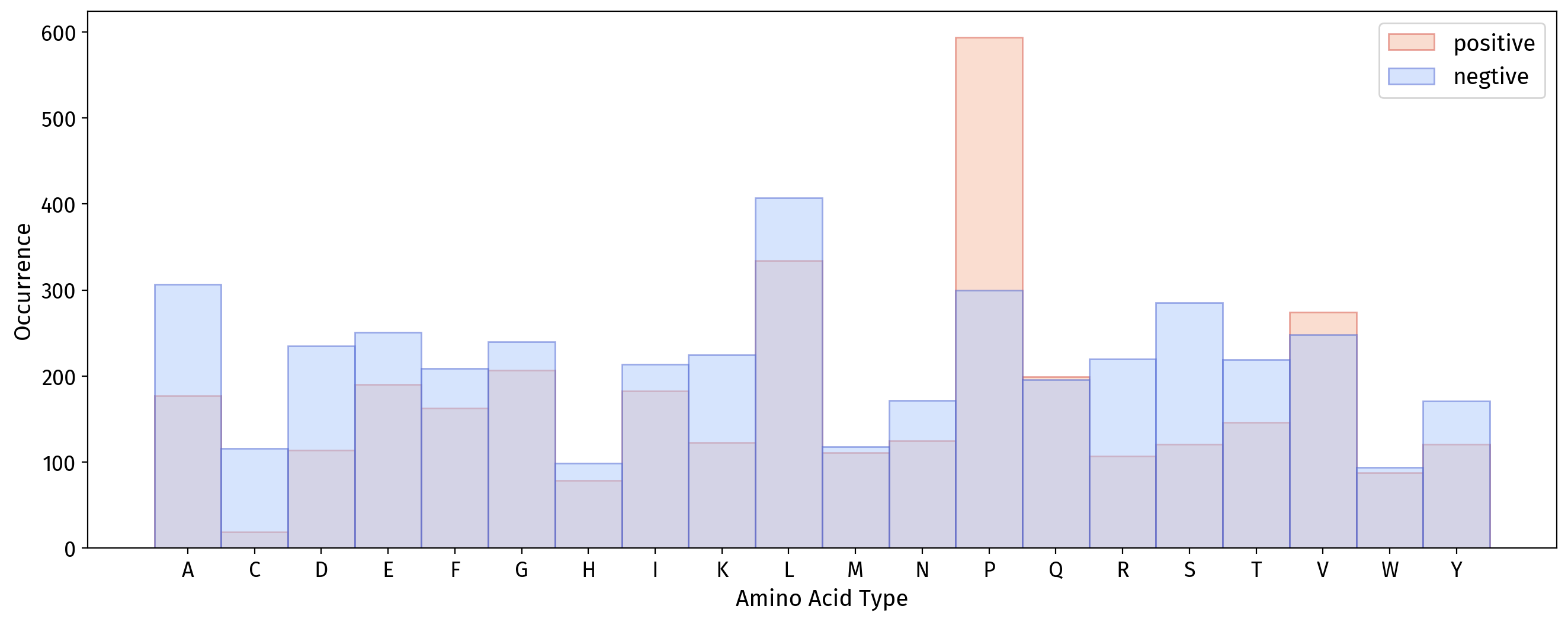}
   \captionof{figure}{Amino acid distribution comparison between positive and negative samples for dpp-iv inhibitor dataset.}
\label{fig:aa_dist_dppiv_inhibitors}
\end{center}

\begin{center} 
\centering
\includegraphics[width=0.8\textwidth]{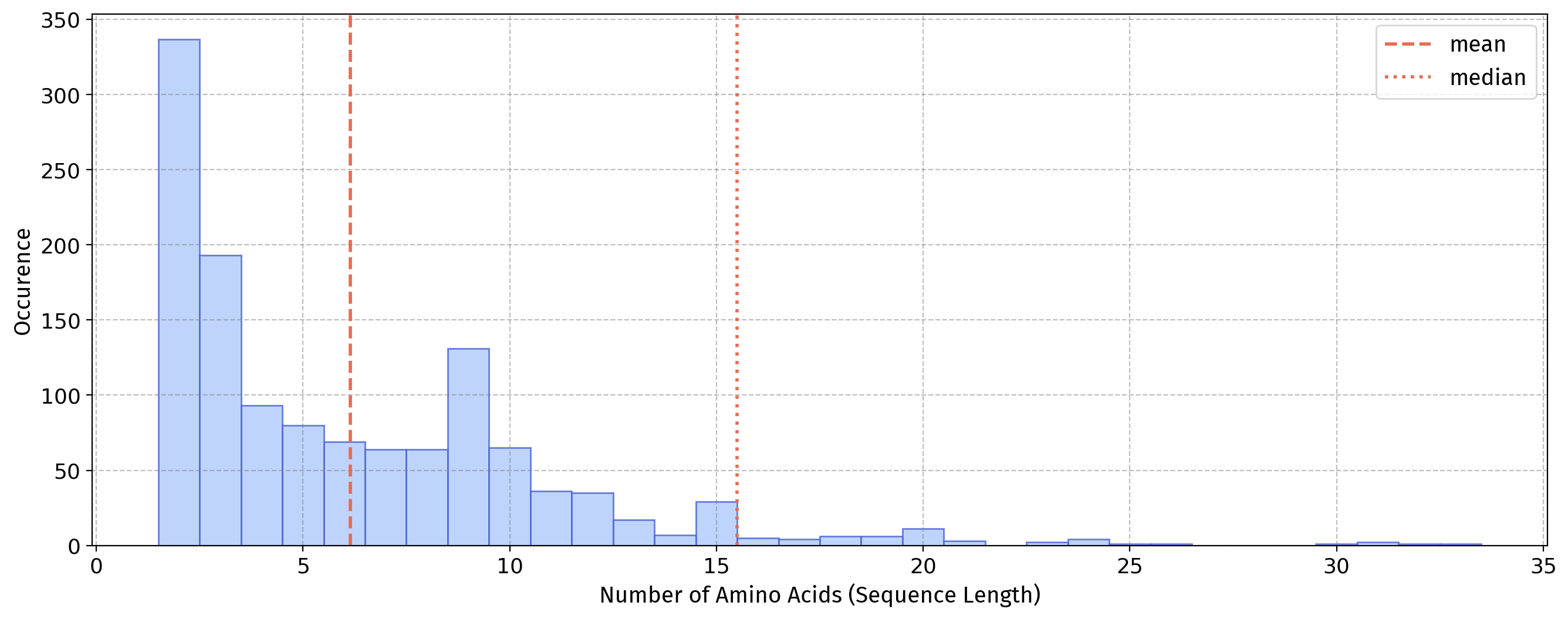}
   \captionof{figure}{Length distribution of dpp-iv inhibitor dataset.}
\label{fig:length_dist_dppiv_inhibitors}
\end{center}

\begin{center} 
\centering
\includegraphics[width=0.8\textwidth]{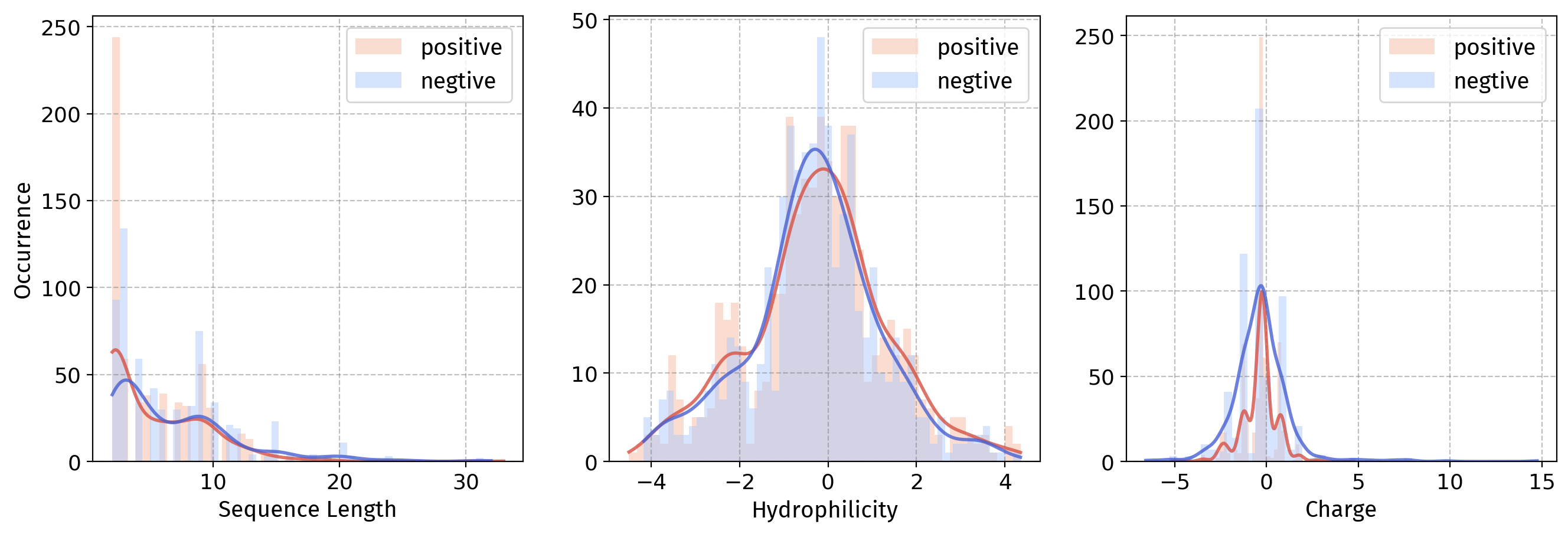}
   \captionof{figure}{Property comparison between positive and negative samples for dpp-iv inhibitor dataset.}
\label{fig:property_comp_dppiv_inhibitors}
\end{center}

\subsubsection{antidiabetic}
\fielditem{Property and Application} 
The antidiabetic peptide dataset captures peptides that modulate insulin signaling pathways and other mechanisms involved in glucose homeostasis. These peptides contribute to lowering blood glucose levels, improving insulin sensitivity, and inhibiting diabetes-related metabolic dysfunctions.

\fielditem{Data Source} 
The positive samples are collected from~\cite{yue2024discovery}, combined with antidiabetic peptides recorded in Peptipedia. The literature collected a total of 1,786 antidiabetic peptides related to Type 1 Diabetes Mellitus (T1DM) and 756 related to Type 2 Diabetes Mellitus (T2DM) from the BioDADPep database~\citep{roy2019biodadpep}. Peptides containing non-standard residues are removed, and redundancy is addressed using the CD-HIT are eliminated.

\paragraph{Experiment Negative Samples} 
In dipeptides, the main activity related to diabetes is DPP-IV inhibition. Therefore, dipeptides with DPPIV inhibition rate <5\% from ~\cite{hikida2013systematic} were added as negative samples to the dataset.
After removing 23 duplicate sequences between positive and negative samples in the merged dataset, the final dataset contains 76 dipeptide negative samples.

\fielditem{Dataset Statistics} 
The dataset contains 3,028 datapoints with sequences ranging from 2 to 46 amino acids (average length 10.41) in length.

\textbf{Task: Classification; Split: Hybrid; Evaluation: ROC-AUC}

\begin{center} 
\centering
\includegraphics[width=0.8\textwidth]{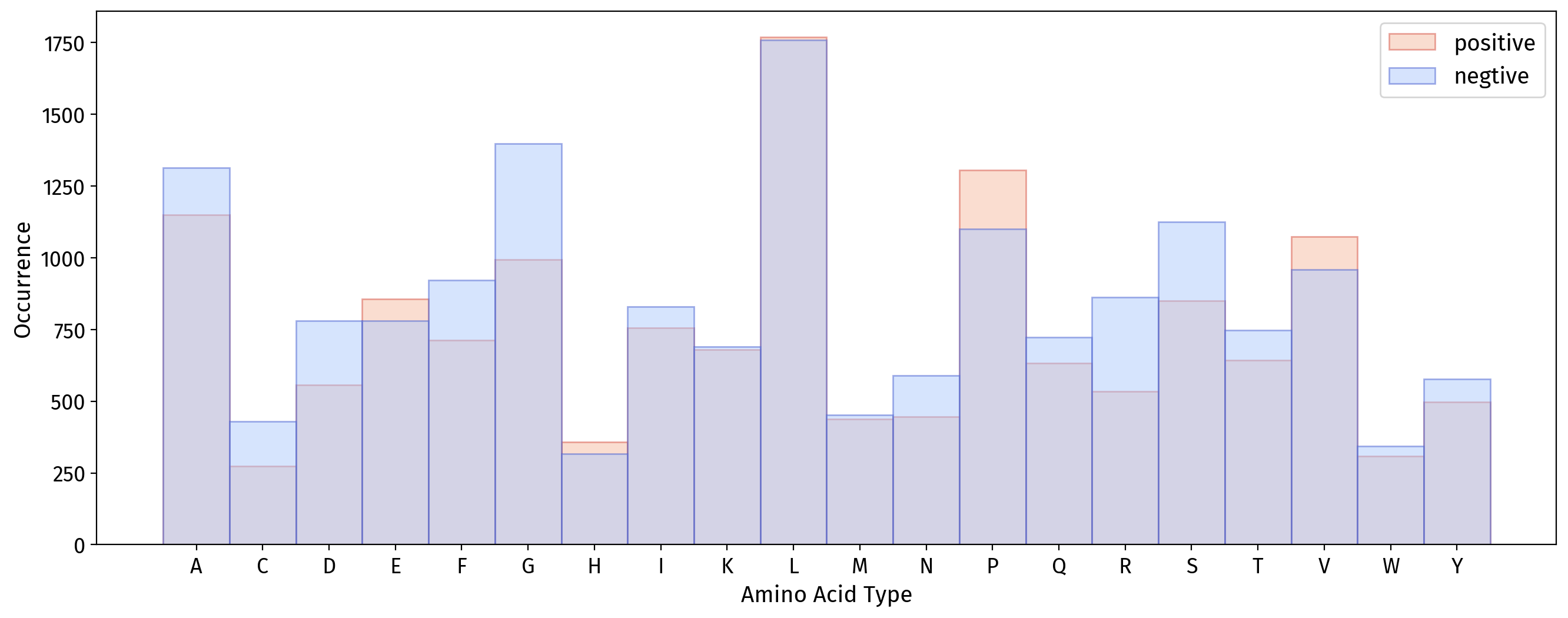}
   \captionof{figure}{Amino acid distribution comparison between positive and negative samples for antidiabetic dataset.}
\label{fig:aa_dist_antidiabetic}
\end{center}

\begin{center} 
\centering
\includegraphics[width=0.8\textwidth]{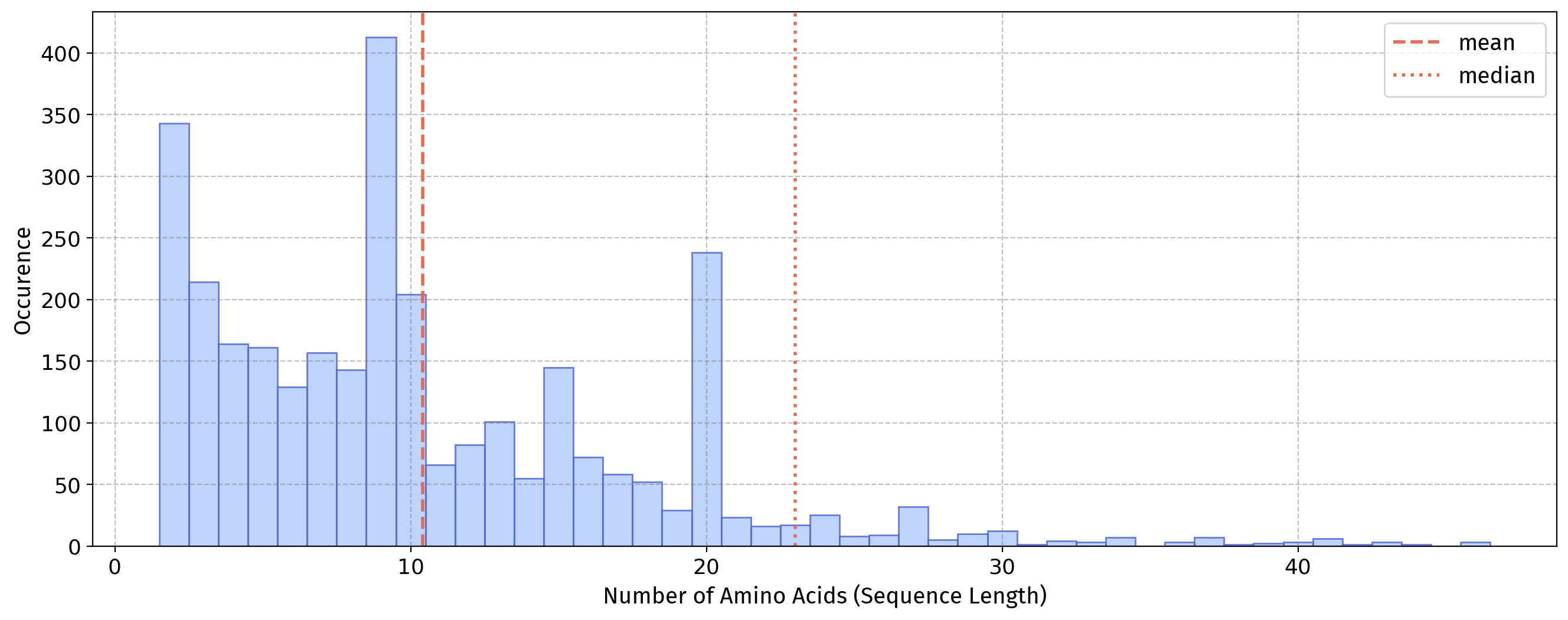}
   \captionof{figure}{Length distribution of antidiabetic dataset.}
\label{fig:length_dist_antidiabetic}
\end{center}

\begin{center} 
\centering
\includegraphics[width=0.8\textwidth]{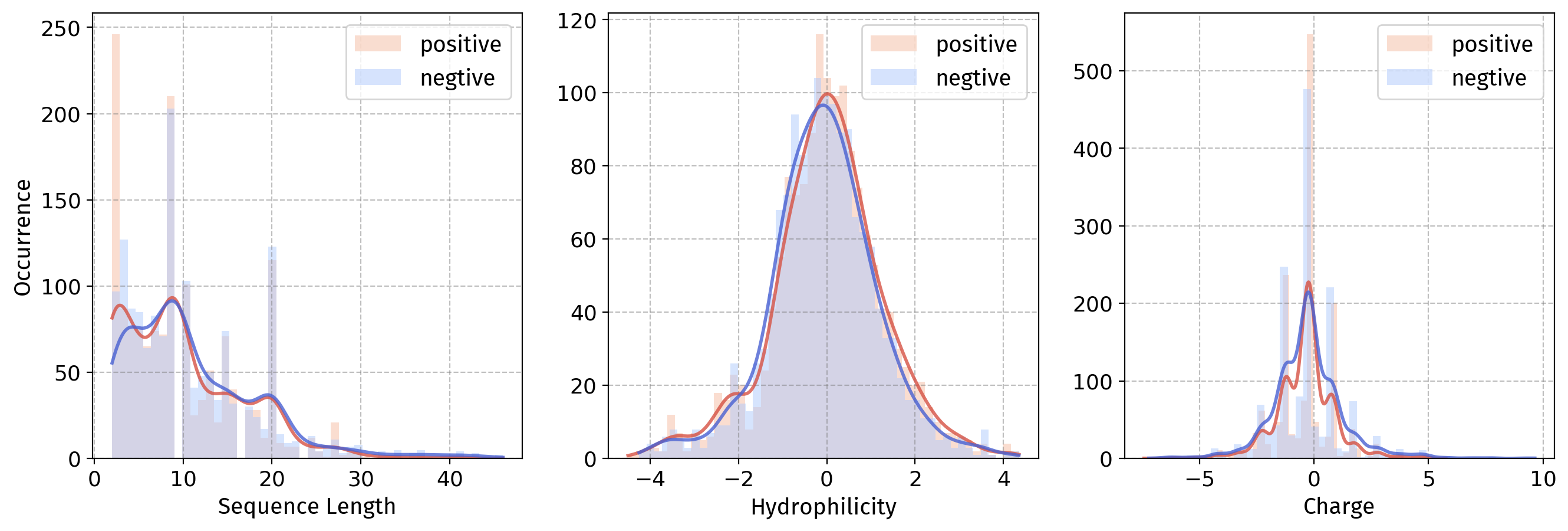}
   \captionof{figure}{Property comparison between positive and negative samples for antidiabetic dataset.}
\label{fig:property_comp_antidiabetic}
\end{center}

\subsection{Others Group}

\begin{datasetbox}
\paragraph{Definition.} 
Datasets in this group include a variety of peptide datasets with biological activities relevant to chronic disease management, which do not fall into the major categories described above.
\paragraph{Impact.} 
Peptides targeting chronic diseases beyond major categories offer therapeutic options for conditions that require long-term management. Their high target specificity, favorable biocompatibility, and reduced off-target effects make them particularly suitable for prolonged treatment, potentially improving patient adherence and minimizing cumulative side effects.
\paragraph{Pipeline.} Activity Modeling
\end{datasetbox}

\subsubsection{antiaging}
\fielditem{Property and Application} 
The antiaging dataset contains peptides that promote healthy lifespan by mitigating adverse health consequences commonly observed in the elderly population. These peptides may target various aging-related pathways and cellular processes.

\fielditem{Data Source} 
The positive samples are all collected from Peptipedia.

\fielditem{Dataset Statistics} 
The dataset contains 558 datapoints with sequences ranging from 2 to 80 amino acids (average length 10.93) in length.

\textbf{Task: Classification; Split: Hybrid; Evaluation: ROC-AUC}

\begin{center} 
\centering
\includegraphics[width=0.8\textwidth]{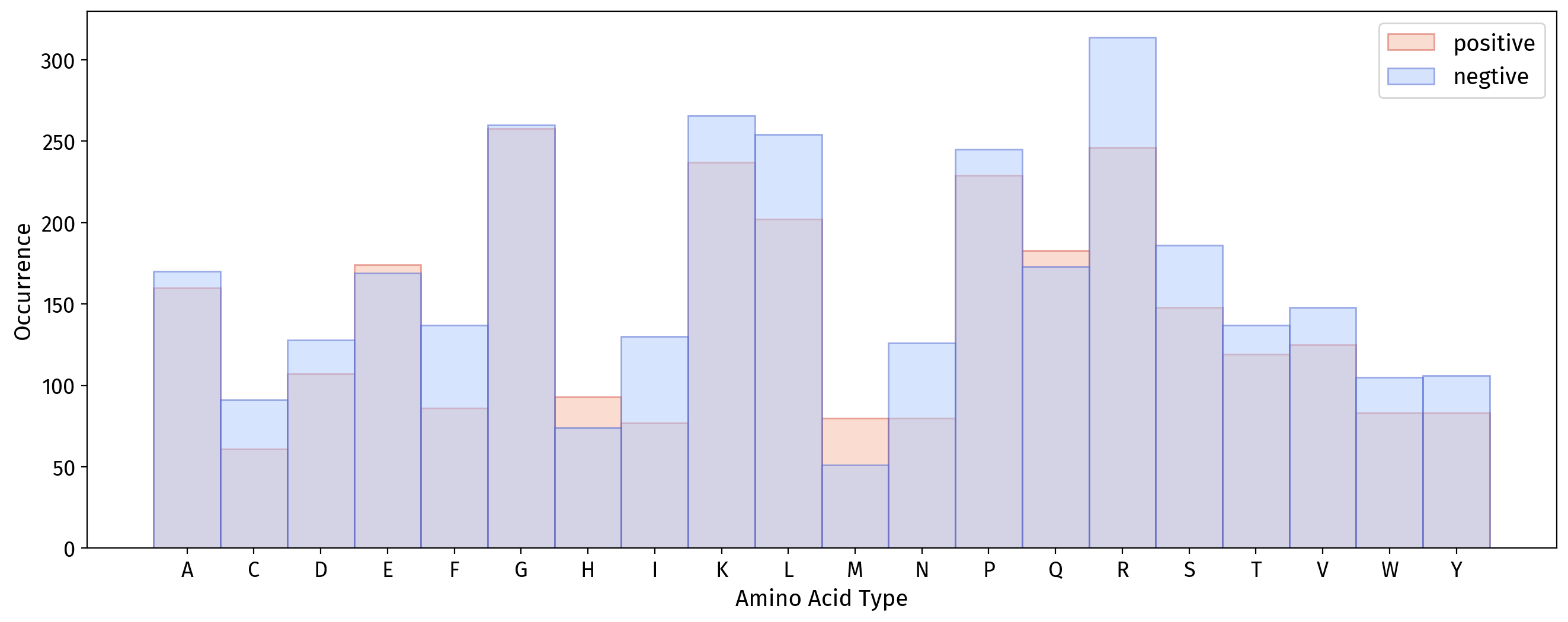}
   \captionof{figure}{Amino acid distribution comparison between positive and negative samples for antiaging dataset.}
\label{fig:aa_dist_antiaging}
\end{center}

\begin{center} 
\centering
\includegraphics[width=0.8\textwidth]{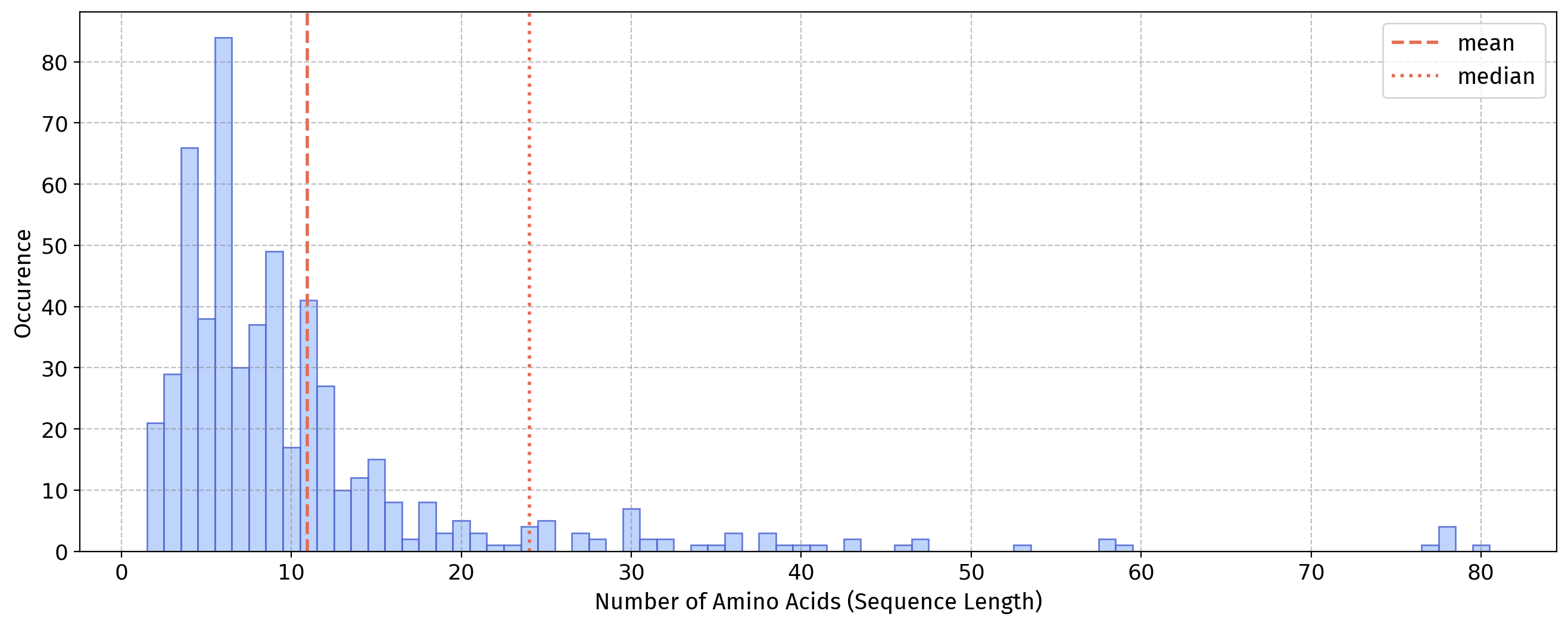}
   \captionof{figure}{Length distribution of antiaging dataset.}
\label{fig:length_dist_antiaging}
\end{center}

\begin{center} 
\centering
\includegraphics[width=0.8\textwidth]{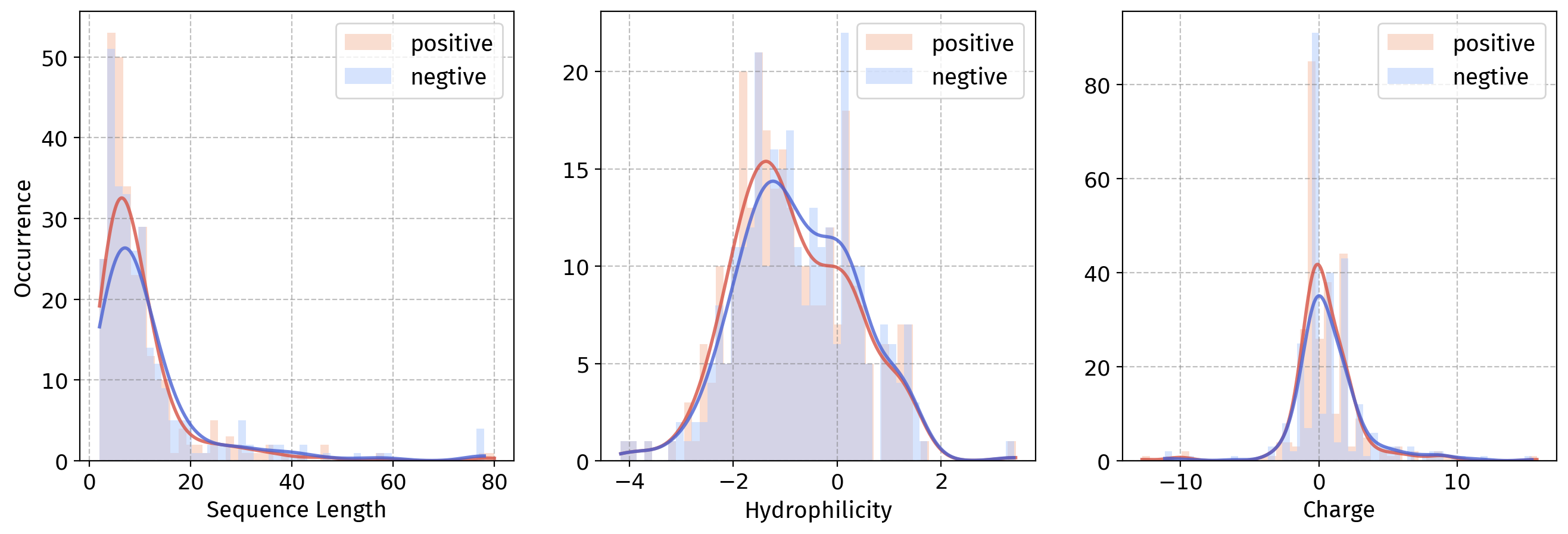}
   \captionof{figure}{Property comparison between positive and negative samples for antiaging dataset.}
\label{fig:property_comp_antiaging}
\end{center}

\subsubsection{anti-inflammatory}
\fielditem{Property and Application} 
the anti-inflammatory dataset contains peptides that inhibit the activity of pro-inflammatory cytokines and signaling pathways, offering therapeutic potential for autoimmune disorders, chronic inflammatory conditions, and inflammatory tissue damage.

\fielditem{Data Source} 
The positive samples are all collected from Peptipedia.

\fielditem{Dataset Statistics} 
The dataset contains 7,750 datapoints with sequences ranging from 2 to 107 amino acids (average length 16.95) in length.

\textbf{Task: Classification; Split: Hybrid; Evaluation: ROC-AUC}

\begin{center} 
\centering
\includegraphics[width=0.8\textwidth]{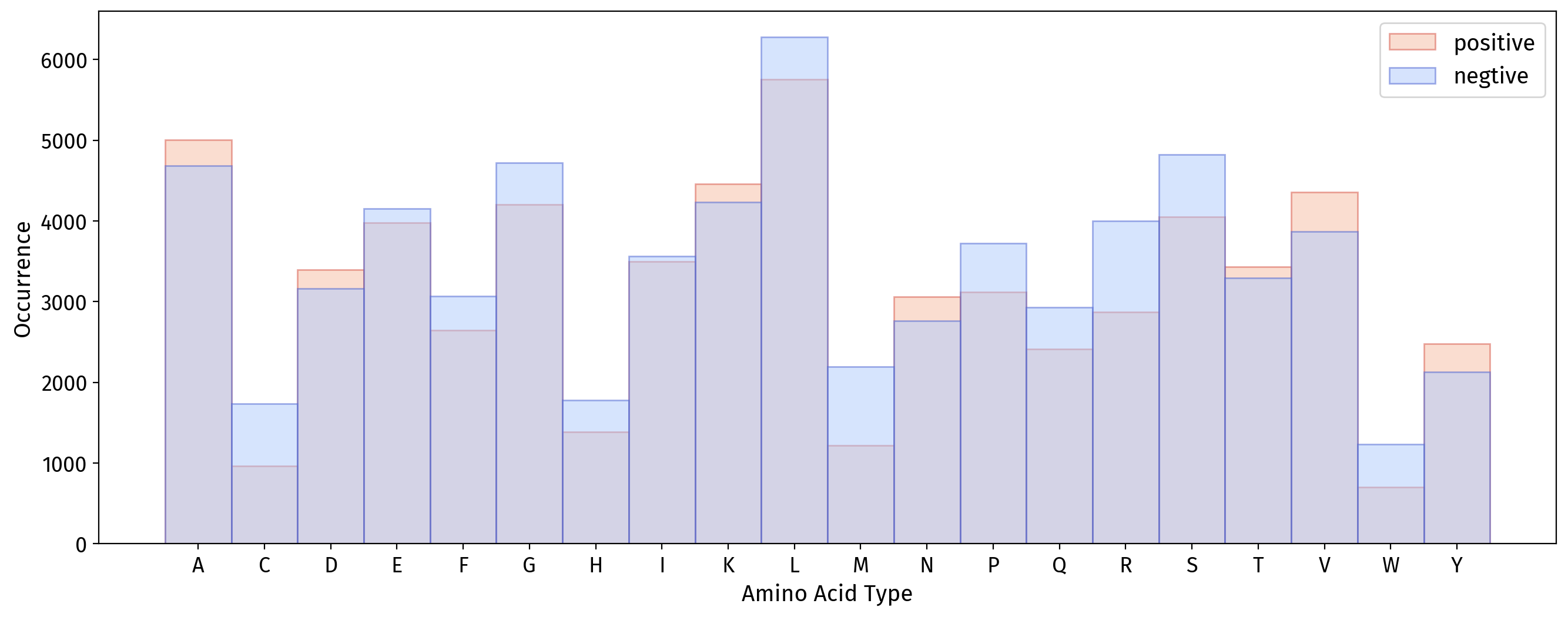}
   \captionof{figure}{Amino acid distribution comparison between positive and negative samples for anti-inflammatory dataset.}
\label{fig:aa_dist_antiinflamatory}
\end{center}

\begin{center} 
\centering
\includegraphics[width=0.8\textwidth]{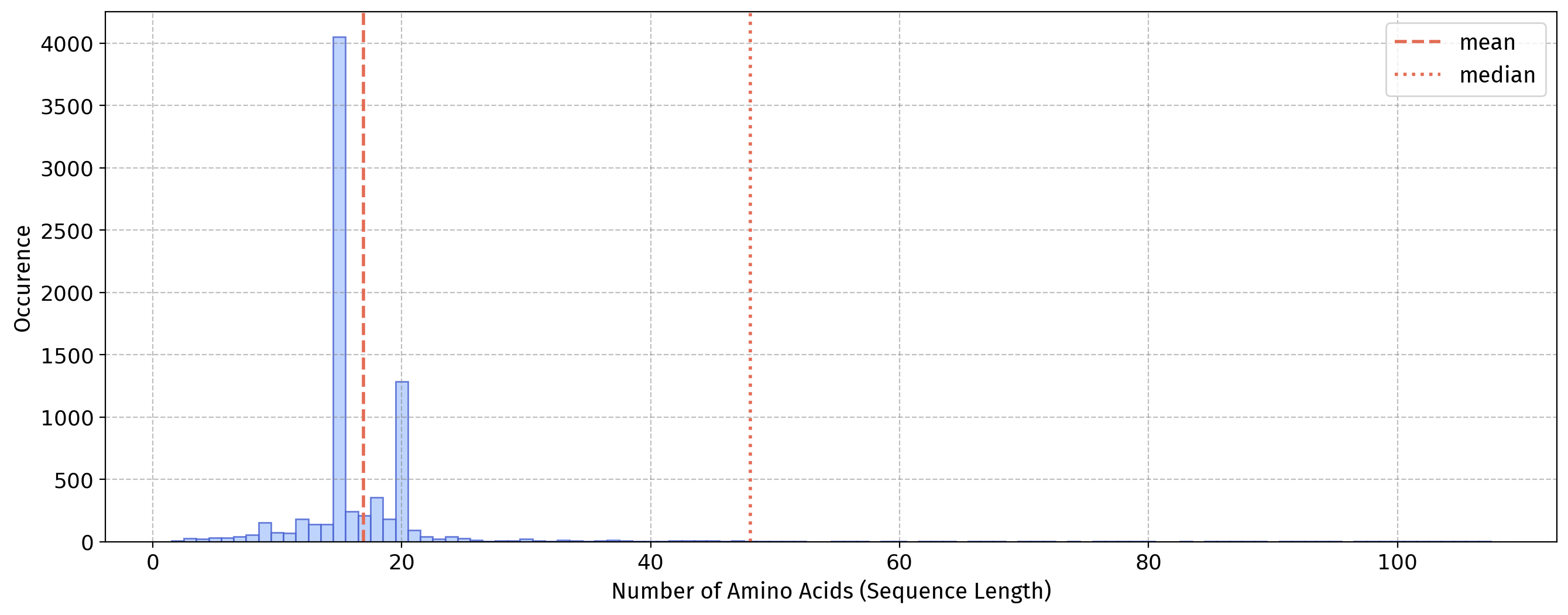}
   \captionof{figure}{Length distribution of anti-inflammatory dataset.}
\label{fig:length_dist_antiinflamatory}
\end{center}

\begin{center} 
\centering
\includegraphics[width=0.8\textwidth]{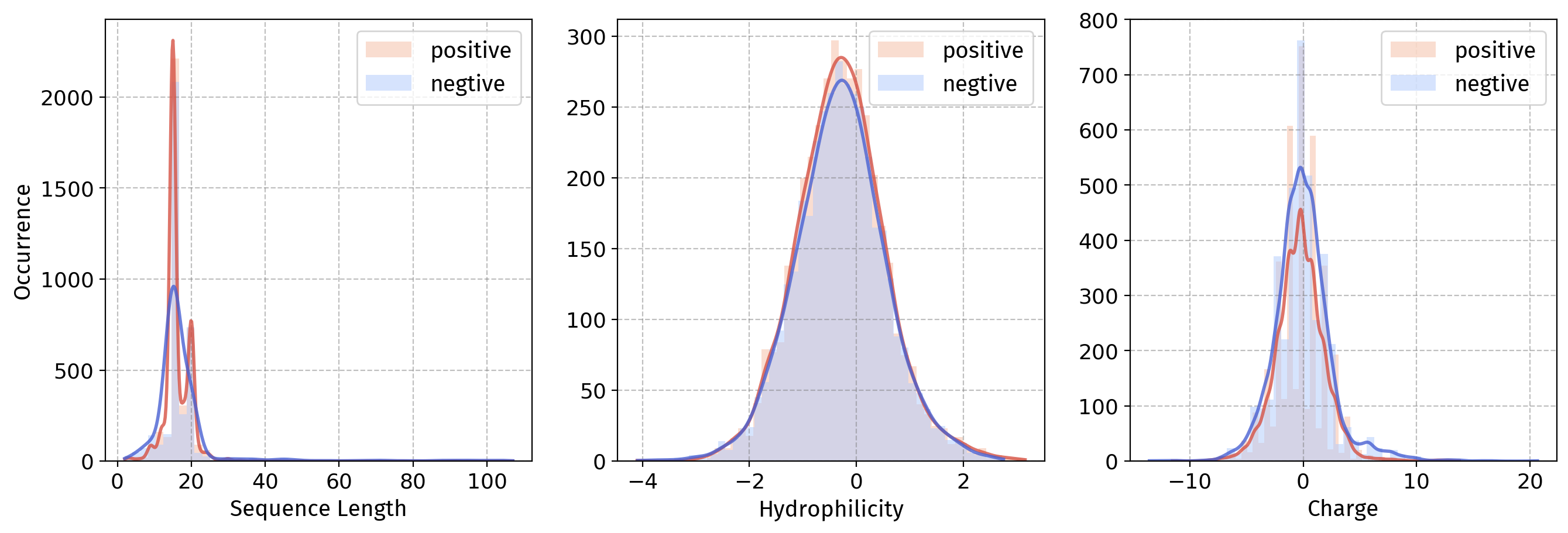}
   \captionof{figure}{Property comparison between positive and negative samples for anti-inflammatory dataset.}
\label{fig:property_comp_antiinflamatory}
\end{center}

\subsubsection{antioxidant}
\fielditem{Property and Application} 
The antioxidant dataset contains peptides that can scavenge reactive oxygen species and free radicals, protecting cells from oxidative damage. Such activity is critical for mitigating cellular injury associated with aging, chronic inflammation, and other oxidative stress-related diseases.

\fielditem{Data Source} 
The dataset is sourced from AnOxPePred~\citep{olsen2020anoxpepred}. The literature constructs a benchmark dataset by extracting data from published articles and the BIOPEP-UWM database
~\citep{iwaniak2024biopep}. Each peptide is binary labelled for two classes: free radical scavenger (FRS) and chelator. The classes are labeled 1 (positive) if their source have measured/indicated an activity and otherwise 0 (negative). Peptides with either class being positive are considered antioxidant peptides. 
To diminish homology bias while training, sequences are filtered using the Needleman–Wunsch algorithm so that no pair shared more than 90\% identity, leaving 436 positive samples. 
In addition, 732 antioxidant peptides from Peptipedia are merged.

\paragraph{Experiment Negative Samples} 
Negative Samples are also collected from AnOxPePred~\citep{olsen2020anoxpepred}. Peptides with both classes negative are considered non-antioxidant. To reduce homology bias, negative peptides are filtered using the Needleman–Wunsch algorithm so that no pair shared more than 90\% identity.
After merging the two source datasets and removing 22 sequences that existed in both positive and negative samples, the final dataset contains 195 negative samples.

\fielditem{Dataset Statistics} 
The dataset contains 2,242 datapoints with sequences ranging from 2 to 11 amino acids (average length 4.10) in length.

\textbf{Task: Classification; Split: Hybrid; Evaluation: ROC-AUC}

\begin{center} 
\centering
\includegraphics[width=0.8\textwidth]{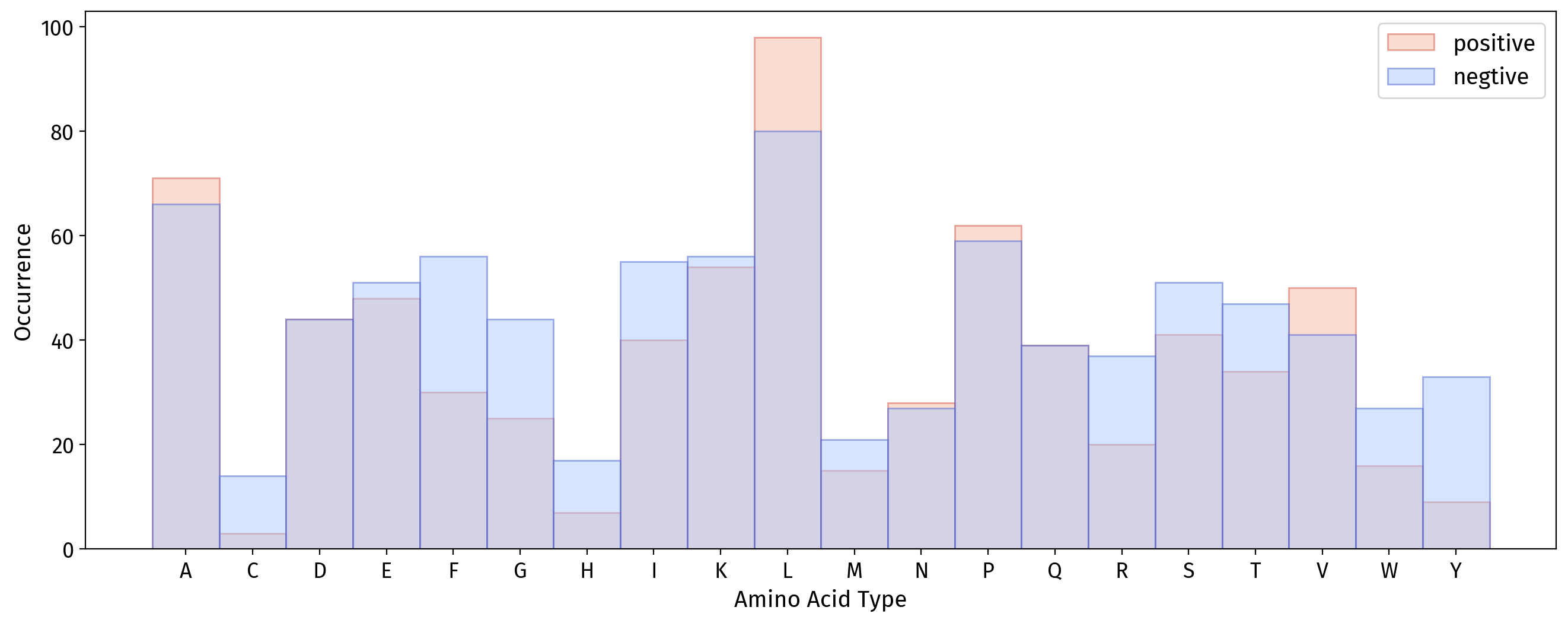}
   \captionof{figure}{Amino acid distribution comparison between positive and negative samples for antioxidant dataset.}
\label{fig:aa_dist_antioxidant}
\end{center}

\begin{center} 
\centering
\includegraphics[width=0.8\textwidth]{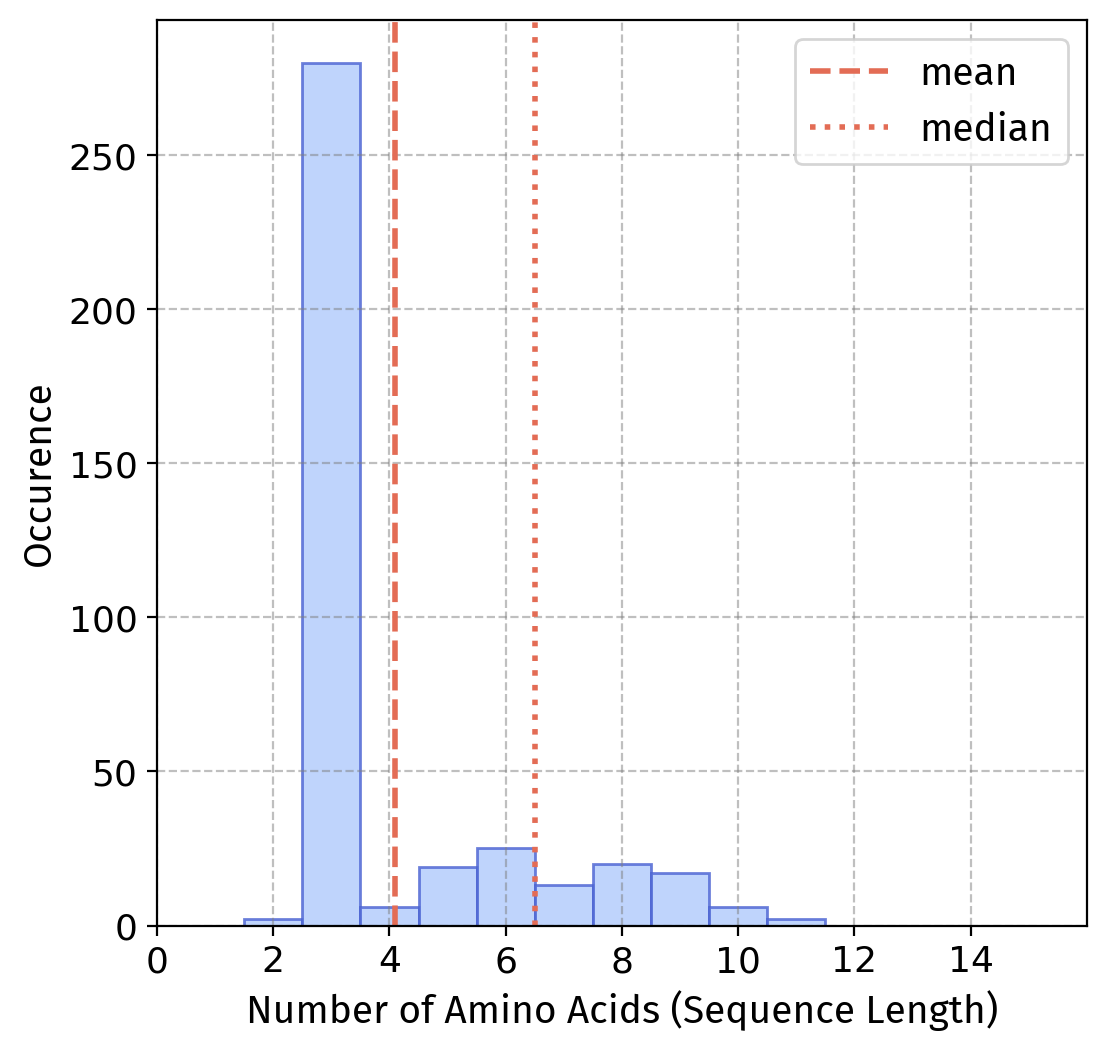}
   \captionof{figure}{Length distribution of antioxidant dataset.}
\label{fig:length_dist_antioxidant}
\end{center}

\begin{center} 
\centering
\includegraphics[width=0.8\textwidth]{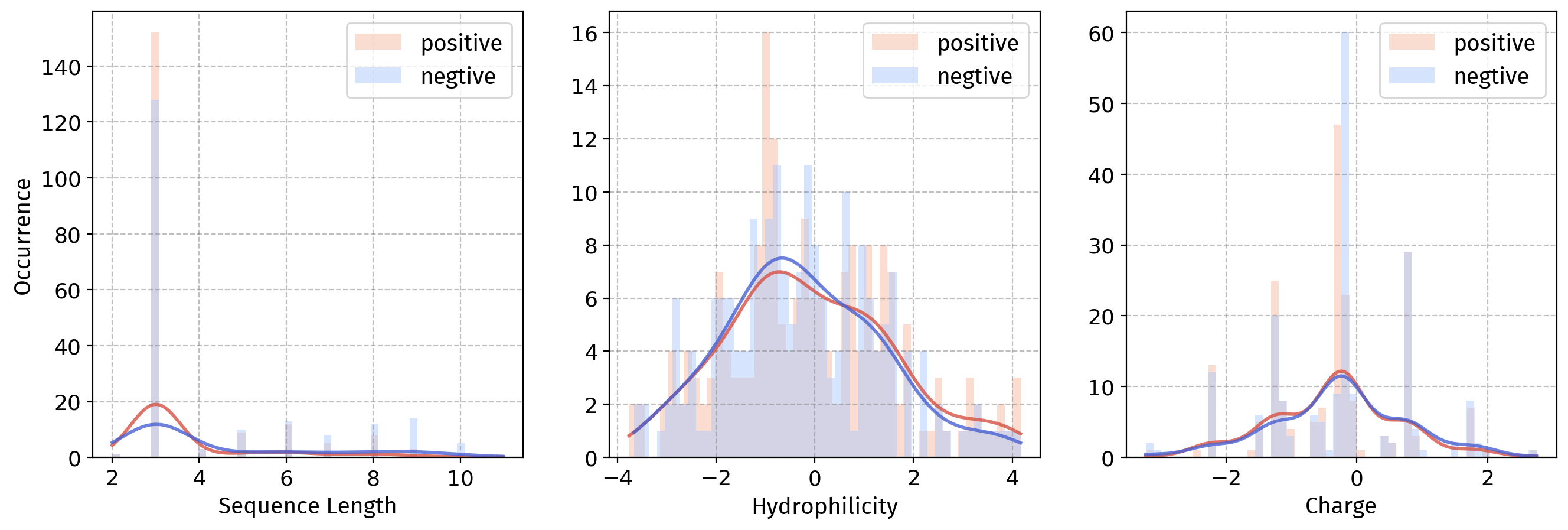}
   \captionof{figure}{Property comparison between positive and negative samples for antioxidant dataset.}
\label{fig:property_comp_antioxidant}
\end{center}

\subsubsection{neuropeptide}
\fielditem{Property and Application} 
The neuropeptides dataset contains peptides that regulate neurotransmitter systems and neural signaling pathways, contributing to the treatment of neurological disorders including depression, epilepsy, and cognitive dysfunction.

\fielditem{Data Source} 
NeuroPeptides (NPs) are sourced from ~\cite{bin2020prediction}. The study derives experimentally validated NPs from the comprehensive resource of NeuroPep~\citep{wang2024neuropep}. The positive dataset is created by three processes: (1) deleting NPs with more than 100 residues; (2) abandoning sequences with unnatural amino acids (B, J, O, U, X, and Z); and (3) removing samples with more than 90\% sequence identity using CD-HIT, leaving 2425 positive samples.
In addition, 2911 neuropeptides from Peptipedia are merged.

\fielditem{Dataset Statistics} 
The dataset contains 9,254 datapoints with sequences ranging from 2 to 150 amino acids (average length 19.57) in length.

\textbf{Task: Classification; Split: Hybrid; Evaluation: ROC-AUC}

\begin{center} 
\centering
\includegraphics[width=0.8\textwidth]{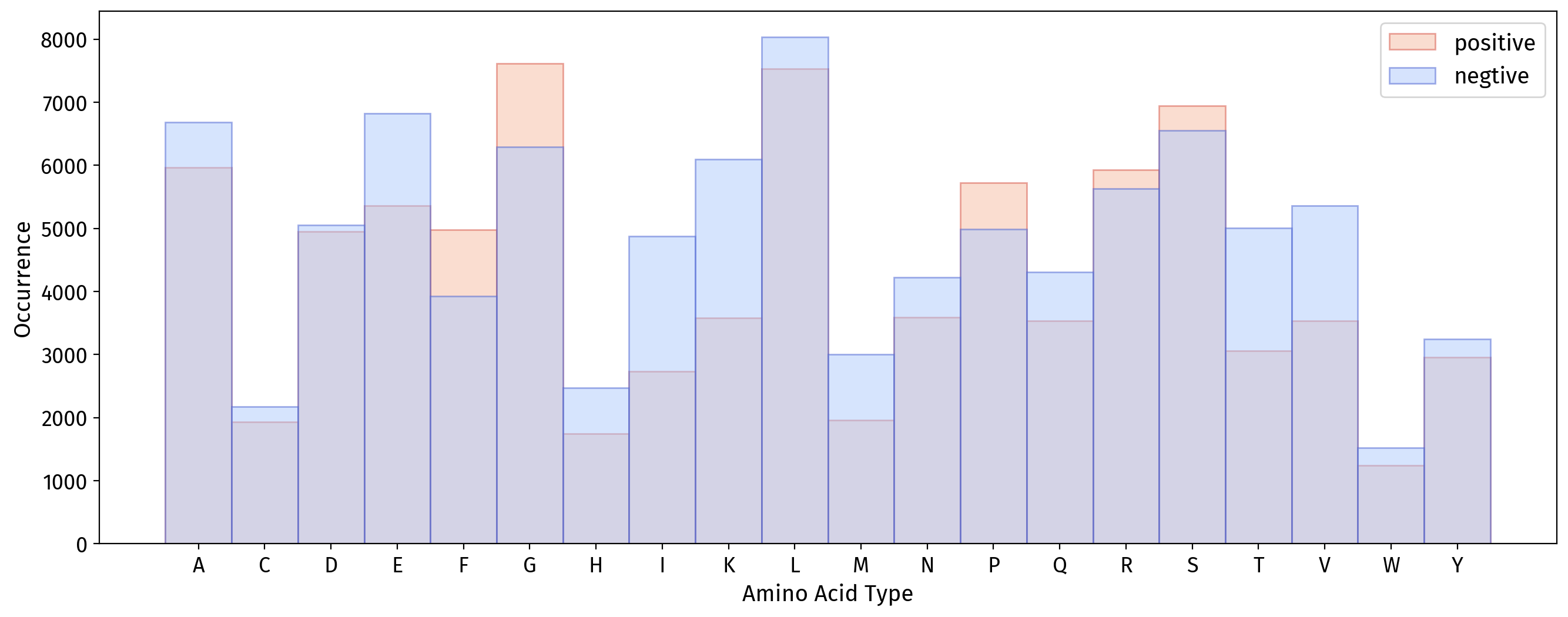}
   \captionof{figure}{Amino acid distribution comparison between positive and negative samples for neuropeptide dataset.}
\label{fig:aa_dist_neuropeptide}
\end{center}

\begin{center} 
\centering
\includegraphics[width=0.8\textwidth]{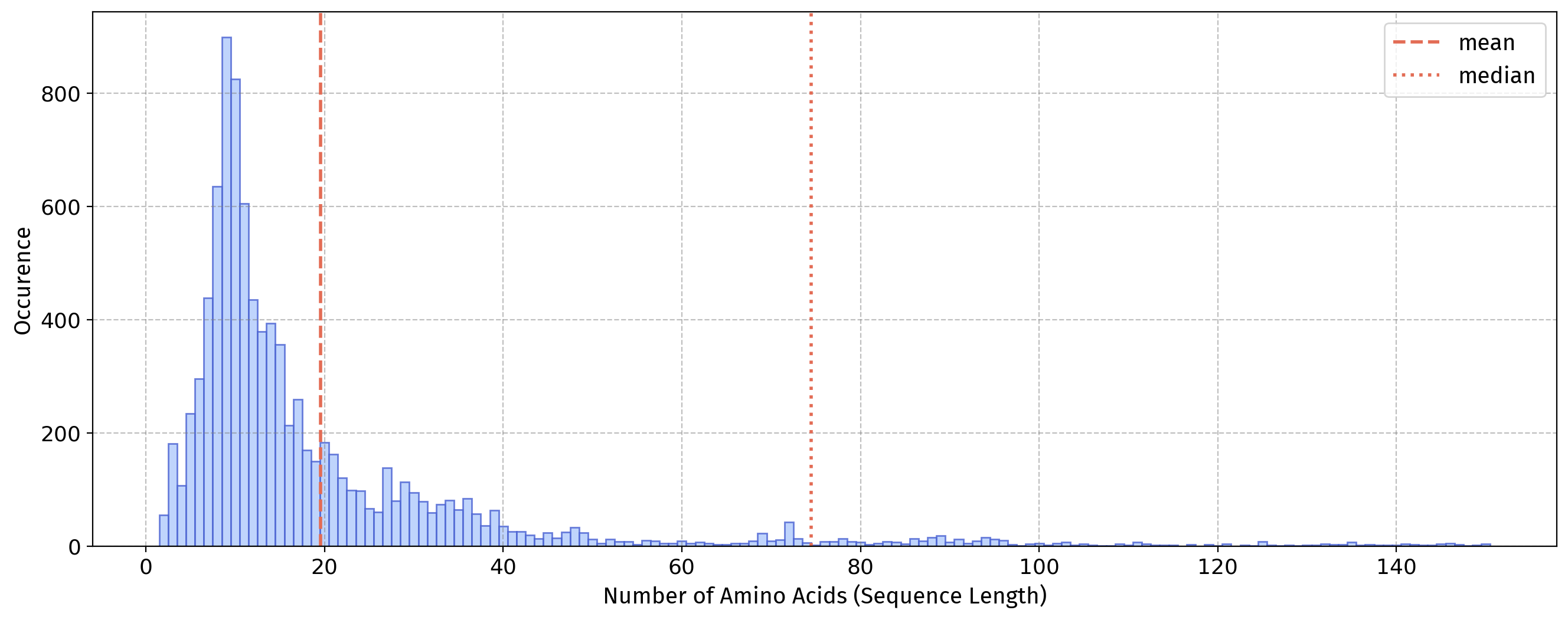}
   \captionof{figure}{Length distribution of neuropeptide dataset.}
\label{fig:length_dist_neuropeptide}
\end{center}

\begin{center} 
\centering
\includegraphics[width=0.8\textwidth]{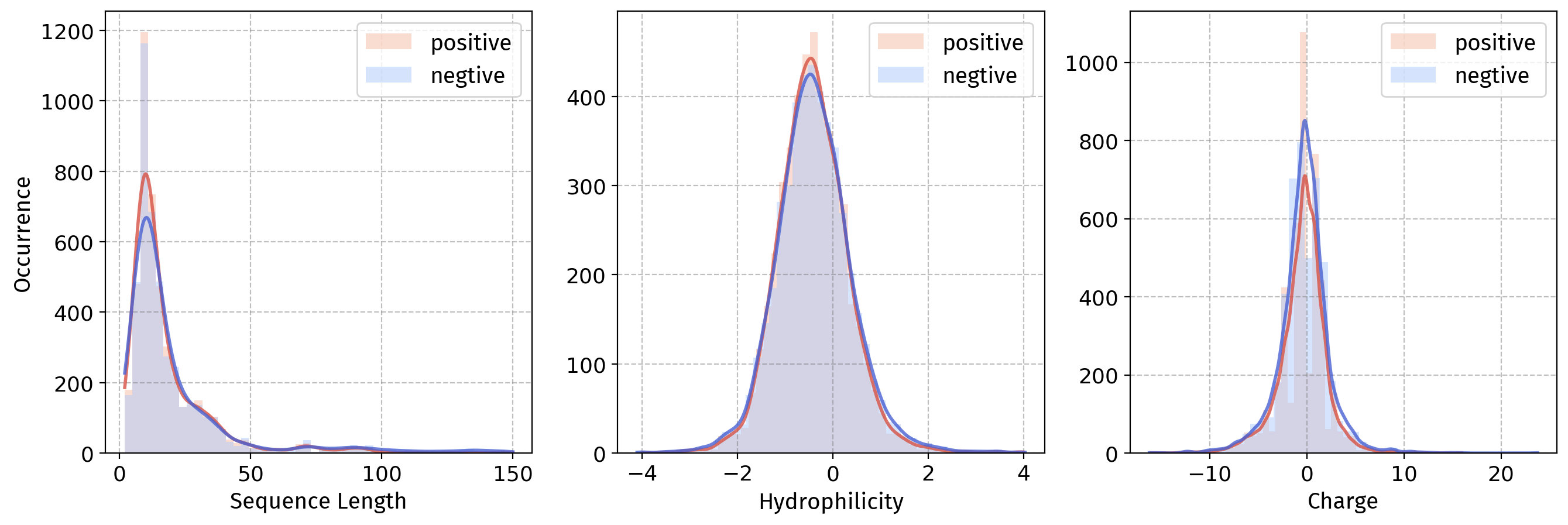}
   \captionof{figure}{Property comparison between positive and negative samples for neuropeptide dataset.}
\label{fig:property_comp_neuropeptide}
\end{center}

\subsubsection{quorum sensing}
\fielditem{Property and Application} 
Quorum sensing peptides (QSPs) regulate communication between bacteria for colony formation, which may affect biofilm formation and host immune response. These peptides are important for understanding bacterial behavior and developing anti-biofilm strategies.

\fielditem{Data Source} 
QSPs are extracted from ~\cite{rajput2015prediction}. The literature extractes 231 entries reported from 1955–2012 from the Quorumpeps database~\citep{wynendaele2013quorumpeps}. Additionally, PubMed is searched and 10 more entries are collected. 100\% identical peptides are removed, leaving 218 positive samples. Additionally, 47 QS peptides from Peptipedia are merged.

\fielditem{Dataset Statistics} 
The dataset contains 490 datapoints with sequences ranging from 3 to 48 amino acids (average length 10.83) in length.

\textbf{Task: Classification; Split: Hybrid; Evaluation: ROC-AUC}

\begin{center} 
\centering
\includegraphics[width=0.8\textwidth]{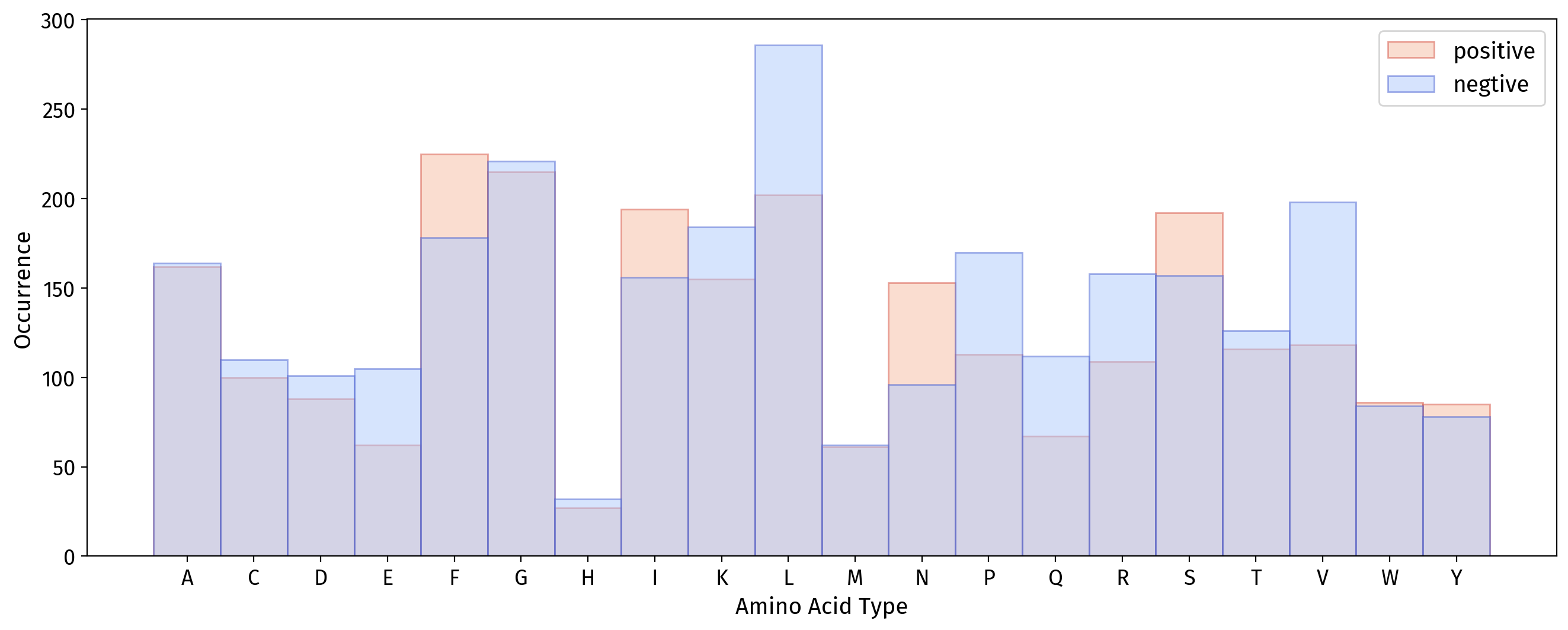}
   \captionof{figure}{Amino acid distribution comparison between positive and negative samples for quorum sensing dataset.}
\label{fig:aa_dist_quorum_sensing}
\end{center}

\begin{center} 
\centering
\includegraphics[width=0.8\textwidth]{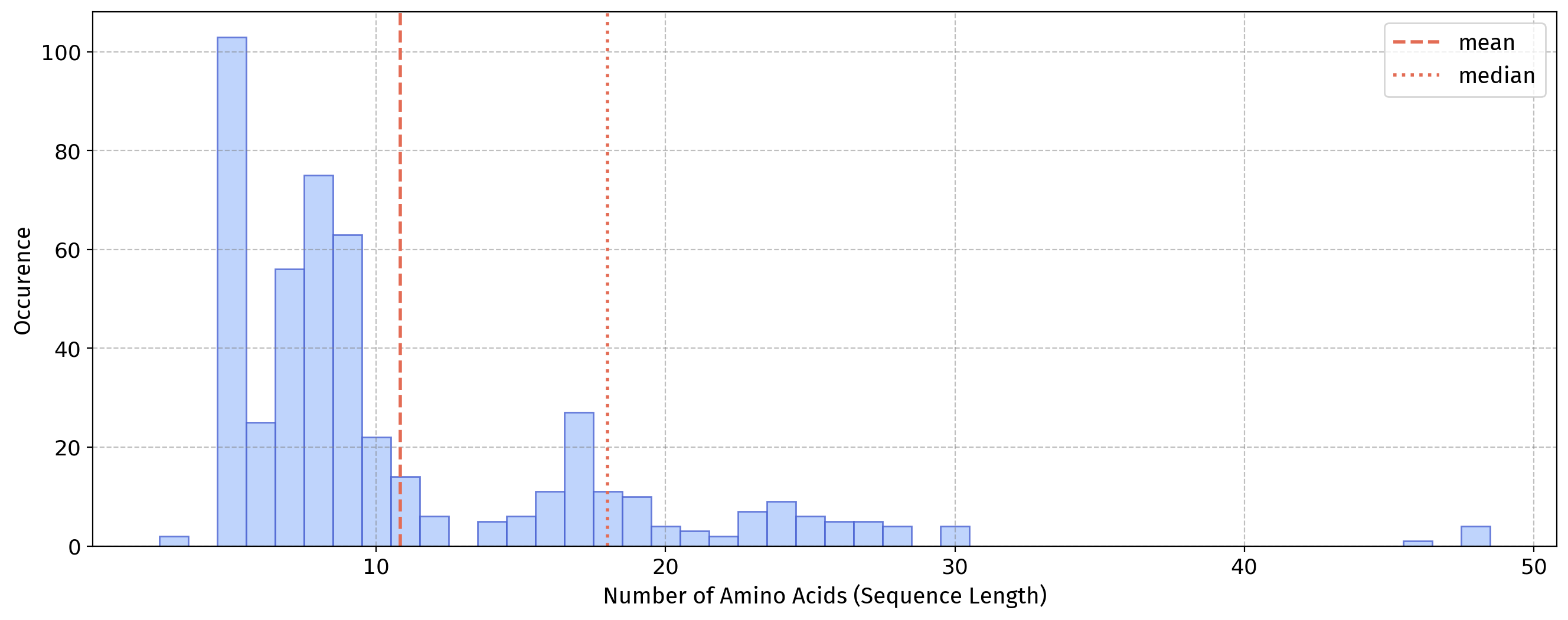}
   \captionof{figure}{Length distribution of quorum sensing dataset.}
\label{fig:length_dist_quorum_sensing}
\end{center}

\begin{center} 
\centering
\includegraphics[width=0.8\textwidth]{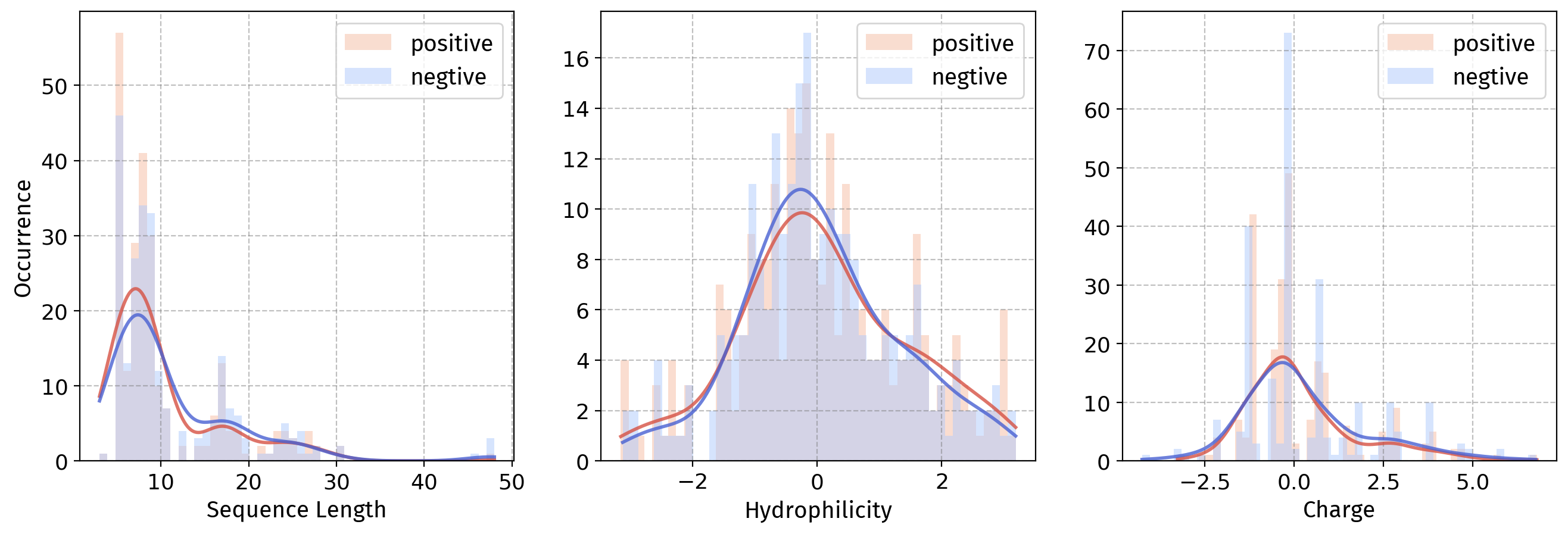}
   \captionof{figure}{Property comparison between positive and negative samples for quorum sensing dataset.}
\label{fig:property_comp_quorum_sensing}
\end{center}

\subsection{PepPI Group}

\begin{datasetbox}
\paragraph{Definition.} 
The PepPI group comprises datasets of peptide–protein interactions.
\paragraph{Impact.} 
Peptides generally offer potential advantages over small-molecule drugs in target engagement, including higher binding specificity and affinity, due to their larger interaction surfaces and structural adaptability.
\paragraph{Pipeline.} Activity Modeling
\end{datasetbox}

\subsubsection{PpI}
\fielditem{Property and Application} 
The PpI dataset collects peptide-protein interaction pairs, providing a foundation for studying molecular recognition mechanisms. These data enable precision drug design by identifying peptide ligands capable of binding specific protein targets with therapeutic relevance.

\fielditem{Data Source} 
The protein--peptide interaction pairs are extracted from~\cite{bhat2025novo} and PepNN \citep{abdin2022pepnn}, which are both derived from the RCSB PDB by selecting peptide--protein complexes with a buried surface area $\geq 400~\AA^2$, peptides $\leq 25$ amino acids, and proteins $\geq 30$ amino acids.

\fielditem{Dataset Statistics} 
The dataset contains 44,148 datapoints with peptide sequences ranging from 5 to 25 amino acids (average length 12.8) in length.

\textbf{Task: Classification; Split: Cold-start; Evaluation: ROC-AUC}

\begin{center} 
  \centering
  \includegraphics[width=0.8\textwidth]{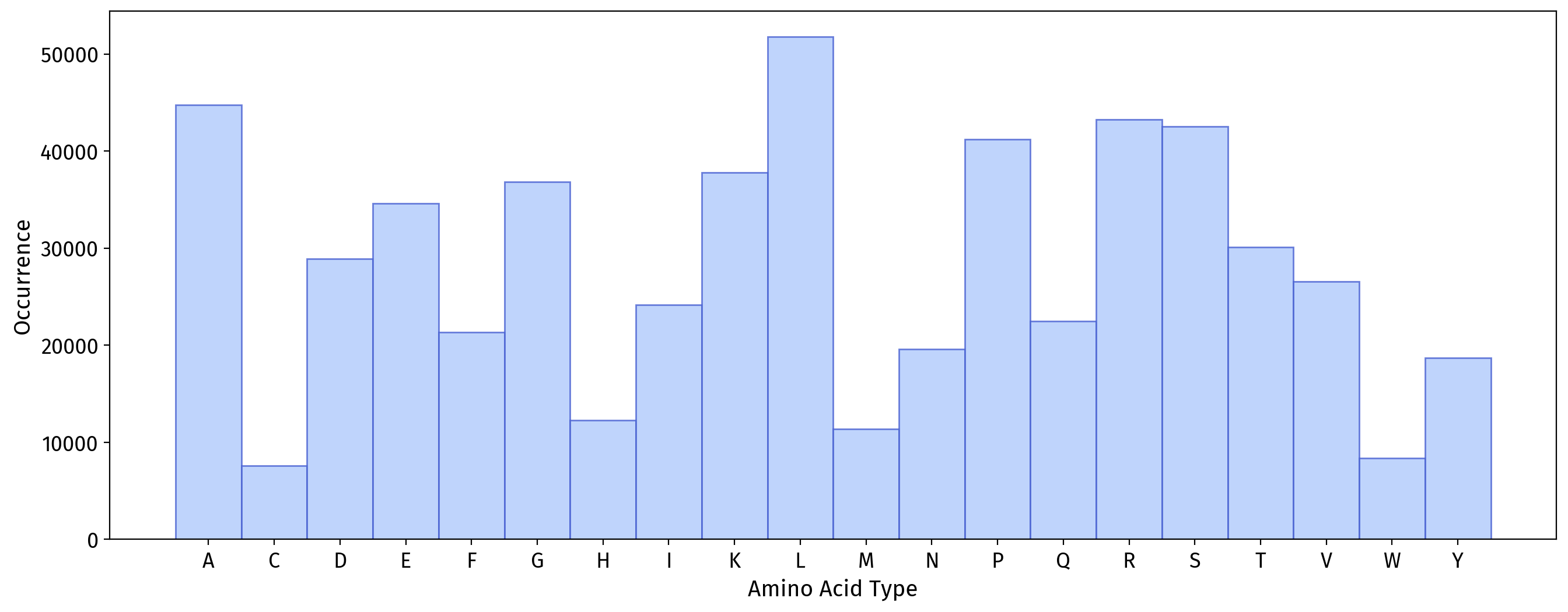}
     \captionof{figure}{Amino acid distribution for PpI dataset.}
  \label{fig:aa_comp_PpI}
\end{center}

\begin{center} 
  \centering
  \includegraphics[width=0.8\textwidth]{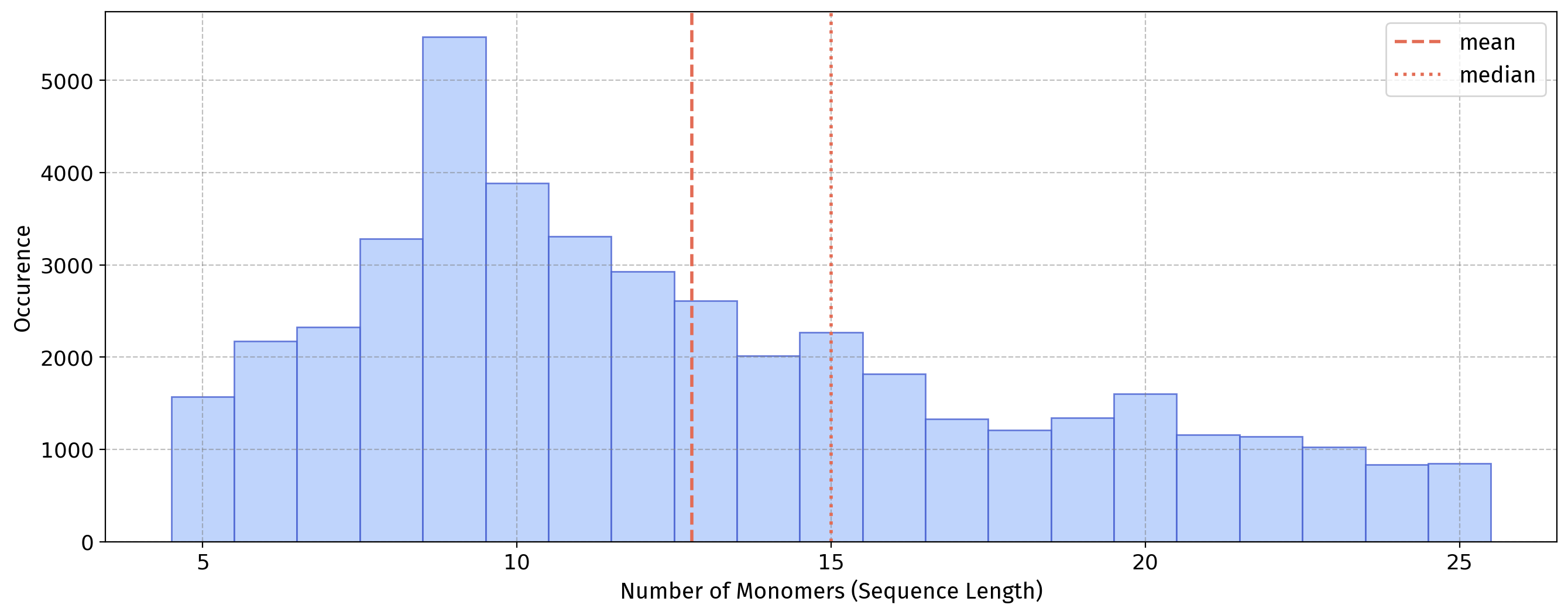}
     \captionof{figure}{Length distribution of PpI dataset.}
  \label{fig:len_comp_PpI}
\end{center}

\subsubsection{Peptide-Protein Binding Affinity (PpI\_ba)}
\fielditem{Property and Application} 
The PpI\_ba dataset provides binding affinity measurements for peptide-protein interactions, reported using metrics -lgKd(M). These quantitative values allow evaluation of interaction strength and support predictive modeling of high-affinity peptide ligands.

\fielditem{Data Source} 
The dataset is sourced from ~\cite{zhang2025pepland}, who constructs a benchmark for peptide–protein binding affinity prediction by following the workflow of ~\cite{lei2021deep}. The affinity data are derived from PDBbind v2019, which provides a high-quality collection of protein–ligand complex structures with experimentally measured binding affinities, all originating from the RCSB PDB.

\fielditem{Dataset Statistics} 
The dataset contains 1,433 datapoints with peptide sequences ranging from 5 to 50 amino acids (average length 16.14) in length.

\textbf{Task: Regression; Split: Cold-start; Evaluation: MAE}

\begin{center} 
  \centering
  \includegraphics[width=0.8\textwidth]{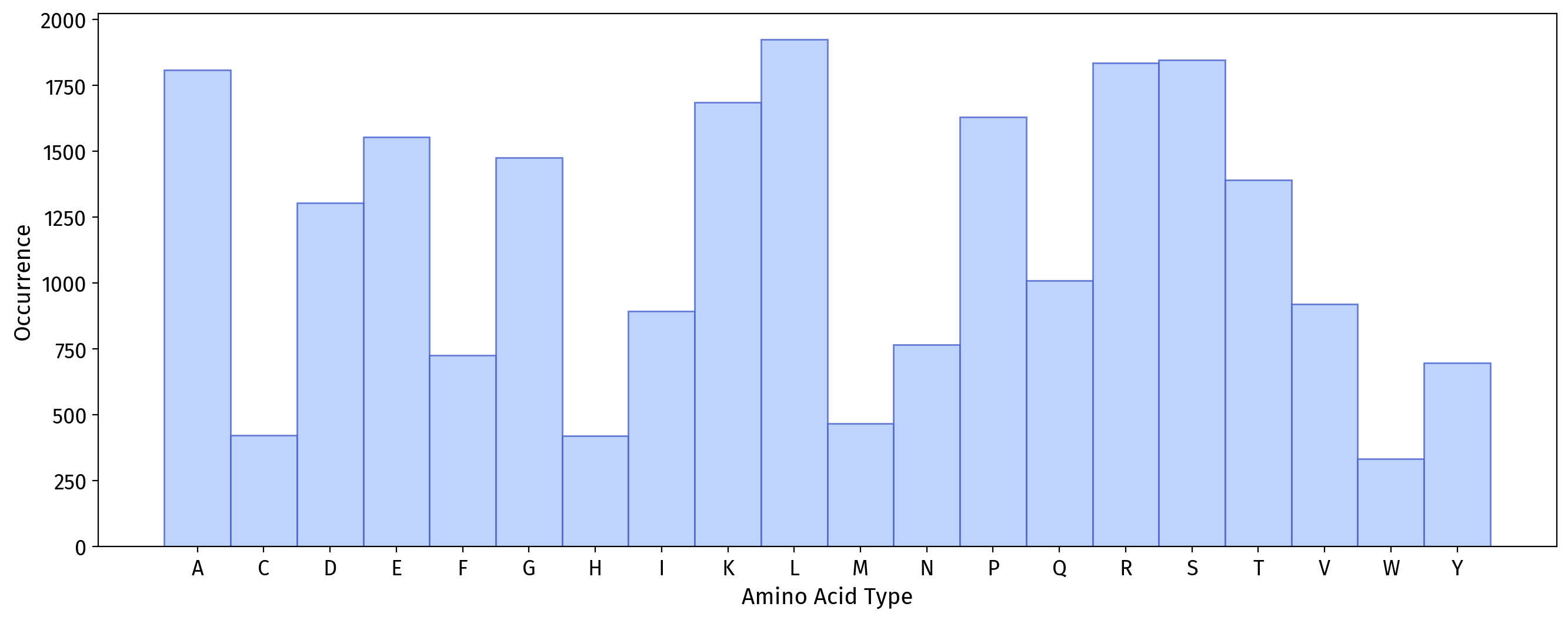}
     \captionof{figure}{Amino acid distribution for PpI\_ba dataset.}
  \label{fig:aa_comp_PpI_ba}
\end{center}

\begin{center} 
  \centering
  \includegraphics[width=0.8\textwidth]{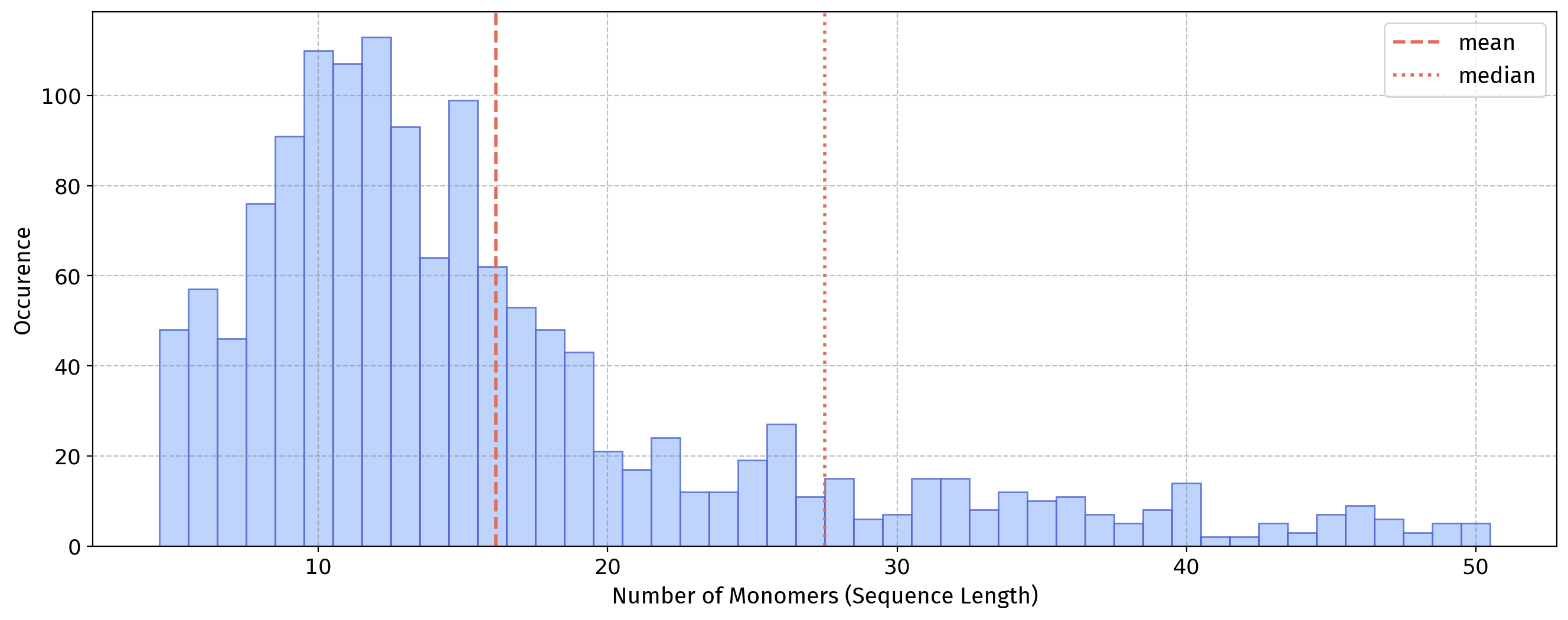}
     \captionof{figure}{Length distribution of PpI\_ba dataset.}
  \label{fig:len_comp_PpI_ba}
\end{center}

\subsubsection{Non-canonical Peptide-Protein Binding Affinity (nc-PpI\_ba)}
\fielditem{Property and Application} 
The nc-PpI\_ba dataset extends binding affinity data to non-natural or modified peptides, reported using metrics -lgKd(M), enabling investigation of how chemical modifications affect peptide-protein recognition. It supports the design of synthetic ligands with improved binding properties and stability.

\fielditem{Data Source} 
The dataset is sourced from ~\cite{zhang2025pepland}. The construction of the non-canonical dataset follows the same workflow as the canonical dataset, but with a focus on peptides containing non-canonical amino acids.

\fielditem{Dataset Statistics} 
The dataset contains 277 datapoints with peptide sequences ranging from 5 to 19 amino acids (average length 8.93) in length.

\textbf{Task: Regression; Split: Cold-start; Evaluation: MAE}

\subsection{Tox Group}

\begin{datasetbox}
\paragraph{Definition.} 
Datasets in this group collect datasets addressing safety-related properties of peptides. Peptide toxicity may manifest as unintended immune activation, neurotoxicity, or off-target cytotoxicity (e.g., hemolysis). 
\paragraph{Impact.} 
This grouping is indispensable in peptide drug development, as safety liabilities often halt clinical translation even after efficacy has been demonstrated. 
\paragraph{Pipeline.} Safety Assessment
\end{datasetbox}

\subsubsection{allergen}
\fielditem{Property and Application} 
The allergen dataset contains peptides that have been identified as immune-triggering epitopes, which are essential for understanding and preventing hypersensitivity reactions.

\fielditem{Data Source} 
The dataset is sourced from Peptipedia.

\fielditem{Dataset Statistics} 
The dataset contains 3,354 datapoints with sequences ranging from 4 to 150 amino acids (average length 39.97) in length.

\textbf{Task: Classification; Split: Hybrid; Evaluation: ROC-AUC}

\begin{center} 
\centering
\includegraphics[width=0.8\textwidth]{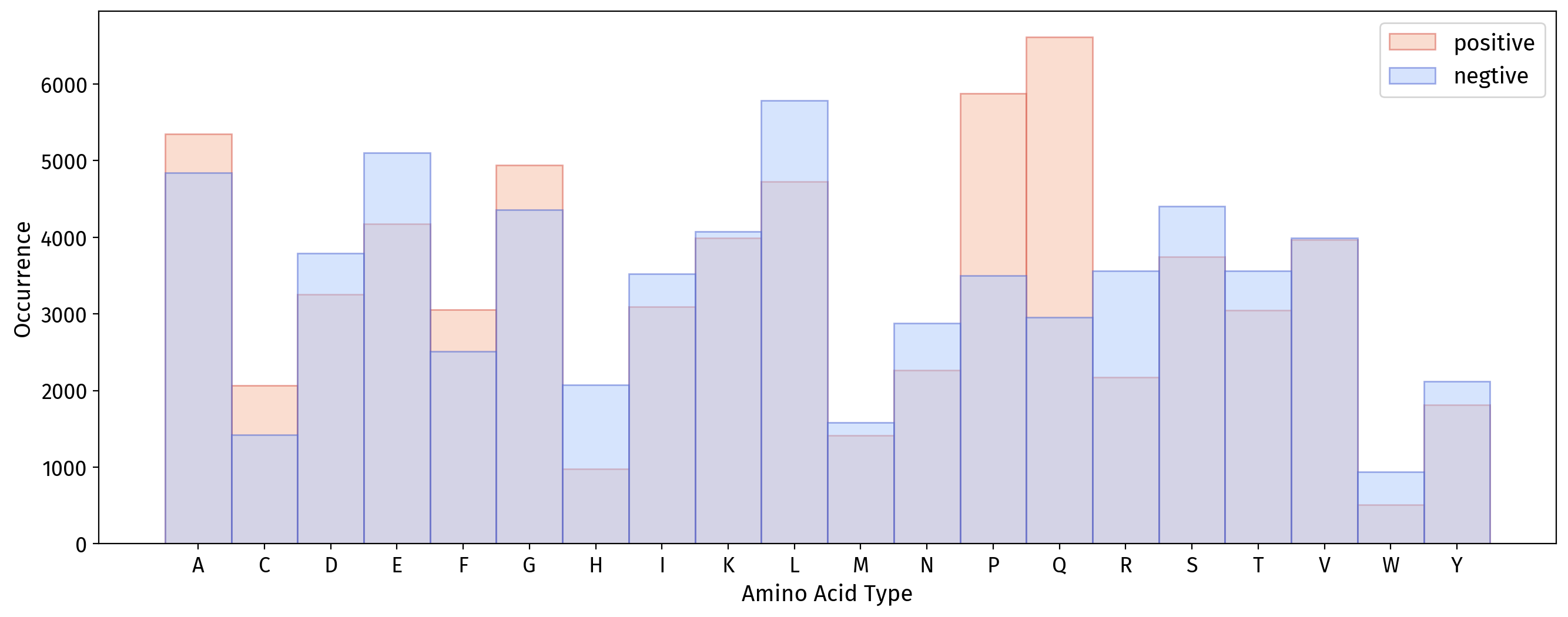}
   \captionof{figure}{Amino acid distribution comparison between positive and negative samples for allergen dataset.}
\label{fig:aa_dist_allergen}
\end{center}

\begin{center} 
\centering
\includegraphics[width=0.8\textwidth]{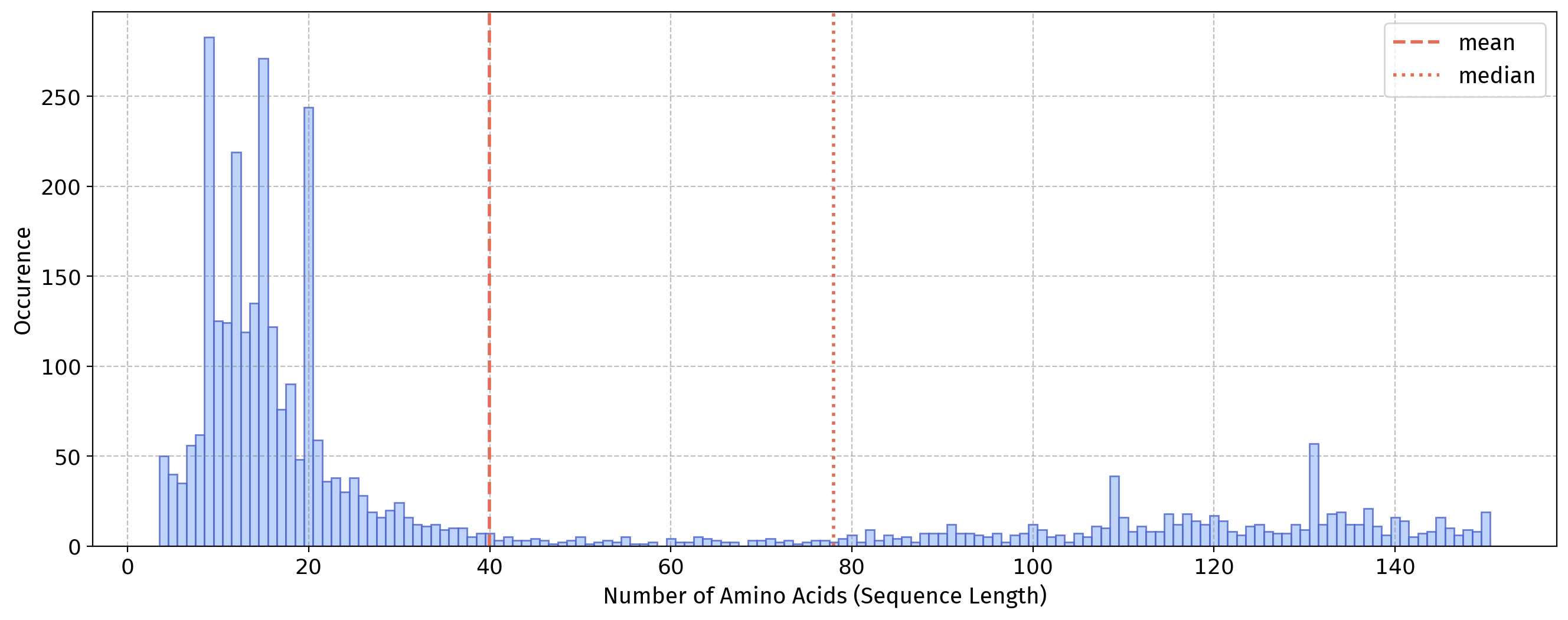}
   \captionof{figure}{Length distribution of allergen dataset.}
\label{fig:length_dist_allergen}
\end{center}

\begin{center} 
\centering
\includegraphics[width=0.8\textwidth]{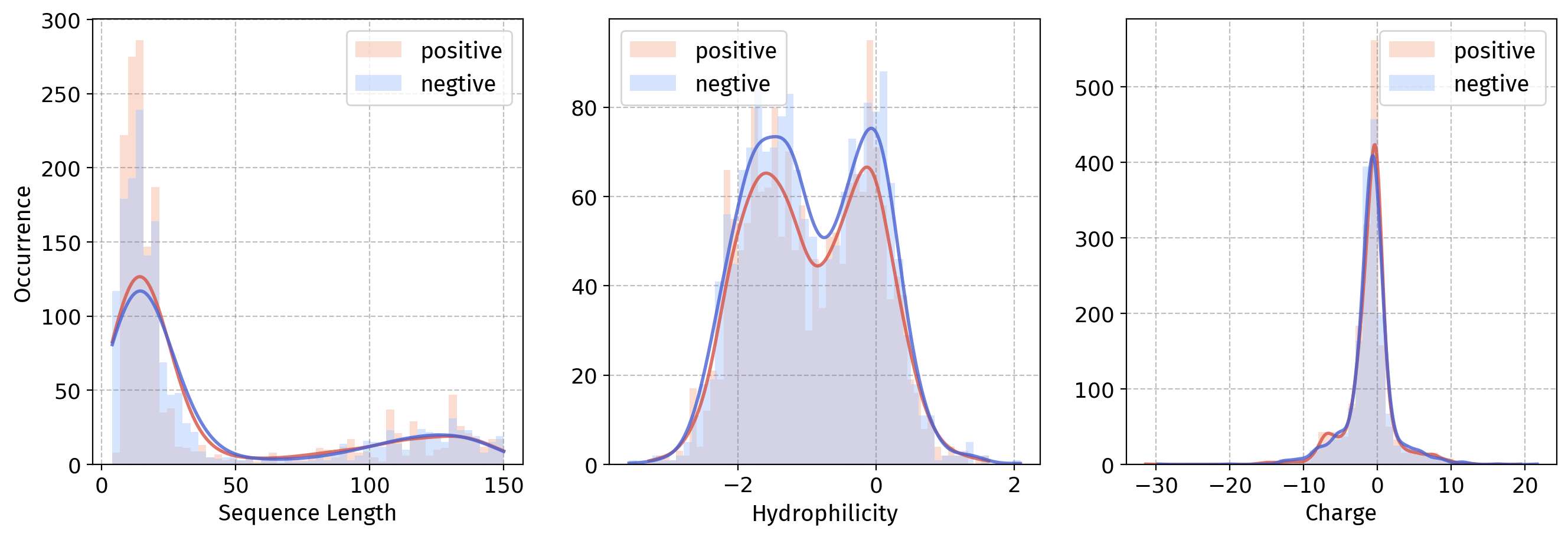}
   \captionof{figure}{Property comparison between positive and negative samples for allergen dataset.}
\label{fig:property_comp_allergen}
\end{center}

\subsubsection{hemolytic\_hc50}
\fielditem{Property and Application} 
The hemolytic\_hc50 dataset reports hemolytic activity as HC50 values (the concentration causing 50\% lysis of red blood cells). This quantitative dataset allows dose–response modeling and helps assess the severity of hemolytic potential in peptide candidates.

\fielditem{Data Source} 
The dataset is sourced from ~\citet{rathore2025prediction}. The study collectes 3,147 peptides from DBAASP and 560 peptides from the Hemolytik database~\citep{singh2025hemolytik2}, whose HC50 values are available. Peptides containing non-natural amino acids and peptides containing less than six residues are removed. In cases where a peptide sequence has multiple HC50 values or a range of HC50 values, the average of these values is computed. The mean activity measure represents the overall hemolytic activity of the peptide under various experimental conditions. By averaging, the model captures the general behavior of the peptide's hemolytic activity rather than specific instances. HC50 values are standardized by converting them into a uniform measurement unit ($\mu$M). Following this, these HC50 values are transformed into pHC50 values.

\fielditem{Dataset Statistics} 
The dataset contains 1,926 datapoints with sequences ranging from 6 to 39 amino acids (average length 18.40) in length.

\textbf{Task: Regression; Split: Cold-start; Evaluation: MAE}

\begin{center} 
\centering
\includegraphics[width=0.8\textwidth]{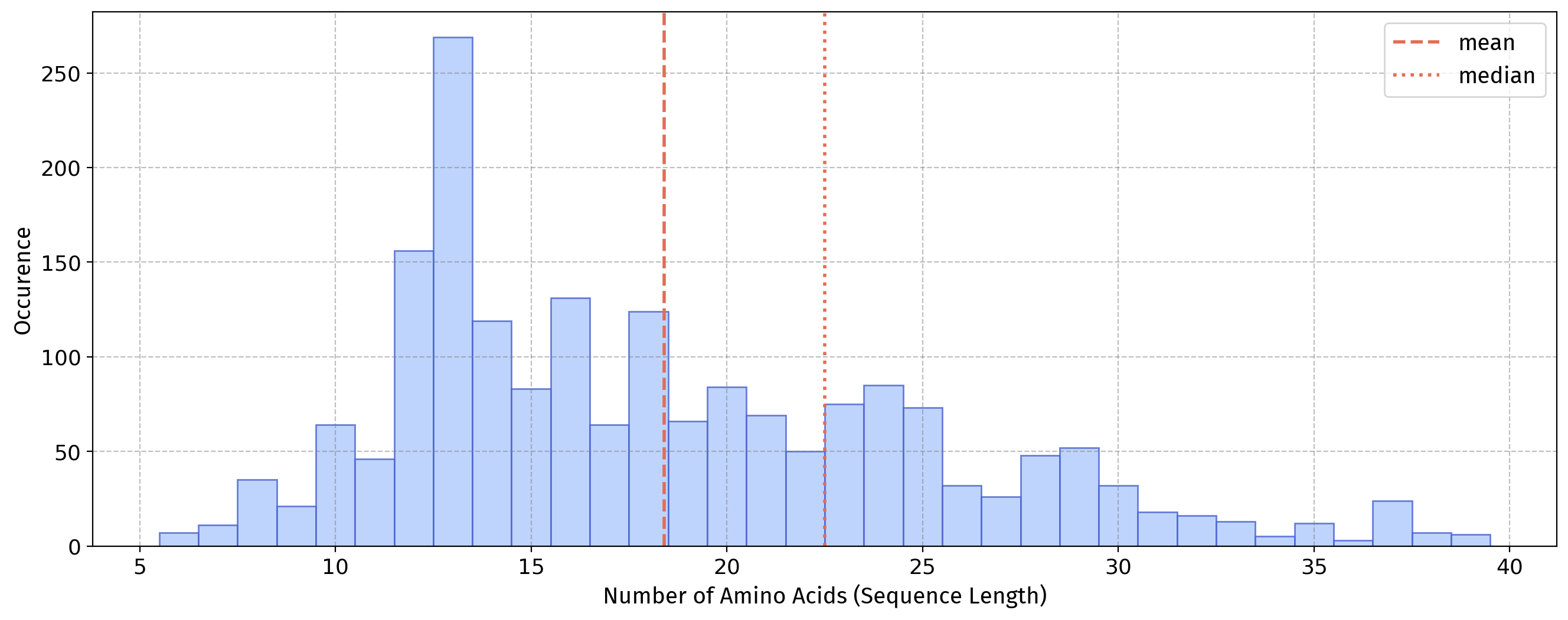}
   \captionof{figure}{Length distribution of hemolytic hc50 dataset.}
\label{fig:length_dist_hemolytic_hc50}
\end{center}

\begin{center} 
\centering
\includegraphics[width=0.8\textwidth]{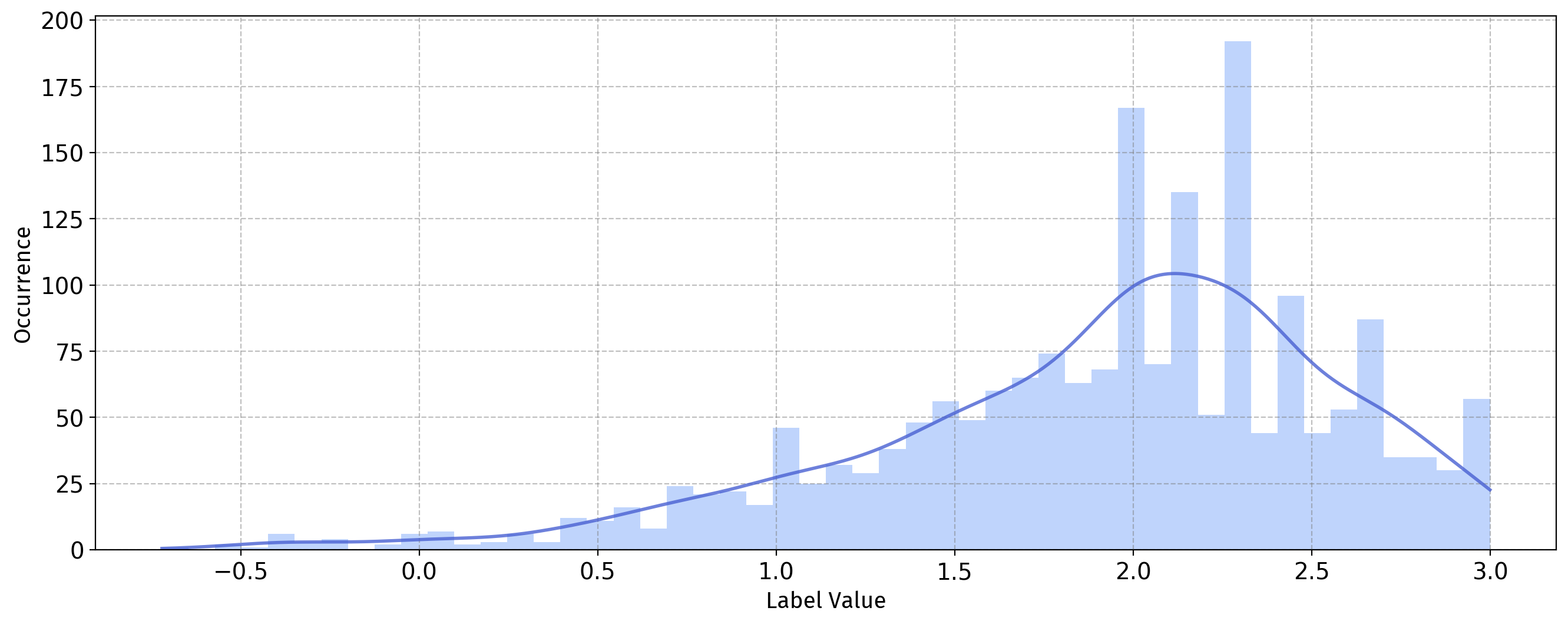}
   \captionof{figure}{Label distribution of hemolytic HC50 dataset.}
\label{fig:label_dist_hemolytic_hc50}
\end{center}

\begin{center} 
\centering
\includegraphics[width=0.8\textwidth]{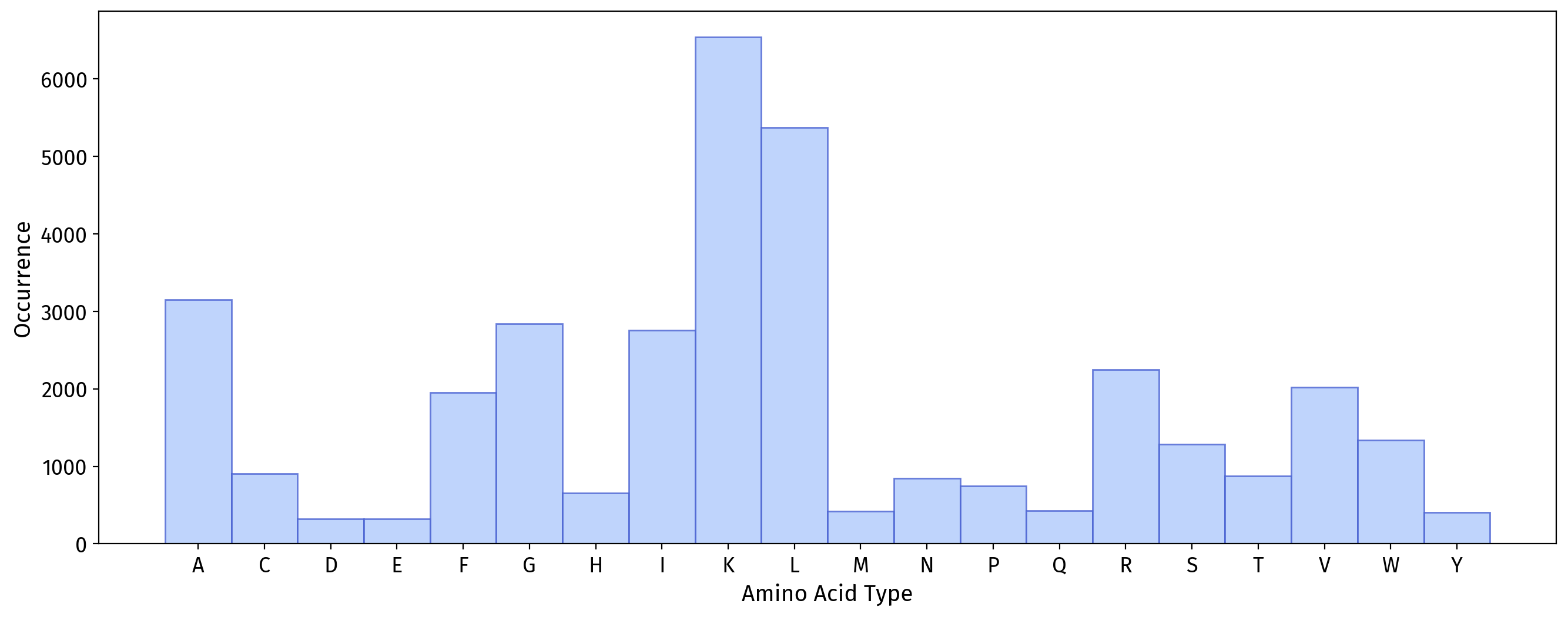}
   \captionof{figure}{Amino acid distribution of hemolytic hc50 dataset.}
\label{fig:aa_dist_hemolytic_hc50}
\end{center}

\subsubsection{hemolytic}
\fielditem{Property and Application} 
The hemolytic dataset evaluates the ability of peptides to cause red blood cell lysis, serving as an indirect indicator of blood compatibility and systemic safety. It integrates sequences with hemolytic annotations from multiple sources, providing a reference for identifying peptides with undesirable cytotoxicity.

\fielditem{Data Source} 
Data sources include: (1) Natural peptides stored in Hemolytik 2~\citep{singh2025hemolytik2}: Data in Hemolytik 2 were manually collected from published literature and various databases. Hemolytic peptides are included in the database if those are found to be evaluated experimentally using hemolysis assay. The database gives each sequence a label (possibly "hemolytic", "low-hemolytic" and "Non-hemolytic"). Since we want to construct a relatively broad classification dataset for preliminary screening of sequences that may have hemolytic properties, "hemolytic" and "low-hemolytic" are defined as "hemolytic", and "Non-hemolytic" is defined as "non-hemolytic". Some sequences in the dataset have conflicting labels of "hemolytic", "low-hemolytic" and "Non-hemolytic". For sequences with conflicting labels, majority voting is used (the most frequent label is used as the final label; sequences with equal positive/negative label counts are removed); (2) Since 100 $\mu$M is usually used as a threshold for distinguishing compound biological activity in machine learning modeling, we additionally include sequences from hemolytic\_hc50 with HC50 <100 $\mu$M as positive samples; (3) Hemolytic peptides recorded in Peptipedia. Duplicates are removed from different sources of positive samples.

\paragraph{Experiment Negative Samples} 
Negative samples are samples from source (1) that are labeled as "non-hemolytic" after processing.

\fielditem{Dataset Statistics} 
The dataset contains 4,512 datapoints with sequences ranging from 2 to 144 amino acids (average length 23.69) in length.

\textbf{Task: Classification; Split: Hybrid; Evaluation: ROC-AUC}

\begin{center} 
\centering
\includegraphics[width=0.8\textwidth]{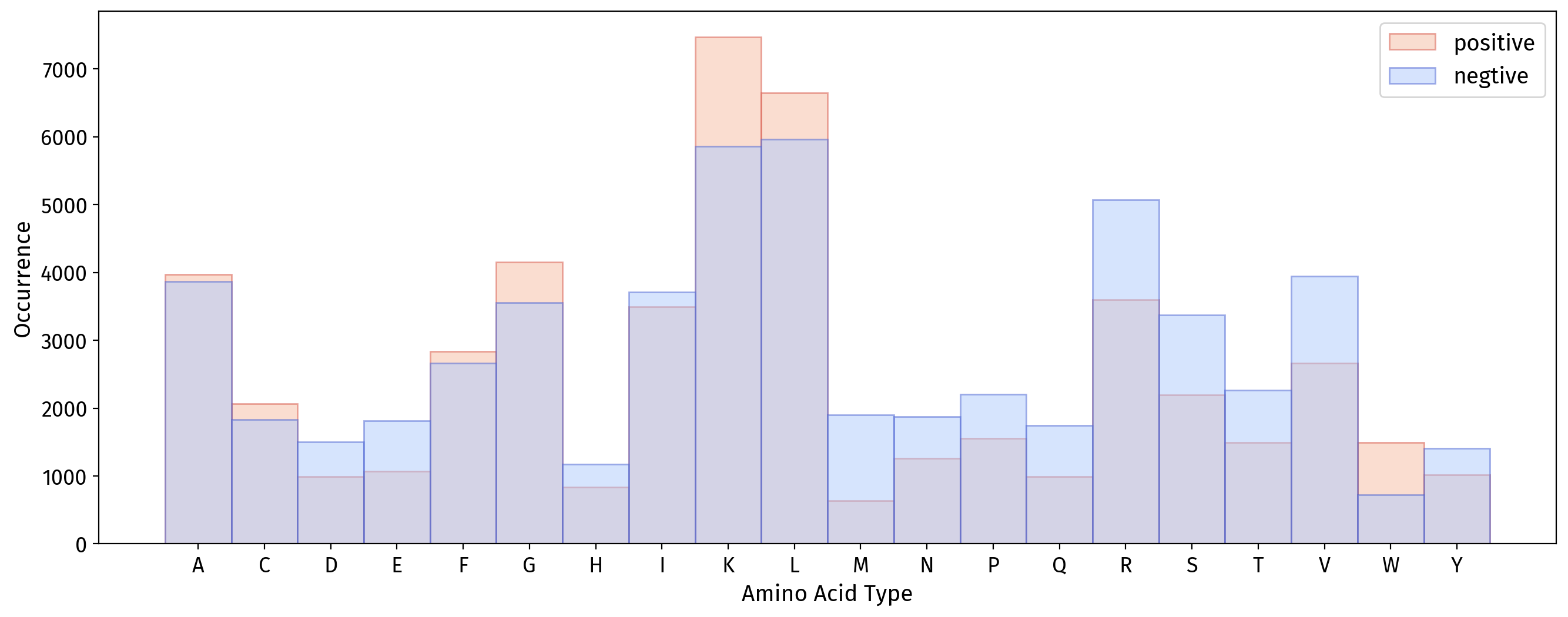}
   \captionof{figure}{Amino acid distribution comparison between positive and negative samples for hemolytic dataset.}
\label{fig:aa_dist_hemolytic}
\end{center}

\begin{center} 
\centering
\includegraphics[width=0.8\textwidth]{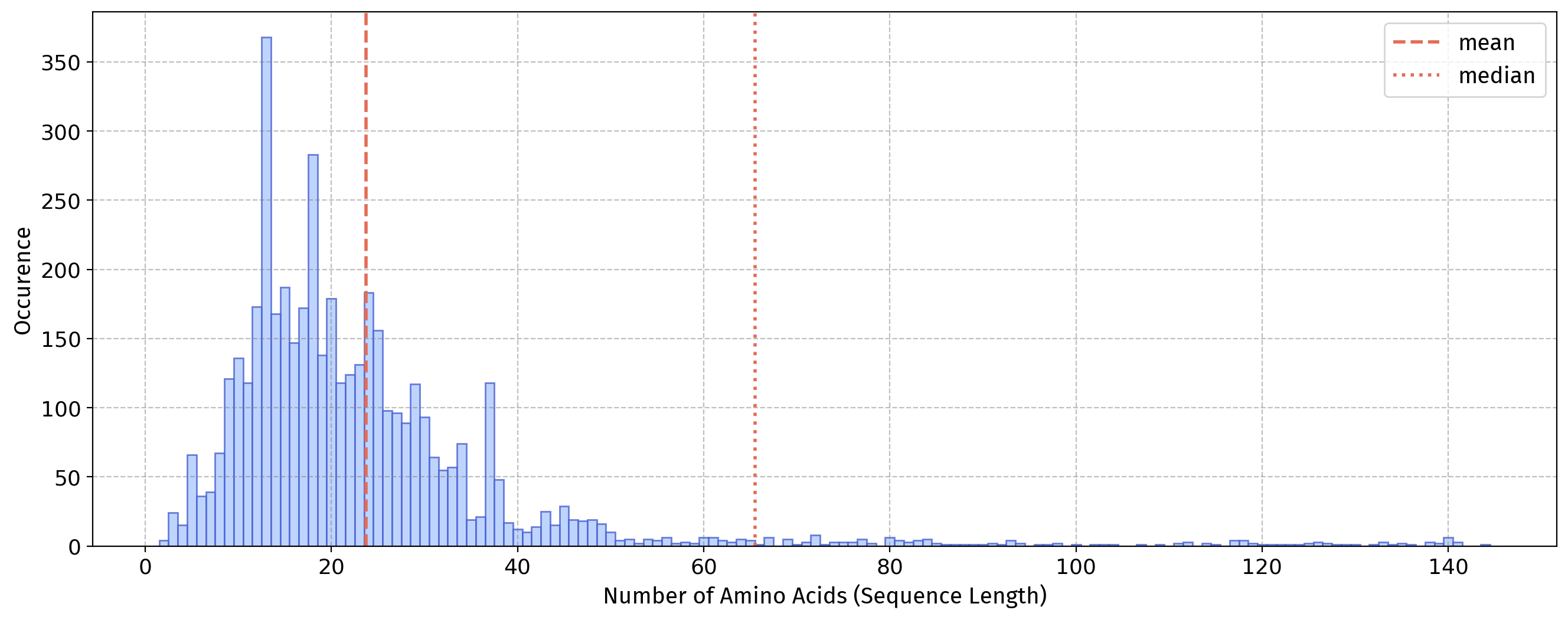}
   \captionof{figure}{Length distribution of hemolytic dataset.}
\label{fig:length_dist_hemolytic}
\end{center}

\begin{center} 
\centering
\includegraphics[width=0.8\textwidth]{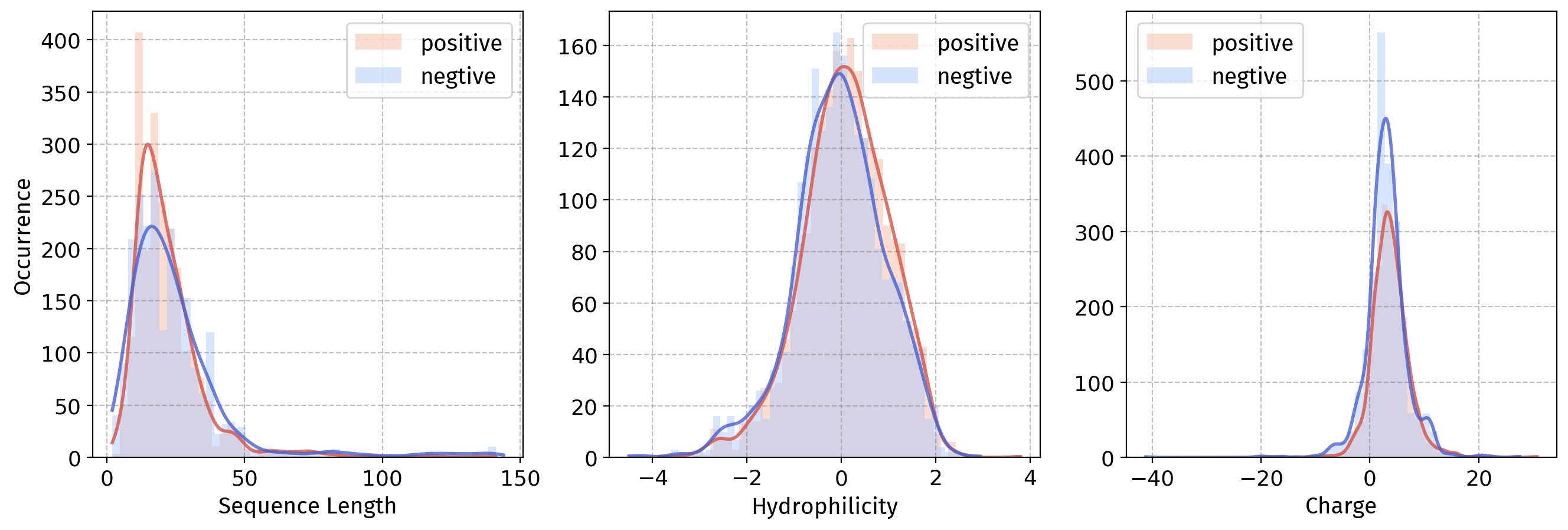}
   \captionof{figure}{Property comparison between positive and negative samples for hemolytic dataset.}
\label{fig:property_comp_hemolytic}
\end{center}

\subsubsection{nc-hemolytic}
\fielditem{Property and Application} 
The nc-hemolytic dataset extends the hemolytic dataset to include non-canonical amino acids, providing a comprehensive dataset for evaluating the hemolytic activity of non-canonical amino acids.

\fielditem{Data Source} 
The dataset sourced from Hemolytik 2~\citep{singh2025hemolytik2}, The determination of data labels refers to the hemolytic dataset.

\fielditem{Dataset Statistics} 

\textbf{Task: Classification; Split: ECFP-based; Evaluation: ROC-AUC}

\begin{center} 
\centering
\includegraphics[width=0.8\textwidth]{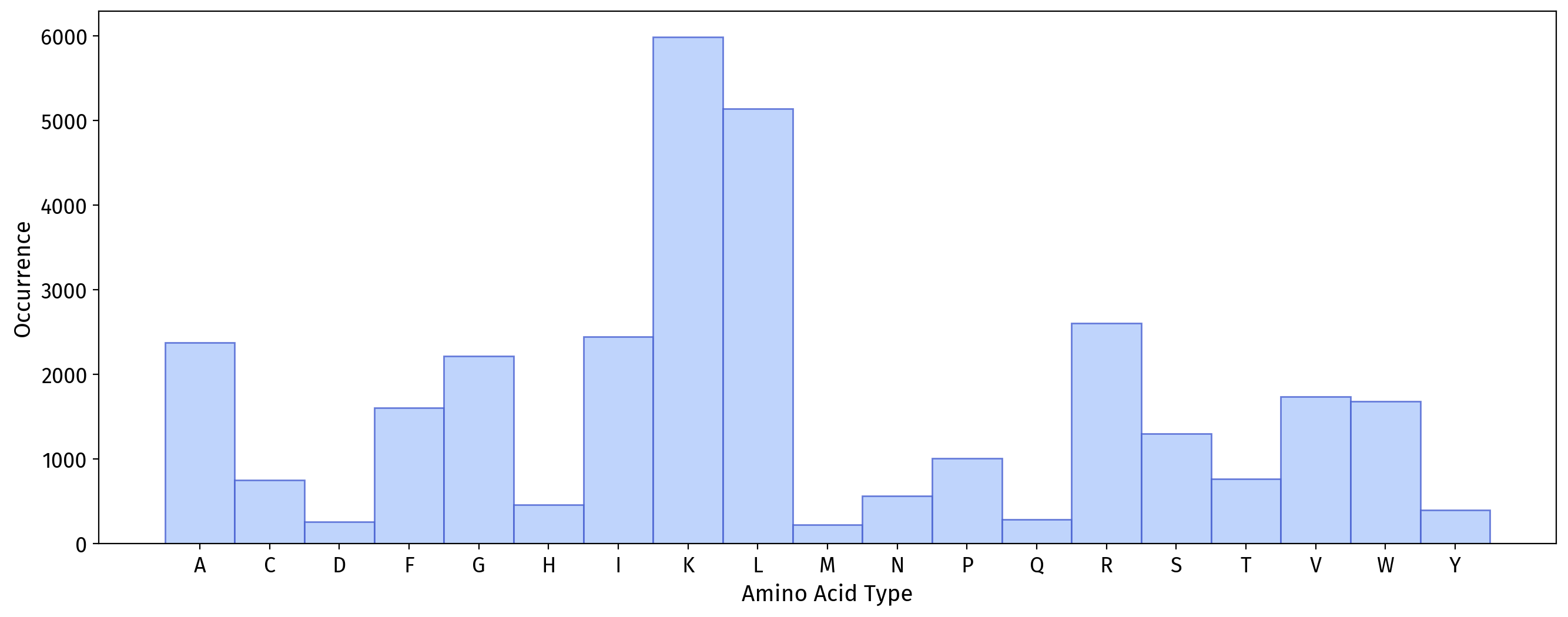}
   \captionof{figure}{Canonical Amino acid distribution comparison between positive and negative samples for nc-hemolytic dataset.}
\label{fig:aa_dist_nc-hemolytic}
\end{center}

\begin{center} 
\centering
\includegraphics[width=0.8\textwidth]{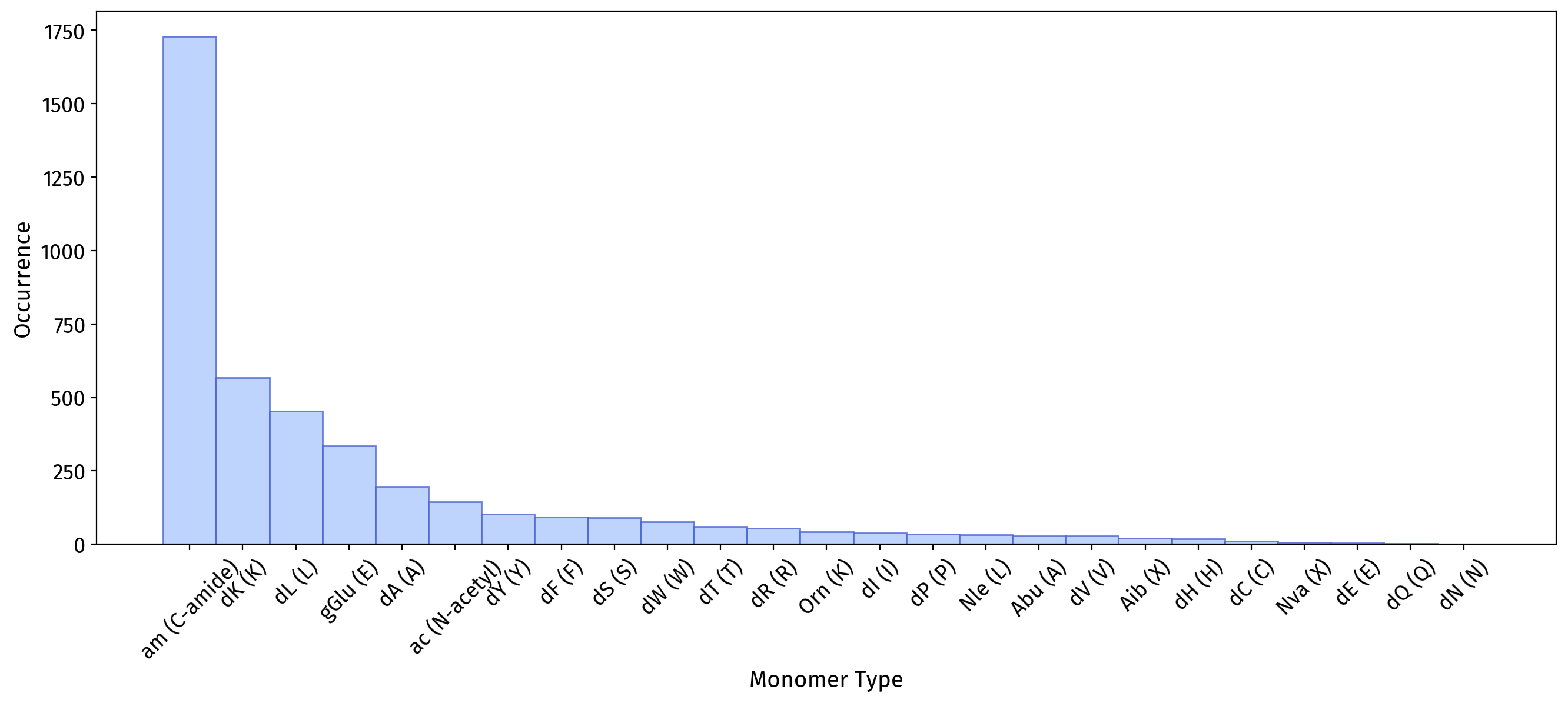}
   \captionof{figure}{non-canonical Amino acid distribution comparison between positive and negative samples for nc-hemolytic dataset.}
\label{fig:ncaa_dist_nc-hemolytic}
\end{center}

\begin{center} 
\centering
\includegraphics[width=0.8\textwidth]{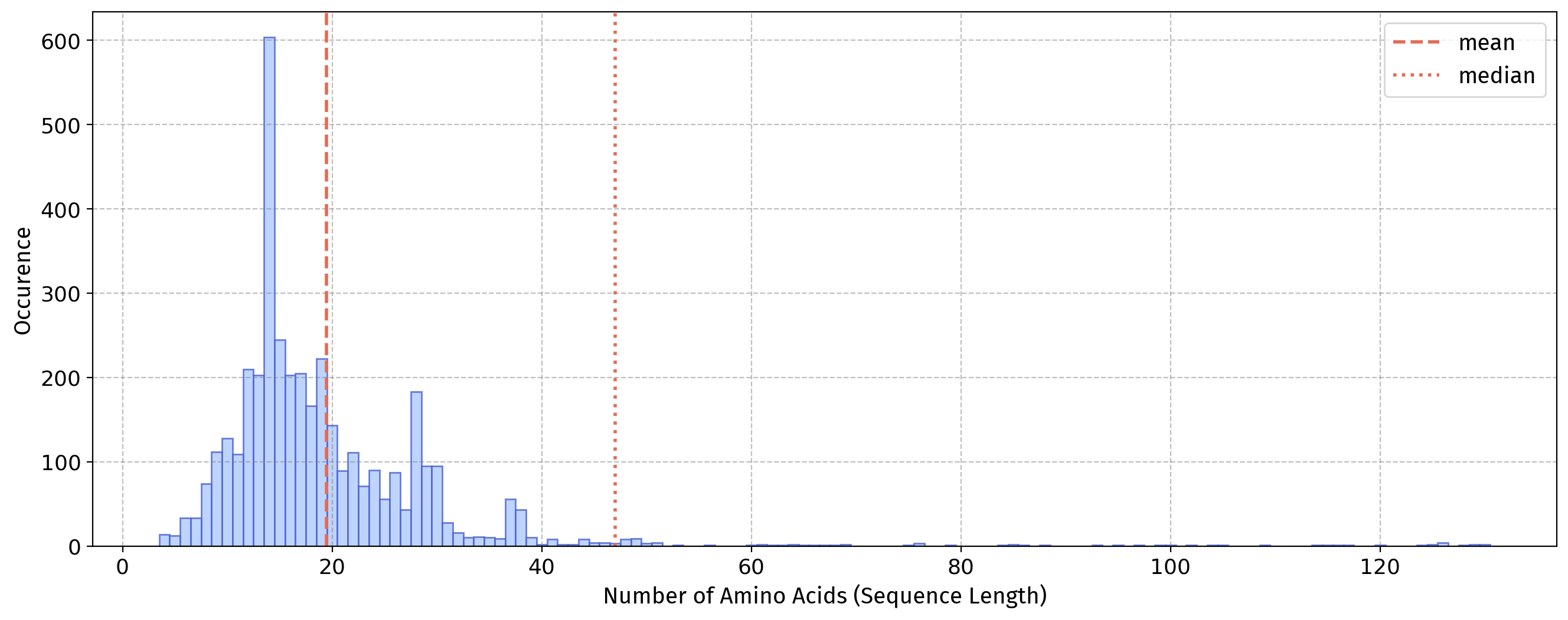}
   \captionof{figure}{Length distribution of nc-hemolytic dataset.}
\label{fig:length_dist_nc-hemolytic}
\end{center}

\subsubsection{neurotoxin}
\fielditem{Property and Application} 
The neurotoxin dataset consists of peptides with neurotoxic activity, which disrupt nervous system function by blocking ion channels or interfering with neurotransmitter release. These peptides can impair neuronal signaling and cause severe adverse effects, making them critical safety liabilities in therapeutic development.

\fielditem{Data Source} 
The dataset is sourced from Peptipedia.

\fielditem{Dataset Statistics} 
The dataset contains 3,506 datapoints with sequences ranging from 7 to 138 amino acids (average length 39.24) in length.

\textbf{Task: Classification; Split: Hybrid; Evaluation: ROC-AUC}

\begin{center} 
\centering
\includegraphics[width=0.8\textwidth]{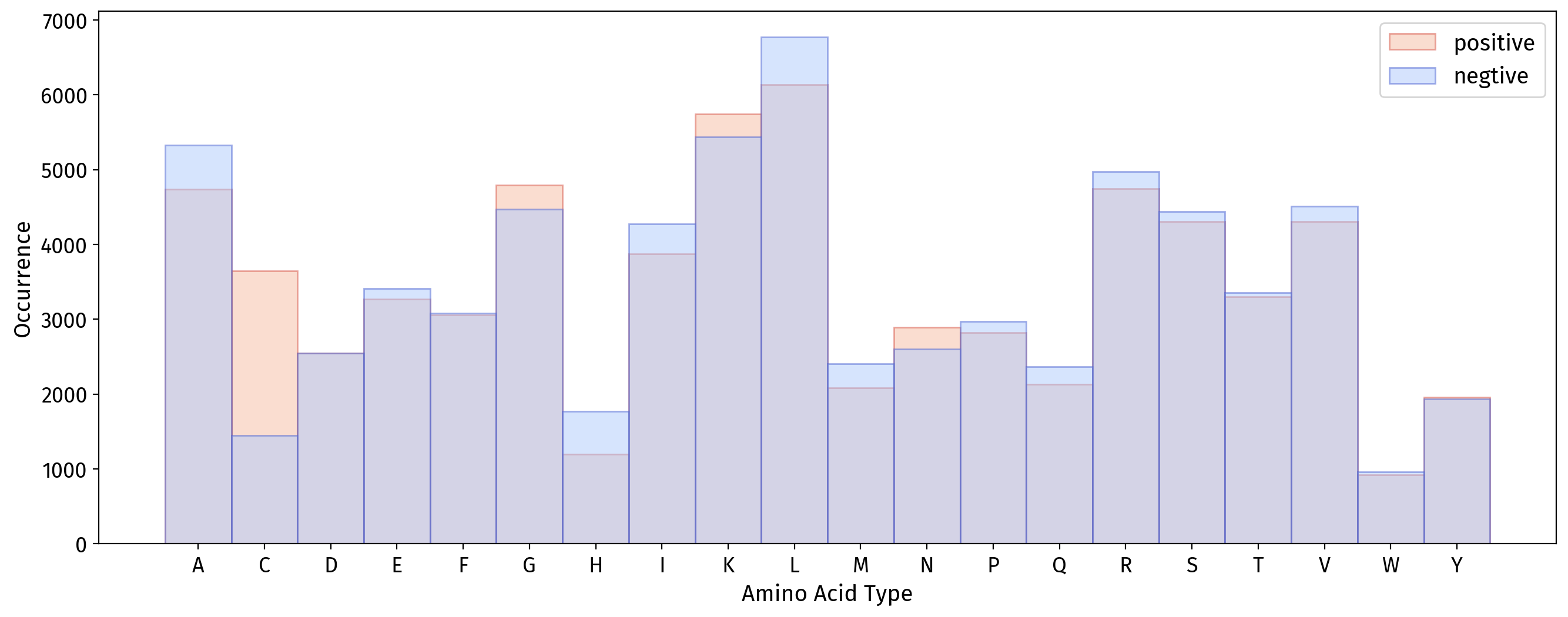}
   \captionof{figure}{Amino acid distribution comparison between positive and negative samples for neurotoxin dataset.}
\label{fig:aa_dist_neurotoxin}
\end{center}

\begin{center} 
\centering
\includegraphics[width=0.8\textwidth]{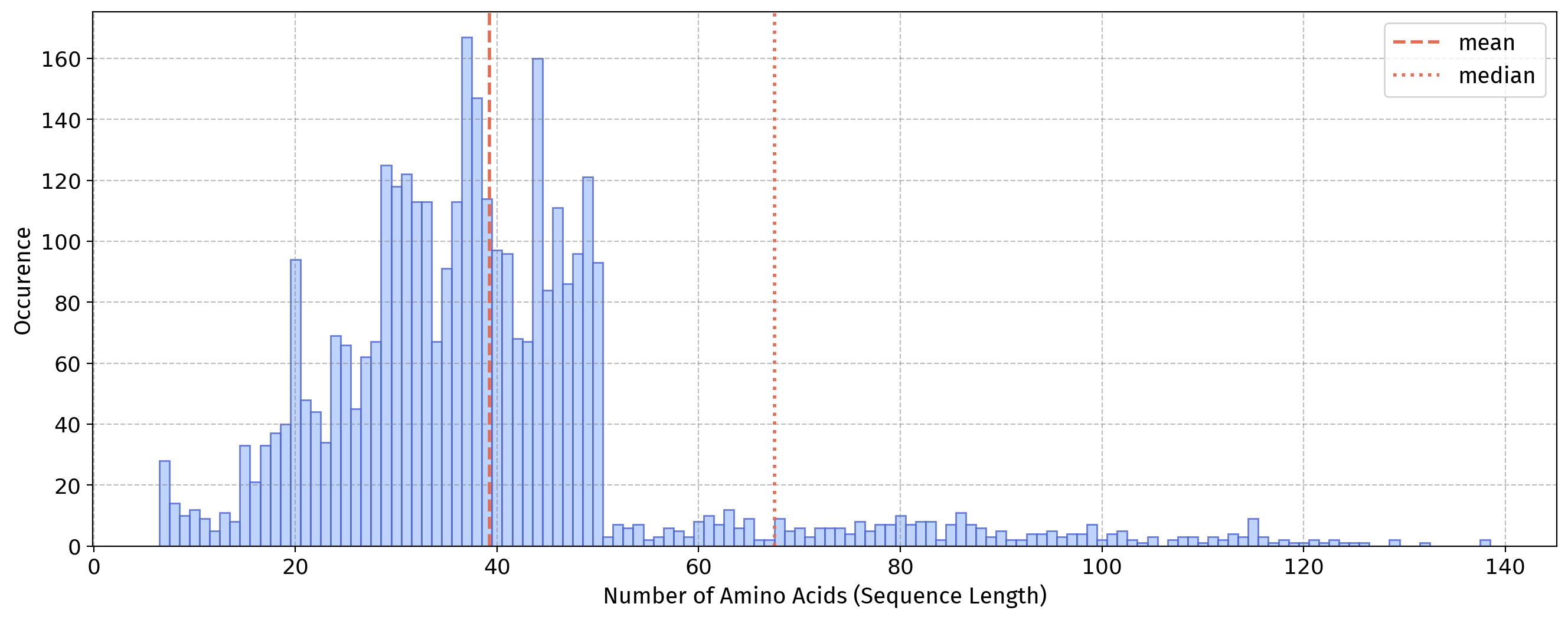}
   \captionof{figure}{Length distribution of neurotoxin dataset.}
\label{fig:length_dist_neurotoxin}
\end{center}

\begin{center} 
\centering
\includegraphics[width=0.8\textwidth]{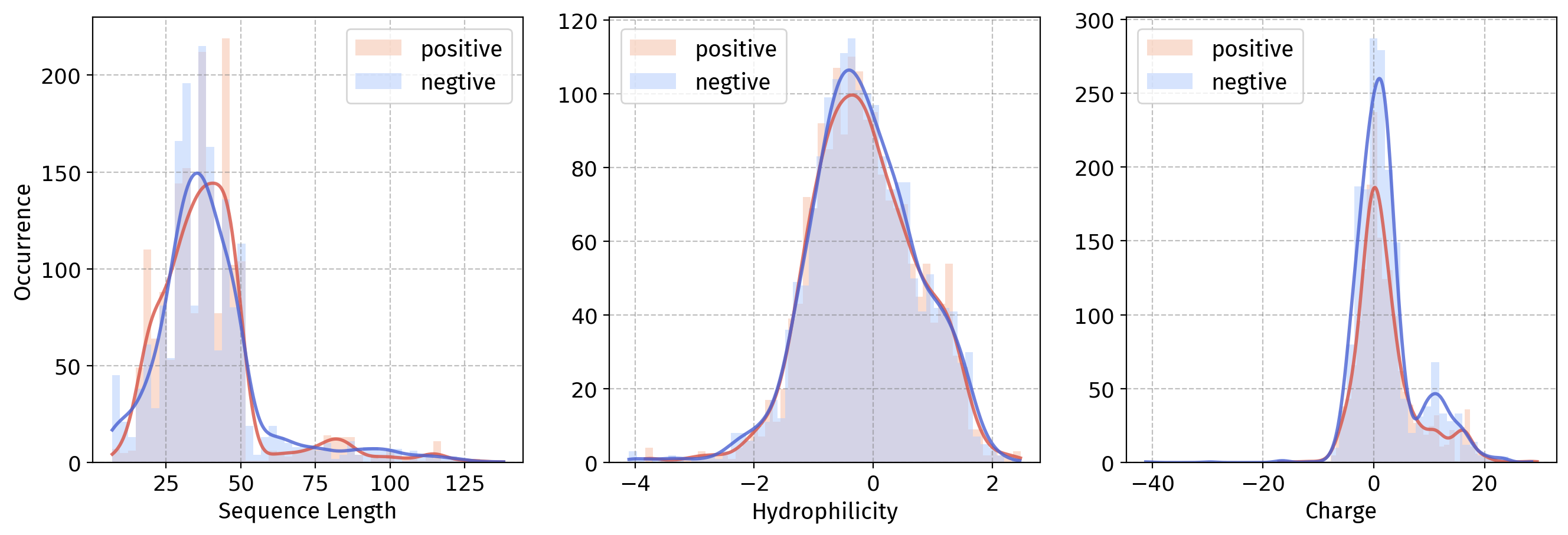}
   \captionof{figure}{Property comparison between positive and negative samples for neurotoxin dataset.}
\label{fig:property_comp_neurotoxin}
\end{center}

\subsubsection{toxicity}
\fielditem{Property and Application} 
This dataset specifically refers to UniProt keyword 0800 annotations for peptides. These toxins are produced by animals (e.g., snakes, scorpions, spiders, cone snails), plants, fungi, and pathogenic bacteria. They act through diverse mechanisms—including neurotoxicity and ion channel disruption—and represent a broad spectrum of natural defense and predation strategies.

\fielditem{Data Source} 
The dataset is sourced from ~\cite{wang2025integrating}. Positive samples consist of experimentally validated toxic peptides collected from three publicly available databases: ConoServer~\cite{kaas2008conoserver}, ArachnoServer~\cite{pineda2018arachnoserver}, and SwissProt. In the SwissProt database, toxic peptides are identified using the keyword "KW-0800". These toxic peptides range in length from 10 to 50 amino acids. After removing duplicate sequences across the three databases, a total of 3,992 toxic peptides are obtained. To further reduce model bias caused by high sequence similarity, CD-HIT is employed to remove sequences with more than 90\% similarity within both the toxic and nontoxic peptide sets. This process results in a final set of 1,932 toxic peptides as positive samples. Since neurotoxic peptides are mechanistically a type of toxicity, neurotoxin peptides from Peptipedia are merged.

\fielditem{Dataset Statistics} 
The dataset contains 4,408 datapoints with sequences ranging from 7 to 138 amino acids (average length 38.41) in length.

\textbf{Task: Classification; Split: Hybrid; Evaluation: ROC-AUC}

\begin{center} 
\centering
\includegraphics[width=0.8\textwidth]{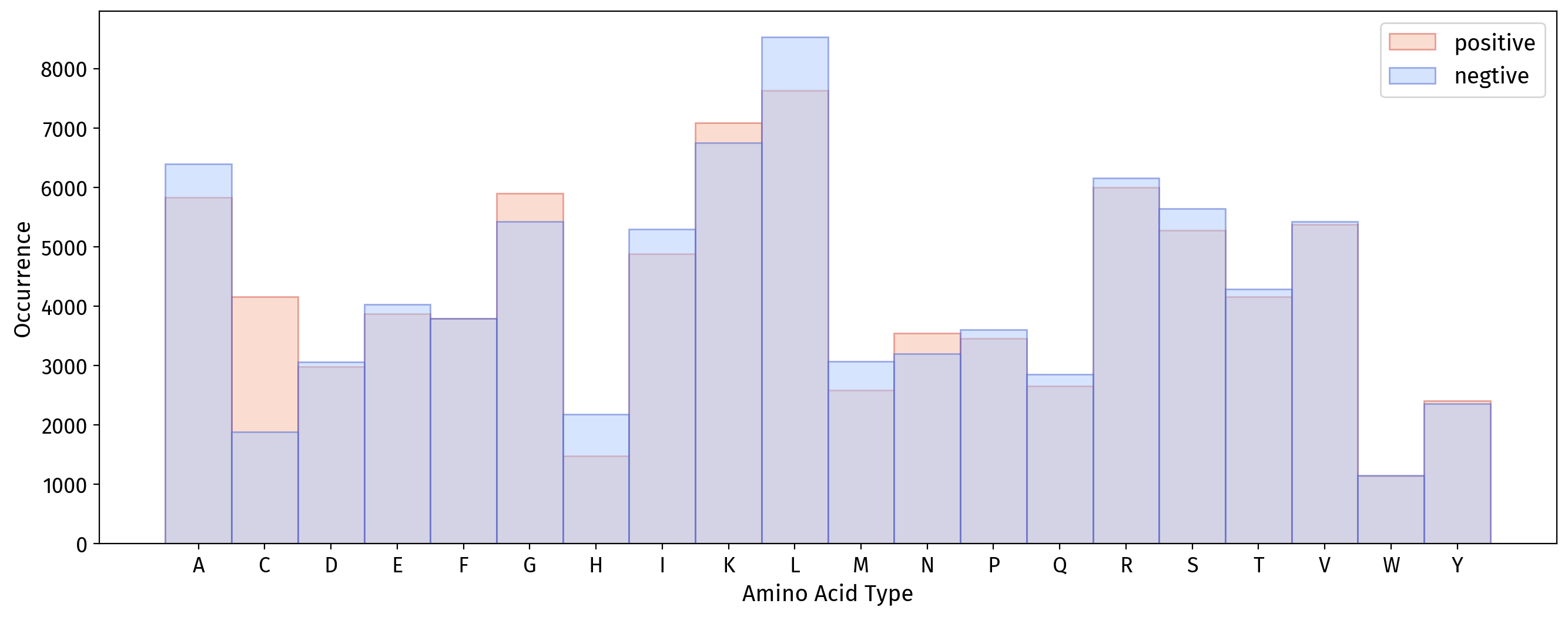}
   \captionof{figure}{Amino acid distribution comparison between positive and negative samples for toxicity dataset.}
\label{fig:aa_dist_toxicity}
\end{center}

\begin{center} 
\centering
\includegraphics[width=0.8\textwidth]{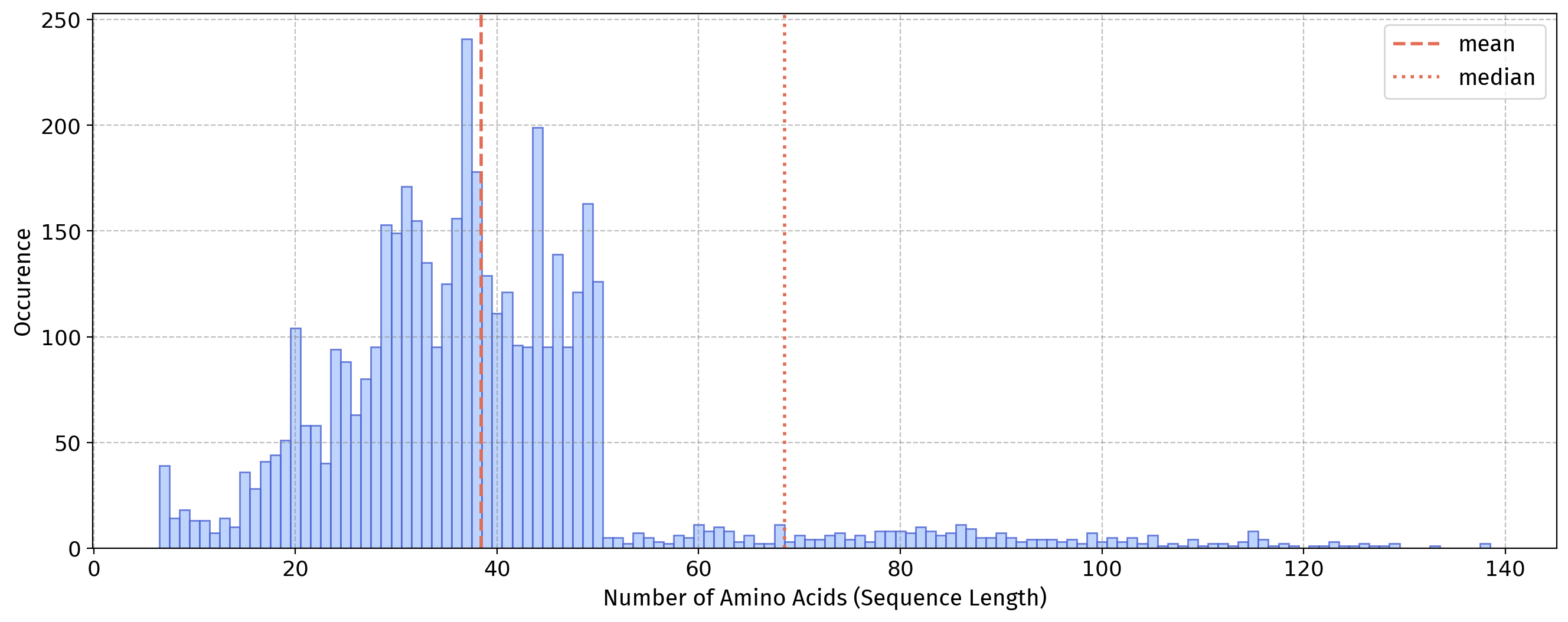}
   \captionof{figure}{Length distribution of toxicity dataset.}
\label{fig:length_dist_toxicity}
\end{center}

\begin{center} 
\centering
\includegraphics[width=0.8\textwidth]{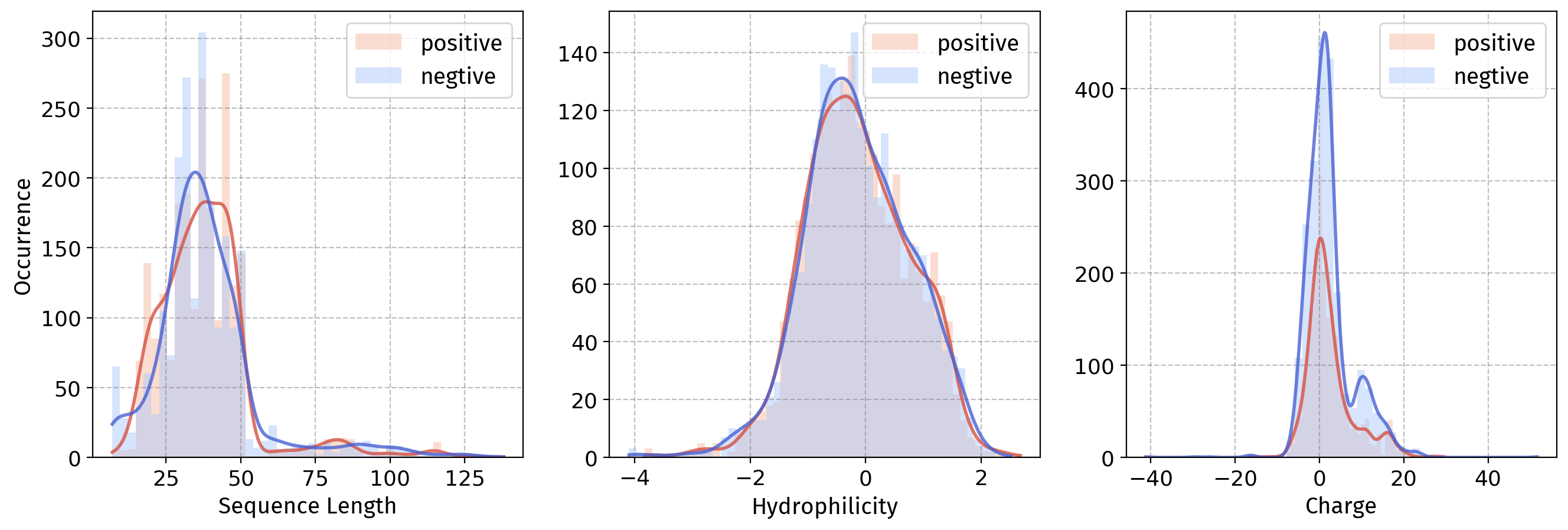}
   \captionof{figure}{Property comparison between positive and negative samples for toxicity dataset.}
\label{fig:property_comp_toxicity}
\end{center}

\subsection{Length Limit}  

In general, peptides are defined as sequences with fewer than 50 amino acids. However, in the literature, some sequences longer than 50 residues are also ambiguously referred to as peptides. Our analysis of the collected datasets revealed the following:  
\begin{enumerate}
    \item In 12 datasets, all sequences have lengths below 50;  
    \item In 18 datasets, fewer than 10\% of the sequences exceed a length of 50;  
    \item In 5 datasets, at least 10\% of the sequences are longer than 50, with maximum lengths reaching up to 150 (including datasets on antifungal, antimicrobial, antiparasitic, anticancer, allergen, neurotoxin, and toxicity).  
\end{enumerate}  

For the benchmark, we believe it is necessary to define a unified length criterion. From the perspective of drug development experts, different research scenarios may focus on peptides of different lengths. For example, peptides with certain properties tend to be relatively short (e.g. ddpiv\_inhibitors), whereas others are generally longer.  

From a model development perspective, sequence length also influences model design. If the focus is only on very short sequences (e.g., fewer than 20 amino acids), methods originally developed for small molecules (e.g., graph neural networks on SMILES at the atomic level) may be applicable. Conversely, for longer sequences, their properties are more protein-like, and pretrained protein models can already capture such features effectively.  

\begin{enumerate}
    \item \texttt{PepBenchData-150}: with a maximum length set to 150, including datasets in which at least 10\% of the sequences are longer than 50;  
    \item \texttt{PepBenchData-50}: with a maximum length set to 50.  
\end{enumerate}


\section{Data Processing Overview}
\label{app:sec:data_processing}

Normalized data processing is crucial for establishing a reliable benchmark. While specific strategies may vary depending on research objectives, the goal of our study is to develop a benchmark that ensures fair and consistent model comparisons and maximizes the distinguishability between different models.

To achieve this goal, we have analyzed potential issues in the data processing procedures of previous studies that result in an oversimplified final dataset. Based on these findings, we systematically design comprehensive processing procedures for each stage. To clearly differentiate the datasets used in previous studies and the official datasets we release, we append the suffix “\textit{raw}” to their names. Table~\ref{tab:rawdatasets} provides a summary of the raw datasets considered in this study.

\begin{table}[h]
\centering
\small
\caption{Summary of the datasets constructed in prior studies employed in analysis.}
\label{tab:rawdatasets}
\resizebox{\textwidth}{!}{%
\begin{tabular}{lclll}
\toprule
\textbf{Dataset Name} & \textbf{Positive Samples} & \textbf{Positive Sample Source} & \textbf{Negative Sampling Method} & \textbf{Used in Articles} \\
\midrule
ace\_inhibitory\_raw   & 1053  & ~\citet{manavalan2019mahtpred}     & Bioactive peptides              & APML~\citet{fernandez2024autopeptideml}                                                \\
antibacterial\_raw     & 6010  & ~\citet{pinacho2021alignment}      & Bioactive peptides              & APML~\citet{fernandez2024autopeptideml}                                              \\
anticancer\_raw        & 861   & ~\citet{Agrawal2021AntiCP}         & Bioactive peptides              & APML~\citet{fernandez2024autopeptideml}                                                \\
antidiabetic\_raw      & 418   & ~\cite{yue2024discovery}          & Random peptides                 & ~\cite{yue2024discovery}                                    \\
antifungal\_raw        & 993   & ~\cite{pinacho2021alignment}      & Bioactive peptides              & APML~\cite{fernandez2024autopeptideml}                                                \\
antimicrobial\_raw     & 22176 & ~\cite{wang2025artificial}        & Inactive peptides               & ~\cite{wang2025artificial}                                \\
antioxidant\_raw       & 436   & ~\cite{olsen2020anoxpepred}       & Bioactive peptides              & APML~\cite{fernandez2024autopeptideml}                                                \\
antiparasitic\_raw     & 301   & ~\cite{zhang2022predapp}          & Bioactive peptides              & APML~\cite{fernandez2024autopeptideml}                                                \\
antiviral\_raw         & 2944  & ~\cite{pinacho2021alignment}      & Bioactive peptides              & APML~\cite{fernandez2024autopeptideml}                                                \\
bbp\_raw               & 119   & ~\cite{dai2021bbppred}            & Bioactive peptides              & APML~\cite{fernandez2024autopeptideml}                                                \\
cpp\_raw               & 1162  & ~\cite{zhang2025pepland}          & Random peptides                 & Pepland~\cite{zhang2025pepland}, PepTune~\cite{tang2025peptune}                                   \\
dppiv\_inhibitors\_raw & 664   & ~\cite{charoenkwan2020idppiv}     & Bioactive peptides              & APML~\cite{fernandez2024autopeptideml}                                                \\
hemolytic\_raw         & 1826  & ~\cite{guntuboina2023peptidebert} & Experimental data               & PepDoRA~\cite{wang2024pepdora}, PepTune~\cite{tang2025peptune} \\
neuropeptide\_raw      & 2425  & ~\cite{bin2020prediction}         & Bioactive peptides              & APML~\cite{fernandez2024autopeptideml}                                                \\
nonfouling\_raw        & 3600  & ~\cite{guntuboina2023peptidebert} & Insoluble and hemolytic peptides & PepDoRA~\cite{wang2024pepdora}, PepTune~\cite{tang2025peptune}\\
quorum\_sensing\_raw   & 218   & ~\cite{rajput2015prediction}      & Bioactive peptides              & APML~\cite{fernandez2024autopeptideml}                                                \\  
toxicity\_raw          & 1932  & ~\cite{wei2021atse}               & Bioactive peptides              & APML~\cite{fernandez2024autopeptideml}                                                \\
ttca\_raw              & 592   & ~\cite{charoenkwan2020ittca}      & Bioactive peptides              & APML~\cite{fernandez2024autopeptideml}                                                \\
\bottomrule
\end{tabular}%
}
\end{table}

\section{Construction Pipeline of Classification Datasets}
\label{app:pipeline:sec:cls}

\begin{figure}[h]
\centering
\includegraphics[width=0.8\textwidth]{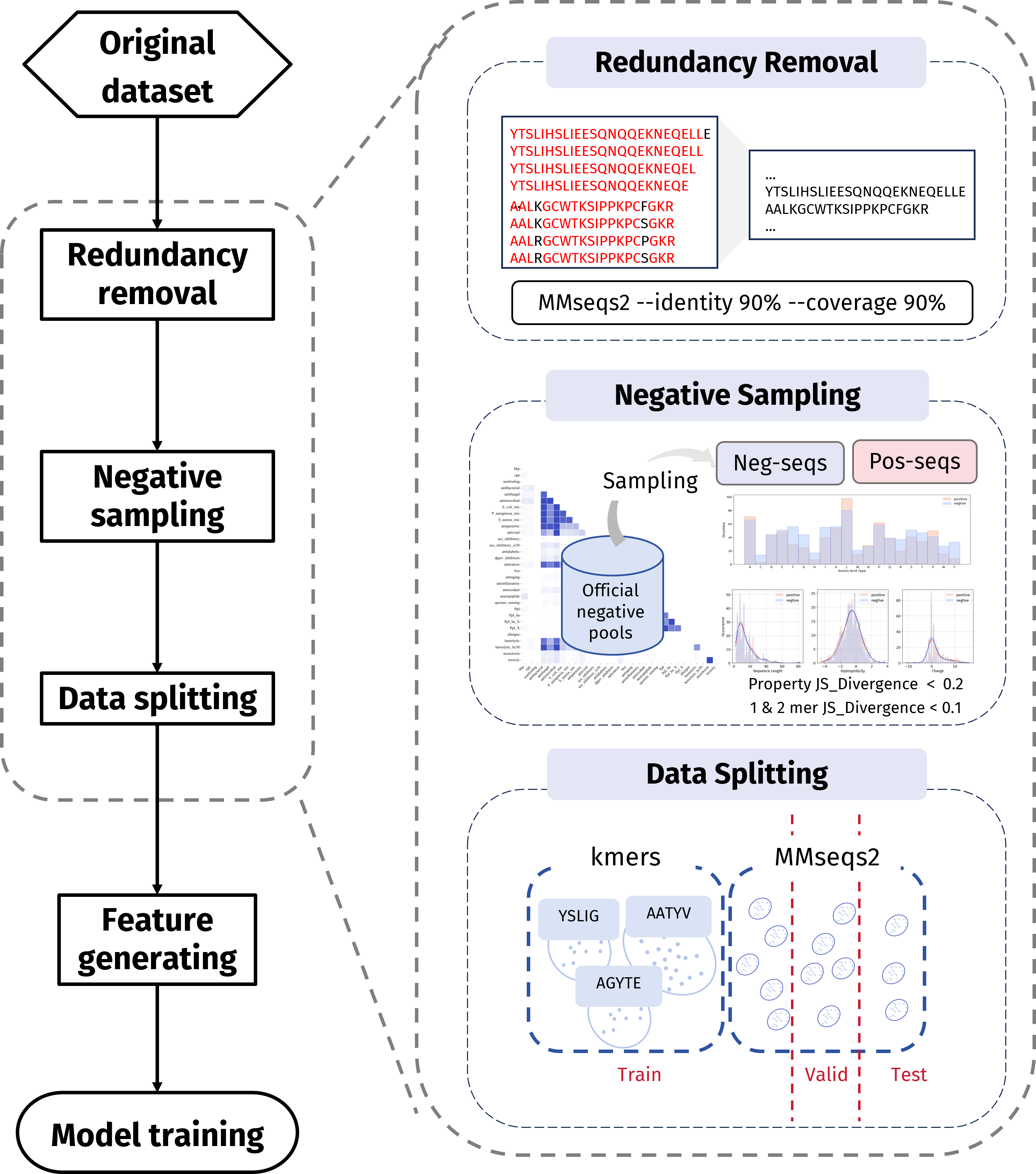}
\caption{Overview of the construction pipeline for peptide classification datasets}
\label{app:dp:fig:cls_overview}
\end{figure}

As illustrated in Figure~\ref{app:dp:fig:cls_overview}, we outline the pipeline for constructing classification datasets. 
In particular, we emphasize three key stages: \textit{sequence redundancy removal}, \textit{negative sampling}, and \textit{dataset splitting}.

\subsection{Sequence Redundancy Removal}

 In typical drug discovery pipelines, researchers often begin with a lead peptide of known bioactivity and systematically generate derivatives through point mutations, truncations, or elongations to probe critical residues and evaluate structural stability. This process inevitably results in \textbf{near-duplicate peptides}, where candidates differ by only a few amino acids. For example, within the \texttt{antifungal} dataset, several positive samples are nearly identical:

\begin{verbatim}
CIKNGNGCQPDGSQGNCCSRYCHKEPGWVAGYCR,1
CIANRNGCQPDGSQGNCCSGYCHKEPGWVAGYCR,1
CIKNGNGCQPNGSQGNCCSGCHKQPGWVAGYCRRK,1
CIKNGNGCQPNGSQGNCCSGYCHKQPGWVAGYCRRK,1
\end{verbatim}
Near-duplicate peptides introduce a substantial source of bias in machine learning models: they promote memorization, exacerbate overfitting to trivial sequence variations, and ultimately compromise generalization. Consequently, redundancy removal constitutes a critical preprocessing step for ensuring reliable performance estimation and for improving the robustness of downstream models. Nevertheless, this issue is often overlooked in existing studies, thereby leading to inflated and potentially misleading performance. As summarized in Table~\ref{tab:dedup_performance}, there are 9 benchmark datasets analyzed in this work exhibit considerable redundancy, with rates exceeding 5\% at a 90\% sequence identity threshold as determined by MMseqs2. To quantify the impact of such redundancy, we trained and evaluated a fingerprint-based Random Forest (RF) model on both the original and deduplicated datasets. The consistent decline in performance following deduplication demonstrates that the previously reported high performances were, at least in part, driven by model memorization of redundant sequences. This finding underscores the necessity of incorporating redundancy removal into peptide modeling pipelines. Notably, the hemolytic dataset displays an exceptionally high redundancy rate of 47\%. After deduplication, its ROC--AUC decreases by 17.39\%, highlighting the pronounced and detrimental influence of redundant sequences on model performance.

\begin{table}[h]
\centering
\small
\caption{Dataset redundancy and model performance before and after deduplication. 
Deduped\_ratio: fraction of redundant positive sequences removed; 
ROC--AUC: model performance on the original dataset; 
de\_ROC--AUC: performance on the deduplicated dataset; 
Perf\_drop: relative decrease in ROC--AUC after deduplication.}

\label{tab:dedup_performance}
\begin{tabular}{ccccc}
\hline
Dataset           & Deduped\_ratio & ROC--AUC & de\_ROC--AUC & Perf\_drop (\%) \\
\hline
hemolytic\_raw         & 0.470          & 0.805$_{(0.022)}$ & 0.665$_{(0.037)}$ & 17.391 \\
dppiv\_inhibitors\_raw & 0.050          & 0.830$_{(0.049)}$ & 0.799$_{(0.024)}$ & 3.735 \\
antibacterial\_raw     & 0.290          & 0.889$_{(0.006)}$ & 0.858$_{(0.017)}$ & 3.487 \\
cpp\_raw               & 0.270          & 0.929$_{(0.005)}$ & 0.904$_{(0.018)}$ & 2.691 \\
quorum\_sensing\_raw   & 0.140          & 0.930$_{(0.024)}$ & 0.913$_{(0.026)}$ & 1.828 \\
anticancer\_raw        & 0.190          & 0.938$_{(0.017)}$ & 0.918$_{(0.010)}$ & 2.132 \\
antifungal\_raw        & 0.280          & 0.955$_{(0.019)}$ & 0.944$_{(0.019)}$ & 1.151 \\
antiviral\_raw         & 0.280          & 0.956$_{(0.004)}$ & 0.942$_{(0.007)}$ & 1.464 \\
antimicrobial\_raw     & 0.370          & 0.977$_{(0.003)}$ & 0.962$_{(0.002)}$ & 1.535 \\
\hline
\end{tabular}%
\end{table}

Here, we applied \texttt{MMseqs2} clustering with a set of stringent parameters specifically tailored for short peptides (sequence length $\leq 50$ amino acids). 
The minimum sequence identity was set to \texttt{0.90} (\texttt{--min-seq-id 0.9}), thereby excluding peptide pairs that differed by more than $\sim$10\% of residues. 
To ensure reliable alignments, we imposed a coverage threshold of \texttt{0.9} (\texttt{-c 0.9}) in combination with \texttt{--cov-mode 0}, which enforced that the alignment spanned at least 90\% of the longer sequence in each pair and thus prevented spurious clustering of short fragments into longer peptides. 
Low-complexity masking was disabled (\texttt{--mask 0}) to preserve functionally relevant Cys- and Lys-rich motifs. 
For alignment, we employed the most informative mode (\texttt{--alignment-mode 3}) and defined sequence identity according to the aligned region without terminal gaps (\texttt{--seq-id-mode 2}), which provides robustness to local insertions and deletions. 
Furthermore, we enabled high sensitivity (\texttt{-s 8}) and set the number of sampled kmers per sequence to \texttt{50} (\texttt{--kmer-per-seq 50}) to maximize the detection of near-duplicate peptides. 
This configuration effectively clustered highly similar variants (e.g., sequences differing by single-residue substitutions, insertions, or deletions) while retaining genuinely distinct peptides.  

In our sequence filtering strategy, the use of \texttt{--cov-mode 0} together with \texttt{-c 0.9} required that at least 90\% of both the query and target sequences be aligned. Since the aligned region cannot exceed the length of the shorter sequence, this condition implicitly ensures that the shorter sequence spans at least 90\% of the longer one, thereby preventing spurious alignments of short fragments to only a small portion of longer peptides. 
In addition, by combining \texttt{--min-seq-id 0.9} with \texttt{--seq-id-mode 2}, sequence identity was defined as the number of identical residues divided by the length of the shorter sequence. 
Under these parameters, an alignment must exhibit at least 90\% identity across the shorter sequence, permitting at most $\lfloor 0.1 \times L_b \rfloor$ mismatches or gaps for a sequence of length $L_b$ (see Table~\ref{tab:mmseqs2_params}). 
Taken together, these stringent criteria ensure that only sequence pairs of comparable length and high similarity were retained for downstream analysis.

\begin{table}[h]
\centering
\small
\caption{MMseqs2 clustering parameters for redundancy removal in peptide datasets.}
\label{tab:mmseqs2_params}
\resizebox{\textwidth}{!}{
\begin{tabular}{lll}
\toprule
\textbf{Parameter} & \textbf{Value} & \textbf{Meaning / Rationale} \\
\midrule
\texttt{--min-seq-id} & 0.90 & Minimum sequence identity (90\%); removes peptides differing by $>$10\%. \\
\texttt{-c}           & 0.9  & Coverage threshold (90\%). \\
\texttt{--cov-mode}   & 0    & Coverage relative to longer sequence; avoids clustering of short fragments into longer peptides. \\
\texttt{--mask}       & 0    & Disable low-complexity masking; retain functional motifs (e.g., Cys- and Lys-rich). \\
\texttt{--alignment-mode} & 3 & Full alignment with start, end, and identity reported. \\
\texttt{--seq-id-mode} & 2   & Identity based on aligned region without terminal gaps; robust to local indels. \\
\texttt{-s}           & 8    & High sensitivity search; minimizes missed near-duplicates. \\
\texttt{--kmer-per-seq} & 50 & Extract up to 50 kmers per sequence; maximizes sensitivity for short peptides. \\
\bottomrule
\end{tabular}%
}
\end{table}

We apply redundancy removal to the canonical classification datasets using \texttt{MMseqs2} with the parameters described above. 
Table~\ref{tab:dataset_statistics} summarizes the dataset sizes before and after redundancy filtering.

\begin{table}[ht]
    \centering
    \small
    \caption{Dataset Statistics: Original and Filtered Sample Counts}
    \label{tab:dataset_statistics}
    \begin{tabular}{ccccccl}
    \hline
    Dataset   Name    & Origin\_pos & New\_pos & Filt\_ratio & Exp\_neg & New\_exp\_neg & \multicolumn{1}{c}{Filt\_ratio} \\
    \hline
    ace\_inhibitory   & 1833        & 1780      & 0.029       & --       & --        & --                              \\
    allergen          & 2405        & 1677      & 0.303       & --       & --        & --                              \\
    antiaging         & 282         & 279       & 0.011       & --       & --        & --                              \\
    antibacterial     & 21220       & 15838     & 0.254       & --       & --        & --                              \\
    anticancer        & 9022        & 6926      & 0.232       & --       & --        & --                              \\
    antidiabetic      & 1599        & 1514      & 0.053       & 75       & 75        & \multicolumn{1}{c}{0.000}       \\
    antifungal        & 11087       & 8349      & 0.247       & --       & --        & --                              \\
    antiinflamatory   & 3902        & 3875      & 0.007       & --       & --        & --                              \\
    antimicrobial     & 42800       & 30752     & 0.281       & --       & --        & --                              \\
    antioxidant       & 1146        & 1121       & 0.022       & 195      & 195       & \multicolumn{1}{c}{0.000}       \\
    antiparasitic     & 6041        & 4316      & 0.286       & --       & --        & --                              \\
    antiviral         & 5210        & 4134      & 0.207       & --       & --        & --                              \\
    bbp               & 358         & 336       & 0.061       & 13       & 13        & \multicolumn{1}{c}{0.000}       \\
    cpp               & 1318        & 1162      & 0.118       & --       & --        & --                              \\
    dppiv\_inhibitors & 650         & 634       & 0.025       & 85       & 85        & \multicolumn{1}{c}{0.000}       \\
    hemolytic         & 3096        & 2256      & 0.271       & 544      & 475       & \multicolumn{1}{c}{0.127}       \\
    neuropeptide      & 5336        & 4627      & 0.133       & --       & --        & --                              \\
    neurotoxin        & 2509        & 1753      & 0.301       & --       & --        & --                              \\
    nonfouling        & 3600        & 3600      & 0.000       & --       & --        & --                              \\
    quorum\_sensing   & 265         & 245       & 0.075       & --       & --        & --                              \\
    toxicity          & 2509        & 2204      & 0.122       & --       & --        & --                              \\
    ttca              & 592         & 591       & 0.002       & --       & --        & --                              \\
    \hline
    \end{tabular}
    \end{table}

\subsection{Negative Sampling}
\subsubsection{Limitations of Existing Negative Sampling Methods}
In peptide classification tasks, most datasets lack experimentally validated negative samples, and there is currently no consensus on negative sampling strategies. Existing approaches for negative sampling often suffer from inherent limitations that bias the training process:

\begin{enumerate}
    \item \textbf{Random sampling or truncation:} Negative samples are constructed by randomly selecting peptide fragments from proteins or databases such as UniProt or SwissProt \citep{Agrawal2021AntiCP,pinacho2021alignment,bin2020prediction}. However, the sampled sequences often differ substantially from biologically active peptides, which may lead models to learn the trivial distinction between \textit{active peptides vs. random sequences} rather than features truly associated with the target property.
    
    \item \textbf{Sampling from other bioactive peptides:} Some studies treat peptides with unrelated activities as negative samples. For example, antimicrobial peptides have been used as negatives for anticancer peptides~\citep{Agrawal2021AntiCP}, or antifungal/antiviral peptides as negatives for antibacterial peptides \citep{fernandez2025generalize}. The potential issue is that models may learn activity-specific differences between the two classes instead of generalizable rules relevant to the target property. APML~\citep{fernandez2024autopeptideml} proposed drawing negative peptides from a database containing multiple bioactivities, but no explicit rule was defined to exclude all activities potentially overlapping with the property of interest. Even if the target property itself is excluded, mislabeling of peptides as negatives (i.e., false negatives) is likely for mechanistically related properties (e.g., antimicrobial vs. anticancer), increasing the risk of biased learning.
\end{enumerate}

Another important source of bias introduced by negative sampling is uncontrolled differences between positive and negative samples. In some studies, overall distribution differences (e.g., peptide length, physicochemical properties) between positive and negative samples are not controlled. In the \textit{nonfouling\_raw} and \textit{ttca\_raw} datasets, the length distributions of positive and negative samples differ markedly, while in the \textit{antimicrobial} dataset, the isoelectric points of positive and negative samples show significant differences (~\ref{fig:pos_neg}). Such disparities allow models to rely on shortcut features rather than true biological signals.

\subsubsection{Biologically Informed and Distribution-Controlled Negative Sampling Strategy}

Based on the aforementioned issues, we propose two guiding principles for negative sampling:  
\begin{enumerate}
    \item Principle 1: The differences between positive and negative samples should reflect generalizable biological properties rather than dataset-specific artifacts (e.g., clear distributional differences, inherent contrasts between active and inactive peptides).  
    \item Principle 2·: False negatives should be avoided as much as possible.  
\end{enumerate} 
Based on these principles, we introduce \textbf{Biologically Informed and Distribution-Controlled Negative Sampling (BDNegSamp)}. The procedure involves three key steps:

\paragraph{Step 1: Construction of the Biologically Informed Negative Sample Pool}

The initial negative pool is defined as all collected bioactive peptides. Using bioactive sequences as candidate negatives avoids the trivial distinction between \textit{active peptides vs. random sequences}, which is a major limitation of random sampling. However, not all bioactive datasets are suitable for inclusion. Many peptide classes are mechanistically related, which raises the risk of introducing false negatives if they are sampled as negatives. For example, antimicrobial peptides (AMPs) and anticancer peptides (ACPs) are strongly correlated: some ACPs have been directly identified from antimicrobial datasets. Therefore, using AMPs as negatives for ACPs would violate Principle~2 by introducing mislabeled samples. To mitigate this risk, we exclude datasets with high biological relatedness to the target activity when constructing the negative pool. This strategy ensures that negative samples are still bioactive peptides, thereby enhancing generalizability, while minimizing the chance of sampling potential positives.  

\begin{figure}[h]
\centering
\includegraphics[width=0.8\linewidth]{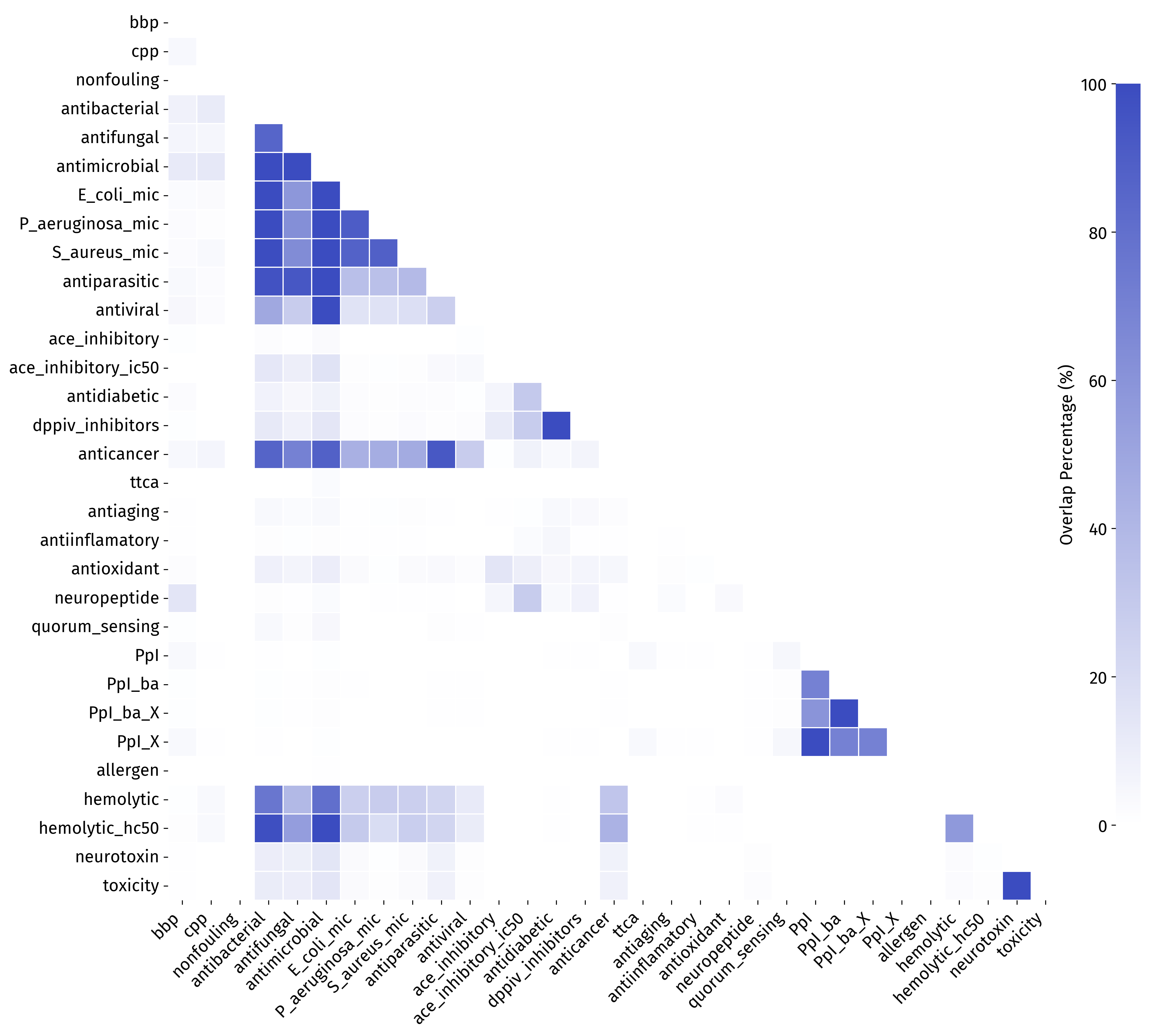}
\caption{Dataset sequence overlap heatmap. The color intensity represents the proportion of shared sequences relative to the smaller of the two datasets.}
\label{fig:nature_sequence_overlap_heatmap}
\end{figure}

To identify task-relatedness between datasets, we adopt two complementary approaches:

\begin{enumerate}
    \item \textbf{Expert knowledge.} Based on mechanistic insights and existing literature, we grouped peptide datasets into three clusters of highly related activities:
    \begin{itemize}
        \item \textit{Membrane-interaction-related activities:} antimicrobial, anticancer, hemolytic, and transmembrane peptides. Their biological functions are all closely associated with interactions at the cell membrane interface.  
        Specifically:  
        (i) AMPs with strong membrane-disruptive ability often also display hemolytic activity, with up to 70\% of AMPs reported as toxic to human red blood cells~\citep{qiu2025amplyze}, largely due to their cationic and amphipathic structures that insert into and perturb membranes.  
        (ii) AMPs and ACPs overlap substantially, as many ACPs share physicochemical features with AMPs, enabling them to disrupt cancer cell membranes~\citep{roudi2017antimicrobial}.  
        (iii) AMPs also resemble transmembrane peptides (TMPs) in their ability to insert into lipid bilayers. At sufficient concentrations, AMPs may adopt TMP-like orientations to form transient pores, although TMPs typically form stable transport channels rather than inducing direct lysis~\citep{pirtskhalava2013transmembrane}.  

        \item \textit{Glucose-regulating peptides:} including DPP-IV inhibitors, antidiabetic, antioxidant, and anti-inflammatory peptides. These activities are mechanistically interrelated. For instance, DPP-IV inhibition represents a primary mechanism underlying antidiabetic peptides. Antioxidant and anti-inflammatory peptides can also contribute to glucose homeostasis by reducing oxidative stress and suppressing pro-inflammatory pathways that impair insulin secretion and action~\citep{leo2016biopeptides}.  

        \item \textit{Neuroactive peptides:} peptides that penetrate the blood-brain barrier or interact with neural receptors/ion channels. Their structural features (e.g., high positive charge, amphipathic motifs) promote transmembrane transport and often result in overlapping neuro-related activities.  
    \end{itemize}

    \item \textbf{Sequence overlap statistics.} To complement expert knowledge, we compute the overlap ratio between datasets, defined as the number of shared sequences divided by the size of the smaller dataset. Datasets with an overlap ratio $>0.05$ are considered related. The overlap analysis is visualized in Figure~\ref{fig:nature_sequence_overlap_heatmap}. In most cases, this statistical assessment is consistent with expert-driven groupings. In cases where datasets show high overlap but lack clear mechanistic explanation, we conservatively treat them as ``highly related,'' thereby excluding them from negative sampling to avoid false negatives.  
\end{enumerate}

\paragraph{Step 2: Redundancy Removal and Similarity Filtering} 

The biologically informed negative pool may still contain a high level of redundancy, where many sequences are highly similar variants of one another. Excessive redundancy reduces the effective diversity of sampled negatives and can bias the training process. To address this, we remove redundant sequences using MMseqs2, applying clustering parameters identical to those summarized in Table~\ref{tab:mmseqs2_params}. This ensures that the remaining pool preserves broad sequence diversity and avoids over-representing particular motifs or sequence families.  

In addition to redundancy reduction, we further filter the negative pool to reduce the risk of false negatives. Based on the general principle that \textit{similar sequences tend to share similar biological properties}, sequences in the negative pool that are highly similar to the positive set are excluded. To achieve this, we apply MMseqs2 with the same clustering parameters, except that the sequence identity threshold is lowered to 0.6 using the \texttt{--min-seq-id} parameter. This threshold is stringent enough to exclude close homologs of positive peptides while retaining more distantly related sequences that are less likely to exhibit the target activity.

\paragraph{Step 3: Distribution-Controlled Sampling and Validation} 

After redundancy removal and similarity filtering, we obtain a refined negative pool. However, if negatives are sampled directly from this pool, they may still differ substantially from positives in overall distributional properties (e.g., sequence length, physicochemical features). Such mismatches would violate Principle~1 by allowing models to exploit trivial dataset-specific artifacts rather than learning biologically meaningful distinctions. Here, we implement a distribution-controlled sampling procedure that explicitly balances positive and negative sets across five key properties: length, net charge, hydrophobicity, amino acid composition, and dipeptide composition. For length, charge, and hydrophobicity, values are discretized into 30 bins and Jensen--Shannon (JS) divergence is constrained below 0.2. For amino acid (1-mer) and dipeptide (2-mer) distributions, stricter thresholds are applied, with JS divergence constrained below 0.05 and 0.15, respectively. This procedure ensures that the sampled negatives are closely matched to the positives in multiple dimensions, reducing the risk of models exploiting trivial distributional artifacts. The results are shown in Table~\ref{tab:js_divergence_summary}.

\begin{table}[ht!]
\centering
\caption{Jensen-Shannon divergence between positive and negative samples after negative sampling for each dataset. \textbf{Length\_js} measures differences in sequence length distributions; \textbf{Charge\_js} quantifies differences in net charge distributions; \textbf{Hydrophobicity\_js} evaluates differences in hydrophobicity distributions; \textbf{1mers\_js} assesses differences in amino acid composition; and \textbf{2mers\_js} measures differences in dipeptide composition.}
\small
\begin{tabular}{lccccc}
\toprule
\textbf{Dataset}           & \textbf{Length\_js} & \textbf{Charge\_js} & \textbf{Hydrophobicity\_js} & \textbf{1mers\_js} & \textbf{2mers\_js} \\
\midrule
ace\_inhibitory   & 0.115      & 0.077      & 0.075              & 0.026     & 0.069     \\
allergen          & 0.249      & 0.075      & 0.054              & 0.014     & 0.039     \\
antiaging         & 0.122      & 0.111      & 0.074              & 0.004     & 0.072     \\
antibacterial     & 0.082      & 0.100      & 0.041              & 0.012     & 0.025     \\
anticancer        & 0.082      & 0.049      & 0.041              & 0.011     & 0.025     \\
antidiabetic      & 0.104      & 0.095      & 0.063              & 0.004     & 0.031     \\
antifungal        & 0.059      & 0.046      & 0.040              & 0.010     & 0.023     \\
antiinflamatory   & 0.123      & 0.081      & 0.037              & 0.004     & 0.010     \\
antimicrobial     & 0.096      & 0.039      & 0.034              & 0.008     & 0.018     \\
antioxidant       & 0.065      & 0.045      & 0.073              & 0.005     & 0.031     \\
antiparasitic     & 0.081      & 0.067      & 0.040              & 0.010     & 0.024     \\
antiviral         & 0.102      & 0.038      & 0.061              & 0.006     & 0.016     \\
bbp               & 0.088      & 0.106      & 0.086              & 0.006     & 0.047     \\
cpp               & 0.184      & 0.199      & 0.145              & 0.016     & 0.060     \\
dppiv\_inhibitors & 0.167      & 0.175      & 0.122              & 0.023     & 0.115     \\
hemolytic         & 0.144      & 0.106      & 0.074              & 0.013     & 0.034     \\
neuropeptide      & 0.074      & 0.045      & 0.033              & 0.007     & 0.022     \\
neurotoxin        & 0.167      & 0.104      & 0.061              & 0.005     & 0.016     \\
nonfouling        & 0.069      & 0.056      & 0.047              & 0.028     & 0.061     \\
quorum\_sensing   & 0.107      & 0.140      & 0.082              & 0.009     & 0.101     \\
toxicity          & 0.178      & 0.093      & 0.077              & 0.004     & 0.013     \\
ttca              & 0.130      & 0.193      & 0.107              & 0.006     & 0.040     \\
\bottomrule
\end{tabular}%
\label{tab:js_divergence_summary}
\end{table}

\begin{figure}[h!]
\centering
\includegraphics[width=0.8\textwidth]{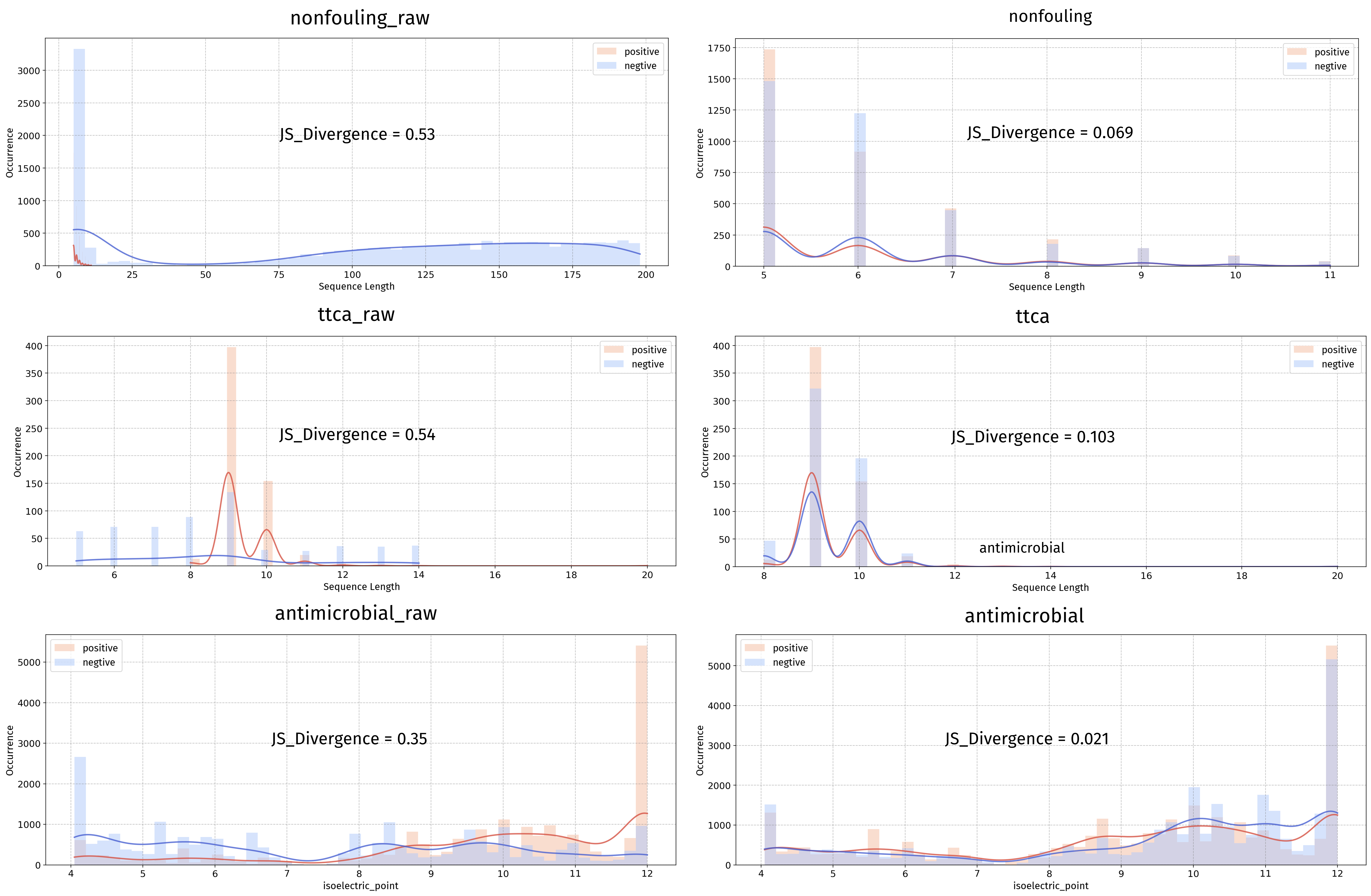}
\caption{Distributional comparison of positive and negative samples before and after sampling.}
\label{fig:pos_neg}
\end{figure}

By enforcing these constraints, the negative set is guaranteed to remain statistically comparable to the positive set across multiple dimensions. Results in Table~\ref{tab:js_divergence_summary} demonstrate that BDNegSamp substantially improves distributional balance compared to the original datasets, particularly for the \texttt{nonfouling}, \texttt{ttca}, and \texttt{antimicrobial} datasets (see Figure~\ref{fig:pos_neg}).

In practice, to realize distribution-controlled sampling we implemented a modular framework that supports multiple distribution-matching strategies. These include random sampling, kernel density estimation (KDE) importance sampling, maximum mean discrepancy (MMD) herding, nearest-neighbor matching, entropy-regularized optimal transport (OT), and moment or histogram-based matching. In addition, hybrid strategies that jointly balance physicochemical properties and k-mer statistics were introduced to further constrain sequence-level distributions. To ensure robustness, BDNegSamp executes multiple strategies in parallel and selects the negative set that achieves the best trade-off between distributional alignment and sample diversity. This flexible design allows BDNegSamp to adaptively select negative samples that not only avoid false negatives but also achieve close alignment to the positive set across multiple distributional dimensions.

In some cases, such as the \texttt{antimicrobial} dataset (30,752 sequences), the bioactive negative pool ($\sim$90,000 sequences in total) is insufficient to satisfy distributional balance constraints. To address this limitation, we adopt a hybrid strategy: additional peptide sequences are drawn from UniProt to expand the candidate pool, and for each sequence length bin in the positive dataset we ensure at least a 10$\times$ coverage of candidate negatives, thereby providing sufficient diversity to achieve balanced sampling. All UniProt-derived sequences are subjected to the same redundancy removal, similarity filtering, and distribution-controlled validation procedures as bioactive negatives, ensuring methodological consistency. This compromise allows BDNegSamp to generate a sufficiently diverse and balanced negative set even for very large datasets, without violating its guiding principles.

\subsection{Data Splitting}
\label{sec:data_splitting}

\subsubsection{Limitations of Existing Splitting Protocols}
\label{sec:limitations_existing_splits}
Prior peptide classification studies typically adopt either random partitioning or homology-based partitioning with \texttt{MMseqs2}. To quantify the impact of these choices, we have evaluated a broad panel of models on several widely used peptide datasets under an 8:1:1 train/validation/test split. The models include three traditional learners have been trained on 1024-bit ECFP6 fingerprints (LightGBM, Random Forest, and XGBoost) and four protein sequence models (ESM2 with 150M and 650M parameters; DPLM with 150M and 650M parameters).

\begin{table}[h]
    \centering  
    \small
    \caption{Model Performance Comparison}
    \label{tab:model_performance} 
    \resizebox{\textwidth}{!}{%
    \begin{tabular}{ccccccccccccccc}
    \toprule
    \multirow{3}{*}{Dataset} & \multicolumn{14}{c}{Models} \\
    \cmidrule(lr){2-15}
                             & \multicolumn{7}{c}{Random Split} & \multicolumn{7}{c}{MMseqs2 Split} \\
    \cmidrule(lr){2-8} \cmidrule(lr){9-15}
                             & LightGBM & RF & XGBoost & ESM2\_150M & ESM2\_650M & DPLM\_150M & DPLM\_650M & LGB & RF & XGB & ESM2\_150M & ESM2\_650M & DPLM\_150M & DPLM\_650M \\
    \midrule
    anticancer               & 0.930 & 0.938 & 0.931 & 0.919 & 0.919 & 0.927 & 0.929 & 0.827 & 0.825 & 0.815 & 0.795 & 0.815 & 0.795 & 0.815 \\
    antifungal               & 0.937 & 0.955 & 0.933 & 0.960 & 0.974 & 0.959 & 0.975 & 0.686 & 0.670 & 0.705 & 0.798 & 0.807 & 0.759 & 0.788 \\
    antimicrobial            & 0.971 & 0.977 & 0.974 & 0.986 & 0.987 & 0.984 & 0.986 & 0.951 & 0.953 & 0.952 & 0.962 & 0.965 & 0.961 & 0.967 \\
    antiviral                & 0.923 & 0.956 & 0.929 & 0.955 & 0.955 & 0.956 & 0.959 & 0.785 & 0.814 & 0.775 & 0.838 & 0.857 & 0.843 & 0.864 \\
    cpp                      & 0.933 & 0.929 & 0.927 & 0.917 & 0.933 & 0.936 & 0.944 & 0.862 & 0.848 & 0.855 & 0.874 & 0.884 & 0.866 & 0.888 \\
    neuropeptide             & 0.925 & 0.918 & 0.923 & 0.920 & 0.930 & 0.930 & 0.938 & 0.922 & \textcolor{red}{\textbf{0.923}} & 0.919 & 0.915 & \textcolor{red}{\textbf{0.932}} & 0.919 & 0.934 \\
    quorum\_sensing          & 0.885 & 0.930 & 0.893 & 0.922 & 0.949 & 0.946 & 0.934 & 0.822 & 0.862 & 0.844 & 0.870 & 0.889 & 0.888 & \textcolor{red}{\textbf{0.903}} \\
    toxicity                 & 0.910 & 0.913 & 0.908 & 0.920 & 0.920 & 0.924 & 0.923 & 0.780 & 0.769 & 0.770 & 0.843 & 0.839 & 0.846 & 0.842 \\
    ttca                     & 0.949 & 0.941 & 0.947 & 0.969 & 0.976 & 0.971 & 0.967 & 0.868 & 0.858 & 0.861 & 0.915 & 0.938 & 0.911 & 0.940 \\
    nonfouling               & 0.930 & 0.929 & 0.926 & 0.924 & 0.923 & 0.921 & 0.921 & 0.903 & 0.903 & 0.900 & 0.903 & 0.898 & 0.893 & 0.893 \\
    dppiv\_inhibitory        & 0.844 & 0.830 & 0.839 & 0.824 & 0.836 & 0.854 & 0.847 & \textcolor{red}{\textbf{0.857}} & \textcolor{red}{\textbf{0.843}} & \textcolor{red}{\textbf{0.846}} & \textcolor{red}{\textbf{0.846}} & \textcolor{red}{\textbf{0.858}} & \textcolor{red}{\textbf{0.862}} & \textcolor{red}{\textbf{0.860}} \\
    ace\_inhibitory\_raw     & 0.844 & 0.837 & 0.834 & 0.856 & 0.857 & 0.868 & 0.864 & \textcolor{red}{\textbf{0.853}} & \textcolor{red}{\textbf{0.852}} & \textcolor{red}{\textbf{0.851}} & \textcolor{red}{\textbf{0.863}} & \textcolor{red}{\textbf{0.873}} & \textcolor{red}{\textbf{0.879}} & \textcolor{red}{\textbf{0.892}} \\
    antioxidant              & 0.653 & 0.645 & 0.647 & 0.691 & 0.680 & 0.693 & 0.672 & \textcolor{red}{\textbf{0.693}} & \textcolor{red}{\textbf{0.689}} & \textcolor{red}{\textbf{0.680}} & \textcolor{red}{\textbf{0.708}} & \textcolor{red}{\textbf{0.711}} & \textcolor{red}{\textbf{0.705}} & \textcolor{red}{\textbf{0.696}} \\
    bbp                      & 0.549 & 0.620 & 0.572 & 0.650 & 0.635 & 0.697 & 0.646 & \textcolor{red}{\textbf{0.696}} & \textcolor{red}{\textbf{0.684}} & \textcolor{red}{\textbf{0.718}} & \textcolor{red}{\textbf{0.689}} & 0.591 & 0.620 & \textcolor{red}{\textbf{0.673}} \\
    \bottomrule
    \end{tabular}%
    }
   
    \end{table}
    
As shown in Table~\ref{tab:model_performance}, under random splitting many datasets exhibit very high ROC--AUC scores (often exceeding 0.9), where different models show almost indistinguishable performance (e.g., \textit{antimicrobial\_raw}, \textit{anticancer\_raw}, \textit{antiviral\_raw}). MMseqs2-based homology splitting alleviates this issue to some extent, as it is generally considered a more challenging evaluation protocol that reduces the similarity between training and test sets. However, we observed that in some datasets (e.g., \textit{ace\_inhibitory\_raw}, \textit{dppiv\_inhibitory\_raw}), performance under MMseqs2 splitting is even higher than under random splitting. We attribute this counter-intuitive phenomenon to an overlooked issue in previous studies, namely \textit{k}-mer leakage.

Specifically, we observes that in most peptide datasets, certain k-mers appear with very high frequency among positive samples, and these k-mers are often dataset-specific. We hypothesize that such k-mers may correspond to activity-related motifs that experts intentionally introduce during peptide design. Unfortunately, this poses a serious problem for machine learning: models can achieve inflated scores by memorizing these local shortcuts rather than learning biologically meaningful patterns, resulting in poor generalization to new active motifs. Importantly, \texttt{MMseqs2} cannot resolve this issue, since two sequences may share the same k-mer while exhibiting low overall sequence similarity, and thus will not be clustered together.

\subsection{Detection of k-mer Leakage via Enrichment Analysis}

We rigorously assess potential k-mer leakage by developing an enrichment analysis framework grounded in Fisher’s exact test. For each candidate k-mer, a $2 \times 2$ contingency table have been constructed to quantify its distribution across positive and negative samples, and the statistical significance of its enrichment in the positive class have been evaluated. Multiple hypothesis testing was addressed using the Benjamini--Hochberg procedure for false discovery rate (FDR) control. Additional stringency criteria were imposed to ensure that only robust and biologically meaningful signals were retained. Each motif was required to surpass a minimum sample support threshold, exhibit a sufficient number of occurrences within positive sequences, and simultaneously satisfy both a stringent $p$-value cutoff and an odds ratio requirement. This multi-layered filtering strategy prioritized k-mers with strong statistical association to the activity label, while reducing the likelihood of confounding artifacts arising from dataset-specific peptide design biases rather than genuine biological patterns. We set $k=5$ for datasets with mean sequence length $>15$ and $k=3$ otherwise. 

\subsubsection{A k-mer–Aware Partitioning Strategy}
\label{sec:kmer_partition}
Motivated by these findings, we introduce a partitioning protocol that explicitly suppresses k-mer leakage by allocating \emph{motif clusters}, rather than individual sequences, to data splits. The procedure is:
\begin{enumerate}
    \item Run the enrichment analysis to identify significantly enriched k-mers (henceforth \emph{motifs}).
    \item Group sequences into clusters such that any two sequences sharing at least one enriched motif are placed in the same cluster.
    \item Assign clusters (not sequences) to train/validation/test according to the target ratio (8:1:1 in our experiments).
\end{enumerate}

By construction, this protocol prevents any enriched motif from appearing across partitions, thereby forcing models to generalize beyond dataset-specific shortcuts and yielding a more faithful estimate of cross-motif generalization difficulty.

\subsubsection{Hybrid-split: Combining k-mer and Homology Constraints}
\label{sec:hybrid_split}
\begin{figure}[h!]
\centering
\includegraphics[width=0.8\textwidth]{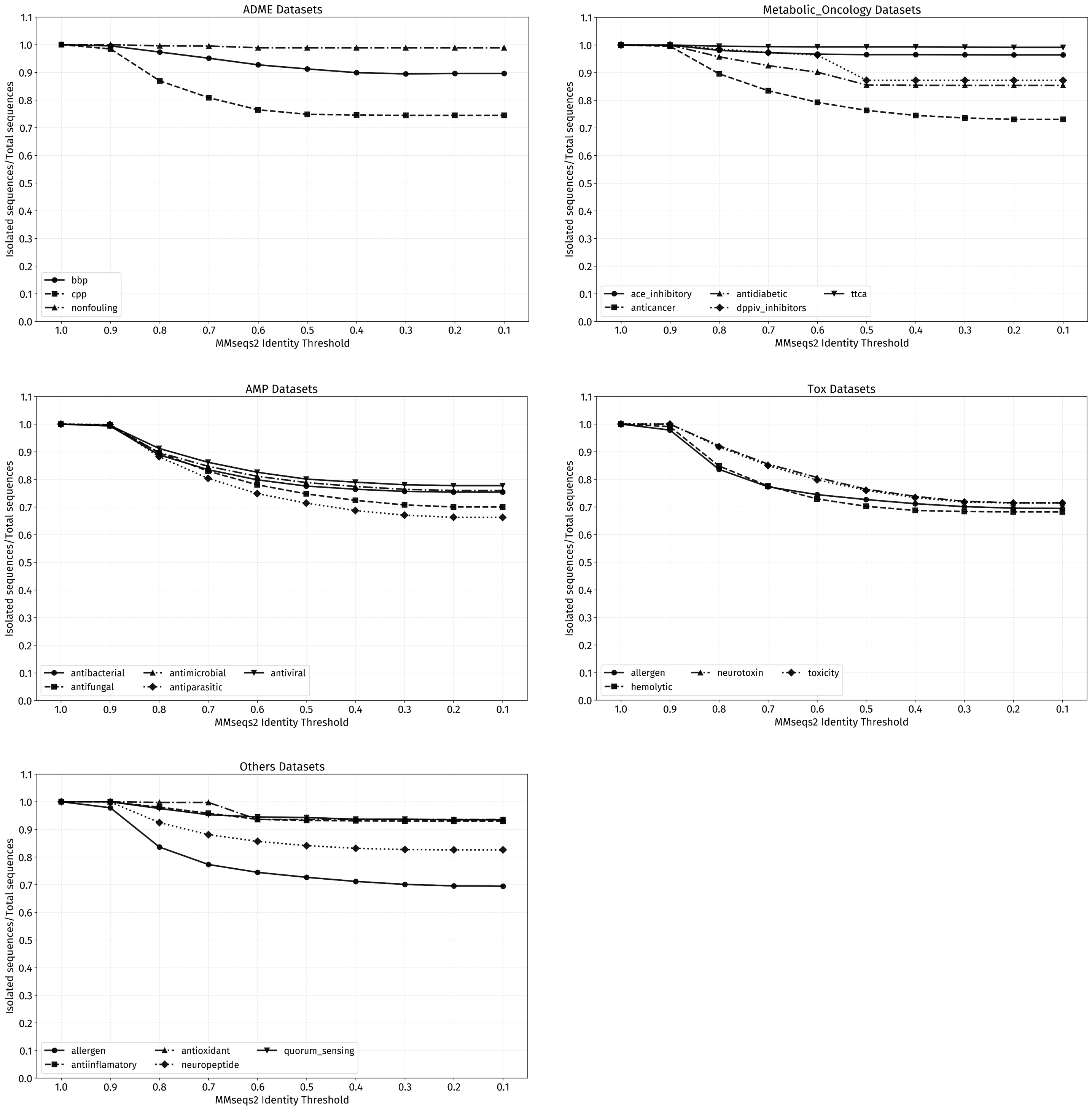}
\caption{Isolated Sequences Ratio vs MMseqs2 Identity Threshold (from 0.1 to 1.0) Across classification Datasets.}
\label{fig:mmseqs_cla}
\end{figure}

While the kmer–aware partitioning addresses motif leakage, it does not exploit information about broader sequence homology among motif-free sequences. We therefore propose \textbf{hybrid-split}, a hybrid protocol that enforces both constraints:
\begin{enumerate}
    \item Apply the kmer–aware procedure in \autoref{sec:kmer_partition} to identify motif clusters and allocate them to splits.
    \item For sequences that do not contain any enriched motif, apply \texttt{MMseqs2} to form homology clusters and allocate these clusters to the existing splits, maintaining the desired proportions.
\end{enumerate}
For MMseqs2-based homology splitting, parameter settings also need to be standardized. By examining the changes in the number of isolated sequences under different identity thresholds (see Figure~\ref{fig:mmseqs_cla}), we find that when the threshold is set to 0.3, the sequences in the dataset can be clustered to the greatest extent. Therefore, in our study, 0.3 is selected as the clustering threshold for MMseqs2.

Hybrid-split simultaneously blocks kmer leakage and reduces global homology between splits, providing a more stringent and biologically grounded evaluation. In practice, we find that hybrid-split mitigates shortcut exploitation and yields more discriminative performance estimates than random or homology-only partitioning, particularly on datasets where enriched motifs are prevalent.

\paragraph{Practical Considerations.}
Stricter partitioning can reduce effective training size—especially in small datasets—and may lower absolute scores. However, this trade-off reflects a closer alignment with real-world generalization, where future peptides may lack the same dataset-specific motifs or share low global similarity with training examples. We therefore recommend hybrid-split as a default evaluation protocol for peptide classification tasks, and we report all results under both conventional splits and hybrid-split to facilitate transparent comparison. The experimental results under different partitioning methods can be found in \ref{app:exp:sec:addition_results}.

\subsection{Pipeline of Non-Canonical Dataset Construction}

The workflow described above applies to canonical peptide datasets. 
For non-canonical peptides, however, several important distinctions must be considered. We analyzed the non-canonical datasets with respect to the same three key issues as in canonical datasets.

\paragraph{Sequence Redundancy}  
Non-canonical peptides are usually generated from canonical peptides through subtle chemical modifications 
(e.g., N-terminal acetylation, C-terminal amidation), resulting in sequences that are highly similar to their canonical counterparts. 
Removing redundancy in this case would discard a substantial number of meaningful samples. 
Therefore, we do not apply redundancy removal to non-canonical peptides. 
Instead, to mitigate potential data leakage, we relied on carefully designed data-splitting strategies described below.

\paragraph{Negative Sampling}  
Similar to canonical datasets, non-canonical peptide classification datasets lack experimentally validated negative samples. 
Unfortunately, the number of available non-canonical sequences is too limited to serve as a sufficiently large negative sampling pool. 
To address this challenge, we have trained a generative model using 9{,}512 known non-canonical peptides (see Appendix~\ref{app:nc_gen:sec}). This model takes a canonical peptide as input and outputs a chemically modified non-canonical peptide, 
allowing us to transform a canonical peptide--based negative pool into a non-canonical negative pool.  

Concretely, we first mapped non-canonical peptides to their canonical equivalents 
by replacing each non-canonical residue with its closest natural homolog 
(e.g., N-methyl-L-alanine $\rightarrow$ Alanine), 
and then applied the negative sampling strategy established for canonical peptides. 
Using the more than 9{,}500 non-canonical sequences in our benchmark, 
we trained a converter capable of mapping canonical sequences back into their non-canonical forms. 
This converter was subsequently used to transform the sampled canonical negatives into non-canonical negative samples. 
Since non-canonical amino acids generally share similar physicochemical properties with their canonical homologs, 
we argue that controlling the distribution of homologous canonical sequences 
effectively ensures that the property distribution of positive and negative non-canonical peptides is also well controlled.

\paragraph{Data Splitting}
Non-canonical peptides (with unnatural residues, stereochemical inversions, covalent modifications, cyclizations, capping, etc.) are not faithfully handled by sequence-only formats and tools that assume the standard 20-letter alphabet. To prevent near-duplicate leakage under this setting, we perform a ECFP-based split. Concretely, we compute ECFP fingerpring for each peptide and use a Tanimoto similarity \(s_{ij}\) between every pair \((i,j)\). We then construct a similarity graph \(G=(V,E_\tau)\) where \((i,j)\in E_\tau\) iff \(s_{ij}\ge \tau\). The connected components \(\{C_k\}\) of \(G\) are treated as clusters, and we assign clusters---rather than individual samples---to the training, validation, and test sets. The similarity threshold is set to \(\tau=0.95\). This cluster-level assignment preserves distributional balance while minimizing cross-split redundancy. Given our dataset size, the \(O(N^2)\) similarity computation is tractable.

For \textbf{larger datasets}, a scalable alternative is to map each non-canonical site to an ambiguity token ``X'' and then apply standard sequence-based partitioning pipelines to the \emph{X-collapsed} sequences (e.g., MMseqs2-based splitting or hybrid-style cluster splitting). This leverages mature sequence tools for efficiency, albeit at the cost of potentially conflating chemistry that is distinct but collapsed by ``X''. We provide a quantitative comparison between the ECFP-based split and the X-collapsed, sequence-based splits (MMseqs2-split, hybrid-split) in Appendix~\ref{app:exp:sec:addition_results}.


\section{Construction Pipeline of Regression Datasets}
\label{app:pipeline:sec:reg}

\begin{figure}[h]
\centering
\includegraphics[width=0.8\textwidth]{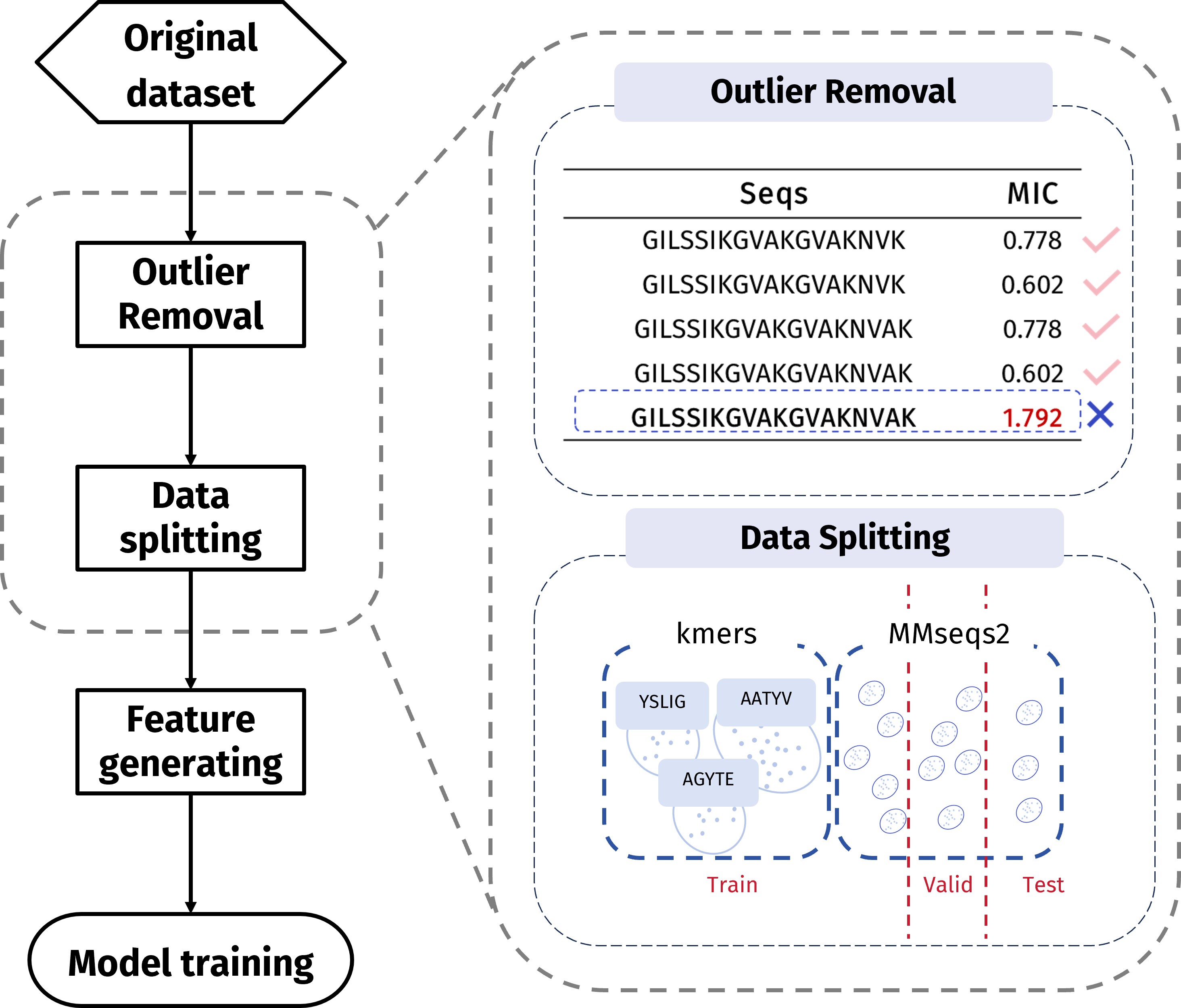}
\caption{Overview of the construction pipeline for peptide regression datasets}
\label{fig:construction_of_regression_datasets}
\end{figure}

As illustrated in Figure~\ref{fig:construction_of_regression_datasets}, we outline the pipeline for constructing regression datasets, focusing on three critical issues: \textit{outlier removal}, \textit{redundancy}, and \textit{data splitting}.

.

\subsection{Outlier Removal}

Many datasets are collected from heterogeneous literature sources, where the same peptide sequence may correspond to multiple experimental measurements due to variations in experiment conditions. For regression tasks, outlier detection is often neglected, and existing studies usually compute the mean of all available measurements~\citep{huang2023identification}. We argue that this approach is inappropriate.  

For example, in the dataset \texttt{S.aureus\_mic\_raw}, the sequence \texttt{GILSSIKGVAKGVAKNVAAQLLDTLKCKITGC} has five records: 0.778, 0.602, 0.778, 1.792, and 0.602. The value 1.792 is clearly an outlier; therefore, it should be removed before averaging the remaining values. Without such filtering, outliers may degrade data quality and compromise model reliability. This issue is particularly prevalent in \textbf{antimicrobial peptide (AMP) datasets} and \textbf{Peptide--protein interaction datasets}, where minimum inhibitory concentration (MIC) and peptide--protein binding affinity (Kd) are often repeatedly measured.  

To address this, we apply an \textbf{interquartile range (IQR)-based method} for regression datasets. Specifically, values outside the range
\[
[Q1 - 1.5 \times IQR, \, Q3 + 1.5 \times IQR]
\]
are removed, and the mean of the remaining values is taken as the final label. We apply this procedure to all regression datasets containing duplicate samples, which exclude many potentially confounding values (Table~\ref{tab:outlier}).

\begin{table}[h]
\centering
\small
\caption{Outlier removal statistics for regression datasets. Outlier\_remove: number of data points identified as outliers.}
\label{tab:outlier}
\resizebox{\textwidth}{!}{%
\begin{tabular}{lccc}
\hline
Dataset           & Sample\_count & Outlier\_remove & Unique\_sample\_count \\
\hline
S.aureus\_mic          & 5,070         & 1,167           & 2,900 \\
E.coli\_mic       & 5,465         & 1,231           & 3,312 \\
P.aeruginosa\_mic & 2,523         &   500           & 1,618 \\
PpI\_ba\_X        & 1,806         &     0           & 1,709 \\
nc-PpI\_ba\_X     &   286         &     0           &   277 \\
\hline
\end{tabular}%
}
\end{table}

\subsection{Redundancy in Regression Datasets}

Regression datasets also contain a large number of nearly identical peptide sequences. However, unlike classification tasks, we argue that \textbf{redundancy removal is not necessary for regression tasks}. This is because although highly similar sequences often show different measured values, these differences provide informative variations that benefit model learning. In other words, retaining similar sequences helps the model to capture subtle sequence–activity relationships rather than discarding them.  

\subsection{Data Splitting}
\begin{figure}[h!]
\centering
\includegraphics[width=0.8\textwidth]{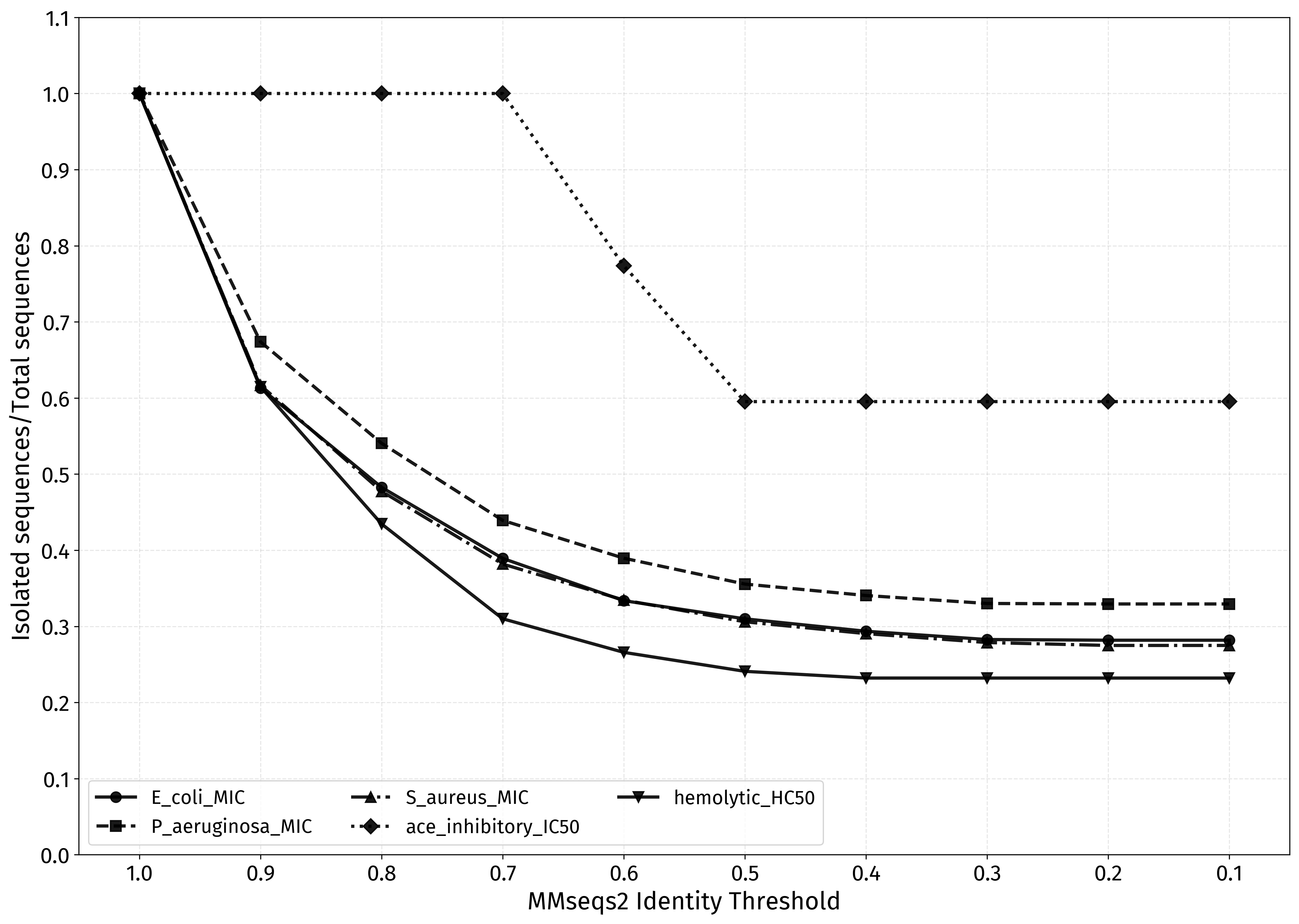}
\caption{Isolated Sequences Ratio vs MMseqs2 Identity Threshold (from 0.1 to 1.0) Across Regression Datasets.}
\label{fig:mmseqs_reg}
\end{figure}

In our collected datasets, the labels cover a very large number of values, so there is no obvious risk of k-mer leakage as in typical classification tasks. However, high-frequency k-mers can still carry meaningful information. For this reason, using a k-mer–aware hybrid-split remains valuable. In our experiments, we also adopt the hybrid-split strategy. Similar to the classification datasets, we find that when the MMseqs2 identity threshold was set to 0.3, the sequences in the regression datasets can be clustered to the greatest extent (~\ref{fig:mmseqs_reg}). Therefore, 0.3 was selected as the clustering threshold for MMseqs2.

\label{}
\section{Construction Pipeline of of Peptide-Protein Interaction Datasets}
\label{app:pipeline:sec:ppi}
We describe here the data processing procedures for the Peptide–Protein Interaction (PPI) datasets.
Before processing, we recognized that the datasets contained special amino acids that do not belong to the 20 canonical amino acids, such as X (unknown amino acid) and U (selenocysteine). For example, \texttt{PpI\_X} and \texttt{PpI\_ba\_X} contain 1,211 and 276 samples with "X," respectively. In addition, \texttt{PpI\_X} includes one peptide sequence containing "U" (GIVEQCUASVCSLYQLENYCNM). While \texttt{lm-base} models can accept such special symbols (treating them as "unknown" tokens), methods that rely on molecular fingerprints or descriptors computed from canonical amino acids cannot handle this type of input.

To address this issue, we provide two versions of the datasets. Since only \texttt{PpI\_X} contains a single sequence with "U" we directly removed this sequence. For "X" we generated paired datasets: \texttt{PpI\_X} and \texttt{PpI}, as well as \texttt{PpI\_ba\_X} and \texttt{PpI\_ba}, where the suffix "X" indicates the inclusion of sequences containing "X" (Table~\ref{tab:datasets_specialAA}). As the proportion of sequences with "X" is relatively small, we recommend using the versions without "X" i.e., \texttt{PpI} and \texttt{PpI\_ba}, in most cases.  
\begin{table}[t]
\centering
\small
\caption{Statistics of datasets with special amino acids}
\label{tab:datasets_specialAA}
\begin{tabular}{cc}
\hline
Dataset    & Size \\ \hline
PpI\_X     & 8570 \\
PpI        & 7358 \\
PpI\_ba    & 1433 \\
PpI\_ba\_X & 1709 \\ \hline
\end{tabular}
\end{table}

\begin{figure}[h!]
\centering
\includegraphics[width=0.8\textwidth]{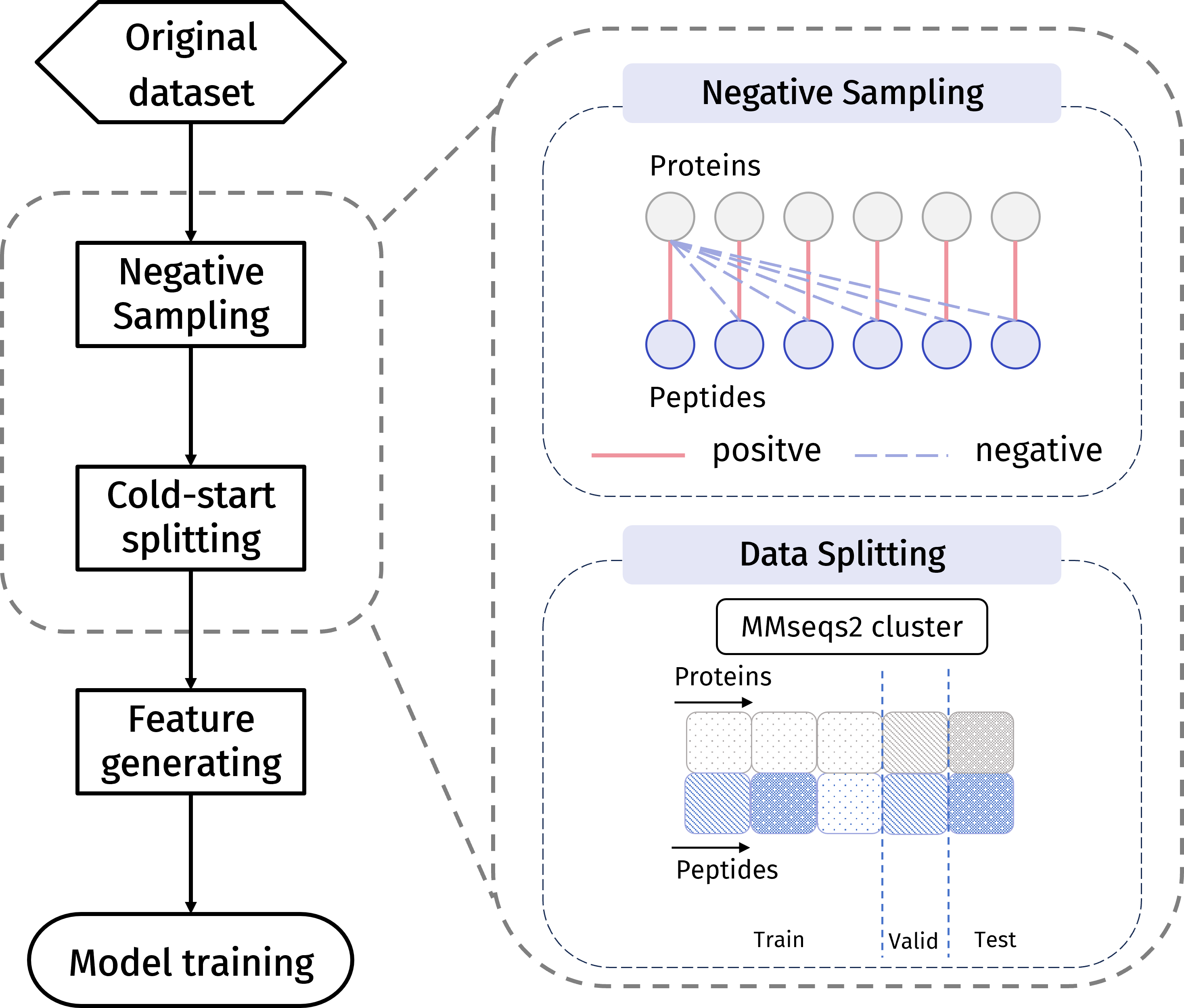}
\caption{Schematic diagram of the classification dataset construction.}
\label{fig:construction of ppi datasets}
\end{figure}.

Both canonical and non-canonical Peptide-Protein Interaction Datasets were the processed using the following steps(illustrated in Figure~\ref{fig:construction of ppi datasets}).

\subsection{Negative Sampling}  
For the collected datasets, only experimentally validated binding pairs were used to construct positive samples. 
Following the approach of CAMP~\citep{lei2021deep}, negative samples were generated by randomly shuffling the protein--peptide pairs. 
Each positive sample was paired with five corresponding negative samples.  

\subsection{Cold-start Splitting}  
For the \texttt{PepPI} task, we follow the standard paradigm in prior studies, where proteins serve as the basis for data splitting. 
Specifically, proteins are clustered based on sequence homology using \texttt{MMseqs2}, and a cold-start setting is enforced: 
proteins appearing in the training set are excluded from the test set.


\section{Experimental Details}

\subsection{Hyperparameter Settings for Four Model Families}

Given the scale of our experiments—35 datasets, 20 models, and 5 repetitions per experiment—exhaustive hyperparameter optimization for every dataset–model pair is computationally infeasible. Instead, we adopt a pragmatic strategy: for each family of models, we select a subset of representative datasets and conduct hyperparameter optimization using \texttt{Optuna}, which combines TPE-based probabilistic modeling with pruning to efficiently explore configurations under limited budgets. The resulting hyperparameters are then applied uniformly across all datasets.

\paragraph{PLM-based and SMILES-based Models.}
For PLM-based and SMILES-based models, we adopt the unified configuration in Table~\ref{tab:plm_smiles_hyperparams}.

\begin{table}[h]
\centering
\caption{Hyperparameters for PLM-based and SMILES-based models}
\label{tab:plm_smiles_hyperparams}
\begin{tabular}{lll}
\hline
\textbf{Parameter} & \textbf{Value} & \textbf{Description} \\
\hline
epochs & 50 & Maximum number of training epochs \\
batch\_size & 64 & Training batch size \\
learning\_rate & 5e-5 & Learning rate \\
early\_stopping\_patience & 5 & Stop if validation loss stagnates for 5 epochs \\
weight\_decay & 0.0 & No regularization \\
\hline
\end{tabular}
\end{table}

\paragraph{GNN-based Models.}
The hyperparameter settings for GNN-based models are shown in Table~\ref{tab:gnn_hyperparams}.

\begin{table}[h]
\centering
\caption{Hyperparameters for GNN-based models}
\label{tab:gnn_hyperparams}
\begin{tabular}{lll}
\hline
\textbf{Parameter} & \textbf{Value} & \textbf{Description} \\
\hline
epochs & 50 & Maximum number of training epochs \\
batch\_size & 64 & Training batch size \\
num\_layers & 3 & Number of GNN layers \\
emb\_dim & 300 & Dimension of hidden embeddings \\
learning\_rate & 0.001 & Learning rate \\
\hline
\end{tabular}
\end{table}

\paragraph{Traditional Machine Learning Models.}
For traditional models (Random Forest, XGBoost, and LightGBM), the default hyperparameters already provided competitive performance, so we did not perform additional tuning. The software versions are:
\begin{itemize}
    \item LightGBM 4.6.0
    \item XGBoost 3.0.2
    \item scikit-learn 1.7.0
\end{itemize}

\paragraph{PPI Task Settings.}
For the PPI prediction task, we apply a unified configuration across all models, summarized in Table~\ref{tab:ppi_hyperparams}.

\begin{table}[h]
\centering
\caption{Hyperparameters for PPI task}
\label{tab:ppi_hyperparams}
\begin{tabular}{lll}
\hline
\textbf{Parameter} & \textbf{Value} & \textbf{Description} \\
\hline
epochs & 50 & Maximum number of training epochs \\
batch\_size & 16 & Smaller batch size due to memory usage \\
learning\_rate & 1e-4 & Learning rate \\
\hline
\end{tabular}
\end{table}

\subsection{Fine-tuning ESM-150M-F} 
\label{app:exp:esm_150_f}

The ESM2 model is pretrained on UniRef~\cite{Suzek2015UniRefClusters}, spanning $\sim$43M UniRef50 clusters and $\sim$138M UniRef90 sequences, totaling about 65M unique proteins. However, short peptides are severely underrepresented: in UniRef50, only \textbf{2.8\%} of sequences are shorter than 50 residues, while most fall between 51–600. Consequently, model updates are dominated by medium- and long-length peptides, leading to poor modeling of short peptides.

\begin{figure}[htb]
	\centering
	\includegraphics[width=0.8\linewidth]{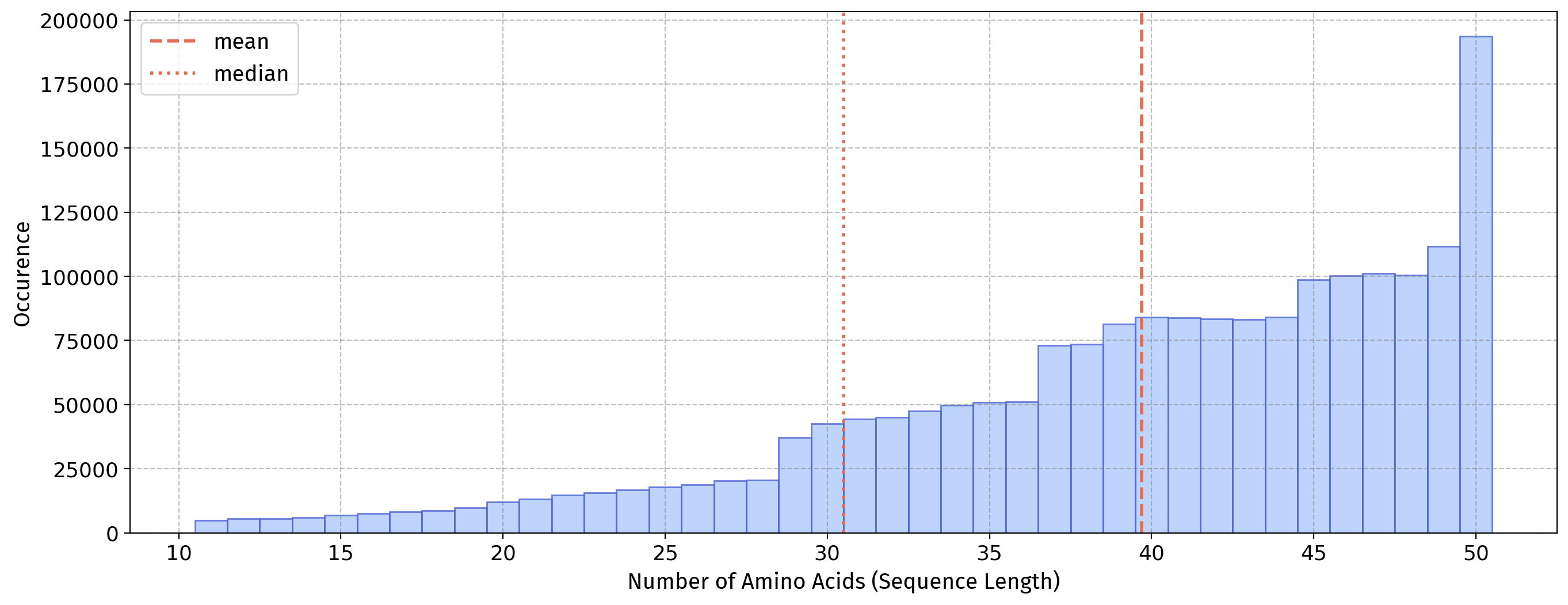}
	\caption{Length distribution of the \texttt{uniref50\_50} dataset.}
	\label{app:exp:uniref50_length}
\end{figure}

\begin{figure}[htb]
	\centering
	\includegraphics[width=0.8\linewidth]{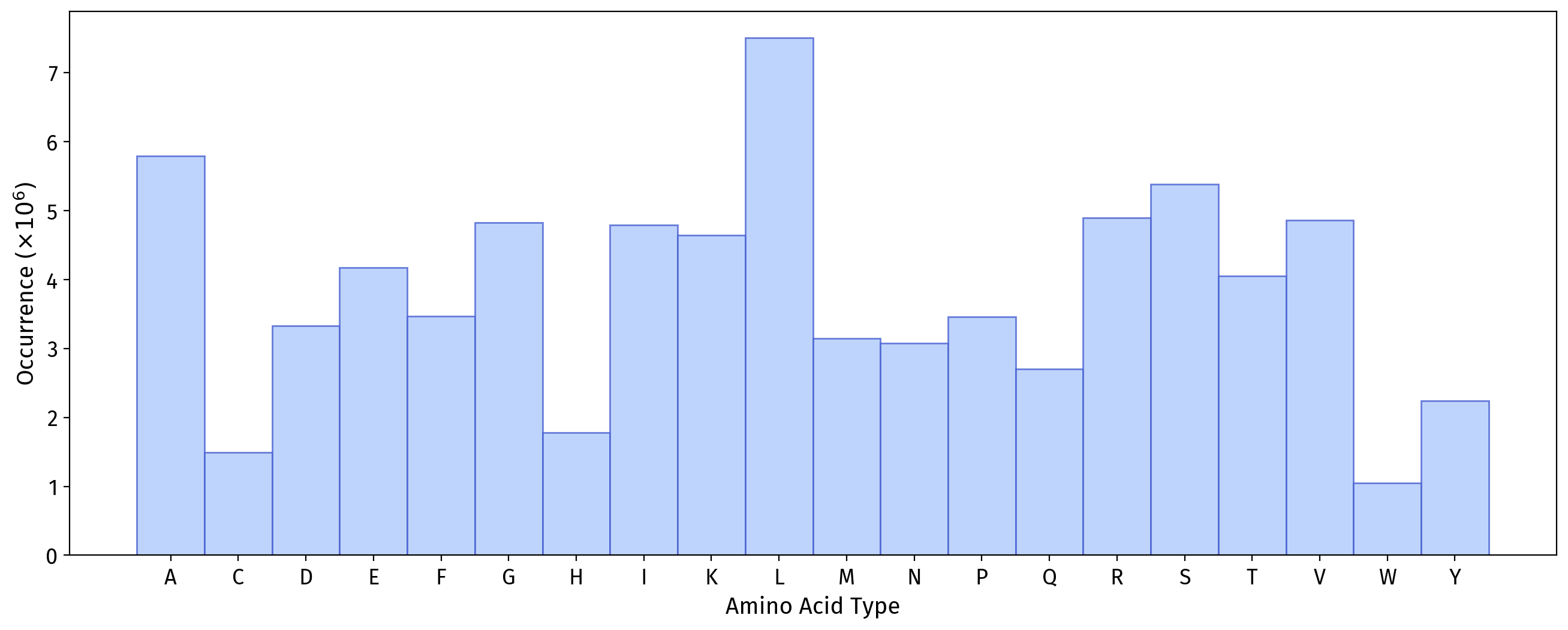}
	\caption{Amino acid distribution of the \texttt{uniref50\_50} dataset.}
	\label{app:exp:uniref50_aa}
\end{figure}

\paragraph{Dataset Preparation.}
To address this, we constructed a short-peptide dataset by truncating UniRef50 (as of April 2025) to sequences of $\leq$50 residues, removing invalid entries, and obtaining 1,932,360 sequences. This dataset, denoted \texttt{uniref50\_50}, was split into training and validation sets at a 9:1 ratio. Figure~\ref{app:exp:uniref50_length} and Figure~\ref{app:exp:uniref50_aa} illustrate the sequence length and amino acid distribution.

\paragraph{Training Setup.}
We fine-tuned ESM-150M using DeepSpeed for distributed training and BF16 mixed precision for efficiency. Training was conducted on 8 A800 GPUs with a per-device batch size of 512 (effective batch size: 4096). The hyperparameters are summarized in Table~\ref{tab:esm150_hyperparams}.

\begin{table}[h]
\centering
\caption{Hyperparameters for fine-tuning ESM-150M (ESM-150M-F)}
\label{tab:esm150_hyperparams}
\begin{tabular}{lll}
\hline
\textbf{Parameter} & \textbf{Value} & \textbf{Description} \\
\hline
learning\_rate & 4e-4 & Learning rate \\
num\_train\_epochs & 500 & Training epochs \\
per\_device\_train\_batch\_size & 512 & Batch size per device \\
gradient\_accumulation\_steps & 1 & Gradient accumulation steps \\
warmup\_steps & 2000 & Learning rate warm-up steps \\
weight\_decay & 0.01 & Weight decay \\
max\_grad\_norm & 5.0 & Gradient clipping threshold \\
optimizer & AdamW & Optimizer type \\
adam\_beta1 & 0.9 & Adam $\beta_1$ parameter \\
adam\_beta2 & 0.98 & Adam $\beta_2$ parameter \\
adam\_epsilon & 1e-8 & Adam $\epsilon$ parameter \\
lr\_scheduler\_type & linear & Learning rate scheduler \\
eval\_steps & 500 & Evaluation frequency \\
save\_steps & 500 & Checkpoint saving frequency \\
distributed\_type & DeepSpeed & Distributed backend \\
num\_processes & 8 & Number of processes (GPUs) \\
num\_machines & 1 & Number of machines \\
zero\_stage & 2 & ZeRO optimization stage \\
bf16 & true & Use BF16 mixed precision \\
seed & 42 & Random seed \\
\hline
\end{tabular}
\end{table}

\paragraph{Evaluation Metric.}
We employed \textbf{pseudo-perplexity (PseudoPPL)} to assess model performance on peptides. Unlike autoregressive models, where perplexity is computed as
\[
\text{PPL}(x) = \exp \left( -\frac{1}{L} \sum_{i=1}^{L} \log p(x_i \mid x_{<i}) \right),
\]
masked language models require the following variant:
\[
\text{PseudoPPL}(x) = \exp \left( -\frac{1}{L} \sum_{i=1}^{L} \log p(x_i \mid x_{j \neq i}) \right).
\]

\begin{figure}[htb]
	\centering
	\includegraphics[width=0.8\linewidth]{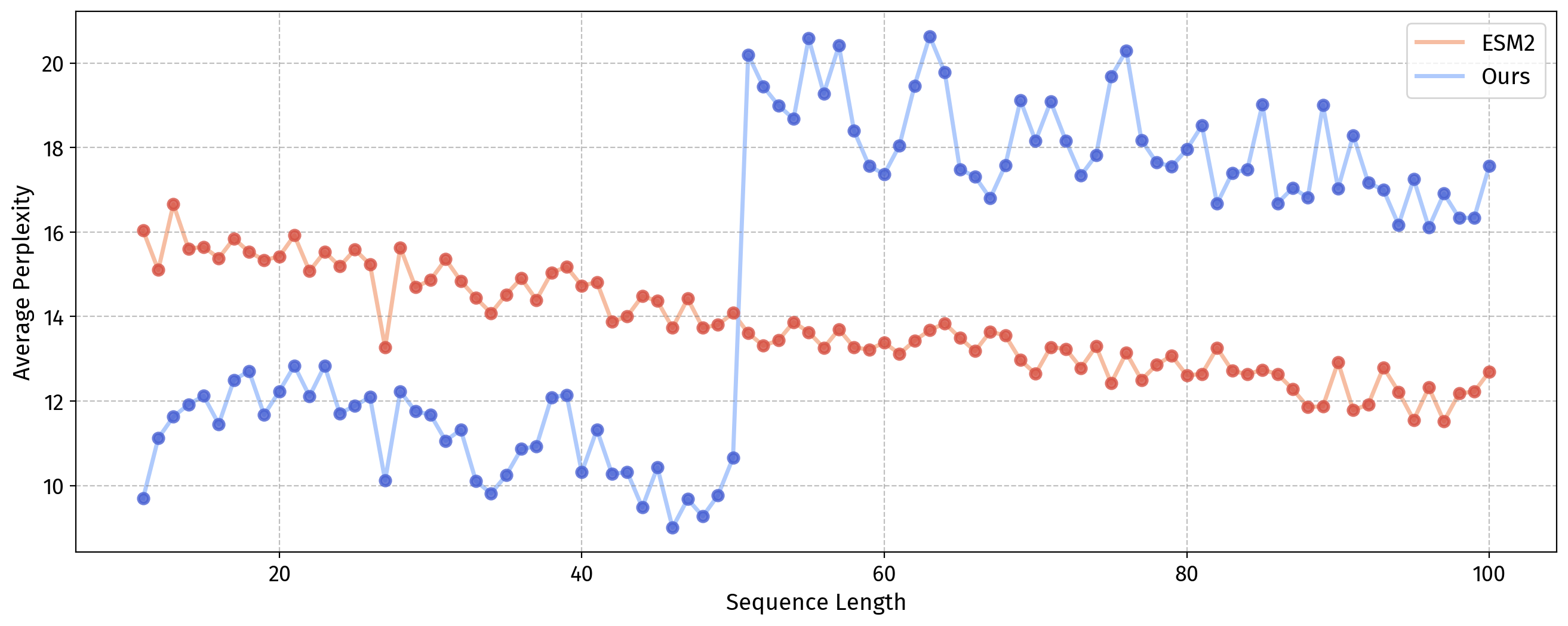}
	\caption{Perplexity of ESM-150M (ESM2) and fine-tuned ESM-150M-F (ours) across peptide lengths.}
	\label{app:exp:fig:perplexity}
\end{figure}

We used 85,113 unique peptide sequences absent from UniRef100 for evaluation. As shown in Fig.~\ref{app:exp:fig:perplexity}, ESM-150M suffers from high perplexity on short peptides, with performance deteriorating as length decreases. In contrast, ESM-150M-F consistently achieves lower perplexity, confirming the effectiveness of fine-tuning.

\paragraph{Summary.}
Fine-tuning on short peptides substantially improves ESM-150M’s ability to model short sequences, as evidenced by reduced perplexity and overall better performance.

\section{Additional Results}
\label{app:exp:sec:addition_results}

In this section, we provide detailed experimental results, including:  

\begin{enumerate}
    \item Results on 22 canonical peptide classification datasets under four splitting strategies: 
    \textit{hybrid-split} (Table~\ref{app:exp:tab:natural_cls_hydra}), 
    \textit{MMseqs2-split} (Table~\ref{app:exp:tab:natural_cls_mmseqs2}), 
    \textit{k-mer–split} (Table~\ref{app:exp:tab:natural_cls_kmer}), and 
    \textit{random-split} (Table~\ref{app:exp:tab:natural_cls_random}). 
    
    \item Results on 4 non-canonical peptide regression datasets under four splitting strategies: 
    \textit{hybrid-split} (Table~\ref{app:exp:tab:natural_reg_hydra}), 
    \textit{MMseqs2-split} (Table~\ref{app:exp:tab:natural_reg_mmseqs2}), 
    \textit{k-mer–split} (Table~\ref{app:exp:tab:natural_reg_kmer}), and 
    \textit{random-split} (Table~\ref{app:exp:tab:natural_reg_random}). 

    \item Results on 5 non-canonical peptide datasets under five splitting strategies: 
    \textit{ECFP-split} (Table~\ref{app:exp:tab:non_natural_ecfp}), 
    \textit{hybrid-split} (Table~\ref{app:exp:tab:non_natural_hydra}), 
    \textit{MMseqs2-split} (Table~\ref{app:exp:tab:non_natural_mmseqs2}), 
    \textit{k-mer–split} (Table~\ref{app:exp:tab:non_natural_kmer}), and 
    \textit{random-split} (Table~\ref{app:exp:tab:non_natural_random}). 

    \item Results on 3 PepPI datasets without freezing the protein encoder 
    (Table~\ref{app:exp:tab:ppi}).
    
    \item Results on the \textit{PepBenchData-150} benchmark 
    (Table~\ref{app:exp:tab:150}). For most datasets in \textit{PepBenchData-150}, 
    the proportion of sequences longer than 50 residues is below 10\% 
    (see Table~\ref{app:des:tab:pepbenchmark_cls} and 
    Table~\ref{app:des:tab:pepbenchmark_reg}). Therefore, we consider these datasets 
    to be largely comparable to the \textit{PepBenchData-50} version. 
    Only five datasets contain more than 50\% sequences longer than 50 residues, 
    and we conducted additional experiments on these datasets.
\end{enumerate}

By analyzing these results, we can draw the following conclusions:
\begin{enumerate}
\item The conclusions in the main text hold across all partitioning strategies.
\item The performance of kmer-split is significantly lower than that of random-split, indicating that the issue of kmer leakage is severe. The hybrid partition combines the advantages of MMseqs2-split and kmer-split, making it a more challenging partitioning strategy.
\item In the PepBenchData-150 version, the fine-tuned ESM2-150-F performs worse than ESM2-150M. This result is expected, as ESM2-150-F is fine-tuned only on peptide data and thus forgets the protein pre-training knowledge. As shown in Figure~\ref{app:exp:fig:perplexity}, ESM2-150M-F exhibits higher perplexity than ESM2-150M on sequences longer than 50.
\item For the PepPI task, it remains uncertain whether freezing the protein encoder is necessary.
\end{enumerate}



\begin{table}[h!]
\centering
\caption{Performance of models on canonical peptide classification (ROC-AUC$\uparrow$, \%) with hybrid-split. Dataset sizes are shown separately; results are mean$_{\pm std}$. Best and second-best scores per row are in \textbf{bold} and gray shadow.}

\label{app:exp:tab:natural_cls_hydra}
\resizebox{\textwidth}{!}{%
\definecolor{lightgray}{gray}{0.9}
\begin{tabular}{l|c|ccccc|cccc|ccc}
\toprule
\multirow{2}{*}{Dataset} & \multirow{2}{*}{Size} & \multicolumn{5}{c|}{FP-based models} & \multicolumn{4}{c|}{GNN-based models} & \multicolumn{3}{c}{SMILES-based models} \\
\cmidrule(lr){3-7} \cmidrule(lr){8-11} \cmidrule(lr){12-14}
 & & RF & XGBoost & LightGBM & GradBoost & AdaBoost & GCN & GAT & GIN & Pepland & ChemBERTa & PeptideCLM & PepDoRA \\
\midrule

nonfouling & 7200  & $76.4_{\pm1.5}$ & $75.4_{\pm2.1}$ & $76.4_{\pm1.4}$ & $77.0_{\pm1.4}$ & $76.3_{\pm1.5}$ & $76.8_{\pm1.0}$ & $76.1_{\pm0.9}$ & $76.7_{\pm1.4}$ & $75.3_{\pm2.3}$ & $76.2_{\pm1.0}$ & $75.6_{\pm2.1}$ & $70.7_{\pm2.1}$ \\
cpp & 2296  & $82.1_{\pm2.9}$ & $82.3_{\pm4.9}$ & $82.1_{\pm3.4}$ & $79.3_{\pm3.8}$ & $74.7_{\pm3.4}$ & $59.7_{\pm5.4}$ & $53.6_{\pm5.2}$ & $57.7_{\pm5.5}$ & $68.2_{\pm7.2}$ & $72.2_{\pm8.2}$ & $58.5_{\pm3.8}$ & $58.9_{\pm4.3}$ \\
bbp & 665  & $\cellcolor{lightgray}{69.3_{\pm6.4}}$ & $68.2_{\pm8.4}$ & $69.2_{\pm7.5}$ & $65.5_{\pm10.5}$ & $60.9_{\pm5.0}$ & $62.5_{\pm5.6}$ & $62.8_{\pm5.0}$ & $61.2_{\pm3.4}$ & $56.1_{\pm9.3}$ & $61.2_{\pm7.1}$ & $61.9_{\pm6.9}$ & $54.7_{\pm7.2}$ \\
antimicrobial & 52941  & $87.6_{\pm0.4}$ & $87.8_{\pm0.5}$ & $87.5_{\pm0.6}$ & $85.2_{\pm0.6}$ & $82.8_{\pm0.8}$ & $81.7_{\pm0.4}$ & $73.2_{\pm1.1}$ & $80.5_{\pm1.9}$ & $67.5_{\pm1.6}$ & $89.9_{\pm0.8}$ & $83.3_{\pm1.6}$ & $75.4_{\pm0.9}$ \\
antibacterial & 28591  & $88.4_{\pm0.9}$ & $88.3_{\pm0.8}$ & $87.6_{\pm0.7}$ & $85.6_{\pm0.8}$ & $84.1_{\pm0.9}$ & $83.2_{\pm1.0}$ & $75.9_{\pm1.3}$ & $82.2_{\pm1.8}$ & $62.7_{\pm3.4}$ & $88.9_{\pm2.2}$ & $77.0_{\pm5.4}$ & $73.8_{\pm2.6}$ \\
antifungal & 12887  & $87.3_{\pm0.8}$ & $87.3_{\pm0.8}$ & $87.1_{\pm0.8}$ & $85.3_{\pm1.0}$ & $83.3_{\pm1.2}$ & $76.0_{\pm2.5}$ & $75.2_{\pm1.1}$ & $78.5_{\pm3.6}$ & $53.5_{\pm3.5}$ & $87.0_{\pm1.1}$ & $59.0_{\pm2.2}$ & $70.3_{\pm1.2}$ \\
\bottomrule
\end{tabular}}
\vspace{1.5mm}
\resizebox{\textwidth}{!}{%
\definecolor{lightgray}{gray}{0.9}
\begin{tabular}{l|c|ccccccccc}
\toprule
\multirow{2}{*}{Dataset} & \multirow{2}{*}{Size} & \multicolumn{9}{c}{PLM-based models} \\
\cmidrule(lr){3-11}
 & & ESM2-8M & ESM2-35M & ESM2-150M & ESM2-650M & DPLM-150M & DPLM-650M & ProtBERT & ESM2-8M-S & ESM2-150M-F \\
\midrule

nonfouling & 7200  & $77.3_{\pm1.3}$ & $77.6_{\pm1.1}$ & $77.9_{\pm1.5}$ & $77.4_{\pm1.0}$ & $\cellcolor{lightgray}{78.0_{\pm0.8}}$ & $78.0_{\pm1.0}$ & $59.8_{\pm7.3}$ & $77.0_{\pm1.2}$ & $\mathbf{78.2_{\pm1.0}}$ \\
cpp & 2296  & $80.5_{\pm3.2}$ & $82.2_{\pm2.3}$ & $\cellcolor{lightgray}{83.2_{\pm3.1}}$ & $80.1_{\pm6.0}$ & $81.9_{\pm2.8}$ & $\mathbf{83.8_{\pm5.8}}$ & $75.6_{\pm2.1}$ & $79.6_{\pm8.0}$ & $81.9_{\pm3.1}$ \\
bbp & 665  & $66.8_{\pm9.5}$ & $61.8_{\pm15.7}$ & $62.4_{\pm12.2}$ & $64.6_{\pm9.0}$ & $64.6_{\pm8.9}$ & $66.8_{\pm6.2}$ & $61.1_{\pm7.1}$ & $\mathbf{71.6_{\pm5.7}}$ & $64.6_{\pm12.5}$ \\
antimicrobial & 52941  & $92.3_{\pm0.3}$ & $92.4_{\pm0.4}$ & $92.3_{\pm0.6}$ & $\cellcolor{lightgray}{92.6_{\pm1.1}}$ & $92.4_{\pm0.2}$ & $92.5_{\pm0.4}$ & $78.0_{\pm22.1}$ & $92.0_{\pm0.6}$ & $\mathbf{93.2_{\pm0.3}}$ \\
antibacterial & 28591  & $92.0_{\pm0.9}$ & $92.5_{\pm0.8}$ & $\cellcolor{lightgray}{93.0_{\pm1.2}}$ & $93.0_{\pm0.6}$ & $92.3_{\pm1.2}$ & $92.7_{\pm0.8}$ & $78.7_{\pm16.9}$ & $90.8_{\pm0.6}$ & $\mathbf{93.2_{\pm0.5}}$ \\
antifungal & 12887  & $90.6_{\pm0.6}$ & $90.7_{\pm1.1}$ & $\cellcolor{lightgray}{91.0_{\pm1.0}}$ & $90.6_{\pm0.8}$ & $90.7_{\pm1.4}$ & $90.6_{\pm0.9}$ & $86.2_{\pm4.1}$ & $87.8_{\pm2.2}$ & $\mathbf{91.1_{\pm0.7}}$ \\
\bottomrule
\end{tabular}}
\end{table}


\begin{table}[h!]
\centering
\caption{Performance of models on canonical peptide classification (ROC-AUC$\uparrow$, \%) with kmer-split. Dataset sizes are shown separately; results are mean$_{\pm std}$. Best and second-best scores per row are in \textbf{bold} and gray shadow.} 
\label{app:exp:tab:natural_cls_kmer}
\resizebox{\textwidth}{!}{%
\definecolor{lightgray}{gray}{0.9}
\begin{tabular}{l|c|ccccc|cccc|ccc}
\toprule
\multirow{2}{*}{Dataset} & \multirow{2}{*}{Size} & \multicolumn{5}{c|}{FP-based models} & \multicolumn{4}{c|}{GNN-based models} & \multicolumn{3}{c}{SMILES-based models} \\
\cmidrule(lr){3-7} \cmidrule(lr){8-11} \cmidrule(lr){12-14}
 & & RF & XGBoost & LightGBM & GradBoost & AdaBoost & GCN & GAT & GIN & Pepland & ChemBERTa & PeptideCLM & PepDoRA \\
\midrule

nonfouling & 7200  & $77.0_{\pm1.4}$ & $76.9_{\pm1.6}$ & $77.4_{\pm1.2}$ & $77.6_{\pm1.4}$ & $77.0_{\pm1.9}$ & $77.7_{\pm1.7}$ & $76.6_{\pm2.5}$ & $78.4_{\pm1.5}$ & $77.0_{\pm2.5}$ & $77.2_{\pm1.9}$ & $75.6_{\pm2.6}$ & $70.9_{\pm2.0}$ \\
cpp & 2296  & $\mathbf{85.8_{\pm3.1}}$ & $83.9_{\pm3.2}$ & $84.0_{\pm1.9}$ & $81.2_{\pm1.7}$ & $76.5_{\pm2.4}$ & $60.8_{\pm7.1}$ & $54.2_{\pm3.4}$ & $64.5_{\pm2.4}$ & $70.3_{\pm2.9}$ & $74.5_{\pm4.1}$ & $60.1_{\pm3.4}$ & $59.5_{\pm4.5}$ \\
bbp & 665  & $\cellcolor{lightgray}{69.8_{\pm11.1}}$ & $67.4_{\pm9.0}$ & $\mathbf{69.8_{\pm9.2}}$ & $62.8_{\pm4.8}$ & $59.3_{\pm5.4}$ & $59.3_{\pm8.3}$ & $59.3_{\pm5.8}$ & $58.6_{\pm10.7}$ & $69.1_{\pm7.0}$ & $51.5_{\pm11.4}$ & $59.3_{\pm6.9}$ & $47.5_{\pm5.0}$ \\
antimicrobial & 52941  & $87.8_{\pm0.1}$ & $87.9_{\pm0.3}$ & $87.5_{\pm0.2}$ & $85.3_{\pm0.3}$ & $82.8_{\pm0.5}$ & $81.9_{\pm0.7}$ & $73.5_{\pm1.3}$ & $81.8_{\pm0.3}$ & $68.8_{\pm2.2}$ & $90.2_{\pm0.6}$ & $83.9_{\pm1.1}$ & $76.1_{\pm2.0}$ \\
antibacterial & 28591  & $89.6_{\pm1.0}$ & $89.3_{\pm1.1}$ & $88.7_{\pm0.9}$ & $86.3_{\pm0.9}$ & $84.5_{\pm0.8}$ & $82.6_{\pm1.9}$ & $77.3_{\pm1.2}$ & $83.8_{\pm0.8}$ & $63.5_{\pm3.7}$ & $89.9_{\pm0.7}$ & $83.2_{\pm0.9}$ & $75.6_{\pm1.7}$ \\
antifungal & 12887  & $90.2_{\pm0.8}$ & $89.8_{\pm0.9}$ & $89.9_{\pm1.0}$ & $87.4_{\pm0.7}$ & $85.0_{\pm1.0}$ & $78.1_{\pm1.9}$ & $77.7_{\pm1.4}$ & $82.2_{\pm3.3}$ & $54.1_{\pm5.9}$ & $88.9_{\pm1.2}$ & $66.1_{\pm12.9}$ & $69.0_{\pm2.1}$ \\
\bottomrule
\end{tabular}}
\vspace{1.5mm}
\resizebox{\textwidth}{!}{%
\definecolor{lightgray}{gray}{0.9}
\begin{tabular}{l|c|ccccccccc}
\toprule
\multirow{2}{*}{Dataset} & \multirow{2}{*}{Size} & \multicolumn{9}{c}{PLM-based models} \\
\cmidrule(lr){3-11}
 & & ESM2-8M & ESM2-35M & ESM2-150M & ESM2-650M & DPLM-150M & DPLM-650M & ProtBERT & ESM2-8M-S & ESM2-150M-F \\
\midrule

nonfouling & 7200  & $\mathbf{79.2_{\pm1.2}}$ & $78.8_{\pm1.1}$ & $\cellcolor{lightgray}{79.1_{\pm1.6}}$ & $78.8_{\pm1.6}$ & $78.8_{\pm1.1}$ & $78.6_{\pm1.6}$ & $62.4_{\pm17.5}$ & $78.2_{\pm1.6}$ & $79.1_{\pm1.2}$ \\
cpp & 2296  & $84.1_{\pm1.5}$ & $83.1_{\pm3.2}$ & $84.8_{\pm2.4}$ & $83.8_{\pm1.5}$ & $81.7_{\pm2.9}$ & $85.1_{\pm1.3}$ & $79.7_{\pm2.3}$ & $82.0_{\pm2.4}$ & $\cellcolor{lightgray}{85.2_{\pm2.2}}$ \\
bbp & 665  & $64.7_{\pm10.4}$ & $60.7_{\pm13.4}$ & $62.0_{\pm14.8}$ & $65.4_{\pm10.6}$ & $64.7_{\pm12.6}$ & $63.5_{\pm12.6}$ & $64.7_{\pm9.7}$ & $63.9_{\pm12.4}$ & $68.3_{\pm12.4}$ \\
antimicrobial & 52941  & $92.3_{\pm0.3}$ & $92.4_{\pm0.8}$ & $92.8_{\pm0.2}$ & $92.7_{\pm0.6}$ & $92.4_{\pm0.6}$ & $\cellcolor{lightgray}{92.9_{\pm1.0}}$ & - & $92.3_{\pm0.3}$ & $\mathbf{93.5_{\pm0.2}}$ \\
antibacterial & 28591  & $92.8_{\pm0.4}$ & $93.5_{\pm0.3}$ & $93.6_{\pm0.7}$ & $\mathbf{93.9_{\pm0.7}}$ & $93.8_{\pm0.4}$ & $93.4_{\pm0.7}$ & $82.8_{\pm7.4}$ & $91.8_{\pm1.1}$ & $\cellcolor{lightgray}{93.9_{\pm0.5}}$ \\
antifungal & 12887  & $92.9_{\pm0.4}$ & $93.2_{\pm0.7}$ & $93.5_{\pm0.5}$ & $93.1_{\pm0.8}$ & $93.1_{\pm1.1}$ & $\mathbf{93.7_{\pm0.9}}$ & $82.4_{\pm15.5}$ & $90.3_{\pm1.0}$ & $\cellcolor{lightgray}{93.6_{\pm1.2}}$ \\
\bottomrule
\end{tabular}}
\end{table}



\begin{table}[h!]
\centering
\caption{Performance of models on canonical peptide classification (ROC-AUC$\uparrow$, \%) with MMseqs2-split. Dataset sizes are shown separately; results are mean$_{\pm std}$. Best and second-best scores per row are in \textbf{bold} and gray shadow.}
\label{app:exp:tab:natural_cls_mmseqs2}
\resizebox{\textwidth}{!}{%
\definecolor{lightgray}{gray}{0.9}
\begin{tabular}{l|c|ccccc|cccc|ccc}
\toprule
\multirow{2}{*}{Dataset} & \multirow{2}{*}{Size} & \multicolumn{5}{c|}{FP-based models} & \multicolumn{4}{c|}{GNN-based models} & \multicolumn{3}{c}{SMILES-based models} \\
\cmidrule(lr){3-7} \cmidrule(lr){8-11} \cmidrule(lr){12-14}
 & & RF & XGBoost & LightGBM & GradBoost & AdaBoost & GCN & GAT & GIN & Pepland & ChemBERTa & PeptideCLM & PepDoRA \\
\midrule

nonfouling & 7200  & $77.7_{\pm1.4}$ & $76.4_{\pm2.4}$ & $77.3_{\pm2.0}$ & $77.7_{\pm1.8}$ & $76.8_{\pm1.8}$ & $77.9_{\pm1.9}$ & $76.7_{\pm1.5}$ & $77.8_{\pm1.8}$ & $77.8_{\pm1.6}$ & $77.6_{\pm1.8}$ & $75.4_{\pm1.9}$ & $71.5_{\pm2.7}$ \\
cpp & 2296  & $85.8_{\pm2.2}$ & $84.3_{\pm2.8}$ & $84.2_{\pm2.4}$ & $81.2_{\pm1.3}$ & $78.1_{\pm1.3}$ & $69.0_{\pm3.0}$ & $63.5_{\pm2.7}$ & $67.2_{\pm5.8}$ & $69.7_{\pm9.8}$ & $80.1_{\pm2.8}$ & $68.3_{\pm2.1}$ & $68.0_{\pm1.8}$ \\
bbp & 665  & $\mathbf{73.5_{\pm5.8}}$ & $70.7_{\pm5.4}$ & $71.6_{\pm5.5}$ & $71.0_{\pm7.0}$ & $63.9_{\pm3.9}$ & $65.4_{\pm4.5}$ & $63.4_{\pm4.4}$ & $65.9_{\pm6.4}$ & $67.6_{\pm4.5}$ & $58.2_{\pm7.8}$ & $64.1_{\pm6.8}$ & $58.9_{\pm7.4}$ \\
antimicrobial & 52941  & $91.9_{\pm0.4}$ & $91.6_{\pm0.5}$ & $91.1_{\pm0.5}$ & $88.8_{\pm0.5}$ & $86.0_{\pm0.7}$ & $85.9_{\pm0.9}$ & $78.8_{\pm1.1}$ & $85.6_{\pm0.8}$ & $72.0_{\pm5.5}$ & $92.9_{\pm0.5}$ & $87.1_{\pm1.8}$ & $83.7_{\pm1.2}$ \\
antibacterial & 28591  & $93.3_{\pm0.3}$ & $92.6_{\pm0.3}$ & $92.2_{\pm0.3}$ & $90.2_{\pm0.4}$ & $88.2_{\pm0.5}$ & $88.3_{\pm0.4}$ & $83.6_{\pm0.6}$ & $88.3_{\pm0.6}$ & $70.1_{\pm4.2}$ & $93.6_{\pm0.5}$ & $87.0_{\pm0.9}$ & $82.4_{\pm2.1}$ \\
antifungal & 12887  & $91.6_{\pm0.7}$ & $90.8_{\pm0.6}$ & $90.8_{\pm0.6}$ & $88.7_{\pm1.0}$ & $86.2_{\pm0.5}$ & $81.6_{\pm0.9}$ & $80.2_{\pm1.3}$ & $85.6_{\pm1.9}$ & $58.5_{\pm3.8}$ & $89.8_{\pm0.7}$ & $61.4_{\pm11.6}$ & $73.0_{\pm1.0}$ \\
\bottomrule
\end{tabular}}
\vspace{1.5mm}
\resizebox{\textwidth}{!}{%
\definecolor{lightgray}{gray}{0.9}
\begin{tabular}{l|c|ccccccccc}
\toprule
\multirow{2}{*}{Dataset} & \multirow{2}{*}{Size} & \multicolumn{9}{c}{PLM-based models} \\
\cmidrule(lr){3-11}
 & & ESM2-8M & ESM2-35M & ESM2-150M & ESM2-650M & DPLM-150M & DPLM-650M & ProtBERT & ESM2-8M-S & ESM2-150M-F \\
\midrule

nonfouling & 7200  & $78.6_{\pm2.4}$ & $78.8_{\pm1.7}$ & $78.9_{\pm2.5}$ & $78.9_{\pm1.5}$ & $\cellcolor{lightgray}{79.1_{\pm1.7}}$ & $78.6_{\pm1.6}$ & $62.5_{\pm11.5}$ & $77.8_{\pm1.8}$ & $\mathbf{79.4_{\pm2.4}}$ \\
cpp & 2296  & $\cellcolor{lightgray}{85.8_{\pm1.8}}$ & $85.4_{\pm2.5}$ & $85.5_{\pm2.4}$ & $85.4_{\pm2.3}$ & $85.3_{\pm3.0}$ & $\mathbf{87.4_{\pm3.0}}$ & $78.9_{\pm7.0}$ & $84.7_{\pm1.4}$ & $85.6_{\pm1.3}$ \\
bbp & 665  & $70.5_{\pm7.5}$ & $70.3_{\pm3.9}$ & $\cellcolor{lightgray}{73.1_{\pm2.5}}$ & $67.2_{\pm7.5}$ & $70.2_{\pm3.8}$ & $69.9_{\pm5.4}$ & $64.2_{\pm5.9}$ & $68.1_{\pm4.7}$ & $68.4_{\pm3.4}$ \\
antimicrobial & 52941  & $94.8_{\pm0.4}$ & $94.8_{\pm0.4}$ & $94.7_{\pm0.4}$ & $95.1_{\pm0.4}$ & $95.0_{\pm0.4}$ & $\cellcolor{lightgray}{95.2_{\pm0.4}}$ & $72.4_{\pm15.8}$ & $94.1_{\pm0.3}$ & $\mathbf{95.6_{\pm0.5}}$ \\
antibacterial & 28591  & $95.0_{\pm0.6}$ & $95.5_{\pm0.4}$ & $\cellcolor{lightgray}{95.6_{\pm0.6}}$ & $95.6_{\pm0.8}$ & $95.6_{\pm0.3}$ & $95.5_{\pm0.6}$ & $89.8_{\pm7.2}$ & $94.3_{\pm0.5}$ & $\mathbf{95.8_{\pm0.4}}$ \\
antifungal & 12887  & $93.7_{\pm0.7}$ & $94.2_{\pm0.5}$ & $\mathbf{94.3_{\pm0.6}}$ & $94.3_{\pm0.4}$ & $94.2_{\pm0.6}$ & $\cellcolor{lightgray}{94.3_{\pm0.4}}$ & $90.1_{\pm3.0}$ & $91.0_{\pm0.9}$ & $93.8_{\pm1.0}$ \\
\bottomrule
\end{tabular}}
\end{table}



\begin{table}[h!]
\centering
\caption{Performance of models on canonical peptide classification (ROC-AUC$\uparrow$, \%) with random-split. Dataset sizes are shown separately; results are mean$_{\pm std}$. Best and second-best scores per row are in \textbf{bold} and gray shadow.}
\label{app:exp:tab:natural_cls_random}
\resizebox{\textwidth}{!}{%
\definecolor{lightgray}{gray}{0.9}
\begin{tabular}{l|c|ccccc|cccc|ccc}
\toprule
\multirow{2}{*}{Dataset} & \multirow{2}{*}{Size} & \multicolumn{5}{c|}{FP-based models} & \multicolumn{4}{c|}{GNN-based models} & \multicolumn{3}{c}{SMILES-based models} \\
\cmidrule(lr){3-7} \cmidrule(lr){8-11} \cmidrule(lr){12-14}
 & & RF & XGBoost & LightGBM & GradBoost & AdaBoost & GCN & GAT & GIN & Pepland & ChemBERTa & PeptideCLM & PepDoRA \\
\midrule

nonfouling & 7200  & $77.1_{\pm1.5}$ & $76.9_{\pm1.6}$ & $77.4_{\pm1.2}$ & $77.6_{\pm1.5}$ & $77.0_{\pm1.9}$ & $78.1_{\pm1.7}$ & $76.7_{\pm1.7}$ & $78.4_{\pm1.6}$ & $77.5_{\pm2.7}$ & $77.3_{\pm1.7}$ & $75.6_{\pm2.5}$ & $70.9_{\pm2.0}$ \\
cpp & 2296  & $91.9_{\pm2.8}$ & $91.7_{\pm3.0}$ & $91.7_{\pm2.9}$ & $88.4_{\pm2.4}$ & $81.8_{\pm3.2}$ & $78.6_{\pm3.4}$ & $71.3_{\pm5.0}$ & $73.7_{\pm7.8}$ & $77.0_{\pm5.4}$ & $85.4_{\pm2.6}$ & $66.4_{\pm5.3}$ & $71.8_{\pm2.6}$ \\
bbp & 665  & $\mathbf{76.5_{\pm4.3}}$ & $\cellcolor{lightgray}{75.5_{\pm4.4}}$ & $75.0_{\pm3.1}$ & $71.2_{\pm7.1}$ & $59.7_{\pm6.2}$ & $59.7_{\pm6.8}$ & $58.8_{\pm7.3}$ & $60.0_{\pm7.3}$ & $62.3_{\pm7.5}$ & $54.9_{\pm5.3}$ & $60.8_{\pm7.7}$ & $47.1_{\pm6.9}$ \\
antimicrobial & 52941  & $93.9_{\pm0.4}$ & $93.2_{\pm0.3}$ & $92.6_{\pm0.4}$ & $90.4_{\pm0.5}$ & $87.5_{\pm0.7}$ & $88.1_{\pm0.6}$ & $80.5_{\pm0.9}$ & $88.0_{\pm0.7}$ & $76.2_{\pm0.7}$ & $94.3_{\pm0.4}$ & $88.5_{\pm0.8}$ & $85.0_{\pm0.6}$ \\
antibacterial & 28591  & $95.0_{\pm0.4}$ & $94.5_{\pm0.4}$ & $93.9_{\pm0.4}$ & $91.7_{\pm0.4}$ & $89.3_{\pm0.7}$ & $89.5_{\pm0.5}$ & $83.8_{\pm1.3}$ & $89.7_{\pm0.6}$ & $70.2_{\pm3.1}$ & $95.2_{\pm0.2}$ & $88.8_{\pm0.6}$ & $84.0_{\pm1.0}$ \\
antifungal & 12887  & $94.9_{\pm0.6}$ & $94.1_{\pm0.7}$ & $94.1_{\pm0.7}$ & $91.7_{\pm0.8}$ & $89.2_{\pm0.8}$ & $87.9_{\pm2.9}$ & $83.0_{\pm1.1}$ & $89.1_{\pm1.2}$ & $55.2_{\pm4.4}$ & $94.0_{\pm0.6}$ & $78.1_{\pm11.5}$ & $79.1_{\pm1.5}$ \\
\bottomrule
\end{tabular}}
\vspace{1.5mm}
\resizebox{\textwidth}{!}{%
\definecolor{lightgray}{gray}{0.9}
\begin{tabular}{l|c|ccccccccc}
\toprule
\multirow{2}{*}{Dataset} & \multirow{2}{*}{Size} & \multicolumn{9}{c}{PLM-based models} \\
\cmidrule(lr){3-11}
 & & ESM2-8M & ESM2-35M & ESM2-150M & ESM2-650M & DPLM-150M & DPLM-650M & ProtBERT & ESM2-8M-S & ESM2-150M-F \\
\midrule

nonfouling & 7200  & $\cellcolor{lightgray}{79.0_{\pm1.4}}$ & $78.6_{\pm1.0}$ & $78.8_{\pm1.2}$ & $78.8_{\pm1.3}$ & $78.9_{\pm0.6}$ & $78.7_{\pm1.4}$ & $59.5_{\pm12.9}$ & $78.7_{\pm1.2}$ & $\mathbf{79.3_{\pm0.9}}$ \\
cpp & 2296  & $92.2_{\pm2.6}$ & $\cellcolor{lightgray}{93.0_{\pm2.0}}$ & $91.9_{\pm3.3}$ & $92.5_{\pm3.5}$ & $91.6_{\pm2.8}$ & $\mathbf{93.2_{\pm2.3}}$ & $86.9_{\pm6.7}$ & $90.8_{\pm2.0}$ & $91.5_{\pm2.8}$ \\
bbp & 665  & $70.8_{\pm7.1}$ & $70.1_{\pm6.5}$ & $71.9_{\pm5.6}$ & $67.5_{\pm3.7}$ & $73.9_{\pm4.8}$ & $70.4_{\pm5.5}$ & $61.0_{\pm12.1}$ & $69.0_{\pm3.8}$ & $73.5_{\pm6.5}$ \\
antimicrobial & 52941  & $96.1_{\pm0.4}$ & $96.3_{\pm0.2}$ & $96.5_{\pm0.2}$ & $\mathbf{96.8_{\pm0.2}}$ & $96.4_{\pm0.2}$ & $96.3_{\pm0.4}$ & $83.4_{\pm3.3}$ & $95.2_{\pm0.2}$ & $\cellcolor{lightgray}{96.7_{\pm0.1}}$ \\
antibacterial & 28591  & $96.7_{\pm0.2}$ & $96.9_{\pm0.2}$ & $97.2_{\pm0.1}$ & $\mathbf{97.3_{\pm0.3}}$ & $97.1_{\pm0.3}$ & $97.2_{\pm0.2}$ & $91.3_{\pm6.3}$ & $96.0_{\pm0.2}$ & $\cellcolor{lightgray}{97.2_{\pm0.3}}$ \\
antifungal & 12887  & $96.6_{\pm0.7}$ & $96.9_{\pm0.6}$ & $97.0_{\pm0.5}$ & $97.0_{\pm0.6}$ & $\mathbf{97.1_{\pm0.4}}$ & $96.9_{\pm0.5}$ & $88.3_{\pm5.0}$ & $94.9_{\pm1.0}$ & $\cellcolor{lightgray}{97.1_{\pm0.4}}$ \\
\bottomrule
\end{tabular}}
\end{table}

\begin{table}[h!]
\centering
\caption{Performance of models on canonical peptide regression (MAE$\downarrow$, \%) with hybrid-split. Dataset sizes are shown separately; results are mean$_{\pm std}$. Best and second-best scores per row are in \textbf{bold} and gray shadow.}
\label{app:exp:tab:natural_reg_hydra}
\resizebox{\textwidth}{!}{
\definecolor{lightgray}{gray}{0.9}
\begin{tabular}{l|c|ccccc|cccc|ccc}
\toprule
\multirow{2}{*}{Dataset} & \multirow{2}{*}{Size} & \multicolumn{5}{c|}{FP-based models} & \multicolumn{4}{c|}{GNN-based models} & \multicolumn{3}{c}{SMILES-based models} \\
\cmidrule(lr){3-7} \cmidrule(lr){8-11} \cmidrule(lr){12-14}
 & & RF & XGBoost & LightGBM & GradBoost & AdaBoost & GCN & GAT & GIN & Pepland & ChemBERTa & PeptideCLM & PepDoRA \\
\midrule
E.coli\_mic & 3204 & $0.593_{\pm0.015}$ & $0.604_{\pm0.019}$ & $0.584_{\pm0.019}$ & $0.629_{\pm0.010}$ & $0.677_{\pm0.013}$ & $0.670_{\pm0.039}$ & $0.665_{\pm0.023}$ & $0.631_{\pm0.015}$ & $0.650_{\pm0.062}$ & $0.603_{\pm0.007}$ & $0.621_{\pm0.012}$ & $0.650_{\pm0.013}$ \\
S.aureus\_mic & 2822 & $0.569_{\pm0.035}$ & $0.588_{\pm0.022}$ & $0.572_{\pm0.025}$ & $0.613_{\pm0.029}$ & $0.651_{\pm0.022}$ & $0.643_{\pm0.017}$ & $0.639_{\pm0.028}$ & $0.635_{\pm0.032}$ & $0.648_{\pm0.016}$ & $0.627_{\pm0.011}$ & $0.659_{\pm0.045}$ & $0.657_{\pm0.029}$ \\
P.aeruginosa\_mic & 1490 & $0.540_{\pm0.030}$ & $0.559_{\pm0.025}$ & $0.550_{\pm0.026}$ & $0.555_{\pm0.019}$ & $0.586_{\pm0.006}$ & $0.612_{\pm0.047}$ & $0.579_{\pm0.027}$ & $0.537_{\pm0.032}$ & $0.566_{\pm0.037}$ & $0.520_{\pm0.041}$ & $0.529_{\pm0.035}$ & $0.546_{\pm0.029}$ \\
hemolytic\_hc50 & 1926 & $0.517_{\pm0.042}$ & $0.527_{\pm0.038}$ & $0.523_{\pm0.035}$ & $0.546_{\pm0.036}$ & $0.619_{\pm0.048}$ & $0.603_{\pm0.077}$ & $0.539_{\pm0.023}$ & $0.528_{\pm0.036}$ & $0.516_{\pm0.033}$ & $0.498_{\pm0.040}$ & $0.539_{\pm0.020}$ & $0.535_{\pm0.027}$ \\
\midrule
Average & - & 0.555 & 0.569 & 0.557 & 0.586 & 0.633 & 0.632 & 0.605 & 0.583 & 0.595 & 0.562 & 0.587 & 0.597 \\
\bottomrule
\end{tabular}}
\vspace{2mm}
\resizebox{\textwidth}{!}{
\definecolor{lightgray}{gray}{0.9}
\begin{tabular}{l|c|ccccccccc}
\toprule
\multirow{2}{*}{Dataset} & \multirow{2}{*}{Size} & \multicolumn{9}{c}{PLM-based models} \\
\cmidrule(lr){3-11}
 & & ESM2-8M & ESM2-35M & ESM2-150M & ESM2-650M & DPLM-150M & DPLM-650M & ProtBERT & ESM2-8M-S & ESM2-150M-F \\
\midrule
E.coli\_mic & 3204 & $0.539_{\pm0.023}$ & $0.527_{\pm0.004}$ & $\cellcolor{lightgray}{0.509_{\pm0.010}}$ & $\mathbf{0.488_{\pm0.008}}$ & $0.525_{\pm0.006}$ & $0.528_{\pm0.010}$ & $0.602_{\pm0.087}$ & $0.548_{\pm0.024}$ & $0.517_{\pm0.023}$ \\
S.aureus\_mic & 2822 & $0.545_{\pm0.030}$ & $0.549_{\pm0.022}$ & $\cellcolor{lightgray}{0.527_{\pm0.032}}$ & $\mathbf{0.522_{\pm0.022}}$ & $0.563_{\pm0.023}$ & $0.552_{\pm0.018}$ & $0.649_{\pm0.071}$ & $0.585_{\pm0.016}$ & $0.533_{\pm0.028}$ \\
P.aeruginosa\_mic & 1490 & $0.506_{\pm0.046}$ & $0.497_{\pm0.042}$ & $0.496_{\pm0.046}$ & $\mathbf{0.471_{\pm0.043}}$ & $0.524_{\pm0.036}$ & $0.493_{\pm0.042}$ & $0.586_{\pm0.012}$ & $0.516_{\pm0.031}$ & $\cellcolor{lightgray}{0.483_{\pm0.024}}$ \\
hemolytic\_hc50 & 1926 & $0.422_{\pm0.042}$ & $\cellcolor{lightgray}{0.400_{\pm0.021}}$ & $0.413_{\pm0.047}$ & $\mathbf{0.394_{\pm0.017}}$ & $0.404_{\pm0.028}$ & $0.419_{\pm0.014}$ & $0.497_{\pm0.071}$ & $0.463_{\pm0.011}$ & $0.412_{\pm0.020}$ \\
\midrule
Average & - & 0.503 & 0.493 & 0.486 & \textbf{0.469} & 0.504 & 0.498 & 0.584 & 0.528 & \cellcolor{lightgray}{0.486} \\
\bottomrule
\end{tabular}}
\end{table}
\begin{table}[h!]
\centering
\caption{Performance of models on canonical peptide regression (MAE$\downarrow$, \%) with kmer-split. Dataset sizes are shown separately; results are mean$_{\pm std}$. Best and second-best scores per row are in \textbf{bold} and gray shadow.}
\label{app:exp:tab:natural_reg_kmer}
\resizebox{\textwidth}{!}{
\definecolor{lightgray}{gray}{0.9}
\begin{tabular}{l|c|ccccc|cccc|ccc}
\toprule
\multirow{2}{*}{Dataset} & \multirow{2}{*}{Size} & \multicolumn{5}{c|}{FP-based models} & \multicolumn{4}{c|}{GNN-based models} & \multicolumn{3}{c}{SMILES-based models} \\
\cmidrule(lr){3-7} \cmidrule(lr){8-11} \cmidrule(lr){12-14}
 & & RF & XGBoost & LightGBM & GradBoost & AdaBoost & GCN & GAT & GIN & Pepland & ChemBERTa & PeptideCLM & PepDoRA \\
\midrule
E.coli\_mic & 3204 & $0.465_{\pm0.022}$ & $0.484_{\pm0.014}$ & $0.476_{\pm0.016}$ & $0.532_{\pm0.023}$ & $0.597_{\pm0.027}$ & $0.593_{\pm0.038}$ & $0.607_{\pm0.025}$ & $0.577_{\pm0.011}$ & $0.557_{\pm0.026}$ & $0.577_{\pm0.026}$ & $0.582_{\pm0.021}$ & $0.635_{\pm0.021}$ \\
S.aureus\_mic & 2822 & $0.450_{\pm0.025}$ & $0.467_{\pm0.022}$ & $0.450_{\pm0.014}$ & $0.513_{\pm0.029}$ & $0.575_{\pm0.036}$ & $0.572_{\pm0.016}$ & $0.573_{\pm0.029}$ & $0.559_{\pm0.020}$ & $0.576_{\pm0.027}$ & $0.551_{\pm0.029}$ & $0.554_{\pm0.022}$ & $0.586_{\pm0.033}$ \\
P.aeruginosa\_mic & 1490 & $0.435_{\pm0.018}$ & $0.440_{\pm0.032}$ & $0.439_{\pm0.024}$ & $0.466_{\pm0.020}$ & $0.510_{\pm0.015}$ & $0.517_{\pm0.024}$ & $0.539_{\pm0.024}$ & $0.504_{\pm0.014}$ & $0.521_{\pm0.017}$ & $0.493_{\pm0.032}$ & $0.511_{\pm0.016}$ & $0.533_{\pm0.029}$ \\
hemolytic\_hc50 & 1926 & $0.383_{\pm0.024}$ & $0.392_{\pm0.029}$ & $0.391_{\pm0.029}$ & $0.428_{\pm0.018}$ & $0.512_{\pm0.035}$ & $0.561_{\pm0.098}$ & $0.483_{\pm0.028}$ & $0.475_{\pm0.009}$ & $0.454_{\pm0.021}$ & $0.439_{\pm0.025}$ & $0.501_{\pm0.039}$ & $0.490_{\pm0.025}$ \\
\midrule
Average & - & 0.433 & 0.446 & 0.439 & 0.485 & 0.548 & 0.561 & 0.550 & 0.529 & 0.527 & 0.515 & 0.537 & 0.561 \\
\bottomrule
\end{tabular}}
\vspace{2mm}
\resizebox{\textwidth}{!}{
\definecolor{lightgray}{gray}{0.9}
\begin{tabular}{l|c|ccccccccc}
\toprule
\multirow{2}{*}{Dataset} & \multirow{2}{*}{Size} & \multicolumn{9}{c}{PLM-based models} \\
\cmidrule(lr){3-11}
 & & ESM2-8M & ESM2-35M & ESM2-150M & ESM2-650M & DPLM-150M & DPLM-650M & ProtBERT & ESM2-8M-S & ESM2-150M-F \\
\midrule
E.coli\_mic & 3204 & $0.430_{\pm0.027}$ & $0.418_{\pm0.029}$ & $0.431_{\pm0.043}$ & $\cellcolor{lightgray}{0.403_{\pm0.026}}$ & $0.430_{\pm0.021}$ & $0.423_{\pm0.027}$ & $0.478_{\pm0.111}$ & $0.494_{\pm0.035}$ & $\mathbf{0.403_{\pm0.022}}$ \\
S.aureus\_mic & 2822 & $0.454_{\pm0.021}$ & $0.438_{\pm0.032}$ & $\mathbf{0.401_{\pm0.018}}$ & $\cellcolor{lightgray}{0.409_{\pm0.022}}$ & $0.431_{\pm0.017}$ & $0.428_{\pm0.009}$ & $0.510_{\pm0.066}$ & $0.496_{\pm0.015}$ & $0.410_{\pm0.018}$ \\
P.aeruginosa\_mic & 1490 & $0.426_{\pm0.020}$ & $\cellcolor{lightgray}{0.400_{\pm0.031}}$ & $0.415_{\pm0.031}$ & $0.401_{\pm0.031}$ & $0.438_{\pm0.029}$ & $0.416_{\pm0.029}$ & $0.481_{\pm0.059}$ & $0.480_{\pm0.016}$ & $\mathbf{0.395_{\pm0.023}}$ \\
hemolytic\_hc50 & 1926 & $0.336_{\pm0.025}$ & $0.326_{\pm0.026}$ & $\mathbf{0.309_{\pm0.026}}$ & $0.318_{\pm0.015}$ & $0.347_{\pm0.016}$ & $0.324_{\pm0.011}$ & $0.399_{\pm0.062}$ & $0.407_{\pm0.016}$ & $\cellcolor{lightgray}{0.316_{\pm0.010}}$ \\
\midrule
Average & - & 0.412 & 0.396 & 0.389 & \cellcolor{lightgray}{0.383} & 0.412 & 0.398 & 0.467 & 0.469 & \textbf{0.381} \\
\bottomrule
\end{tabular}}
\end{table}
\begin{table}[h!]
\centering
\caption{Performance of models on canonical peptide regression (MAE$\downarrow$, \%) with MMseqs2-split. Dataset sizes are shown separately; results are mean$_{\pm std}$. Best and second-best scores per row are in \textbf{bold} and gray shadow.}
\label{app:exp:tab:natural_reg_mmseqs2}
\resizebox{\textwidth}{!}{
\definecolor{lightgray}{gray}{0.9}
\begin{tabular}{l|c|ccccc|cccc|ccc}
\toprule
\multirow{2}{*}{Dataset} & \multirow{2}{*}{Size} & \multicolumn{5}{c|}{FP-based models} & \multicolumn{4}{c|}{GNN-based models} & \multicolumn{3}{c}{SMILES-based models} \\
\cmidrule(lr){3-7} \cmidrule(lr){8-11} \cmidrule(lr){12-14}
 & & RF & XGBoost & LightGBM & GradBoost & AdaBoost & GCN & GAT & GIN & Pepland & ChemBERTa & PeptideCLM & PepDoRA \\
\midrule
E.coli\_mic & 3204 & $0.573_{\pm0.035}$ & $0.583_{\pm0.036}$ & $0.578_{\pm0.034}$ & $0.620_{\pm0.027}$ & $0.676_{\pm0.031}$ & $0.662_{\pm0.040}$ & $0.639_{\pm0.027}$ & $0.618_{\pm0.043}$ & $0.639_{\pm0.042}$ & $0.583_{\pm0.013}$ & $0.608_{\pm0.027}$ & $0.630_{\pm0.021}$ \\
S.aureus\_mic & 2822 & $0.568_{\pm0.026}$ & $0.592_{\pm0.024}$ & $0.577_{\pm0.026}$ & $0.620_{\pm0.022}$ & $0.644_{\pm0.026}$ & $0.631_{\pm0.012}$ & $0.624_{\pm0.025}$ & $0.616_{\pm0.022}$ & $0.638_{\pm0.043}$ & $0.615_{\pm0.023}$ & $0.636_{\pm0.027}$ & $0.644_{\pm0.033}$ \\
P.aeruginosa\_mic & 1490 & $0.518_{\pm0.030}$ & $0.523_{\pm0.032}$ & $0.526_{\pm0.023}$ & $0.532_{\pm0.020}$ & $0.577_{\pm0.023}$ & $0.575_{\pm0.049}$ & $0.565_{\pm0.019}$ & $0.558_{\pm0.036}$ & $0.561_{\pm0.013}$ & $0.547_{\pm0.036}$ & $0.561_{\pm0.042}$ & $0.568_{\pm0.028}$ \\
hemolytic\_hc50 & 1926 & $0.547_{\pm0.034}$ & $0.554_{\pm0.047}$ & $0.543_{\pm0.039}$ & $0.561_{\pm0.038}$ & $0.627_{\pm0.032}$ & $0.617_{\pm0.075}$ & $0.548_{\pm0.034}$ & $0.532_{\pm0.028}$ & $0.540_{\pm0.017}$ & $0.532_{\pm0.045}$ & $0.556_{\pm0.023}$ & $0.538_{\pm0.031}$ \\
\midrule
Average & - & 0.552 & 0.563 & 0.556 & 0.583 & 0.631 & 0.621 & 0.594 & 0.581 & 0.594 & 0.569 & 0.590 & 0.595 \\
\bottomrule
\end{tabular}}
\vspace{2mm}
\resizebox{\textwidth}{!}{
\definecolor{lightgray}{gray}{0.9}
\begin{tabular}{l|c|ccccccccc}
\toprule
\multirow{2}{*}{Dataset} & \multirow{2}{*}{Size} & \multicolumn{9}{c}{PLM-based models} \\
\cmidrule(lr){3-11}
 & & ESM2-8M & ESM2-35M & ESM2-150M & ESM2-650M & DPLM-150M & DPLM-650M & ProtBERT & ESM2-8M-S & ESM2-150M-F \\
\midrule
E.coli\_mic & 3204 & $0.517_{\pm0.018}$ & $0.497_{\pm0.022}$ & $0.489_{\pm0.022}$ & $\mathbf{0.472_{\pm0.018}}$ & $0.523_{\pm0.012}$ & $0.500_{\pm0.031}$ & $0.607_{\pm0.097}$ & $0.543_{\pm0.018}$ & $\cellcolor{lightgray}{0.488_{\pm0.022}}$ \\
S.aureus\_mic & 2822 & $0.569_{\pm0.007}$ & $0.548_{\pm0.036}$ & $0.539_{\pm0.021}$ & $\cellcolor{lightgray}{0.526_{\pm0.034}}$ & $0.566_{\pm0.051}$ & $0.536_{\pm0.021}$ & $0.622_{\pm0.021}$ & $0.578_{\pm0.023}$ & $\mathbf{0.523_{\pm0.018}}$ \\
P.aeruginosa\_mic & 1490 & $0.523_{\pm0.026}$ & $0.514_{\pm0.040}$ & $0.495_{\pm0.044}$ & $\mathbf{0.471_{\pm0.023}}$ & $0.528_{\pm0.044}$ & $0.497_{\pm0.053}$ & $0.557_{\pm0.054}$ & $0.529_{\pm0.022}$ & $\cellcolor{lightgray}{0.476_{\pm0.030}}$ \\
hemolytic\_hc50 & 1926 & $0.431_{\pm0.029}$ & $0.432_{\pm0.038}$ & $\mathbf{0.414_{\pm0.013}}$ & $0.424_{\pm0.039}$ & $0.422_{\pm0.025}$ & $\cellcolor{lightgray}{0.421_{\pm0.023}}$ & $0.491_{\pm0.045}$ & $0.468_{\pm0.022}$ & $0.432_{\pm0.020}$ \\
\midrule
Average & - & 0.510 & 0.498 & 0.484 & \textbf{0.473} & 0.510 & 0.488 & 0.569 & 0.530 & \cellcolor{lightgray}{0.480} \\
\bottomrule
\end{tabular}}
\end{table}

\begin{table}[h!]
\centering
\caption{Performance of models on canonical peptide regression (MAE$\downarrow$, \%) with random-split. Dataset sizes are shown separately; results are mean$_{\pm std}$. Best and second-best scores per row are in \textbf{bold} and gray shadow.}
\label{app:exp:tab:natural_reg_random}
\resizebox{\textwidth}{!}{
\definecolor{lightgray}{gray}{0.9}
\begin{tabular}{l|c|ccccc|cccc|ccc}
\toprule
\multirow{2}{*}{Dataset} & \multirow{2}{*}{Size} & \multicolumn{5}{c|}{FP-based models} & \multicolumn{4}{c|}{GNN-based models} & \multicolumn{3}{c}{SMILES-based models} \\
\cmidrule(lr){3-7} \cmidrule(lr){8-11} \cmidrule(lr){12-14}
 & & RF & XGBoost & LightGBM & GradBoost & AdaBoost & GCN & GAT & GIN & Pepland & ChemBERTa & PeptideCLM & PepDoRA \\
\midrule
E.coli\_mic & 3204 & $0.468_{\pm0.022}$ & $0.484_{\pm0.014}$ & $0.476_{\pm0.016}$ & $0.532_{\pm0.023}$ & $0.593_{\pm0.025}$ & $0.587_{\pm0.028}$ & $0.613_{\pm0.025}$ & $0.570_{\pm0.025}$ & $0.565_{\pm0.017}$ & $0.551_{\pm0.045}$ & $0.577_{\pm0.028}$ & $0.640_{\pm0.021}$ \\
S.aureus\_mic & 2822 & $0.448_{\pm0.020}$ & $0.467_{\pm0.022}$ & $0.450_{\pm0.014}$ & $0.513_{\pm0.029}$ & $0.577_{\pm0.036}$ & $0.572_{\pm0.028}$ & $0.573_{\pm0.036}$ & $0.554_{\pm0.021}$ & $0.581_{\pm0.021}$ & $0.558_{\pm0.040}$ & $0.552_{\pm0.018}$ & $0.587_{\pm0.033}$ \\
P.aeruginosa\_mic & 1490 & $0.433_{\pm0.021}$ & $0.440_{\pm0.032}$ & $0.439_{\pm0.024}$ & $0.465_{\pm0.020}$ & $0.508_{\pm0.020}$ & $0.532_{\pm0.037}$ & $0.529_{\pm0.021}$ & $0.499_{\pm0.015}$ & $0.513_{\pm0.021}$ & $0.491_{\pm0.033}$ & $0.513_{\pm0.032}$ & $0.534_{\pm0.032}$ \\
hemolytic\_hc50 & 1926 & $0.377_{\pm0.026}$ & $0.392_{\pm0.029}$ & $0.391_{\pm0.029}$ & $0.428_{\pm0.018}$ & $0.519_{\pm0.022}$ & $0.515_{\pm0.058}$ & $0.488_{\pm0.035}$ & $0.468_{\pm0.021}$ & $0.454_{\pm0.023}$ & $0.433_{\pm0.019}$ & $0.494_{\pm0.034}$ & $0.490_{\pm0.025}$ \\
\midrule
Average & - & 0.432 & 0.446 & 0.439 & 0.485 & 0.549 & 0.551 & 0.551 & 0.523 & 0.528 & 0.508 & 0.534 & 0.563 \\
\bottomrule
\end{tabular}}
\vspace{2mm}
\resizebox{\textwidth}{!}{
\definecolor{lightgray}{gray}{0.9}
\begin{tabular}{l|c|ccccccccc}
\toprule
\multirow{2}{*}{Dataset} & \multirow{2}{*}{Size} & \multicolumn{9}{c}{PLM-based models} \\
\cmidrule(lr){3-11}
 & & ESM2-8M & ESM2-35M & ESM2-150M & ESM2-650M & DPLM-150M & DPLM-650M & ProtBERT & ESM2-8M-S & ESM2-150M-F \\
\midrule
E.coli\_mic & 3204 & $0.441_{\pm0.026}$ & $0.445_{\pm0.021}$ & $\cellcolor{lightgray}{0.406_{\pm0.027}}$ & $\mathbf{0.402_{\pm0.024}}$ & $0.423_{\pm0.039}$ & $0.428_{\pm0.038}$ & $0.534_{\pm0.113}$ & $0.477_{\pm0.022}$ & $0.411_{\pm0.024}$ \\
S.aureus\_mic & 2822 & $0.438_{\pm0.011}$ & $0.418_{\pm0.014}$ & $0.414_{\pm0.014}$ & $\mathbf{0.403_{\pm0.019}}$ & $0.444_{\pm0.035}$ & $0.437_{\pm0.037}$ & $0.579_{\pm0.084}$ & $0.495_{\pm0.020}$ & $\cellcolor{lightgray}{0.412_{\pm0.018}}$ \\
P.aeruginosa\_mic & 1490 & $0.425_{\pm0.039}$ & $0.405_{\pm0.020}$ & $0.411_{\pm0.020}$ & $\mathbf{0.377_{\pm0.012}}$ & $0.430_{\pm0.038}$ & $0.419_{\pm0.027}$ & $0.485_{\pm0.053}$ & $0.476_{\pm0.021}$ & $\cellcolor{lightgray}{0.395_{\pm0.021}}$ \\
hemolytic\_hc50 & 1926 & $0.342_{\pm0.020}$ & $0.341_{\pm0.022}$ & $\cellcolor{lightgray}{0.322_{\pm0.031}}$ & $\mathbf{0.317_{\pm0.031}}$ & $0.324_{\pm0.035}$ & $0.348_{\pm0.016}$ & $0.418_{\pm0.108}$ & $0.395_{\pm0.019}$ & $0.329_{\pm0.016}$ \\
\midrule
Average & - & 0.411 & 0.402 & 0.388 & \textbf{0.375} & 0.405 & 0.408 & 0.504 & 0.461 & \cellcolor{lightgray}{0.387} \\
\bottomrule
\end{tabular}}
\end{table}
\begin{table}[h]
\centering
\caption{Performance of models on non-canonical peptide classification (ROC-AUC$\uparrow$, \%) and regression (MAE$\downarrow$, \%) with ECFP-split. Dataset sizes are shown separately; results are mean$_{\pm std}$. Best and second-best scores per row are in \textbf{bold} and gray shadow.}
\label{app:exp:tab:non_natural_ecfp}
\resizebox{\textwidth}{!}{%
\definecolor{lightgray}{gray}{0.9}
\begin{tabular}{l|l|l|c|ccccc|cccc|ccc}
\toprule
\multirow{2}{*}{Task} & \multirow{2}{*}{Metric} & \multirow{2}{*}{Dataset} & \multirow{2}{*}{Size} & \multicolumn{5}{c|}{FP-based models} & \multicolumn{4}{c|}{GNN-based models} & \multicolumn{3}{c}{SMILES-based models} \\
\cmidrule(lr){5-9} \cmidrule(lr){10-13} \cmidrule(lr){14-16}
 & & & & RF & XGBoost & LightGBM & GradBoost & AdaBoost & GCN & GAT & GIN & Pepland & ChemBERTa & PeptideCLM & PepDoRA \\
\midrule

\multirow{6}{*}{cls} & \multirow{6}{*}{AUC ROC} & nc-antibacterial & 1668& $\mathbf{94.4_{\pm1.4}}$& $93.6_{\pm2.2}$& $\cellcolor{lightgray}{93.8_{\pm1.8}}$& $92.5_{\pm2.8}$& $90.9_{\pm3.4}$& $93.0_{\pm2.6}$& $81.5_{\pm3.5}$& $90.7_{\pm4.9}$& $84.9_{\pm3.9}$& $91.9_{\pm2.7}$& $71.4_{\pm1.8}$& $72.4_{\pm7.4}$ \\
& & nc-antifungal & 407& $\cellcolor{lightgray}{95.4_{\pm3.2}}$& $\mathbf{96.5_{\pm2.9}}$& $94.8_{\pm3.1}$& $95.2_{\pm3.8}$& $95.4_{\pm2.4}$& $70.1_{\pm12.0}$& $78.5_{\pm6.2}$& $86.1_{\pm2.8}$& $83.5_{\pm4.4}$& $78.0_{\pm18.3}$& $65.4_{\pm7.0}$& $65.9_{\pm12.0}$ \\
& & nc-antimicrobial & 2465& $97.6_{\pm0.9}$& $\cellcolor{lightgray}{97.7_{\pm0.9}}$& $\mathbf{97.8_{\pm0.6}}$& $97.3_{\pm0.7}$& $95.3_{\pm1.8}$& $94.9_{\pm1.6}$& $79.2_{\pm4.4}$& $91.5_{\pm2.3}$& $88.0_{\pm0.9}$& $95.2_{\pm1.9}$& $68.3_{\pm4.7}$& $80.0_{\pm1.9}$ \\
& & nc-hemolytic & 3971& $96.1_{\pm1.1}$& $\cellcolor{lightgray}{96.2_{\pm0.5}}$& $\mathbf{96.3_{\pm0.7}}$& $95.3_{\pm0.8}$& $93.7_{\pm1.6}$& $89.5_{\pm3.7}$& $85.7_{\pm2.6}$& $89.0_{\pm4.1}$& $82.8_{\pm2.1}$& $91.4_{\pm1.5}$& $76.1_{\pm4.5}$& $72.1_{\pm7.3}$ \\
\cmidrule(lr){3-16}
& & avg & -& $\cellcolor{lightgray}{95.9}$& $\mathbf{96.0}$& $95.7$& $95.1$& $93.8$& $86.9$& $81.2$& $89.3$& $84.8$& $89.1$& $70.3$& $72.6$ \\
\bottomrule
\end{tabular}}
\end{table}

\begin{table}[h]
\centering
\caption{Performance of models on non-canonical peptide classification (ROC-AUC$\uparrow$, \%) and regression (MAE$\downarrow$, \%) with hybrid-split. Dataset sizes are shown separately; results are mean$_{\pm std}$. Best and second-best scores per row are in \textbf{bold} and gray shadow.}
\label{app:exp:tab:non_natural_hydra}
\resizebox{\textwidth}{!}{%
\definecolor{lightgray}{gray}{0.9}
\begin{tabular}{l|l|l|c|ccccc|cccc|ccc}
\toprule
\multirow{2}{*}{Task} & \multirow{2}{*}{Metric} & \multirow{2}{*}{Dataset} & \multirow{2}{*}{Size} & \multicolumn{5}{c|}{FP-based models} & \multicolumn{4}{c|}{GNN-based models} & \multicolumn{3}{c}{SMILES-based models} \\
\cmidrule(lr){5-9} \cmidrule(lr){10-13} \cmidrule(lr){14-16}
 & & & & RF & XGBoost & LightGBM & GradBoost & AdaBoost & GCN & GAT & GIN & Pepland & ChemBERTa & PeptideCLM & PepDoRA \\
\midrule

\multirow{6}{*}{cls} & \multirow{6}{*}{AUC ROC} & nc-antibacterial & 1668& $\mathbf{94.1_{\pm1.4}}$& $93.8_{\pm1.9}$& $\cellcolor{lightgray}{94.0_{\pm2.2}}$& $\cellcolor{lightgray}{94.0_{\pm2.4}}$& $91.8_{\pm2.6}$& $88.1_{\pm6.6}$& $82.0_{\pm4.0}$& $89.7_{\pm1.9}$& $82.9_{\pm1.6}$& $92.2_{\pm1.8}$& $71.9_{\pm4.5}$& $75.1_{\pm8.8}$ \\
& & nc-antifungal & 407& $\mathbf{97.3_{\pm1.1}}$& $\cellcolor{lightgray}{96.0_{\pm1.9}}$& $96.0_{\pm1.6}$& $94.9_{\pm1.1}$& $96.0_{\pm1.9}$& $70.5_{\pm2.9}$& $77.4_{\pm9.8}$& $81.5_{\pm8.5}$& $84.5_{\pm2.4}$& $87.3_{\pm6.9}$& $69.8_{\pm8.2}$& $61.8_{\pm14.1}$ \\
& & nc-antimicrobial & 2465& $96.8_{\pm1.3}$& $\cellcolor{lightgray}{96.9_{\pm0.7}}$& $\mathbf{97.0_{\pm0.9}}$& $96.7_{\pm0.4}$& $95.6_{\pm1.2}$& $91.8_{\pm1.3}$& $83.5_{\pm2.7}$& $89.2_{\pm3.1}$& $86.9_{\pm2.0}$& $95.2_{\pm0.8}$& $75.4_{\pm2.6}$& $80.2_{\pm4.6}$ \\
& & nc-hemolytic & 3971& $\cellcolor{lightgray}{96.5_{\pm1.3}}$& $96.0_{\pm1.4}$& $\mathbf{96.8_{\pm1.2}}$& $95.7_{\pm1.7}$& $93.8_{\pm2.2}$& $90.9_{\pm3.8}$& $83.9_{\pm5.4}$& $91.2_{\pm2.8}$& $84.3_{\pm4.8}$& $92.0_{\pm2.5}$& $76.1_{\pm9.5}$& $76.8_{\pm7.3}$ \\
\cmidrule(lr){3-16}
& & avg & -& $\mathbf{96.2}$& $95.7$& $\cellcolor{lightgray}{95.9}$& $95.3$& $94.3$& $85.3$& $81.7$& $87.9$& $84.7$& $91.7$& $73.3$& $73.5$ \\
\bottomrule
\end{tabular}}
\end{table}
\begin{table}[h]
\centering
\caption{Performance of models on non-canonical peptide classification (ROC-AUC$\uparrow$, \%) and regression (MAE$\downarrow$, \%) with kmer-split. Dataset sizes are shown separately; results are mean$_{\pm std}$. Best and second-best scores per row are in \textbf{bold} and gray shadow.}
\label{app:exp:tab:non_natural_kmer}
\resizebox{\textwidth}{!}{%
\definecolor{lightgray}{gray}{0.9}
\begin{tabular}{l|l|l|c|ccccc|cccc|ccc}
\toprule
\multirow{2}{*}{Task} & \multirow{2}{*}{Metric} & \multirow{2}{*}{Dataset} & \multirow{2}{*}{Size} & \multicolumn{5}{c|}{FP-based models} & \multicolumn{4}{c|}{GNN-based models} & \multicolumn{3}{c}{SMILES-based models} \\
\cmidrule(lr){5-9} \cmidrule(lr){10-13} \cmidrule(lr){14-16}
 & & & & RF & XGBoost & LightGBM & GradBoost & AdaBoost & GCN & GAT & GIN & Pepland & ChemBERTa & PeptideCLM & PepDoRA \\
\midrule

\multirow{6}{*}{cls} & \multirow{6}{*}{AUC ROC} & nc-antibacterial & 1668& $94.7_{\pm0.9}$& $\cellcolor{lightgray}{94.8_{\pm0.9}}$& $\mathbf{95.2_{\pm1.3}}$& $94.4_{\pm0.8}$& $91.6_{\pm2.3}$& $92.2_{\pm3.0}$& $80.6_{\pm7.4}$& $85.6_{\pm12.0}$& $83.6_{\pm4.9}$& $91.9_{\pm2.8}$& $72.2_{\pm5.7}$& $73.2_{\pm5.6}$ \\
& & nc-antifungal & 407& $\mathbf{98.9_{\pm1.1}}$& $98.7_{\pm1.0}$& $\cellcolor{lightgray}{98.9_{\pm0.6}}$& $98.3_{\pm0.9}$& $98.1_{\pm0.9}$& $71.8_{\pm12.8}$& $77.7_{\pm9.8}$& $89.6_{\pm5.3}$& $77.3_{\pm20.1}$& $91.4_{\pm14.6}$& $65.9_{\pm9.3}$& $57.2_{\pm10.7}$ \\
& & nc-antimicrobial & 2465& $\mathbf{97.9_{\pm0.9}}$& $97.4_{\pm0.9}$& $\cellcolor{lightgray}{97.6_{\pm0.8}}$& $97.6_{\pm0.8}$& $95.7_{\pm1.4}$& $93.5_{\pm1.8}$& $82.6_{\pm5.8}$& $92.2_{\pm3.5}$& $87.8_{\pm1.5}$& $96.1_{\pm0.8}$& $77.9_{\pm3.0}$& $81.1_{\pm1.7}$ \\
& & nc-hemolytic & 3971& $97.1_{\pm0.9}$& $\cellcolor{lightgray}{97.4_{\pm0.9}}$& $\mathbf{97.8_{\pm0.7}}$& $96.7_{\pm0.9}$& $94.0_{\pm0.8}$& $86.9_{\pm4.2}$& $81.8_{\pm0.9}$& $88.3_{\pm3.7}$& $83.2_{\pm2.7}$& $93.2_{\pm1.2}$& $75.1_{\pm3.8}$& $77.9_{\pm3.1}$ \\
\cmidrule(lr){3-16}
& & avg & -& $\cellcolor{lightgray}{97.2}$& $97.1$& $\mathbf{97.4}$& $96.7$& $94.9$& $86.1$& $80.7$& $88.9$& $83.0$& $93.1$& $72.8$& $72.4$ \\
\bottomrule
\end{tabular}}
\end{table}
\begin{table}[h]
\centering
\caption{Performance of models on non-canonical peptide classification (ROC-AUC$\uparrow$, \%) and regression (MAE$\downarrow$, \%) with MMseqs2-split. Dataset sizes are shown separately; results are mean$_{\pm std}$. Best and second-best scores per row are in \textbf{bold} and gray shadow.}
\label{app:exp:tab:non_natural_mmseqs2}
\resizebox{\textwidth}{!}{%
\definecolor{lightgray}{gray}{0.9}
\begin{tabular}{l|l|l|c|ccccc|cccc|ccc}
\toprule
\multirow{2}{*}{Task} & \multirow{2}{*}{Metric} & \multirow{2}{*}{Dataset} & \multirow{2}{*}{Size} & \multicolumn{5}{c|}{FP-based models} & \multicolumn{4}{c|}{GNN-based models} & \multicolumn{3}{c}{SMILES-based models} \\
\cmidrule(lr){5-9} \cmidrule(lr){10-13} \cmidrule(lr){14-16}
 & & & & RF & XGBoost & LightGBM & GradBoost & AdaBoost & GCN & GAT & GIN & Pepland & ChemBERTa & PeptideCLM & PepDoRA \\
\midrule

\multirow{6}{*}{cls} & \multirow{6}{*}{AUC ROC} & nc-antibacterial & 1668& $\mathbf{97.9_{\pm1.2}}$& $97.7_{\pm0.7}$& $\cellcolor{lightgray}{97.8_{\pm1.1}}$& $97.1_{\pm1.1}$& $95.7_{\pm1.4}$& $96.2_{\pm1.6}$& $89.9_{\pm3.9}$& $95.8_{\pm3.5}$& $90.7_{\pm2.3}$& $96.9_{\pm1.6}$& $84.5_{\pm2.0}$& $86.4_{\pm3.5}$ \\
& & nc-antifungal & 407& $\cellcolor{lightgray}{95.4_{\pm3.4}}$& $93.5_{\pm2.9}$& $93.6_{\pm3.4}$& $93.2_{\pm4.6}$& $\mathbf{95.4_{\pm2.5}}$& $73.8_{\pm1.7}$& $79.8_{\pm5.6}$& $85.2_{\pm4.3}$& $84.3_{\pm5.9}$& $92.2_{\pm7.0}$& $75.3_{\pm5.0}$& $61.5_{\pm5.1}$ \\
& & nc-antimicrobial & 2465& $\mathbf{98.0_{\pm0.4}}$& $97.8_{\pm0.4}$& $\cellcolor{lightgray}{97.9_{\pm0.5}}$& $97.8_{\pm0.5}$& $96.4_{\pm0.6}$& $95.8_{\pm1.1}$& $87.4_{\pm3.6}$& $95.6_{\pm1.8}$& $89.4_{\pm1.1}$& $96.7_{\pm1.1}$& $82.1_{\pm1.3}$& $87.5_{\pm2.1}$ \\
& & nc-hemolytic & 3971& $\cellcolor{lightgray}{98.2_{\pm0.4}}$& $98.1_{\pm0.6}$& $\mathbf{98.3_{\pm0.5}}$& $97.5_{\pm0.7}$& $95.7_{\pm0.7}$& $94.4_{\pm1.1}$& $89.8_{\pm2.9}$& $94.9_{\pm1.1}$& $90.8_{\pm0.7}$& $96.2_{\pm1.1}$& $85.4_{\pm2.5}$& $87.9_{\pm1.2}$ \\
\cmidrule(lr){3-16}
& & avg & -& $\mathbf{97.4}$& $96.8$& $\cellcolor{lightgray}{96.9}$& $96.4$& $95.8$& $90.1$& $86.7$& $92.9$& $88.8$& $95.5$& $81.8$& $80.8$ \\
\bottomrule
\end{tabular}}
\end{table}
\begin{table}[h]
\centering
\caption{Performance of models on non-canonical peptide classification (ROC-AUC$\uparrow$, \%) and regression (MAE$\downarrow$, \%) with random-split. Dataset sizes are shown separately; results are mean$_{\pm std}$. Best and second-best scores per row are in \textbf{bold} and gray shadow.}
\label{app:exp:tab:non_natural_random}
\resizebox{\textwidth}{!}{%
\definecolor{lightgray}{gray}{0.9}
\begin{tabular}{l|l|l|c|ccccc|cccc|ccc}
\toprule
\multirow{2}{*}{Task} & \multirow{2}{*}{Metric} & \multirow{2}{*}{Dataset} & \multirow{2}{*}{Size} & \multicolumn{5}{c|}{FP-based models} & \multicolumn{4}{c|}{GNN-based models} & \multicolumn{3}{c}{SMILES-based models} \\
\cmidrule(lr){5-9} \cmidrule(lr){10-13} \cmidrule(lr){14-16}
 & & & & RF & XGBoost & LightGBM & GradBoost & AdaBoost & GCN & GAT & GIN & Pepland & ChemBERTa & PeptideCLM & PepDoRA \\
\midrule

\multirow{6}{*}{cls} & \multirow{6}{*}{AUC ROC} & nc-antibacterial & 1668& $\cellcolor{lightgray}{98.4_{\pm1.1}}$& $98.4_{\pm1.0}$& $\mathbf{98.7_{\pm0.9}}$& $98.1_{\pm1.0}$& $97.0_{\pm1.1}$& $97.1_{\pm0.6}$& $94.1_{\pm1.5}$& $96.6_{\pm2.0}$& $92.5_{\pm2.8}$& $97.8_{\pm1.1}$& $85.5_{\pm4.3}$& $86.0_{\pm3.7}$ \\
& & nc-antifungal & 407& $\mathbf{97.5_{\pm2.5}}$& $\cellcolor{lightgray}{96.6_{\pm2.6}}$& $96.2_{\pm3.2}$& $96.2_{\pm3.1}$& $95.1_{\pm3.6}$& $85.1_{\pm11.6}$& $83.4_{\pm5.8}$& $85.1_{\pm8.5}$& $85.8_{\pm9.7}$& $95.2_{\pm4.9}$& $79.1_{\pm6.4}$& $72.2_{\pm11.6}$ \\
& & nc-antimicrobial & 2465& $\mathbf{99.1_{\pm0.4}}$& $98.9_{\pm0.3}$& $\cellcolor{lightgray}{99.0_{\pm0.3}}$& $98.6_{\pm0.3}$& $97.7_{\pm0.6}$& $96.2_{\pm1.7}$& $91.6_{\pm1.9}$& $97.0_{\pm1.0}$& $92.7_{\pm1.7}$& $97.9_{\pm0.8}$& $88.2_{\pm1.5}$& $85.9_{\pm2.7}$ \\
& & -hemolytic & 3971& $\mathbf{99.0_{\pm0.2}}$& $98.7_{\pm0.2}$& $\cellcolor{lightgray}{98.9_{\pm0.2}}$& $98.2_{\pm0.5}$& $96.6_{\pm0.8}$& $95.7_{\pm0.9}$& $89.1_{\pm1.9}$& $94.3_{\pm3.0}$& $89.1_{\pm1.5}$& $97.6_{\pm0.8}$& $83.6_{\pm2.2}$& $87.2_{\pm2.1}$ \\
\cmidrule(lr){3-16}
& & avg & -& $\mathbf{98.5}$& $98.1$& $\cellcolor{lightgray}{98.2}$& $97.8$& $96.6$& $93.5$& $89.5$& $93.3$& $90.0$& $97.1$& $84.1$& $82.9$ \\
\bottomrule
\end{tabular}}
\end{table}
\begin{table}[h]
    \centering
    \caption{Performance of models on PepPI classification (ROC-AUC$\uparrow$, \%) and regression (MAE$\downarrow$, \%) with cold-start-split. Dataset sizes are shown separately; results are mean$_{\pm std}$. Best and second-best scores per row are in \textbf{bold} and gray shadow.}
    \label{app:exp:tab:ppi}
    \resizebox{\textwidth}{!}{%
    \definecolor{lightgray}{gray}{0.9}
    \begin{tabular}{ll|l|c|cc|cccc|ccc|ccccc}
    \toprule
    \multirow{2}{*}{Task} & \multirow{2}{*}{Metric} & \multirow{2}{*}{Dataset} & \multirow{2}{*}{Size} & \multicolumn{2}{c|}{FP-based models} & \multicolumn{4}{c|}{GNN-based models} & \multicolumn{3}{c|}{SMILES-based models} & \multicolumn{5}{c|}{PLM-based models} \\
    \cmidrule(lr){5-6} \cmidrule(lr){7-10} \cmidrule(lr){11-13} \cmidrule(lr){14-18}
    & & & & ECFP4 & ECFP6 & GCN & GAT & GIN & Pepland & ChemBERTa & PeptideCLM & PepDoRA & ESM2-8M & ESM2-35M & ESM2-150M & ESM2-650M & ESM2-150M-F \\
    \midrule

cls & AUC ROC & ppi & 44148  & $48.7_{\pm3.2}$ & $46.7_{\pm3.9}$ & $47.5_{\pm2.3}$ & $49.5_{\pm4.3}$ & $51.1_{\pm3.9}$ & $50.5_{\pm3.5}$ & $51.9_{\pm4.4}$ & $52.4_{\pm2.9}$ & $51.0_{\pm2.5}$ & $50.6_{\pm6.6}$ & $51.8_{\pm4.4}$ & $52.6_{\pm2.3}$ & $\mathbf{56.0_{\pm3.1}}$ & $\cellcolor{lightgray}{55.0_{\pm2.9}}$ \\
\midrule
\multirow{2}{*}{reg} & \multirow{2}{*}{MAE} & ppi\_ba & 1433& $\cellcolor{lightgray}{1.040_{\pm0.052}}$& $1.052_{\pm0.063}$& $1.103_{\pm0.032}$& $1.128_{\pm0.030}$& $1.112_{\pm0.030}$& $1.091_{\pm0.048}$& $1.053_{\pm0.025}$& $1.124_{\pm0.104}$& $1.125_{\pm0.031}$& $1.059_{\pm0.043}$& $1.052_{\pm0.062}$& $1.062_{\pm0.060}$& $1.082_{\pm0.013}$& $\mathbf{1.026_{\pm0.043}}$ \\
& & nc\_ppi\_ba & 278& $\cellcolor{lightgray}{1.425_{\pm0.249}}$& $1.462_{\pm0.177}$& $1.468_{\pm0.183}$& $1.451_{\pm0.218}$& $1.489_{\pm0.208}$& $1.584_{\pm0.144}$& $1.473_{\pm0.217}$& $1.474_{\pm0.129}$& $\mathbf{1.422_{\pm0.100}}$& -& -& -& -& - \\
\bottomrule
\end{tabular}}
\end{table}

\begin{table}[h]
    \centering
    \caption{Performance of models on PepPI classification (ROC-AUC$\uparrow$, \%) and regression (MAE$\downarrow$, \%) with cold-start-split. Dataset sizes are shown separately; results are mean$_{\pm std}$. Best and second-best scores per row are in \textbf{bold} and gray shadow.}
    \label{app:exp:tab:ppi_emb}
    \resizebox{\textwidth}{!}{%
    \definecolor{lightgray}{gray}{0.9}
    \begin{tabular}{ll|l|c|cc|cc|ccc|ccccc}
    \toprule
    \multirow{2}{*}{Task} & \multirow{2}{*}{Metric} & \multirow{2}{*}{Dataset} & \multirow{2}{*}{Size} & \multicolumn{2}{c|}{FP-based models} & \multicolumn{2}{c|}{GNN-based models} & \multicolumn{3}{c|}{SMILES-based models} & \multicolumn{5}{c}{PLM-based models} \\
    \cmidrule(lr){5-6} \cmidrule(lr){7-8} \cmidrule(lr){9-11} \cmidrule(lr){12-16}
    & & & & ECFP4 & ECFP6 & GIN & Pepland & ChemBERTa & PeptideCLM & PepDoRA & ESM2-8M & ESM2-35M & ESM2-150M & ESM2-650M & ESM2-150M-F \\
    \midrule
    
    cls & AUC ROC & ppi & 44148  & $54.4_{\pm2.3}$ & $53.7_{\pm2.3}$ & $61.3_{\pm7.2}$ & $59.6_{\pm2.7}$ & $52.0_{\pm4.4}$ & $51.4_{\pm3.6}$ & $59.4_{\pm2.8}$ & $\mathbf{60.2_{\pm6.3}}$ & \cellcolor{lightgray}$57.6_{\pm7.2}$ & $55.4_{\pm3.5}$ & $51.9_{\pm2.4}$ & $56.0_{\pm3.1}$ \\
    \midrule
    \multirow{2}{*}{reg} & \multirow{2}{*}{MAE} & ppi\_ba & 1433 & $1.043_{\pm0.050}$& $1.043_{\pm0.060}$& $1.189_{\pm0.140}$& $1.176_{\pm0.152}$& $\mathbf{1.034_{\pm0.013}}$& $1.128_{\pm0.042}$& $1.084_{\pm0.044}$& $1.061_{\pm0.043}$& $1.056_{\pm0.092}$& $1.079_{\pm0.038}$& $1.051_{\pm0.072}$& \cellcolor{lightgray}$1.038_{\pm0.030}$ \\
    & & nc\_ppi\_ba & 278 & $1.665_{\pm0.260}$& $1.647_{\pm0.286}$& $1.741_{\pm0.359}$& $1.705_{\pm0.366}$ & $1.613_{\pm0.189}$& \cellcolor{lightgray}$1.580_{\pm0.142}$& $\mathbf{1.465_{\pm0.234}}$& -& -& -& -& - \\

\bottomrule
\end{tabular}}
\end{table}

\begin{table}[h!]
\centering
\caption{Models' classification (ROC-AUC$\uparrow$, \%) performance on the canonical peptide dataset (length $>$ 150) using hybrid-split. Dataset sizes are shown separately; results are mean$_{\pm std}$. Best and second-best scores per row are in \textbf{bold} and gray shadow.}
\label{app:exp:tab:150}
\resizebox{\textwidth}{!}{%
\definecolor{lightgray}{gray}{0.9}
\begin{tabular}{l|c|ccccc|cccc}
\toprule
\multirow{2}{*}{Dataset} & \multirow{2}{*}{Size} & \multicolumn{5}{c|}{FP-based models} & \multicolumn{4}{c|}{GNN-based models} \\
\cmidrule(lr){3-7} \cmidrule(lr){8-11}
 & & RF & XGBoost & LightGBM & GradBoost & AdaBoost & GCN & GAT & GIN & Pepland \\
\midrule

antimicrobial & 52941  & $88.0_{\pm0.3}$ & $87.5_{\pm0.5}$ & $87.3_{\pm0.4}$ & $85.1_{\pm0.3}$ & $82.5_{\pm0.4}$ & $81.4_{\pm0.7}$ & $72.3_{\pm1.5}$ & $80.2_{\pm1.6}$ & $80.2_{\pm1.3}$ \\
antifungal & 12887  & $87.0_{\pm1.0}$ & $86.0_{\pm0.7}$ & $85.6_{\pm0.8}$ & $83.8_{\pm0.6}$ & $82.0_{\pm0.8}$ & $76.0_{\pm4.4}$ & $73.5_{\pm1.6}$ & $78.9_{\pm2.9}$ & $75.2_{\pm2.4}$ \\
antiparasitic & 6755  & $86.8_{\pm1.2}$ & $86.2_{\pm1.7}$ & $86.5_{\pm0.8}$ & $84.5_{\pm1.1}$ & $82.8_{\pm1.6}$ & $75.6_{\pm1.5}$ & $75.8_{\pm1.4}$ & $74.4_{\pm2.6}$ & $74.8_{\pm2.3}$ \\
anticancer & 12013  & $86.9_{\pm0.8}$ & $86.4_{\pm0.6}$ & $86.0_{\pm0.5}$ & $83.3_{\pm0.8}$ & $81.0_{\pm0.8}$ & $75.0_{\pm4.1}$ & $73.7_{\pm1.0}$ & $77.3_{\pm4.3}$ & $78.1_{\pm1.2}$ \\
allergen & 2538  & $76.5_{\pm2.9}$ & $82.2_{\pm2.0}$ & $82.2_{\pm1.4}$ & $79.4_{\pm1.7}$ & $66.8_{\pm2.8}$ & $63.3_{\pm1.6}$ & $59.9_{\pm4.7}$ & $67.2_{\pm2.7}$ & $68.8_{\pm2.5}$ \\
\midrule
Average & -  & 85.0 & 85.6 & 85.5 & 83.2 & 79.0 & 74.3 & 71.0 & 75.6 & 75.4 \\
\bottomrule
\end{tabular}}
\vspace{1.5mm}
\resizebox{\textwidth}{!}{%
\definecolor{lightgray}{gray}{0.9}
\begin{tabular}{l|c|ccc|cccc}
\toprule
\multirow{2}{*}{Dataset} & \multirow{2}{*}{Size} & \multicolumn{3}{c|}{SMILES-based models} & \multicolumn{4}{c}{PLM-based models} \\
\cmidrule(lr){3-5} \cmidrule(lr){6-9}
 & & ChemBERTa & PeptideCLM & PepDoRA & ESM2-8M & ESM2-35M & ESM2-150M & ESM2-150M-F \\
\midrule

antimicrobial & 52941  & $88.0_{\pm1.1}$ & $52.7_{\pm10.4}$ & $77.9_{\pm3.4}$ & $91.8_{\pm0.8}$ & $91.8_{\pm0.4}$ & $\cellcolor{lightgray}{92.4_{\pm0.4}}$ & $\mathbf{93.3_{\pm1.2}}$ \\
antifungal & 12887  & $84.1_{\pm1.5}$ & $50.7_{\pm6.7}$ & $70.3_{\pm2.2}$ & $90.9_{\pm0.5}$ & $91.5_{\pm1.6}$ & $\mathbf{92.6_{\pm0.9}}$ & $\cellcolor{lightgray}{91.8_{\pm0.7}}$ \\
antiparasitic & 6755  & $82.0_{\pm1.6}$ & $59.2_{\pm4.7}$ & $69.2_{\pm1.8}$ & $91.6_{\pm0.7}$ & $91.9_{\pm1.2}$ & $\mathbf{92.8_{\pm1.2}}$ & $\cellcolor{lightgray}{92.6_{\pm0.8}}$ \\
anticancer & 12013  & $87.0_{\pm1.6}$ & $55.5_{\pm1.3}$ & $70.9_{\pm2.3}$ & $92.4_{\pm0.9}$ & $\cellcolor{lightgray}{93.1_{\pm0.6}}$ & $\mathbf{93.3_{\pm0.5}}$ & $92.8_{\pm0.7}$ \\
allergen & 2538  & $74.8_{\pm3.1}$ & $53.8_{\pm2.3}$ & $55.7_{\pm1.6}$ & $84.0_{\pm1.1}$ & $83.4_{\pm3.1}$ & $\mathbf{87.3_{\pm2.6}}$ & $\cellcolor{lightgray}{85.9_{\pm1.8}}$ \\
\midrule
Average & -  & 83.2 & 54.4 & 68.8 & 90.1 & 90.4 & \textbf{91.7} & \cellcolor{lightgray}{91.3} \\
\bottomrule
\end{tabular}}
\end{table}

\clearpage
\section{Non-natural Peptide Generation Model}
\label{app:nc_gen:sec}

\subsection{Motivation for a Sequence-to-Sequence Generation Model}

Non-natural peptides, which incorporate non-canonical amino acids or modified backbones, represent a significant extension of their natural counterparts. They exhibit enhanced pharmacokinetic properties, such as improved stability and oral bioavailability, which directly address the principal limitations of traditional peptide therapeutics, including rapid \textit{in vivo} degradation. These attributes have expanded their utility in diverse therapeutic areas, from antimicrobial agents to anticancer treatments and membrane translocation.

The systematic design of non-natural peptides, however, faces substantial challenges. The primary obstacle is the pronounced data scarcity in public repositories. This scarcity, compounded by the vast and heterogeneous chemical space of non-canonical monomers, severely constrains the applicability of mainstream deep learning models. Consequently, a methodological gap persists, where prior methods for sequence generation, predominantly tailored for natural peptides or small molecules, lack effective mechanisms to model the complex sequence-property relationships in the non-natural domain. Applying these models without appropriate biophysical and synthetic constraints often yields chemically implausible structures.

This work introduces a sequence-to-sequence framework for the generation of non-natural peptides from natural peptide templates. We employ a large language model, specifically a GPT-2 architecture, to capture the implicit grammatical and semantic correspondences between natural and non-natural sequences. Our approach facilitates the generation of a virtual library of non-natural peptides whose designs are guided by existing natural sequences, thereby enhancing their potential chemical and pharmacological relevance. The framework is designed to improve the controllability and diversity of sequence generation, providing a structured methodology for exploring the non-natural peptide space.

\subsection{Model Architecture, Data, and Training Procedure}

\paragraph{Sequence-to-Sequence Formulation.} We formulate the generation of non-natural peptides as a conditional sequence-to-sequence conversion task, where a natural peptide sequence serves as the input condition from which a corresponding non-natural sequence is generated. We select the GPT-2 architecture as the backbone model, given its established capacity for contextual modeling in analogous translation and generation tasks.

\paragraph{Data Representation and Preprocessing.} To support this formulation, a custom vocabulary was constructed which explicitly defines all canonical amino acids and a set of task-specific symbols, including \texttt{[NATURAL]}, \texttt{[BILN]}, and \texttt{[EOS]}, as discrete tokens. The model's token embedding layer was resized to accommodate this expanded vocabulary, where the \texttt{[EOS]} token also serves as the padding token. The training corpus consists of paired sequences, each structured as \texttt{[NATURAL]} $<$ natural\_seq $>$ \texttt{[BILN]} $<$non-natural\_seq$>$  \texttt{[EOS]}. All sequences were tokenized and processed to a maximum length of 768, where shorter inputs were padded while longer ones were truncated. Data handling was managed using the Hugging Face \texttt{datasets} library.

\paragraph{Model Training and Optimization.} The model was fine-tuned using a causal language modeling (CLM) objective within the Hugging Face \texttt{Trainer} framework. Optimization was performed with the AdamW algorithm, which was configured with a learning rate of $1\times 10^{-4}$ and a weight decay coefficient of 0.01. A linear learning rate warm-up schedule was applied over the initial 500 steps. The model was trained for 500 epochs with a batch size of 32, leveraging mixed-precision (FP16) computation to ensure resource efficiency on a single NVIDIA A800 GPU.

\paragraph{Inference and Post-processing.} For sequence generation, the model checkpoint exhibiting the optimal validation performance was utilized. Inference was initiated with a prompt containing the natural sequence and the \texttt{[BILN]} tag, after which the model autoregressively completed the non-natural sequence via greedy decoding. A rule-based post-processing procedure was subsequently applied to extract the sequence segment between the \texttt{[BILN]} and \texttt{[EOS]} tokens and to remove any residual special symbols, thereby yielding the final non-natural peptide sequence. All implementations were based on the PyTorch and Hugging Face Transformers libraries.

\subsection{Evaluation Metrics and Results for Generation Quality}

\paragraph{Chemical Validity} The chemical validity of the generated sequences was assessed based on the success rates of their conversion to SMILES representations and the subsequent validation thereof by RDKit. The model achieved high rates for both metrics (Table~\ref{tab:eval}), which confirms its capacity to generate chemically plausible molecular structures from the target sequence representation.

\paragraph{Fidelity to Reference Structures} Fidelity was quantified using two complementary metrics, which were the exact match rate against reference sequences and the average Tanimoto similarity of Morgan fingerprints (radius=2, 1024 bits). The model yielded a low exact match rate, whereas the structural similarity remained high. This outcome indicates that the model generalizes beyond simple sequence replication, enabling the generation of structurally analogous yet novel molecules.

\begin{figure}[t]
\centering
\begin{subfigure}[b]{0.95\linewidth}
    \includegraphics[width=\linewidth]{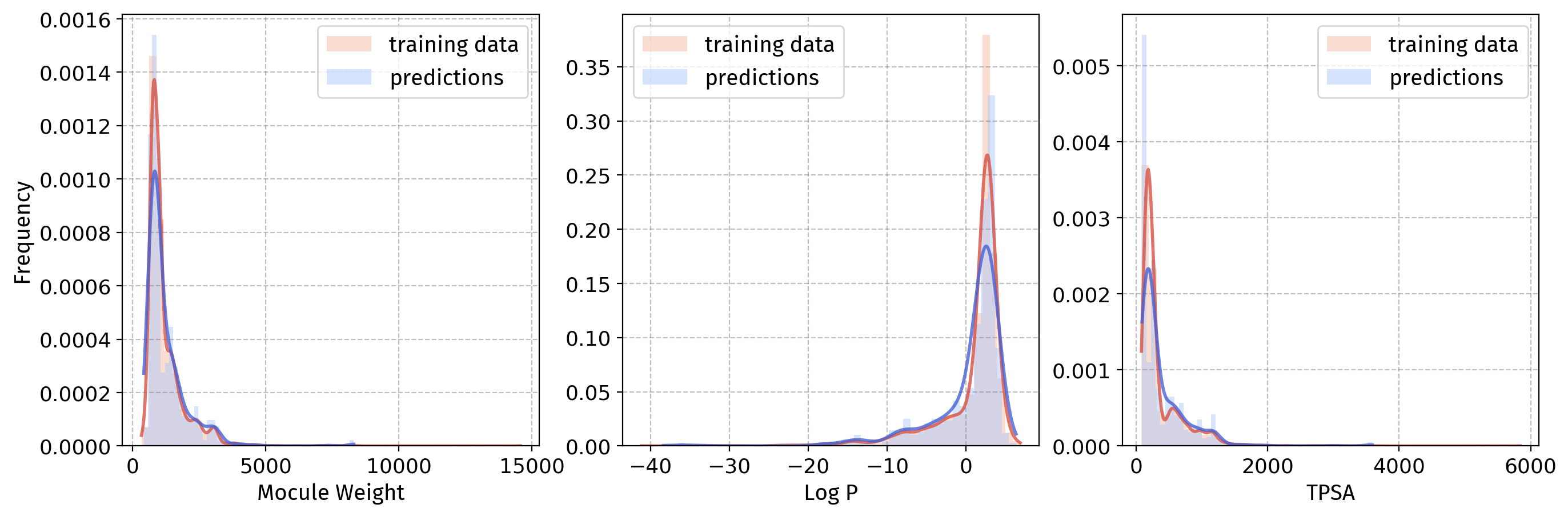}
\end{subfigure}

\vspace{0.5cm} 

\begin{subfigure}[b]{0.65\linewidth}
    \includegraphics[width=\linewidth]{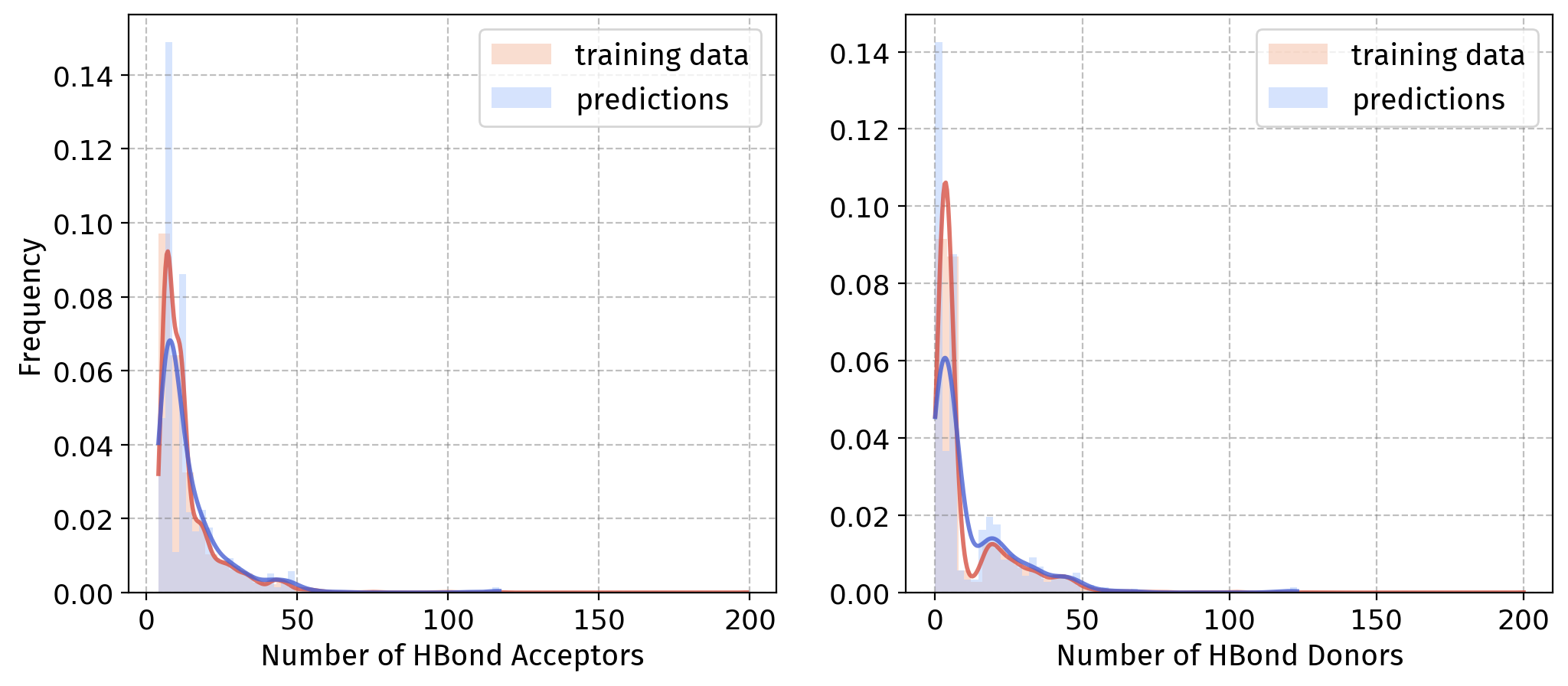}
\end{subfigure}

\caption{Comparison of physicochemical property distributions between generated molecules and the training set.}
\label{fig:dist}
\end{figure}

\paragraph{Physicochemical Property Distribution} A comparison of physicochemical property distributions was conducted between the generated molecular ensemble and the training set. The properties analyzed included molecular weight (MolWt), lipophilicity (LogP), hydrogen bond donor and acceptor counts (NumHDonors, NumHAcceptors), and topological polar surface area (TPSA). The generated molecules exhibit distributions closely aligned with the training set (Figure~\ref{fig:dist}), confirming that the model preserves the underlying statistical characteristics of the reference data.

\paragraph{Diversity and Novelty} Diversity was measured by internal pairwise Tanimoto similarity, while novelty was defined as the fraction of generated molecules absent from the training set. The model produced a generated set with high internal diversity and a substantial novelty ratio (Table~\ref{tab:eval}). These two results, taken together, demonstrate the model's ability to explore the chemical space beyond the training data.

\paragraph{Sequence Modeling Performance} The model's sequence modeling capability was evaluated using perplexity (PPL) on the validation set. A final perplexity of 1.23e3 was achieved, which indicates a robust capture of the statistical patterns inherent in the non-natural peptide sequence representation.


\begin{table}[t]
\centering
\caption{Comprehensive evaluation of generated molecules across multiple dimensions.}
\label{tab:eval}
\begin{tabular}{lcc}
\hline
\textbf{Category} & \textbf{Metric} & \textbf{Value} \\
\hline
\multirow{3}{*}{Validity} 
  & BLIN$\rightarrow$SMILES Success Rate & 89.7\% \\
  & Chemical Validity Rate & 100\% \\
  & Overall Success Rate & 89.7\% \\
\hline
\multirow{2}{*}{Closeness to Ground Truth} 
  & Exact Match Rate & 21.9\% \\
  & Avg. Tanimoto Similarity & 87.9\% \\
\hline
\multirow{2}{*}{Diversity and Novelty} 
  & Internal Diversity & 66.7\% \\
  & Novelty Ratio & 76.5\% \\
\hline
Language Modeling & Perplexity (PPL) & 1232.0 \\
\hline
\end{tabular}
\end{table}

\subsection{Inference Pipeline for Large-Scale Dataset Generation}

\paragraph{Inference Configuration} For generation, the trained model and tokenizer were loaded, with the model set to evaluation mode. The padding token was mapped to the end-of-sequence token to ensure consistent autoregressive decoding. Generation was conditioned on prompts which follow a unified template, \texttt{[NATURAL] {natural\_peptide\_sequence} [BILN]}, thereby maintaining structural consistency with the training format. The maximum number of new tokens to be generated was dynamically adjusted relative to the input sequence length within each batch, a process which, combined with batch processing, improves computational efficiency. The generation process accommodates both greedy decoding for deterministic output and temperature-controlled sampling, where the temperature parameter modulates output stochasticity.

\paragraph{Application and Performance} The described protocol was applied to convert a corpus of 4,323 natural peptide sequences derived from four distinct datasets. We conducted eight independent generation runs using a range of temperature parameters, from which up to five chemically valid non-natural sequences were retained per input. This procedure successfully yielded corresponding non-natural peptide sequences for 4,221 of the inputs, which represents a conversion success rate of 97.6


\section{Toolkit Support}
\label{app:useage:sec}
To facilitate reproducible and standardized dataset construction for the research community, we introduce \texttt{PepBenchmark}, a Python package designed with a modular and extensible architecture. The framework allows users to compose, customize, and extend pipeline components to meet diverse experimental requirements. It integrates peptide-specific utilities for dataset analysis and visualization, and provides six splitting strategies---including \emph{k}-mer, MMseqs2, ECFP, random, yydra, and protein cold-start---to ensure rigorous control of partitioning under sequence similarity and distributional constraints. In addition, \texttt{PepBenchmark} offers negative-sampling management tools that support the construction of dynamic negative pools with adjustable similarity thresholds and physicochemical property matching. A sequence similarity analysis module is also included for redundancy removal and identity-based filtering, thereby guaranteeing strict benchmarking conditions.

\texttt{PepBenchmark} is under active development, and we plan to continuously integrate additional functionalities that are broadly applicable to peptide-related machine learning tasks. The source code is provided in the supplementary material and will be made publicly available.

Below we illustrate typical usage patterns.

\lstset{
  language=Python,
  basicstyle=\ttfamily\small,
  columns=fullflexible,
  frame=single,
  breaklines=true,
  showstringspaces=false,
  captionpos=b
}
\clearpage
\paragraph{Basic Usage}
For users who only need standardized datasets, \texttt{PepBenchmark} offers ready-to-use processed datasets that can be loaded, split, trained, and evaluated with just a few lines of code.

\begin{lstlisting}
# 1. Load a standardized dataset
dataset = SinglePeptideDatasetManager(
    dataset_name="bbp",
    official_feature_names=["fasta", "label"]
)

# 2. Set the split strategy (reproducible with a fixed seed)
dataset.set_official_split_indices(split_type="mmseqs2_split", fold_seed=0)

# 3. Initialize a baseline model
model = PLM(model="facebook/esm2_t30_150M_UR50D", dataset=dataset)

# 4. Train and evaluate
model.run(epochs=50, metrics=["roc-auc"])

# Users may also export processed features to train their own models
train_features, val_features, test_features = dataset.get_train_val_test_features(format="dict")
\end{lstlisting}

\paragraph{Advanced Usage}
For advanced users who wish to build customized datasets, the package exposes modular utilities for redundancy removal, negative sampling, splitting, and feature conversion.

\begin{lstlisting}
# 1. Remove redundancy using MMseqs2 (sequence identity threshold = 0.9)
dedup_seqs, result = remove_redundancy(pos_seq, method="mmseqs2", identity=0.9)

# 2. Negative sampling pipeline
# 2.1 Construct a negative pool from existing datasets
pool_manager = SamplingPoolManager(include_datasets=["cpp", "bbp"])  

# 2.2 Remove sequences in the pool that are too similar to the positive set (identity threshold = 0.6)
pool_manager.remove_similar_to_positives(dedup_seqs, method="mmseqs2", threshold=0.6)  

# 2.3 Initialize a sampler that generates negatives with distributional control
neg_sampler = NegSampler(pool_manager.get_sampling_pool(), dedup_seqs)

# 2.4 Sample negatives while matching key physicochemical properties
neg_seqs = neg_sampler.sample_negatives(
    ratio=1,                     # 1:1 ratio of negatives to positives
    method="bin",                # property-matching method
    properties=["length", "charge"]  # distributions to preserve
)

# 2.5 Combine positive and negative sequences into a single dataset
fasta_list = pos_seq + neg_seqs
labels = [1] * len(pos_seq) + [0] * len(neg_seqs)

# 3. Dataset splitting with hybridSplitter
#    - Splits into train/validation/test while controlling redundancy and motif enrichment
hybrid_splitter = HybridSplitter(
    fasta_list, labels,
    frac_train=0.8,
    frac_valid=0.1,
    frac_test=0.1,
    cluster_distribution_strategy="sort_cluster_size",  # distribute clusters by size
    preserve_cluster_integrity=True,   # ensure clusters remain intact
    seed=42,                           # reproducibility
    ks=5,                              # number of random splits
    test_method="fisher",              # statistical test for motif enrichment
    alternative="greater",
    min_cluster_size=3,                # motifs must cover at least 3 samples
    min_pos=3,                         # motif must appear at 3 times
    mode="all",
    pval_threshold=0.05,               # significance cutoff
    min_score=4.0,                     # enrichment score threshold
    fdr_correct=True,                  # apply FDR correction
    min_support=5,                     # minimum supporting sequences
    min_jaccard=0.6                    # Jaccard similarity cutoff
)
split_res = hybrid_splitter.get_split_indices()

# 4. Feature transformation: convert FASTA sequences into SMILES strings
converter = Fasta2Smiles()
smiles_list = converter(fasta_list)
\end{lstlisting}

\end{document}